\documentclass{article}
\usepackage{amsmath, amssymb, amsthm}
\usepackage{graphicx}
\usepackage{natbib}
\usepackage{hyperref}
\usepackage{geometry}
\usepackage{xcolor}

\usepackage{amsthm}
\usepackage{bm}
\usepackage{comment}
\usepackage{booktabs}

\usepackage{placeins}

\usepackage{subcaption}

\newtheorem{theorem}{Theorem}

\usepackage{algorithmic} 
\usepackage[ruled,vlined]{algorithm2e}
\usepackage{enumitem}

\usepackage{mathrsfs}

\usepackage{adjustbox}

\newtheorem{lemma}{Lemma}
\newtheorem{proposition}{Proposition}
\newtheorem{assumption}{Assumption}

\newtheorem{corollary}{Corollary}

\title{Bayesian transfer for offline Q-learning with multi-task Gaussian processes}

\author{
  Wenxing Guo\\
  School of Mathematics, Statistics and Actuarial Science\\
  University of Essex\\
  Colchester, United Kingdom
}

\date{}

\begin{document}

\maketitle

\begin{abstract}
\noindent
Offline reinforcement learning can benefit from data collected in related tasks, yet indiscriminate knowledge transfer may lead to negative transfer under task mismatch and limited target-data coverage. We propose a target-aware Bayesian transfer framework for offline Q-learning based on multi-task Gaussian processes that jointly model task-specific Q-functions. Source tasks are assessed using target predictive evidence and maximum mean discrepancy (MMD), while posterior uncertainty and behavior-supported action sets guide conservative Bellman updates and policy extraction. To account for the dependence induced by Bellman bootstrapping, we distinguish Bellman innovations from bootstrap errors and establish stability and predictive source-selection results under explicit regularity conditions. We further derive selection-adaptive posterior contraction and offline policy-performance bounds, identifying the structural conditions under which source information can be pooled to achieve minimax-optimal rates in a shared-function Gaussian sequence experiment. Simulation studies with heterogeneous source tasks demonstrate improved target Q-function estimation and reduced material negative transfer compared with indiscriminate source pooling. An application to human decision-making data further shows higher average expected immediate payoff than both target-only learning and matched-size random source selection. These results highlight the benefits of combining target-aware source selection with uncertainty-aware conservative learning for reliable and sample-efficient offline reinforcement learning.
\end{abstract}


\section{Introduction}
\label{sec:introduction}

Transfer learning in reinforcement learning (RL) seeks to accelerate learning in a target task by leveraging knowledge acquired from related source tasks. Rather than learning each decision-making problem in isolation, an agent may reuse experience, value functions, policies, or representations across tasks. This is particularly attractive when collecting target-task interactions is expensive or only a limited offline dataset is available. However, the usefulness of transferred knowledge depends on the relationship between the source and target tasks. Differences in rewards, transition dynamics, and state--action coverage can make apparently relevant experience misleading, resulting in negative transfer \citep{taylor2009transfer,lazaric2008transfer}. The central question is therefore not merely how to transfer knowledge, but how to identify which sources are useful for a particular target and how to account for uncertainty in that decision.
This question becomes especially important in offline RL. A learner operating on a fixed dataset cannot freely collect additional interactions to correct errors caused by inappropriate transfer. Moreover, increasing the amount of training data does not necessarily improve decision quality when the additional data change the distribution on which value estimates are fitted. Extrapolation beyond dataset support can undermine offline value learning \citep{fujimoto2019off}, while indiscriminate sharing across tasks can further amplify distributional shift \citep{yu2021conservative}. These observations motivate a transfer mechanism that considers task relevance, predictive uncertainty, and the target decision problem jointly.
Existing research studies several complementary forms of knowledge reuse. \citet{lazaric2008transfer} transfer experience samples according to source-task compliance and sample relevance, explicitly considering differences between task models. \citet{tirinzoni2018importance} develop an importance-weighted sample-transfer approach and provide a finite-sample analysis of the resulting batch RL procedure. Another line of work transfers structured value representations: successor features separate reward-dependent components from shared dynamics and support generalized policy improvement across tasks with common transition dynamics \citep{barreto2017successor}. 
More recently, \cite{chen2026batchq}
developed a transfer fitted Q-iteration framework for
batch reinforcement learning and established statistical
error bounds that account for task discrepancies and
source and target sample sizes.
\cite{luo2026direct} proposed a joint
estimation approach that incorporates source information
into target Q-function learning through an
$\ell_1$-penalized objective, with theoretical guarantees
for target estimation and policy learning under
homogeneous or heterogeneous transition dynamics.
These approaches study knowledge transfer through
regression-based estimation and task-specific
discrepancy correction.
Our framework instead models cross-task dependence
probabilistically through a multi-task Gaussian process
and investigates target-aware source selection under
Bellman-generated regression responses.
These methods illustrate different ways of exploiting task relationships. Our focus is instead on representing cross-task dependence probabilistically at the level of Q-functions and using target-specific predictive evidence to guide source selection.

Gaussian processes (GPs) provide a nonparametric framework for specifying distributions over functions and obtaining posterior predictions under an explicit observation model \citep{rasmussen2006gaussian}. GP-based RL methods have used this machinery for value estimation and control \citep{engel2005reinforcement}. \citet{grande2014sample} further analyze sample-efficient model-based and model-free GP algorithms, demonstrating that learning guarantees depend on the algorithmic treatment of uncertainty rather than on the use of a GP alone. For multiple related outputs, \citet{bonilla2007multitask} combine an input covariance function with a learned task covariance matrix, while \citet{alvarez2012kernels} describe the broader connection between multi-output GPs and vector-valued reproducing kernel Hilbert spaces (RKHSs). 
\cite{chowdhary2014offpolicy}
developed an off-policy Gaussian-process Q-learning
framework and established convergence conditions for
batch learning.
\cite{wang2025nonstationary} proposed a nonstationary multi-output Gaussian process with
dynamic spike-and-slab priors on task correlations
to identify informative sources and mitigate negative
transfer, including an application to reinforcement
learning.
These studies demonstrate the usefulness of
Gaussian processes for value-function learning
and adaptive information sharing.
Our focus is on target-specific predictive source
selection within offline Q-learning, together with
statistical analysis of the dependence and errors
introduced by Bellman bootstrapping.
These constructions offer a natural language for modeling similarities among task-specific value functions.
A particularly close predecessor is the Bayesian multi-task RL framework of \citet{lazaric2010bayesian}. Their approach combines Gaussian-process temporal-difference modeling with hierarchical Bayesian priors, sharing information among related value functions and transferring the learned structure to a newly observed task. Building on this foundation, the present work focuses on target-aware source-subset selection within iterative Q-learning. In particular, we study source usefulness through target prediction risk and explicitly account for the fact that the regression responses are constructed through Bellman updates rather than supplied as independent observations of the true Q-function.

Our approach also relates to methods that control the consequences of uncertainty in offline RL. Batch-constrained learning restricts action selection to reduce extrapolation error \citep{fujimoto2019off}, whereas conservative Q-learning regularizes value estimates to mitigate overestimation under distribution shift \citep{kumar2020conservative}. \citet{jin2021pessimism} provide a theoretical treatment of pessimistic value iteration using uncertainty-dependent penalties. In the multi-task setting, conservative data sharing routes experience according to its usefulness for individual tasks instead of pooling all datasets indiscriminately \citep{yu2021conservative}. 
\cite{bai2024pessimistic} developed an
uncertainty-based multi-task data-sharing framework
that applies pessimistic value iteration to pooled
offline datasets without explicit data selection.
Their analysis highlights the role of data coverage
and uncertainty in controlling policy suboptimality.
In contrast, our approach explicitly evaluates
candidate source tasks using target-predictive
evidence and combines source selection with
Gaussian-process uncertainty penalties and
behavior-supported action maximization.
More recent model-based work relates uncertainty penalties to disagreement in Bellman estimates across learned dynamics models \citep{sun2023model}. Our method follows the conservative-learning motivation but uses posterior GP uncertainty together with task selection and behavior-supported action maximization. The resulting pessimistic score is a decision rule; its interpretation as a valid lower confidence bound requires the corresponding statistical assumptions.

Domain-adaptation theory distinguishes distributional similarity from compatibility of the prediction tasks themselves. For example, \citet{bendavid2010theory} relate target generalization to source performance, domain divergence, and a joint approximation term. In kernel methods, maximum mean discrepancy (MMD) compares distributions through differences between their RKHS mean embeddings \citep{gretton2012kernel}. This provides a useful way to quantify mismatch between source and target state--action distributions. Nevertheless, a small discrepancy between these marginals does not by itself establish that two MDPs have compatible rewards or transition dynamics: the same state--action distribution can accompany different Bellman operators. Motivated by this distinction, we treat MMD-based penalties as complementary distributional information for source assessment, rather than as a stand-alone certificate of beneficial transfer.
The statistical analysis of transfer Q-learning must also address the recursive nature of its training targets. Finite-time analyses of fitted value iteration connect regression error, function-class approximation, and error propagation through Bellman updates \citep{munos2008finite}. Separately, GP posterior contraction theory relates convergence to RKHS approximation and small-ball probabilities of the prior \citep{vandervaart2008rates}. These results provide useful foundations, but applying regression contraction arguments to Q-learning requires additional control of trajectory dependence and bootstrap error. 

We investigate these issues in a Bayesian transfer framework that models the target and source Q-functions jointly. For task indices $i,j$ and shared state--action representations $x,x'$, we use the separable covariance structure
\(
K\bigl((i,x),(j,x')\bigr)
=K_{\mathrm{task}}(i,j)\,k_{\mathrm{in}}(x,x').
\)
The input kernel describes within-task regularity, while the task kernel represents cross-task dependence. We consider both source datasets with Q-function labels and source transition datasets whose labels are constructed iteratively. Bayesian source-subset selection is formulated around target predictive performance, with sparsity-inducing inclusion indicators allowing sources to be excluded. The selection objective is target-specific: it concerns the usefulness of a source subset for the target, rather than recovery of every relationship among all tasks.
For offline policy learning, posterior predictions are combined through a pessimistic score
\(
Q^{-}(s,a)=\mu_{\mathrm{post}}(s,a)
-\alpha_{\mathrm{pess}}\sigma_{\mathrm{post}}(s,a),
\)
and actions are optimized over a behavior-supported set. This combines two distinct safeguards: source selection addresses potentially misleading cross-task information, while pessimism and action restrictions address uncertainty in the resulting target estimates. On the theoretical side, Bellman-generated responses are decomposed into the optimal Q-function, Bellman innovation noise, and a bootstrap remainder. This decomposition organizes our investigation of how one-step regression accuracy, source selection, and target posterior concentration interact. Here, safe transfer refers to comparison with a target-only learning strategy under the stated assumptions, not to an unconditional guarantee of improvement for every finite dataset.

This paper makes four main contributions.
First, we develop a target-aware Bayesian transfer framework for offline Q-learning based on multi-task Gaussian processes. By jointly modeling task-specific Q-functions through a separable covariance structure, the proposed framework accommodates both directly labeled source data and transition-only source data. It integrates target-predictive source-subset selection, posterior uncertainty quantification, and conservative policy extraction through uncertainty penalties and behavior-supported action sets.
Second, we establish a theoretical foundation for source selection under Bellman bootstrapping. By decomposing Bellman-generated responses into conditionally centered innovations and bootstrap errors, we derive uniform Bellman residual bounds and establish bootstrap consistency under explicit regularity conditions. We further analyze how bootstrap perturbations affect target-predictive scores and derive posterior concentration results for predictively near-optimal source subsets. Under an additional calibration condition linking predictive scores to target risk, these results also imply target-risk near-oracle selection.
Third, we investigate transfer-adaptive posterior contraction and the statistical benefits of source information. We establish conditions under which predictive source selection and control of Bellman errors yield contraction of the model-averaged target posterior. Our analysis distinguishes the target-information bottleneck associated with full-rank task covariance from the precision pooling available under a rank-one shared-function model. In the latter Gaussian sequence experiment, we establish minimax-optimal contraction and estimation rates. We also derive offline policy-performance bounds that explicitly account for Q-function estimation error, action coverage, and uncertainty-aware pessimism.
Fourth, we evaluate the proposed method through controlled simulations and an application to real human decision-making data. The simulations examine heterogeneous source tasks, target-aware selection, negative-transfer behavior, Q-function estimation accuracy, and the contribution of pessimistic policy extraction. The real-data analysis further evaluates cross-participant transfer and source selection against target-only learning and matched-size random source selection. Across these settings, the results demonstrate improved average estimation or decision performance while highlighting the conditions and limitations of safe knowledge transfer.
Together, these contributions provide a unified statistical framework for understanding when source information can improve offline Q-learning, how transferable sources can be identified, and how uncertainty and task mismatch affect the reliability of transfer.

Section~2 introduces the problem setup, multi-task GP model, and learning algorithms.
Section~3 studies Bellman bootstrap stability and target-aware source selection.
Section~4 investigates posterior contraction and its connections to transfer gains and offline policy performance.
Section~5 presents the simulation design and experimental results.
Finally, Section~6 provides a real-data analysis to evaluate cross-participant transfer and target-aware source selection in human decision-making tasks.

\section{Preliminary and problem setup}

\subsection{Notation}
\label{subsec:notation}

For a positive integer $J$, let $[J]=\{1,\ldots,J\}$, and let $|S|$
denote the cardinality of a finite set $S$. We use boldface symbols
for finite-dimensional vectors and matrices, and write
$\mathbf{A}^{\top}$ for the transpose of $\mathbf{A}$.
The identity matrix of dimension $n$ is denoted by $\mathbf{I}_n$.
For a vector $\mathbf{v}$ and a matrix $\mathbf{A}$,
$\|\mathbf{v}\|_2$ and $\|\mathbf{A}\|_{\mathrm{op}}$ denote the
Euclidean and operator norms, respectively. For a symmetric matrix,
$\lambda_{\min}(\mathbf{A})$ and $\lambda_{\max}(\mathbf{A})$ denote
its smallest and largest eigenvalues.
We write $\mathbb{P}$ and
$\mathbb{E}$ for probability and expectation, and
$\mathcal{N}(\mathbf{m},\boldsymbol{\Sigma})$ for a Gaussian
distribution with mean $\mathbf{m}$ and covariance
$\boldsymbol{\Sigma}$.
Task $0$ is the target task, and tasks $k\in[J]$ are source tasks.
A parenthesized superscript $(k)$ indicates a task-specific quantity.
The state and action spaces of task $k$ are denoted by
$\mathcal{S}^{(k)}$ and $\mathcal{A}^{(k)}$, respectively, and
$\gamma\in(0,1)$ is the discount factor. In the common-space setting,
we omit the task superscripts on the spaces and write $x=(s,a)$.
When task-specific representations are used, we write
\[
    x=\phi^{(k)}(s,a)\in\mathcal{X},
    \qquad
    \phi^{(k)}:
    \mathcal{S}^{(k)}\times\mathcal{A}^{(k)}
    \longrightarrow\mathcal{X},
\]
where $\mathcal{X}$ is the shared input space. The symbols
$Q^{(k)}$, $Q^{(k)}_*$, and $\widehat{Q}^{(k)}_e$ denote a generic
Q-function, the optimal Q-function, and its estimate at iteration
$e$, respectively. We use $\mathcal{T}^{(k)}$ to denote the Bellman optimality operator of task $k$.

The dataset for task $k$ is denoted by $\mathcal{D}^{(k)}$ with
size $n_k$, and the target offline transition dataset is denoted by
$\mathcal{D}^{(0)}_{\mathrm{off}}$. Depending on the setting, source
data consist of labeled pairs $(x^{(k)}_t,y^{(k)}_t)$ or transitions
$(s^{(k)}_t,a^{(k)}_t,r^{(k)}_t,s^{\prime(k)}_t)$.
The index $t$ labels observations or trajectory steps, whereas $e$
labels Bellman regression iterations. The quantity $y_{t,e}^{(k)}$ denotes the Bellman pseudo-response
constructed for the $t$th transition of task $k$ at iteration $e$, the iteration
index is suppressed for fixed labels. We collect inputs and
responses as
\[
    \mathbf{X}^{(k)}
    =\bigl[x^{(k)}_1,\ldots,x^{(k)}_{n_k}\bigr]^{\top},
    \qquad
    \mathbf{y}^{(k)}_e
    =\bigl(y^{(k)}_{1,e},\ldots,y^{(k)}_{n_k,e}\bigr)^{\top}.
\]
The symbols $X$ and $Y_e$ denote the collections of task inputs
and iteration-$e$ responses, respectively. For the transition data, uppercase symbols denote random variables, whereas the corresponding lowercase symbols denote their observed realizations.
We use $k_{\mathrm{in}}$ to denote the input kernel and
$\mathbf K_{\mathrm{task}}$ to denote the task covariance matrix,
whose $(i,j)$th entry is $K_{\mathrm{task}}(i,j)$.
For the input collections
$\mathbf X^{(i)}=\{x_a^{(i)}\}_{a=1}^{n_i}$ and
$\mathbf X^{(j)}=\{x_b^{(j)}\}_{b=1}^{n_j}$,
define the cross-task covariance block
$\mathbf K^{(i,j)}\in\mathbb R^{n_i\times n_j}$ by
\[
    \bigl[\mathbf K^{(i,j)}\bigr]_{ab}
    =
    K_{\mathrm{task}}(i,j)\,
    k_{\mathrm{in}}\!\left(x_a^{(i)},x_b^{(j)}\right),
    \qquad
    1\le a\le n_i,\quad 1\le b\le n_j.
\]
The prior mean used in multi-task regression is denoted by
$m^{(k)}$. At iteration $e$, $\mu^{(k)}_e(x)$ denotes the posterior
mean for task $k$, $\Sigma_e$ denotes the joint posterior covariance,
and $\sigma^{(k)}_e(x)$ denotes the posterior standard deviation of
the latent task-$k$ Q-function at $x$. The subscript
$\mathrm{post}$ is used when the iteration index is suppressed.
In contrast, $\sigma_k^2$ denotes an observation-noise variance or
a sub-Gaussian variance proxy according to the observation model
under consideration.

A candidate source subset is denoted by $S\subseteq[J]$, and
$\mathcal{S}_J=2^{[J]}$ is the class of all such subsets.
The empty subset $S=\varnothing$ corresponds to target-only learning.
We use $S_{\mathrm{tr}}$ to denote a specified transferable source
subset and $\widehat S$ to denote a data-dependent selected subset.
The notation $\mathcal S^\star$ is reserved for the oracle subset
class introduced in the selection analysis.
The notation $\Pi(S)$ denotes the prior probability of a subset,
while $\Pi(\,\cdot\mid Y_e,X)$ denotes the relevant posterior
conditional on the inputs and pseudo-responses. We write
$n_{k,\mathrm{eff}}$ for the effective sample size of task $k$;
$N_{\mathrm{eff}}$ and $N_{\mathrm{safe}}$ denote the aggregate
effective sample sizes specified in the selection and safe-transfer
analyses, respectively.
Let $P_X^{(0)}$ denote the target input distribution. For a
real-valued function $f$ on $\mathcal{X}$, define
\[
    \|f\|_{\infty}
    =\sup_{x\in\mathcal{X}}|f(x)|,
    \qquad
    \|f\|_{L^2(P_X^{(0)})}
    =
    \left\{
        \int_{\mathcal{X}} |f(x)|^2\,
        \mathrm{d}P_X^{(0)}(x)
    \right\}^{1/2}.
\]
The target prediction risk of an estimator $\widehat{Q}$ is
written as
\[
    R_0(\widehat{Q})
    =
    \bigl\|\widehat{Q}-Q_*^{(0)}
    \bigr\|_{L^2(P_X^{(0)})}^2.
\]
For a reproducing kernel Hilbert space $\mathcal{H}$, its inner
product and norm are denoted by
$\langle\cdot,\cdot\rangle_{\mathcal{H}}$ and
$\|\cdot\|_{\mathcal{H}}$. In particular,
$\mathcal{H}_{\mathrm{in}}$ and $\mathcal{H}_{\mathrm{MT}}$ denote
the RKHSs associated with the input and multi-task kernels.
We use $H^{\alpha}(\mathcal{X})$ for the Sobolev space of smoothness
$\alpha>0$. To distinguish smoothness from conservative value
penalization, we denote the pessimism coefficient by
$\alpha_{\mathrm{pess}}\geq 0$.

For nonnegative sequences $a_N$ and $b_N$, we write
$a_N\lesssim b_N$ if $a_N\leq Cb_N$ for a constant $C>0$
independent of $N$, and $a_N\asymp b_N$ if both
$a_N\lesssim b_N$ and $b_N\lesssim a_N$.
The notation $O_{\mathbb{P}}(b_N)$ denotes boundedness in
probability after normalization by $b_N$, whereas
$o_{\mathbb{P}}(b_N)$ denotes convergence to zero in probability
after the same normalization.
We use $\widetilde{O}(\cdot)$ for bounds up to logarithmic factors.
The symbols $c$ and $C$ denote positive finite constants that may
differ across displays.

\subsection{Reinforcement learning}
\label{subsec:reinforcement-learning}

We consider an infinite-horizon discounted Markov decision process
(MDP) \citep{sutton1998reinforcement},
\begin{equation}
    \mathcal{M}=(\mathcal{S},\mathcal{A},P,R,\gamma),
    \label{R-1}
\end{equation}
where $\mathcal{S}$ is a measurable state space, $\mathcal{A}$ is a
nonempty finite action space, 
$P(\cdot\mid s,a)$ denotes the conditional probability distribution
of the next state given the current state--action pair, and $\gamma\in(0,1)$ is
the discount factor. The conditional mean reward is defined by
\[
R(s,a)
=
\mathbb{E}\!\left[r_t\mid S_t=s,\ A_t=a\right].
\]
We assume the uniform moment condition
\[
\sup_{(s,a)\in\mathcal S\times\mathcal A}
\mathbb{E}\!\left[
|r_t|
\mid S_t=s,\ A_t=a
\right]
<\infty.
\]
This condition, together with $\gamma<1$, guarantees integrability of
the infinite-horizon discounted return. The finiteness of
$\mathcal A$ ensures that the action maxima in the Bellman operators
are attained.

A stationary Markov policy $\pi(\cdot\mid s)$ assigns a probability
distribution over actions at each state. Its action-value function is
\begin{equation}
    Q^{\pi}(s,a)
    =\mathbb{E}_{\pi}\!\left[
        \sum_{t=0}^{\infty}\gamma^t r_t
        \;\middle|\; S_0=s,\ A_0=a
    \right],
    \label{R-2}
\end{equation}
where the initial action is fixed to $a$ and subsequent actions follow
$\pi$.
The learning objective is to recover the optimal action-value function
\begin{equation}
    Q_*(s,a)=\sup_{\pi}Q^{\pi}(s,a),
    \label{R-3}
\end{equation}
and an optimal deterministic policy can then be chosen as
\[
\pi_*(s)=
\arg\max_{a\in\mathcal A}Q_*(s,a),
\qquad s\in\mathcal S.
\]
The optimal Q-function satisfies the Bellman optimality equation
\begin{equation}
    Q_*(s,a)
    =R(s,a)+\gamma
    \int_{\mathcal{S}}
        \max_{a'\in\mathcal{A}}Q_*(s',a')\,
        P(\mathrm{d}s'\mid s,a).
    \label{R-4}
\end{equation}
For a bounded measurable function $Q$, define the Bellman optimality
operator by
\[
    (\mathcal{T}Q)(s,a)
    =R(s,a)+\gamma\,
    \mathbb{E}_{S'\sim P(\cdot\mid s,a)}
    \!\left[\max_{a'\in\mathcal{A}}Q(S',a')\right].
\]
The contraction property of the Bellman optimality operator is central
to the subsequent analysis, as it controls the propagation of
approximation errors across successive Q-learning iterations.
Specifically, for any two bounded Q-functions $Q$ and $\widetilde Q$,
\begin{equation}
    \|\mathcal T Q-\mathcal T\widetilde Q\|_\infty
    \le
    \gamma\|Q-\widetilde Q\|_\infty.
    \label{R-contraction}
\end{equation}
Hence, $\mathcal T$ is a $\gamma$-contraction on the space of bounded
Q-functions equipped with the supremum norm. Since $\gamma\in(0,1)$,
the contraction mapping theorem implies that $\mathcal T$ admits a
unique bounded fixed point, namely $Q_*$, and that repeated exact
Bellman updates converge geometrically to $Q_*$.
In fitted Q-learning, however, the exact Bellman update is replaced by
a regression estimate constructed from sampled Bellman targets.
This replacement introduces regression and sampling errors at each
iteration.
The contraction property in \eqref{R-contraction} provides the basic
mechanism for controlling the accumulation of these one-step errors
and for establishing convergence of the fitted iterates to $Q_*$
\citep{munos2008finite}.

In the multi-task setting, task $0$ is the target and tasks
$k\in[J]$ are sources. Unless otherwise stated, they share
$\mathcal{S}$, $\mathcal{A}$, and $\gamma$, but may have different
transition kernels $P^{(k)}$ and reward functions $R^{(k)}$.
We denote their Bellman operators and optimal Q-functions by
$\mathcal{T}^{(k)}$ and $Q_*^{(k)}$, respectively.

\subsection{Gaussian process regression}
\label{subsec:gaussian-process-regression}

Gaussian process regression (GPR) provides a nonparametric Bayesian
framework for modeling unknown functions and quantifying predictive
uncertainty \citep{rasmussen2006gaussian}. To model statistical
dependence across tasks, we place a joint Gaussian process prior on
the collection of task-specific Q-functions. Specifically, define
\[
    Q(k,x)=Q^{(k)}(x),
    \qquad
    (k,x)\in\{0,\ldots,J\}\times\mathcal X,
\]
and assume
\(
    Q
    \sim
    \mathcal{GP}(m,K),
\)
with task-specific mean functions
\(
    m(k,x)=m^{(k)}(x).
\)
In the common-state--action setting, $x=(s,a)$. For heterogeneous
tasks, we instead use a task-specific representation
\(
    x=\phi^{(k)}(s,a)\in\mathcal X,
\)
where $\phi^{(k)}$ maps the original state--action space of task $k$
to a shared representation space $\mathcal X$. In this case, the GP
model assumes that the modeled Q-function factors through the
representation $\phi^{(k)}$, whereas the Bellman operator remains
defined on the original state and action spaces.
Using the notation introduced in Section~\ref{subsec:notation},
we adopt the separable multi-task covariance structure
\begin{equation}\nonumber
    K\bigl((i,x),(j,x')\bigr)
    =
    K_{\mathrm{task}}(i,j)\,
    k_{\mathrm{in}}(x,x'),
    \label{G-1}
\end{equation}
which decomposes the covariance into a task-level component and an
input-level component. Specifically, $K_{\mathrm{task}}(i,j)$
governs the prior dependence between tasks $i$ and $j$, whereas
$k_{\mathrm{in}}(x,x')$ characterizes similarity between
state--action inputs
\citep{bonilla2007multitask,alvarez2012kernels}.
The off-diagonal entries of $\mathbf K_{\mathrm{task}}$
characterize prior dependence across tasks.
Any structural prior on $\mathbf K_{\mathrm{task}}$ is restricted
to the cone of positive-semidefinite matrices.

For fixed task indices $i,j$ and inputs $x,x'$, define
\[
    v_i
    =
    K\bigl((i,x),(i,x)\bigr),
    \quad
    v_j
    =
    K\bigl((j,x'),(j,x')\bigr), \quad
    c_{ij}
    =
    K\bigl((i,x),(j,x')\bigr).
\]
The finite-dimensional marginal distribution induced by the GP prior
is then
\begin{equation}\nonumber
    \begin{bmatrix}
        Q^{(i)}(x)\\
        Q^{(j)}(x')
    \end{bmatrix}
    \sim
    \mathcal N\!\left(
        \begin{bmatrix}
            m^{(i)}(x)\\
            m^{(j)}(x')
        \end{bmatrix},
        \begin{bmatrix}
            v_i & c_{ij}\\
            c_{ij} & v_j
        \end{bmatrix}
    \right).
    \label{G-2}
\end{equation}
If $v_i>0$, conditioning on an exact latent evaluation
$Q^{(i)}(x)=q$ gives
\begin{equation}\nonumber
    Q^{(j)}(x')
    \mid
    Q^{(i)}(x)=q
    \sim
    \mathcal N\!\left(
        m^{(j)}(x')
        +
        \frac{c_{ij}}{v_i}
        \bigl(q-m^{(i)}(x)\bigr),
        \;
        v_j-\frac{c_{ij}^2}{v_i}
    \right).
    \label{G-3}
\end{equation}
Thus, information from task $i$ affects predictions for task $j$
through the cross-task covariance $c_{ij}$.
More generally, suppose that task $i$ is observed with Gaussian noise,
\[
    y
    =
    Q^{(i)}(x)+\epsilon,
    \qquad
    \epsilon\sim\mathcal N(0,\sigma_i^2),
\]
where $\epsilon$ is independent of the latent GP. Then
\[
    Q^{(j)}(x')\mid y
    \sim
    \mathcal N\!\left(
        m^{(j)}(x')
        +
        \frac{c_{ij}}{v_i+\sigma_i^2}
        \bigl(y-m^{(i)}(x)\bigr),
        \;
        v_j-
        \frac{c_{ij}^2}{v_i+\sigma_i^2}
    \right).
\]
This expression makes explicit that cross-task information sharing
depends jointly on the cross-covariance $c_{ij}$, the marginal
uncertainty $v_i$, and the observation noise $\sigma_i^2$.
A large task covariance alone therefore does not imply beneficial
transfer; whether source information improves target prediction also
depends on the observation model and the compatibility of the
corresponding tasks.

Throughout this section, GP conditioning is understood conditional on
the kernel and noise hyperparameters, whose dependence is suppressed
from the notation. If these hyperparameters are assigned priors and
integrated out, the resulting predictive distribution is generally a
mixture of Gaussian distributions rather than a single Gaussian.
We use $m^{(k)}$ to denote prior mean functions and $\mu^{(k)}$ to
denote posterior mean functions.
Finally, the GP prior specifies statistical regularity and dependence
among the task-specific Q-functions, but does not by itself impose the
Bellman fixed-point equations. The reinforcement-learning structure
enters through the Bellman-generated responses and the iterative
Q-learning updates developed in the subsequent section.

\subsection{Multi-task transfer Q-learning}
\label{subsec:multitask-transfer-q-learning}

The target task is observed through an offline transition dataset
\[
    \mathcal{D}^{(0)}_{\mathrm{off}}
    =\bigl\{(s_t^{(0)},a_t^{(0)},r_t^{(0)},s_t^{\prime(0)})
      \bigr\}_{t=1}^{n_0}.
\]
We first describe transfer from labeled source data, for which Gaussian
conditioning is exact under the stated observation model, and then
extend the construction to Bellman-generated responses.
For each $k\in[J]$, let
$\mathcal{D}^{(k)}=\{(x_t^{(k)},y_t^{(k)})\}_{t=1}^{n_k}$.
The labeled-source model is
\begin{equation}
    y_t^{(k)}=Q^{(k)}(x_t^{(k)})+\epsilon_t^{(k)},
    \qquad
    \epsilon_t^{(k)}\sim\mathcal{N}(0,\sigma_k^2),
    \label{T-2}
\end{equation}
where $\sigma_k^2>0$. Conditional on the inputs, the noises are
independent across observations and tasks and independent of the
latent functions. Define
\(
    \mathbf{X}^{(k)}
    =[x_1^{(k)},\ldots,x_{n_k}^{(k)}]^{\top}\) and \(
    \mathbf{y}^{(k)}
    =(y_1^{(k)},\ldots,y_{n_k}^{(k)})^{\top},
\)
and let $\mathbf{Q}^{(k)}=Q^{(k)}(\mathbf{X}^{(k)})$ denote the
vector of latent evaluations. Then
\begin{equation}
    \mathbf{y}^{(k)}=\mathbf{Q}^{(k)}+\boldsymbol{\epsilon}^{(k)},
    \qquad
    \boldsymbol{\epsilon}^{(k)}
    \sim\mathcal{N}(\mathbf{0},\sigma_k^2\mathbf{I}_{n_k}).
    \label{T-3}
\end{equation}
Let $\mathbf{X}^{(0)}$ denote the target inputs and set
$\mathbf{Q}^{(0)}=Q^{(0)}(\mathbf{X}^{(0)})$.
For $i,j\in\{0,\ldots,J\}$, define
\[
    \mathbf{K}^{(i,j)}
    =K_{\mathrm{task}}(i,j)
      \mathbf{K}_{\mathrm{in}}
      (\mathbf{X}^{(i)},\mathbf{X}^{(j)}),\qquad
    [\mathbf{K}_{\mathrm{in}}(\mathbf{X},\mathbf{X}')]_{ab}
    =k_{\mathrm{in}}(x_a,x_b').
\]
To condition on all sources simultaneously, stack their responses
and prior means in a fixed task order:
\[
    \mathbf{y}_s
    =\bigl((\mathbf{y}^{(1)})^{\top},\ldots,
             (\mathbf{y}^{(J)})^{\top}\bigr)^{\top},\quad
    \mathbf{m}_s
    =\bigl((\mathbf{m}^{(1)})^{\top},\ldots,
             (\mathbf{m}^{(J)})^{\top}\bigr)^{\top},\quad
    \boldsymbol{\Sigma}_s
    =\operatorname{diag}
      (\sigma_1^2\mathbf{I}_{n_1},\ldots,
       \sigma_J^2\mathbf{I}_{n_J}),
\]
where $\mathbf{m}^{(k)}=m^{(k)}(\mathbf{X}^{(k)})$. Set
\[
    \mathbf{K}^{(s,s)}=[\mathbf{K}^{(i,j)}]_{i,j=1}^{J},\quad
    \mathbf{K}^{(0,s)}
    =[\mathbf{K}^{(0,1)}, \ldots, \mathbf{K}^{(0,J)}],\quad
    \mathbf{K}^{(s,0)}=(\mathbf{K}^{(0,s)})^{\top}.
\]
Writing $X$ for the collection of all input locations, we have
\begin{equation}
    \begin{bmatrix}
        \mathbf{y}_s\\ \mathbf{Q}^{(0)}
    \end{bmatrix}
    \Bigm|X
    \sim\mathcal{N}\!\left(
        \begin{bmatrix}
            \mathbf{m}_s\\ \mathbf{m}^{(0)}
        \end{bmatrix},
        \begin{bmatrix}
            \mathbf{K}^{(s,s)}+\boldsymbol{\Sigma}_s
                &\mathbf{K}^{(s,0)}\\
            \mathbf{K}^{(0,s)}&\mathbf{K}^{(0,0)}
        \end{bmatrix}
    \right).
    \label{T-4}
\end{equation}
Consequently,
$\mathbf{Q}^{(0)}\mid X,\mathbf{y}_s
\sim\mathcal{N}(\boldsymbol{\mu}_{0\mid s},
\boldsymbol{\Sigma}_{0\mid s})$, where
\begin{equation}\nonumber
    \boldsymbol{\mu}_{0\mid s}
    =\mathbf{m}^{(0)}
     +\mathbf{K}^{(0,s)}
      \bigl(\mathbf{K}^{(s,s)}+\boldsymbol{\Sigma}_s\bigr)^{-1}
      (\mathbf{y}_s-\mathbf{m}_s),
    \label{T-5}
\end{equation}
and
\begin{equation}\nonumber
    \boldsymbol{\Sigma}_{0\mid s}
    =\mathbf{K}^{(0,0)}
     -\mathbf{K}^{(0,s)}
      \bigl(\mathbf{K}^{(s,s)}+\boldsymbol{\Sigma}_s\bigr)^{-1}
      \mathbf{K}^{(s,0)}.
    \label{T-6}
\end{equation}
Positive observation-noise variances ensure that the inverted matrix
is positive definite. These expressions retain the cross-source
covariance blocks; conditioning separately on each source does not,
in general, yield the joint posterior. For a selected source subset
$S\subseteq[J]$, the same construction uses only the corresponding
source blocks, while target observations are retained in the
subsequent update. The choice $S=\varnothing$ gives target-only
learning. Source selection is developed in Section~3.

When direct source Q-labels are unavailable, the source information is
provided through transition data. At iteration $e$, we construct the
Bellman pseudo-response for task $k$ as
\begin{equation}\nonumber
    y_{t,e}^{(k)}
    =
    r_t^{(k)}
    +
    \gamma
    \max_{a'\in\mathcal A^{(k)}}
    \widehat Q_{e-1}^{(k)}
    \bigl(s_t^{\prime(k)},a'\bigr),
    \label{T-bellman-target}
\end{equation}
and use the same construction for the target task before introducing
the conservative modification described below. This corresponds to a
fitted Q-iteration step in which the Bellman targets are regressed
through the multi-task GP model \citep{munos2008finite}.
To characterize the statistical error introduced by the Bellman
update, decompose the pseudo-response as
\begin{equation}\nonumber
    y_{t,e}^{(k)}
    =
    Q_*^{(k)}\bigl(x_t^{(k)}\bigr)
    +
    \epsilon_t^{(k)}
    +
    b_{t,e}^{(k)},
    \label{eq:bellman-decomposition}
\end{equation}
where
\begin{align}\nonumber
    \epsilon_t^{(k)}
    &=
    r_t^{(k)}
    +
    \gamma
    \max_{a'\in\mathcal A^{(k)}}
    Q_*^{(k)}\bigl(s_t^{\prime(k)},a'\bigr)
    -
    Q_*^{(k)}\bigl(x_t^{(k)}\bigr),
   % \label{eq:bellman-innovation}
    \\ \nonumber
    b_{t,e}^{(k)}
    &=
    \gamma
    \left[
    \max_{a'\in\mathcal A^{(k)}}
    \widehat Q_{e-1}^{(k)}
    \bigl(s_t^{\prime(k)},a'\bigr)
    -
    \max_{a'\in\mathcal A^{(k)}}
    Q_*^{(k)}
    \bigl(s_t^{\prime(k)},a'\bigr)
    \right].
    \label{eq:bootstrap-error}
\end{align}
The term $\epsilon_t^{(k)}$ represents the Bellman innovation at the
optimal Q-function, whereas $b_{t,e}^{(k)}$ captures the additional
error arising from replacing $Q_*^{(k)}$ by the previous fitted
iterate $\widehat Q_{e-1}^{(k)}$.
\\
By the Bellman optimality equation,
\(
    \mathbb E\!\left[
        \epsilon_t^{(k)}
        \,\middle|\,
        S_t^{(k)},A_t^{(k)}
    \right]
    =0.
\)
Moreover, using the elementary inequality
\[
    \left|
    \max_a f(a)-\max_a g(a)
    \right|
    \le
    \max_a |f(a)-g(a)|,
\]
the bootstrap error satisfies
\begin{align}
    |b_{t,e}^{(k)}|
    \le
    \gamma
    \max_{a'\in\mathcal A^{(k)}}
    \left|
    \widehat Q_{e-1}^{(k)}
    \bigl(s_t^{\prime(k)},a'\bigr)
    -
    Q_*^{(k)}
    \bigl(s_t^{\prime(k)},a'\bigr)
    \right|
    \nonumber
    \le
    \gamma
    \left\|
    \widehat Q_{e-1}^{(k)}
    -
    Q_*^{(k)}
    \right\|_\infty.
    \label{eq:bootstrap-error-bound}
\end{align}
Hence, the Bellman decomposition separates a conditionally centered
innovation term from a bootstrap error whose magnitude is controlled
by the estimation error of the preceding Q-function iterate.
This distinction is important for the GP update. Unlike the
labeled-source model, the Bellman-generated responses are generally
neither conditionally Gaussian nor independent across observations,
and the bootstrap term may depend on the same data used to construct
$\widehat Q_{e-1}^{(k)}$. Accordingly, the Gaussian regression model
applied to these pseudo-responses is viewed as a working
posterior, rather than as the exact Bayesian posterior under the
trajectory-generating mechanism. The dependence structure and the
stability of the bootstrap error are analyzed in
Section~\ref{sec:selection-consistency}.

To avoid relying on actions with insufficient support in the target
offline data, define the behavior-supported action set
\[
    \mathcal C_{\beta}(s)
    =
    \left\{
        a\in\mathcal A:
        \widehat\beta(a\mid s)\ge \tau_{\beta}
    \right\},
\]
where $\widehat\beta(a\mid s)$ is the estimated target behavior policy
and $\tau_{\beta}>0$ is a support threshold. We assume that
$\mathcal C_{\beta}(s)$ is nonempty for every state encountered by the
learning procedure. Restricting Bellman maximization to
$\mathcal C_{\beta}(s)$ limits the use of poorly supported actions,
while posterior uncertainty is incorporated through a pessimistic
value adjustment, following the general principle of conservative
offline reinforcement learning
\citep{fujimoto2019off,jin2021pessimism}.
At iteration $e-1$, define the pessimistic target score
\begin{equation}
    Q_{e-1}^{-}(s,a)
    =
    \mu_{e-1}^{(0)}(s,a)
    -
    \alpha_{\mathrm{pess}}
    \sigma_{e-1}^{(0)}(s,a),
    \label{T-pessimistic-score}
\end{equation}
where $\alpha_{\mathrm{pess}}\ge0$ controls the strength of the
uncertainty penalty. Here
$\sigma_{e-1}^{(0)}(s,a)$ denotes the posterior standard deviation of
the latent target Q-function, rather than the predictive standard
deviation of a noisy observation.
The corresponding conservative Bellman pseudo-response is
\begin{equation}
    y_{t,e}^{(0),\mathrm{cons}}
    =
    r_t^{(0)}
    +
    \gamma
    \max_{a'\in
    \mathcal C_{\beta}(s_t^{\prime(0)})}
    Q_{e-1}^{-}
    \bigl(s_t^{\prime(0)},a'\bigr).
    \label{T-conservative-target}
\end{equation}
Thus, the target update combines two forms of regularization:
action restriction controls extrapolation outside the empirical
support of the target behavior policy, whereas the pessimistic term
penalizes state--action pairs with large posterior uncertainty.

At iteration $e$, let $\widetilde{\mathbf y}_e$ denote the vector
obtained by stacking the source responses and the current target
pseudo-responses. Each observation is paired with its task index, so
we write the augmented input as
\(
    z_{e,u}=(k_u,x_u),~
    u=1,\ldots,N_e,
\)
where $k_u\in\{0,\ldots,J\}$ denotes the task associated with the
$u$th observation.
Define the corresponding prior-mean vector and joint Gram matrix by
\[
    \widetilde{\mathbf m}_e
    =
    \bigl(m^{(k_u)}(x_u)\bigr)_{u=1}^{N_e}\quad
\text{and} \quad
    \mathbf K_e
    =
    \bigl[
        K(z_{e,u},z_{e,v})
    \bigr]_{u,v=1}^{N_e}.
\]
Let
\(
    \boldsymbol\Lambda_e
    =
    \operatorname{diag}
    \bigl(
        \nu_{e,1}^2,\ldots,\nu_{e,N_e}^2
    \bigr)
    \succ0
\)
denote the diagonal working-noise covariance matrix.
For directly observed source labels, the corresponding entries are
given by their observation-noise variances. For a target
pseudo-response constructed at input $x$, we set
\(
    \nu_e^2(x)
    =
    \lambda+
    \bigl[\sigma_{e-1}^{(0)}(x)\bigr]^2,~
    \lambda>0.
\)
Source pseudo-responses in the transition-only setting are assigned
analogous positive working variances. These quantities are used for
regularization in the GP update and are not identified with the true
conditional variances of the Bellman innovations.
For a target query $x$, define the covariance vector
\(
    \mathbf k_{0,e}(x)
    =
    \bigl(
        K(z_{e,u},(0,x))
    \bigr)_{u=1}^{N_e}.
\)
Conditioning the multi-task GP on the stacked observations gives
\begin{equation}
\begin{aligned}
    \mu_e^{(0)}(x)
    &=
    m^{(0)}(x)
    +
    \mathbf k_{0,e}(x)^\top
    (\mathbf K_e+\boldsymbol\Lambda_e)^{-1}
    \bigl(
        \widetilde{\mathbf y}_e
        -
        \widetilde{\mathbf m}_e
    \bigr),
    \\
    \bigl[\sigma_e^{(0)}(x)\bigr]^2
    &=
    K\bigl((0,x),(0,x)\bigr)
    -
    \mathbf k_{0,e}(x)^\top
    (\mathbf K_e+\boldsymbol\Lambda_e)^{-1}
    \mathbf k_{0,e}(x).
\end{aligned}
\label{T-joint-update}
\end{equation}
The same conditioning formulas apply to a source task $k$ after
replacing the query index $0$ by $k$, and we set
\(
    \widehat Q_e^{(k)}
    =
    \mu_e^{(k)}.
\)
In the transition-only setting, source pseudo-responses are updated
at each Bellman iteration. For a given transition, the newly computed
pseudo-response replaces its previous value rather than being treated
as an additional independent observation.

After $E$ iterations, the target policy is defined by
\begin{equation}
    \pi_{\mathrm{off}}(s)
    =
    \arg\max_{a\in\mathcal C_\beta(s)}
    \left\{
        \mu_E^{(0)}(s,a)
        -
        \alpha_{\mathrm{pess}}
        \sigma_E^{(0)}(s,a)
    \right\}.
    \label{T-offline-policy}
\end{equation}
The quantity
$\mu_e^{(0)}-\alpha_{\mathrm{pess}}\sigma_e^{(0)}$
serves as a pessimistic value score. Its interpretation as a formal
lower confidence bound requires additional calibration conditions.
Action restriction, posterior pessimism, and source selection play
distinct roles: they control target-data support, predictive
uncertainty, and potentially harmful cross-task transfer,
respectively. The theoretical analysis therefore treats coverage,
regression accuracy, bootstrap stability, and transfer selection as
separate components.

\begin{algorithm}[H]
\caption{Bayesian Transfer GP Offline Q-Learning with Labeled Sources} 
\KwIn{Target offline dataset $\mathcal D^{(0)}_{\mathrm{off}}=\{(s_i,a_i,r_i,s_i')\}_{i=1}^{n_0}$; target task $\mathcal{M}^{(0)}=(\mathcal{S},\mathcal{A},P,R,\gamma)$; source datasets $\{\mathcal{D}^{(k)}\}_{k=1}^J$; 
input kernel matrix $\textbf k_{\text{in}}$; 
priors for  task similarity kernel matrix $\textbf K_{\text{task}}$ with spike-and-slab indicators $\eta_{ij}$,\, $i,\,j=0,\ldots, J$; 
noise variances $\{\sigma_k^2\}_{k=0}^J$; discount factor $\gamma$; epochs $E$;
batch size $B$;
pessimism coefficient $\alpha \ge 0$;
behavior threshold $\tau_\beta$;
regularization $\lambda>0$.
}

\textbf{Output:} Posterior $(\bm\mu_{\mathrm{post}}, \bm\Sigma_{\mathrm{post}})$;
kernel matrix $\mathbf K_{\mathrm{task}}$;
policy $\pi_{\mathrm{off}}$.

\vspace{2mm}
\textbf{Initialization:}\\
Set prior means $\textbf m^{(k)}$ for all tasks, $k=0,\dots,J$;\\
Sample $\eta_{ij} \sim \text{Bernoulli}(p)$ and $K_{\mathrm{task}}(i,j)|\eta_{ij}\sim (1-\eta_{ij})z_0 + \eta_{ij}\mathcal{N}(0,\tau^2)$;\\
Learn behavior policy $\hat\beta(a|s)$ from $D^{(0)}_{\mathrm{off}}$;\\
Define $C_\beta(s)=\{a:\hat\beta(a|s)\ge\tau_\beta\}$.\\[2mm]

\textbf{Build multi-task GP:}\\
Construct kernel using 
$K\big((i,x),(j,x')\big) = K_{\mathrm{task}}(i,j)\, k_{\mathrm{in}}(x,x')$ with $x=(s,a)$;\\
Form stacked dataset $\textbf X_s
=[\textbf X^{(1)},\dots,\textbf X^{(J)}]$ and
$\textbf y_s=\begin{bmatrix}(\textbf y^{(1)})^{\top},\dots,(\textbf y^{(J)})^{\top}\end{bmatrix}^\top$ from all sources;% with weights $w_k$\;
\\
Compute prior mean vectors:
$
\mathbf{m}^{(0)} = \mathbf m^{(0)}(\mathbf{X}^{(0)}), \quad
\mathbf{m}_s =
\begin{bmatrix}
(\mathbf{m}^{(1)})^\top, \cdots, (\mathbf{m}^{(J)})^\top
\end{bmatrix}^\top$;
\\
Construct noise covariance matrix:
$
\mathbf{\Sigma}_s = \mathrm{diag}(\sigma_1^2 \mathbf{I}_{n_1}, \dots, \sigma_J^2 \mathbf{I}_{n_J}).
$\\
\vspace{2mm}
\textbf{Posterior inference:}\\
The conditional posterior for the target task:\\
$
\bm{\mu}_{\text{post}} = \textbf m^{(0)} + \mathbf{K}^{(0,s)}
(\mathbf{K}^{(s,s)} + \mathbf{\Sigma}_s)^{-1}(\textbf y_s - \textbf m_s)
$
and
$
\bm\Sigma_{\mathrm{post}} = \mathbf{K}^{(0,0)} 
- \mathbf{K}^{(0,s)}(\mathbf{K}^{(s,s)} + \mathbf{\Sigma}_s)^{-1}\mathbf{K}^{(s,0)};
$
\\
Compute the
posterior mean
$
\mu_{\mathrm{post}}(s,a)
$
and posterior standard deviation
$
\sigma_{\mathrm{post}}(s,a)$, for any $(s,a)$.
\vspace{2mm}

\textbf{Offline RL loop:}\\
\textbf{for} $e=1$ to $E$ \textbf{do}\\
\hspace*{1em} Sample minibatch $\mathcal B \subset \mathcal D^{(0)}_{\mathrm{off}}$;\\
\hspace*{1em} Initialize $\mathbf X^{(0)}_{\mathrm{pseudo}}, \mathbf y^{(0)}_{\mathrm{pseudo}}, \bm\Sigma^{(0)}_{\mathrm{pseudo}} \leftarrow \varnothing$;\\

\hspace*{1em} \textbf{for each} $(s,a,r,s') \in \mathcal B$ \textbf{do}\\
\hspace*{2em} Compute pessimistic value:
\[
Q^-(s',a') = \mu_{\mathrm{post}}(s',a') - \alpha \sigma_{\mathrm{post}}(s',a');
\]
\hspace*{2em} Compute target:
\[
y = r + \gamma \max_{a'\in C_\beta(s')} Q^-(s',a');
\]
\hspace*{2em} Set weight:
\[
\nu^2(s,a) = \lambda + \sigma_{\mathrm{post}}^2(s,a);
\]
\hspace*{2em} Append $((s,a),y,\nu^2(s,a))$ to pseudo set;\\
\hspace*{1em} \textbf{end for}\\

\hspace*{1em} Merge:
\[
\tilde{\mathbf X} \leftarrow \mathbf X_s \cup \mathbf X^{(0)}_{\mathrm{pseudo}},\quad
\tilde{\mathbf y} \leftarrow \mathbf y_s \cup \mathbf y^{(0)}_{\mathrm{pseudo}};
\]

\hspace*{1em} Update GP posterior using $(\tilde{\mathbf X},\tilde{\mathbf y})$;\\

\textbf{end for}

\textbf{Policy extraction:}\\
\[
\pi_{\mathrm{off}}(s) = \arg\max_{a\in C_\beta(s)}
\big[\mu_{\mathrm{post}}(s,a) - \alpha \sigma_{\mathrm{post}}(s,a)\big].
\]

\textbf{return} $\bm\mu_{\mathrm{post}}, \bm\Sigma_{\mathrm{post}}, \mathbf K_{\mathrm{task}}, \pi_{\mathrm{off}}$.
\end{algorithm}

\begin{algorithm}[H]
\caption{Bayesian Transfer GP Offline Q-Learning with Unlabeled Sources}
\label{alg:btgpq_unlabeled_joint}
\KwIn{ Target dataset $\mathcal{D}^{(0)}_{\mathrm{off}}=\{(s_i,a_i,r_i,s'_i)\}_{i=1}^{n_0}$;
source datasets $\{\mathcal{D}^{(k)}\}_{k=1}^J$, 
$\mathcal{D}^{(k)}=\{(s_i^{(k)},a_i^{(k)},r_i^{(k)},s_{i}^{\prime (k)})\}_{i=1}^{n_k}$; 
input kernel $\mathbf k_{\mathrm{in}}$; task kernel prior $\mathbf K_{\mathrm{task}}$; 
discount factor $\gamma$; epochs $E$; batch size $B$; 
pessimism coefficient $\alpha \ge 0$; behavior threshold $\tau_\beta$; regularization $\lambda > 0$.
}

\textbf{Output:} 
Posterior $(\bm\mu_{\mathrm{post}}, \bm\Sigma_{\mathrm{post}})$;
kernel matrix $\mathbf K_{\mathrm{task}}$;
policy $\pi_{\mathrm{off}}$.

\textbf{Initialization:} \\
Initialize $Q_0^{(k)}(s,a)$ for all tasks $k=0,\dots,J$; 
estimate behavior policy $\hat\beta(a|s)$ from $\mathcal{D}^{(0)}_{\mathrm{off}}$; 
define $\mathcal{C}_\beta(s)=\{a:\hat\beta(a|s)\ge \tau_\beta\}$.

\textbf{for} $e=1$ \textbf{to} $E$ \textbf{do}

\quad \textbf{Construct pseudo-labels for source tasks:}

\quad \textbf{for} $k=1$ \textbf{to} $J$ \textbf{do}

\qquad \textbf{for each} $(s_i^{(k)},a_i^{(k)},r_i^{(k)},s_{i}^{\prime(k)}) \in \mathcal{D}^{(k)}$ \textbf{do}
\[
y_i^{(k)} \leftarrow r_i^{(k)} + \gamma \max_{a'} Q_{e-1}^{(k)}(s_{i}^{\prime(k)},a')
\]
\qquad \textbf{end for}

\quad \textbf{end for}

\quad Construct source pseudo-dataset
\[
\mathbf X_s = \bigcup_{k=1}^J \mathbf X^{(k)}, 
\qquad
\mathbf y_s = \bigcup_{k=1}^J \mathbf y^{(k)}
\]

\quad \textbf{Construct pseudo-labels for target task:}

\quad Sample minibatch $\mathcal{B} \subset \mathcal{D}^{(0)}_{\mathrm{off}}$; initialize target pseudo-set $\mathbf X_t \leftarrow \emptyset$, $\mathbf y_t \leftarrow \emptyset$

\quad \textbf{for each} $(s,a,r,s') \in \mathcal{B}$ \textbf{do}
\[
Q^{-}_{e-1}(s',a') = Q^{(0)}_{e-1}(s',a') - \alpha\,\widetilde{\sigma}_{e-1}(s',a')
\]
\[
y \leftarrow r + \gamma \max_{a' \in \mathcal{C}_\beta(s')} Q^{-}_{e-1}(s',a')
\]
\[
\nu^2(s,a) \leftarrow \lambda + \widetilde{\sigma}_{e-1}^2(s,a)
\]
\quad Append $(s,a)$ to $\mathbf X_t$ and append $y$ to $\mathbf y_t$.\\
\quad \textbf{end for}

\quad \textbf{Joint posterior update using both source and target pseudo-data:}

\quad Construct joint dataset
\[
\mathbf{\widetilde{X}} \leftarrow \mathbf X_s \cup \mathbf X_t,
\qquad
\mathbf{\widetilde{y}} \leftarrow \mathbf y_s \cup \mathbf y_t
\]

\quad Build kernel
\[
K\bigl((k,x),(l,x')\bigr)=K_{\mathrm{task}}(k,l)\,k_{\mathrm{in}}(x,x')
\]

\quad Compute posterior $(\bm\mu_e,\bm\Sigma_e)$ using $(\mathbf{\widetilde{X}},\mathbf{\widetilde{y}})$.

\quad \textbf{Update Q-functions:}

\quad $Q_e^{(0)}(s,a) \leftarrow \mu_e^{(0)}(s,a)$

\quad \textbf{for} $k=1$ \textbf{to} $J$ \textbf{do}
\[
Q_e^{(k)}(s,a) \leftarrow \mu_e^{(k)}(s,a)
\]
\quad \textbf{end for}

\quad Set $\widetilde{\sigma}_e(s,a) \leftarrow \sqrt{\Sigma_e((0,(s,a)),(0,(s,a)))}$

\textbf{end for}

\textbf{Policy extraction:}

\[
\pi_{\mathrm{off}}(s) = \arg\max_{a \in \mathcal{C}_\beta(s)}
\left[\mu_E^{(0)}(s,a)-\alpha\,\widetilde{\sigma}_E(s,a)\right]
\]

\textbf{return} $\bm\mu_{\mathrm{post}}, \bm\Sigma_{\mathrm{post}}, \mathbf K_{\mathrm{task}}, \pi_{\mathrm{off}}$.
\end{algorithm}

\section{Selection consistency}
\label{sec:selection-consistency}

\subsection{Target-aware transfer selection consistency}
\label{subsec:target-aware-selection}

In this section, we study the consistency of target-aware source selection under
Bellman bootstrapping. Candidate source subsets are evaluated according
to their predictive contribution to the target task, and the primary
question is whether the resulting selection rule asymptotically
eliminates target-risk-suboptimal subsets. Accordingly, the theoretical
objective is concentration on a target-risk oracle class rather than,
in general, recovery of a predetermined source subset. The analysis
requires both control of Bellman-bootstrap perturbations and sufficient
predictive-score separation between oracle and suboptimal subsets.
Exact recovery of a unique source subset additionally requires an
identifying target-risk gap.
For heterogeneous tasks, we assume
that the shared representations preserve the optimal values,
\[
    Q_*^{(k)}(s,a)
    =q_*^{(k)}\bigl(\phi^{(k)}(s,a)\bigr),
    \qquad k=0,\ldots,J,
\]
and retain the convention $Q_*^{(k)}(x)=q_*^{(k)}(x)$ on the represented
domain. Bellman maximization remains defined on the native action
spaces, and we write
$V_*^{(k)}(s)=\max_{a\in\mathcal{A}^{(k)}}Q_*^{(k)}(s,a)$.
All sampling statements below are with respect to the true joint
trajectory law $\mathbb{P}_\star$.

At iteration $e$, every candidate $S\in\mathcal{S}_J$ is evaluated
against the same target response vector $\mathbf{Y}_e^{(0)}$ from
Section~2.4. The response-construction rule and the target working-noise
matrix $\boldsymbol{\Lambda}_{0,e}\succ0$ are held common across
candidates. Let $\mathscr{C}_e$ contain all input locations, source
pseudo-responses, and fitting information treated as fixed at this
selection step.
For fixed GP hyperparameters, let $\boldsymbol{\mu}_{0,S,e}$ and
$\boldsymbol{\Sigma}_{0,S,e}$ be the source-conditioned moments of the
target latent evaluations. These moments are obtained by jointly
conditioning on the source blocks indexed by $S$, retaining the
cross-source covariances \citep{bonilla2007multitask}. The corresponding
target predictive density is
\[
    p_{S,e}(\,\cdot\mid\mathscr{C}_e)
    =\mathcal{N}_{n_0}\!\left(
        \boldsymbol{\mu}_{0,S,e},
        \boldsymbol{\Sigma}_{0,S,e}
        +\boldsymbol{\Lambda}_{0,e}\right).
\]
Given a subset prior $\Pi(S)>0$, define
\begin{equation}
    \Pi_e(S\mid Y_e,X)
    =\frac{\Pi(S)\,
        p_{S,e}(\mathbf{Y}_e^{(0)}\mid\mathscr{C}_e)}
        {\displaystyle\sum_{U\in\mathcal{S}_J}
        \Pi(U)\,p_{U,e}(\mathbf{Y}_e^{(0)}\mid\mathscr{C}_e)}.
    \label{sel:subset-posterior}
\end{equation}
This predictive-score update defines a working subset posterior; it
need not coincide with the posterior under the full trajectory
likelihood.
Let $\widehat{Q}_{S,e}^{\mathrm{MT}}$ denote the target posterior mean
after additionally conditioning on $\mathbf{Y}_e^{(0)}$. Selection is
assessed through the target risk $R_0$ defined in Section~2.1, applied
to this fitted estimator rather than to the source-conditioned
predictive mean. The comparator $\widehat{Q}_e^{\mathrm{ST}}$ is a
separately specified target-only fit with source-free response
construction and tuning. It coincides with the empty-subset fit only
when these components also coincide. The oracle class introduced below
is defined by target risk, not by task-covariance magnitude or marginal
input similarity.

For an oracle reference $S^\star$ and a candidate $S$, define
\begin{equation}
    \mathrm{KL}_{0,e}(S^\star,S\mid\mathscr{C}_e)
    =
    \int_{\mathbb{R}^{n_0}}
    p_{S^\star,e}(\mathbf{y}\mid\mathscr{C}_e)
    \log
    \frac{
        p_{S^\star,e}(\mathbf{y}\mid\mathscr{C}_e)
    }{
        p_{S,e}(\mathbf{y}\mid\mathscr{C}_e)
    }
    \,\mathrm{d}\mathbf{y}.
    \label{KL-1}
\end{equation}
This quantity measures the separation between the two working
predictive distributions under the reference predictive model.
To characterize the corresponding predictive-score separation under
the actual trajectory law $\mathbb{P}_\star$, define, whenever the
conditional expectation exists,
\begin{equation}
    \Delta_{0,e}(S^\star,S\mid\mathscr{C}_e)
    =
    \mathbb{E}_\star\!\left[
        \log
        \frac{
            p_{S^\star,e}(\mathbf{Y}_e^{(0)}\mid\mathscr{C}_e)
        }{
            p_{S,e}(\mathbf{Y}_e^{(0)}\mid\mathscr{C}_e)
        }
        \,\middle|\,\mathscr{C}_e
    \right].
    \label{sel:expected-score-gap}
\end{equation}
By construction, $-\Delta_{0,e}$ is the conditional expected log
predictive odds of $S$ relative to $S^\star$. If
$p_{S^\star,e}(\,\cdot\mid\mathscr{C}_e)$ coincides with the true
conditional law of $\mathbf{Y}_e^{(0)}$, then
$\Delta_{0,e}=\mathrm{KL}_{0,e}$ and is therefore nonnegative.
Under model misspecification, this identity no longer holds, and
$\Delta_{0,e}$ need not be nonnegative. Establishing selection
consistency therefore requires separation in expected predictive score
under $\mathbb{P}_\star$, together with uniform control of stochastic
score fluctuations on the same scale. Predictive KL separation alone
is insufficient unless a valid link to the corresponding expected-score
separation under the true data-generating law is established. Related
likelihood-ratio theory for dependent and misspecified models provides
a useful framework \citep{shalizi2009dynamics}, but the required
conditions and concentration bounds must be verified for the
predictive update in \eqref{sel:subset-posterior}.

Write
$Z_t^{(k)}=(S_t^{(k)},A_t^{(k)},R_t^{(k)},S_{t+1}^{(k)})$,
and let $\epsilon_t^{(k)}$ and $b_{t,e}^{(k)}$ denote the Bellman
innovation and bootstrap remainder of Section~2.4, respectively.
We impose the following sampling conditions.

\begin{assumption}[Geometric absolute regularity]
\label{ass:transfer_mix}
For each $k\in\{0,\ldots,J\}$, the transition process
$\{Z_t^{(k)}\}_{t\geq1}$ is stationary. For sigma-fields
$\mathcal{U}$ and $\mathcal{V}$, define
\[
    \beta(\mathcal{U},\mathcal{V})
    =\frac12\sup_{\{A_i\},\{B_j\}}
        \sum_{i=1}^{I}\sum_{j=1}^{L}
        \left|\mathbb{P}_\star(A_i\cap B_j)
        -\mathbb{P}_\star(A_i)\mathbb{P}_\star(B_j)\right|,
\]
where the supremum is over finite measurable partitions of the
probability space with $A_i\in\mathcal{U}$ and $B_j\in\mathcal{V}$.
With $\mathcal{F}_{a}^{b,(k)}=\sigma(Z_t^{(k)}:a\leq t\leq b)$, set
\[
    \beta_k(\ell)
    =\sup_{t\geq1}
        \beta\!\left(\mathcal{F}_{1}^{t,(k)},
                     \mathcal{F}_{t+\ell}^{\infty,(k)}\right).
\]
There exist task-uniform constants $C_{\mathrm{mix}}<\infty$ and
$\rho\in(0,1)$ such that
$\beta_k(\ell)\leq C_{\mathrm{mix}}\rho^\ell$ for every $\ell\geq1$.
\end{assumption}

\begin{assumption}[Conditionally sub-Gaussian Bellman innovations]
\label{bellman-noise}
For each task $k$, let the pre-transition filtration be
\[
    \mathscr{F}_t^{(k)}
    =\sigma\!\left(
        S_1^{(k)},A_1^{(k)},
        (R_u^{(k)},S_{u+1}^{(k)},A_{u+1}^{(k)})_{1\leq u<t}
        \right).
\]
The innovation $\epsilon_t^{(k)}$ is
$\mathscr{F}_{t+1}^{(k)}$-measurable and, for a finite variance proxy
$\sigma_k^2>0$, satisfies
\[
\begin{aligned}
    \mathbb{E}_\star[\epsilon_t^{(k)}
        \mid\mathscr{F}_t^{(k)}]=0,\quad
    \mathbb{E}_\star\!\left[
        \exp(u\epsilon_t^{(k)})\mid\mathscr{F}_t^{(k)}\right]
        \leq\exp\!\left(\frac{u^2\sigma_k^2}{2}\right),
        \quad u\in\mathbb{R}.
\end{aligned}
\]
\end{assumption}
\noindent
Assumption~\ref{ass:transfer_mix} uses the absolute-regularity
coefficient \citep{bradley2005basic} for the full transition process,
including rewards and successor states. Neither assumption imposes
cross-task independence or state--action coverage. The proxy
$\sigma_k^2$ concerns the innovation at the Bellman fixed point and
is distinct from the GP working-noise variance. In particular,
Assumption~\ref{bellman-noise} does not extend the same centering or
sub-Gaussian condition to $\epsilon_t^{(k)}+b_{t,e}^{(k)}$.

For the conservative target response introduced in Section~2.4, write
\[
    Y_{t,e}^{(0),\mathrm{cons}}
    =
    Q_*^{(0)}(X_t^{(0)})
    +\epsilon_t^{(0)}
    +\widetilde{b}_{t,e}^{(0)},
\]
where $\widetilde{b}_{t,e}^{(0)}$ collects the discrepancy induced by
using the behavior-supported action set and the pessimistic value
estimate in place of the optimal continuation value.
Let $\mathcal{C}_{\beta}^{(0)}(s)$ denote the nonempty
behavior-supported action set for the target task, and define the
corresponding coverage gap by
\[
    c_{\beta}^{(0)}(s)
    =
    V_*^{(0)}(s)
    -
    \max_{a\in\mathcal{C}_{\beta}^{(0)}(s)}
    Q_*^{(0)}(s,a).
\]
Since
$\mathcal{C}_{\beta}^{(0)}(s)\subseteq\mathcal{A}^{(0)}$,
we have $c_{\beta}^{(0)}(s)\geq0$.
The conservative bootstrap remainder then satisfies the pathwise bound
\begin{equation}\nonumber
    |\widetilde{b}_{t,e}^{(0)}|
    \leq
    \gamma\Bigl\{
        \|\mu_{e-1}^{(0)}-Q_*^{(0)}\|_{\infty}
        +c_{\beta}^{(0)}(S_{t+1}^{(0)})
        +\alpha_{\mathrm{pess}}
        \sup_{a'\in
        \mathcal{C}_{\beta}^{(0)}(S_{t+1}^{(0)})}
        \sigma_{e-1}^{(0)}
        (S_{t+1}^{(0)},a')
    \Bigr\}.
\label{sel:conservative-remainder}
\end{equation}
This decomposition shows that the conservative update introduces two
additional sources of error beyond the fitted-value approximation:
loss due to restricted action coverage and the uncertainty penalty
induced by pessimism. Consequently, extending the bootstrap-stability
analysis from the unconstrained Bellman response to the conservative
response requires both terms to be controlled at the relevant
asymptotic scale.

Fix a task $k$. Let $w_t^{(k)}(s,a)$ denote state--action-specific
nonnegative weights satisfying
\[
    w_t^{(k)}(s,a)\geq 0,
    \qquad
    \sum_{t=1}^{n_k} w_t^{(k)}(s,a)=1.
\]
Given a fixed measurable function
$g_k:\mathcal{S}^{(k)}\to\mathbb{R}$ with
$\|g_k\|_{\infty}\leq B$, define the conditional transition estimator
\begin{equation}\nonumber
    \widehat{\mathbb{E}}_{n_k}^{(k)}
    [g_k(S')\mid s,a]
    =
    \sum_{t=1}^{n_k}
    w_t^{(k)}(s,a)\,
    g_k(S_{t+1}^{(k)}).
    \label{sel:conditional-expectation}
\end{equation}
In discrete state--action spaces, a natural choice is to use normalized
visit indicators associated with observations satisfying
$(S_t^{(k)},A_t^{(k)})=(s,a)$. In continuous settings, the weights may
instead be induced by a conditional regression or smoothing procedure,
whose approximation and coverage properties must be specified.
Conditioning on $(s,a)$ is essential, since an unweighted trajectory
average generally estimates a marginal successor-state expectation
rather than the transition expectation conditional on the current
state--action pair.
Using the same weighting scheme, define the empirical Bellman operator
by
\begin{equation}\nonumber
    (\widehat{\mathcal{T}}_{n_k}^{(k)}Q)(s,a)
    =
    \sum_{t=1}^{n_k}
    w_t^{(k)}(s,a)
    \left[
        R_t^{(k)}
        +\gamma
        \max_{a'\in\mathcal{A}^{(k)}}
        Q(S_{t+1}^{(k)},a')
    \right].
    \label{sel:empirical-bellman-operator}
\end{equation}
This operator is used to isolate the sampling component of a Bellman
update from the error introduced by the regression fit. It is therefore
an analytical device and need not coincide with the GP regression
estimator.
Using the population Bellman operator $\mathcal{T}^{(k)}$ defined in
Section~2.2, the one-step Bellman residual admits the decomposition
\begin{equation}
    \widehat{Q}_{e}^{(k)}
    -\mathcal{T}^{(k)}\widehat{Q}_{e-1}^{(k)}
    ={}
    \left(
        \widehat{Q}_{e}^{(k)}
        -\widehat{\mathcal{T}}_{n_k}^{(k)}
         \widehat{Q}_{e-1}^{(k)}
    \right)
    +
    \left(
        \widehat{\mathcal{T}}_{n_k}^{(k)}
        \widehat{Q}_{e-1}^{(k)}
        -\mathcal{T}^{(k)}
         \widehat{Q}_{e-1}^{(k)}
    \right).
\label{sel:bellman-residual-decomposition}
\end{equation}
The first term represents the regression and function-approximation
error of the fitted Bellman update, whereas the second measures the
conditional sampling error arising from estimation of the Bellman
expectation using the observed trajectory
\citep{munos2008finite}.
An additional difficulty arises when the same trajectory is reused
across Bellman iterations. In this setting, the continuation function
\(
    g_{k,e}(s')
    =
    \max_{a'\in\mathcal{A}^{(k)}}
    \widehat{Q}_{e-1}^{(k)}(s',a')
\)
depends on the data used to construct the empirical Bellman operator.
Consequently, a concentration result established only for a fixed
function $g_k$ cannot be applied directly to the fitted iterate.
Uniform concentration over a suitably controlled function class, a
valid sample-splitting construction, or another explicit treatment of
this dependence is therefore required. Any effective sample size
$n_{k,\mathrm{eff}}$ used below is understood to arise from the
particular concentration argument being applied and does not replace
the state--action coverage conditions required for conditional
estimation.

The following results establish control of the two error components in
\eqref{sel:bellman-residual-decomposition}. We first derive
concentration bounds for dependent Markov observations and then obtain
a uniform finite-rank GP regression bound that remains valid for
data-dependent Bellman iterates. Combining these results yields uniform
control of the one-step Bellman residual. This bound is subsequently
propagated through the Bellman contraction to establish bootstrap
stability. The finite-rank analysis requires additional
Markov-contraction and nondegenerate feature-design conditions beyond
geometric $\beta$-mixing and does not imply a dimension-free parametric
rate for a general GP kernel.

\begin{lemma}[Concentration for bounded Markov averages]
\label{lem:markov_bellman_concentration}
Let $U_1,\ldots,U_N$ be a possibly inhomogeneous Markov chain on a
standard Borel space, with an arbitrary initial distribution.
Let $P_i$ denote the transition kernel from $U_i$ to $U_{i+1}$, and
define
\[
    \theta_i
    =
    \sup_{u,v}
    \bigl\|P_i(\cdot\mid u)-P_i(\cdot\mid v)\bigr\|_{\mathrm{TV}},
    \qquad i=1,\ldots,N-1,
\]
where
$\|\mu-\nu\|_{\mathrm{TV}}:=\sup_A|\mu(A)-\nu(A)|$.
Set
\[
    c_{i,N}
    =
    1+\sum_{j=i+1}^{N}
    \prod_{\ell=i}^{j-1}\theta_\ell,
    \qquad
    M_N=\max_{1\leq i\leq N}c_{i,N},
\]
with $c_{N,N}=1$.
Let $h_1,\ldots,h_N$ be deterministic measurable functions satisfying
$\max_{1\leq i\leq N}\|h_i\|_\infty\leq B$ for some $B>0$, and define
\[
    F_N=\frac{1}{N}\sum_{i=1}^{N}h_i(U_i).
\]
Then, for every $u>0$,
\begin{equation}
\begin{aligned}
    \mathbb{P}\bigl(|F_N-\mathbb{E}F_N|\geq u\bigr)
    \leq
    2\exp\!\left\{
        -\frac{N^2u^2}
        {8B^2\sum_{i=1}^{N}c_{i,N}^2}
    \right\} 
    \leq
    2\exp\!\left\{
        -\frac{Nu^2}{8B^2M_N^2}
    \right\}.
\end{aligned}
\label{eq:boot-markov-concentration}
\end{equation}
If $\theta_i\leq\theta$ for all $i$, with $\theta\in[0,1)$, then
\[
    M_N\leq\frac{1}{1-\theta}.
\]
The concentration bound does not require stationarity.
When the chain is stationary and $h_i=h$ for every $i$, the centering
simplifies to
$\mathbb{E}F_N=\mathbb{E}[h(U_1)]$.
\end{lemma}

\noindent The proof is provided in Appendix~\ref{app:proof:lem:markov_bellman_concentration}.
The proof uses a direct Doob-martingale argument related to the
Markov concentration framework of
\citet{kontorovich2008concentration}.
To apply the lemma to task $k$, suppose that the shifted state--action
process
\(
    U_t=(S_{t+1}^{(k)},A_{t+1}^{(k)}),
    ~ t=1,\ldots,N,
\)
is Markov, and define its transition coefficients as in
Lemma~\ref{lem:markov_bellman_concentration}.
For a deterministic measurable function $g$ with
$\|g\|_\infty\leq B$, taking $h_t(s,a)=g(s)$ gives
\[
    F_N
    =
    \frac{1}{N}\sum_{t=1}^{N}g(S_{t+1}^{(k)}),
    \qquad
    \mathbb{E}F_N
    =
    \frac{1}{N}\sum_{t=1}^{N}
    \mathbb{E}[g(S_{t+1}^{(k)})].
\]
The underlying process is taken to be defined through time $N+1$,
although evaluating $F_N$ does not require observing
$A_{N+1}^{(k)}$.
Under stationarity, the centering reduces to the marginal expectation
$\mathbb{E}[g(S_1^{(k)})]$.
The lemma therefore controls a marginal sampling average rather than
the conditional transition estimator at a specified state--action
pair. The uniform bound on $M_N$ applies when the shifted transition
kernels satisfy the stated contraction condition, which is not
implied by geometric $\beta$-mixing alone.
Furthermore, the deterministic-function requirement excludes direct
substitution of a continuation function fitted to the same trajectory.
Lemma~\ref{lem:bellman_residual} addresses this dependence through a
uniform argument over the prescribed function classes.

Consider $J+1$ tasks, each providing $N$ observations. Let
$W_t^{(k)}=(S_t^{(k)},A_t^{(k)})$ denote the native state--action
process and let $X_t^{(k)}=\phi^{(k)}(W_t^{(k)})$ be its shared
representation. The Markov assumption is imposed on $W_t^{(k)}$,
not on the represented process $X_t^{(k)}$.
A fixed measurable feature map $\psi:\mathcal X\to\mathbb R^m$
defines the finite-rank input kernel and taskwise design matrices by
\[
\begin{aligned}
    k_{\mathrm{in}}(x,x')=\psi(x)^\top\psi(x'),\quad
    \Phi_k=[\psi(X_1^{(k)}),\ldots,\psi(X_N^{(k)})]^\top,
    \quad A_k=\Phi_k^\top\Phi_k.
\end{aligned}
\]
The feature map $\psi$ is distinct from the task-specific
representation maps $\phi^{(k)}$. Set
\[
    \Phi=\operatorname{diag}(\Phi_0,\ldots,\Phi_J),
    \quad
    A=\Phi^\top\Phi=\operatorname{diag}(A_0,\ldots,A_J),
\]
and stack the taskwise responses as
$Y=(Y_0^\top,\ldots,Y_J^\top)^\top$, where
$Y_k=(Y_1^{(k)},\ldots,Y_N^{(k)})^\top$.
For a block vector $v=(v_0^\top,\ldots,v_J^\top)^\top$ and a
conformably partitioned matrix $H=[H_{k\ell}]$, define
\[
    \|v\|_{2,\infty}=\max_k\|v_k\|_2,
    \quad
    \|H\|_{\mathrm b}=\max_k\sum_{\ell=0}^{J}
                              \|H_{k\ell}\|_{\mathrm{op}}.
\]
The matrix norm is submultiplicative and satisfies
$\|Hv\|_{2,\infty}\leq\|H\|_{\mathrm b}\|v\|_{2,\infty}$.
Under a zero-mean Gaussian coefficient prior with precision
$\Omega=K_{\mathrm{task}}^{-1}\otimes I_m$, where
$K_{\mathrm{task}}\succ0$, and a common Gaussian working-noise
variance $\lambda_N>0$, the joint posterior mean is
\begin{equation}
\begin{aligned}
    \widehat\vartheta
    =(A+\lambda_N\Omega)^{-1}\Phi^\top Y,\quad
    \widehat f^{(k)}(x)
    =\psi(x)^\top\widehat\vartheta_k.
\end{aligned}
\label{eq:boot-joint-ridge}
\end{equation}
This is the weight-space representation of finite-rank multi-task
GP regression \citep{rasmussen2006gaussian,bonilla2007multitask}.
The working variance $\lambda_N$ is a regularization parameter and
need not equal the variance proxy of the actual observation noise.
The choice $\Omega=I_{(J+1)m}$ gives independent task fits.

\begin{lemma}[Uniform finite-rank GP regression under Markov designs]
\label{lem:posterior_regression}
Suppose that each native design process $\{W_t^{(k)}\}_{t=1}^{N}$
is stationary and satisfies the conditions of
Lemma~\ref{lem:markov_bellman_concentration}, with
$M_N^{(k)}\leq M_N$ for a deterministic bound $M_N$.
Assume that
\[
    \sup_{x\in\mathcal X}\|\psi(x)\|_2\leq\kappa,
    \quad
    \Sigma_k=\mathbb E[\psi(X_t^{(k)})\psi(X_t^{(k)})^\top],
    \quad
    \lambda_{\min}(\Sigma_k)\geq\nu>0.
\]
Let the responses satisfy
\[
    Y_t^{(k)}=\psi(X_t^{(k)})^\top\vartheta_k+\xi_t^{(k)},
    \quad
    \max_k\|\vartheta_k\|_2\leq R,
\]
for deterministic coefficients $\vartheta_k$.
For each task, let $\{\mathscr F_t^{(k)}\}$ be a filtration such
that $X_t^{(k)}$ is $\mathscr F_t^{(k)}$-measurable,
$\xi_t^{(k)}$ is $\mathscr F_{t+1}^{(k)}$-measurable, and
\[
    \mathbb E\!\left[
        \exp(u\xi_t^{(k)})\mid\mathscr F_t^{(k)}
    \right]
    \leq\exp\!\left(\frac{u^2\sigma^2}{2}\right),
    \qquad u\in\mathbb R.
\]
Suppose that $\|\Omega\|_{\mathrm b}\leq\Lambda_N$ and
$\lambda_N\Lambda_N\leq N\nu/4$.
Given $\delta\in(0,1)$, assume
\begin{equation}
    N\geq
    \frac{128\kappa^4M_N^2}{\nu^2}
    \left\{m\log9+\log\frac{4(J+1)}{\delta}\right\}.
\label{eq:boot-design-size}
\end{equation}
Define
\[
    L_N(\delta)=m\log5+\log\frac{4(J+1)}{\delta},
    \quad
    s_N(\delta)=2\sigma\kappa
                      \sqrt{\frac{2L_N(\delta)}{N}}.
\]
Then, with probability at least $1-\delta$,
\begin{equation}
    \max_{0\leq k\leq J}
    \|\widehat f^{(k)}-f^{(k)}\|_\infty
    \leq
    \frac{4\kappa}{\nu}
    \left\{s_N(\delta)+\frac{\lambda_N\Lambda_NR}{N}\right\},
    \quad
    f^{(k)}(x)=\psi(x)^\top\vartheta_k.
\label{eq:boot-regression-bound}
\end{equation}
On the same event, let $\widehat f_d^{(k)}$ be the fit obtained by
replacing $Y$ with $Y+d$ while keeping $\Phi$, $\Omega$, and
$\lambda_N$ fixed. Every perturbation satisfying
$\max_{k,t}|d_t^{(k)}|\leq a$ obeys
\[
    \max_k\|\widehat f_d^{(k)}-\widehat f^{(k)}\|_\infty
    \leq\frac{4\kappa^2a}{\nu}.
\]
Thus \eqref{eq:boot-regression-bound} remains valid for the
perturbed fit after adding $4\kappa^2a/\nu$ to its right-hand side.
The perturbations may depend on the full dataset.
\end{lemma}

\noindent The proof is provided in Appendix~\ref{app:proof:lem:posterior_regression}.

Lemma~\ref{lem:posterior_regression} is uniform over tasks and input
locations for the stated regression model. Its stochastic term and
regularization bias are explicit. In particular,
\(
    m\nu\leq\operatorname{tr}(\Sigma_k)\leq\kappa^2,
\)
so a growing feature dimension requires tracking the dependence of
$\kappa$ and $\nu$ on $m$. The condition $m/N\to0$ alone does not
remove the dimension dependence from the error bound.
The same argument applies to a symmetric positive-semidefinite,
possibly data-dependent penalty $\Omega$ when its norm bound holds
pathwise. In this extension, the inverse in
\eqref{eq:boot-joint-ridge} exists on $\mathcal E_D$.
If it is singular on other realizations, the estimator may be defined
there as zero without changing the high-probability conclusion.
This convention is unnecessary for a positive-definite GP precision.
A singular task covariance, learned features, or observation-specific
adaptive working variances requires a separate analysis.

\begin{lemma}[Uniform Bellman residual control under trajectory reuse]
\label{lem:bellman_residual}
Retain the design, feature, and regularization conditions of
Lemma~\ref{lem:posterior_regression}, including
\eqref{eq:boot-design-size}.
For each task $k$, let $\mathcal Q_k$ be a nonempty deterministic
class of measurable functions on
$\mathcal S^{(k)}\times\mathcal A^{(k)}$ that is totally bounded
in the supremum norm. Suppose
\[
    \sup_{q\in\mathcal Q_k}\|q\|_\infty\leq B,
    \quad
    \|Q_*^{(k)}\|_\infty\leq B.
\]
Assume that the native Markov reward–transition model holds and that the Bellman innovations defined in Section 2.4 satisfy
\[
    \mathbb E\!\left[
        \exp(u\epsilon_t^{(k)})\mid\mathscr F_t^{(k)}
    \right]
    \leq\exp\!\left(\frac{u^2\sigma^2}{2}\right),
    \quad u\in\mathbb R,
\]
where $\mathscr F_t^{(k)}$ is the pre-transition filtration.
For every $q\in\mathcal Q_k$, suppose that a coefficient vector
$\vartheta_k(q)\in\mathbb R^m$ satisfies
\begin{equation}
\begin{aligned}
    \|\vartheta_k(q)\|_2\leq R_N,\quad
    \sup_{s,a}\left|
        (\mathcal T^{(k)}q)(s,a)
        -\psi(\phi^{(k)}(s,a))^\top\vartheta_k(q)
    \right|\leq a_N.
\end{aligned}
\label{eq:boot-bellman-approximation}
\end{equation}
For $\boldsymbol q=(q_0,\ldots,q_J)$ in
$\mathcal Q=\prod_{k=0}^{J}\mathcal Q_k$, define
\[
    Y_t^{(k)}(q_k)
    =R_t^{(k)}+
       \gamma\max_{a\in\mathcal A^{(k)}}q_k(S_{t+1}^{(k)},a).
\]
Let $\widehat\vartheta(\boldsymbol q)$ be the coefficient estimate
in \eqref{eq:boot-joint-ridge} for these stacked responses, and set
\[
    \mathfrak F_N^{(k)}(\boldsymbol q)(s,a)
    =\psi(\phi^{(k)}(s,a))^\top
      \widehat\vartheta_k(\boldsymbol q).
\]
Given a deterministic $\varepsilon>0$, define
\[
\begin{aligned}
    H_N(\varepsilon)
    =\max_k\log\mathcal N
          (\varepsilon,\mathcal Q_k,\|\cdot\|_\infty),\quad
    \sigma_B^2=2\sigma^2+8\gamma^2B^2,\quad
    \ell_N=m\log5+H_N(\varepsilon)
                 +\log\frac{4(J+1)}{\delta},
\end{aligned}
\]
where $\mathcal N(\varepsilon,\mathcal Q_k,\|\cdot\|_\infty)$
is the covering number with centers in $\mathcal Q_k$.
Then, with probability at least $1-\delta$,
\begin{equation}
    \sup_{\boldsymbol q\in\mathcal Q}\max_k
    \|\mathfrak F_N^{(k)}(\boldsymbol q)
                 -\mathcal T^{(k)}q_k\|_\infty
    \leq\overline\eta_N(\delta,\varepsilon),
\label{eq:boot-uniform-bellman}
\end{equation}
where
\begin{equation}\nonumber
\begin{aligned}
    \overline\eta_N(\delta,\varepsilon)
    ={}\left(1+\frac{4\kappa^2}{\nu}\right)a_N
    +\frac{4\kappa}{\nu}
    \left\{
        2\sigma_B\kappa\sqrt{\frac{2\ell_N}{N}}
        +2\gamma\kappa\varepsilon
        +\frac{\lambda_N\Lambda_NR_N}{N}
    \right\}.
\end{aligned}
\label{eq:boot-explicit-residual}
\end{equation}
On this same event, every fitted iteration satisfying
$\widehat{\boldsymbol Q}_{e-1}\in\mathcal Q$ and
$\widehat{\boldsymbol Q}_{e}
 =\mathfrak F_N(\widehat{\boldsymbol Q}_{e-1})$ obeys
\[
    \max_k\|\widehat Q_e^{(k)}
              -\mathcal T^{(k)}\widehat Q_{e-1}^{(k)}\|_\infty
    \leq\overline\eta_N(\delta,\varepsilon).
\]
The event is common to all such iterations, even when the iterates
are constructed from the same trajectories used in the regression.
\end{lemma}

\noindent The proof is provided in Appendix~\ref{app:proof:lem:bellman_residual}.

Condition \eqref{eq:boot-bellman-approximation} requires the
finite-rank feature space to approximate the Bellman image of every
candidate continuation function. It is an approximate
Bellman-completeness condition, with exact completeness when $a_N=0$.
Neither this property nor membership of the fitted iterates in the
prescribed classes follows from bounded rewards alone.
For example, the clipped feature class
\[
    \mathcal Q_k
    =\left\{
        (s,a)\longmapsto
        \operatorname{clip}_{[-B,B]}
        \bigl(\psi(\phi^{(k)}(s,a))^\top\theta\bigr)
        :\|\theta\|_2\leq L
      \right\}
\]
satisfies
\[
    H_N(\varepsilon)
    \leq m\log\!\left(1+\frac{2\kappa L}{\varepsilon}\right).
\]
This follows by covering the coefficient ball at radius
$\varepsilon/\kappa$ and using the nonexpansiveness of clipping.
Applying the lemma with this class requires verifying the coefficient
constraint for the actual iterates or accounting explicitly for a
projection step. Clipping alone ensures bounded output values but
does not establish the coefficient constraint.
If $\|\mathcal T^{(k)}q\|_\infty\leq B$, clipping a fitted output
to $[-B,B]$ cannot increase its distance from $\mathcal T^{(k)}q$,
so it preserves the residual bound. A sufficient condition is
$|R^{(k)}(s,a)|\leq R_{\max}$ and
$B\geq R_{\max}/(1-\gamma)$, because
$\|\mathcal T^{(k)}q\|_\infty\leq R_{\max}+\gamma B\leq B$.
Here the bound is on the conditional mean reward and does not
require bounded realized rewards.
With fixed $m$ and uniformly bounded $\kappa$, $\nu^{-1}$,
$\sigma_B$, $R_N$, $\Lambda_N$, and $L$, setting $a_N=0$,
$\lambda_N=O(1)$, and $\varepsilon=N^{-1}$ gives
\[
    \overline\eta_N(\delta,N^{-1})
    =O\!\left(
        \sqrt{\frac{m\log N+\log((J+1)/\delta)}{N}}
      \right),
\]
subject to \eqref{eq:boot-design-size} and the preceding entropy
bound. When $m$ grows, the dependence on $\kappa$, $\nu$, $R_N$,
and $\Lambda_N$ must also be retained.
The high-probability event is simultaneous over all iterations whose
inputs remain in the prescribed classes. Trajectory reuse therefore
requires uniformity over the continuation functions rather than an
interpretation of successive Bellman responses as independent
observations \citep{munos2008finite}.

The next theorem converts the uniform residual control of
Lemma~\ref{lem:bellman_residual} into consistency of the fitted
Q-iterates. The error-propagation argument is deterministic, while
the statistical guarantee uses the common high-probability event
established in that lemma.
\begin{theorem}[Bootstrap consistency]
\label{the:bootstrap_consistency}
For each task $k$, let $\mathcal T^{(k)}$ be a
$\gamma$-contraction in the supremum norm, with fixed point
$Q_*^{(k)}$, where $\gamma\in(0,1)$. Define
\[
    D_{e,N}
    =
    \max_k
    \|\widehat Q_e^{(k)}-Q_*^{(k)}\|_\infty,
    \quad
    \eta_{e,N}
    =
    \max_k
    \|\widehat Q_e^{(k)}
      -\mathcal T^{(k)}\widehat Q_{e-1}^{(k)}\|_\infty .
\]
Then, for every $e\geq1$,
\begin{equation}
    D_{e,N}
    \leq
    \gamma^eD_{0,N}
    +
    \sum_{j=1}^{e}
    \gamma^{e-j}\eta_{j,N}.
    \label{eq:boot-recursion}
\end{equation}
Suppose further that the conditions of
Lemma~\ref{lem:bellman_residual} hold up to iteration $E_N$ and that
$D_{0,N}\leq D_0$ almost surely. Then, with probability at least
$1-\delta_N$,
\begin{equation}
    D_{e,N}
    \leq
    \gamma^eD_0
    +
    \frac{1-\gamma^e}{1-\gamma}\,
    \overline\eta_N,
    \qquad
    1\leq e\leq E_N,
    \label{eq:boot-high-probability}
\end{equation}
where $\overline\eta_N$ is the uniform one-step residual bound from
Lemma~\ref{lem:bellman_residual}.
If
\[
    \delta_N\to0,
    \qquad
    \overline\eta_N\to0,
    \qquad
    \gamma^{E_N}D_0\to0,
\]
then
\[
    D_{E_N,N}\xrightarrow{\mathbb P}0.
\]
\end{theorem}

\noindent The proof is provided in Appendix~\ref{app:proof:the:bootstrap_consistency}.

\begin{corollary}[Stability of the Bellman bootstrap remainder]
\label{cor:bootstrap_stability}
Under Theorem~\ref{the:bootstrap_consistency}, define, for $e\geq1$,
\[
    b_{t,e}^{(k)}
    =
    \gamma\left\{
        \max_{a\in\mathcal A^{(k)}}
        \widehat Q_{e-1}^{(k)}(S_{t+1}^{(k)},a)
        -
        \max_{a\in\mathcal A^{(k)}}
        Q_*^{(k)}(S_{t+1}^{(k)},a)
    \right\}.
\]
Then, for every $k,t,e$,
\begin{equation}
    |b_{t,e}^{(k)}|
    \leq
    \gamma D_{e-1,N}.
    \label{eq:boot-pathwise-stability}
\end{equation}
Hence, on the event $\mathcal E_N$ of
Theorem~\ref{the:bootstrap_consistency},
\begin{equation}
    \sup_{k,t}|b_{t,e}^{(k)}|
    \leq
    \gamma r_{e-1,N},
    \quad
    1\leq e\leq E_N.
    \label{eq:boot-high-probability-remainder}
\end{equation}
Consequently, for any deterministic sequence $e_N\leq E_N$,
\[
    r_{e_N-1,N}\to0,
    \quad
    \delta_N\to0
\]
imply
\[
    \sup_{k,t}|b_{t,e_N}^{(k)}|
    \xrightarrow{\mathbb P}0.
\]
Suppose additionally that the fitted update is deterministically
restricted to the supremum-norm ball of radius $B$, for example by
clipping or projection, and that
$\|Q_*^{(k)}\|_\infty\leq B$ for every task. Then
\begin{equation}
    \sup_{k,t}
    \mathbb E\!\left[(b_{t,e}^{(k)})^2\right]
    \leq
    \gamma^2
    \left\{
        r_{e-1,N}^2
        +4B^2\delta_N
    \right\},
    \quad
    1\leq e\leq E_N.
    \label{eq:boot-mean-square-stability}
\end{equation}
Therefore,
$r_{e_N-1,N}\to0$ and $\delta_N\to0$ also imply
\[
    \sup_{k,t}
    \mathbb E\!\left[(b_{t,e_N}^{(k)})^2\right]
    \longrightarrow0.
\]
\end{corollary}

\noindent The proof is provided in Appendix~\ref{app:proof:cor:bootstrap_stability}.
The preceding results concern the Bellman remainder associated with the
unconstrained update. For a conservative target update, the same
pathwise error-propagation argument remains valid, but the corresponding
one-step residual must also account for the coverage gap and the
pessimism penalty identified in
\eqref{sel:conservative-remainder}.
Lemma~\ref{lem:markov_bellman_concentration} suggests the quantity
$N/(M_N^{(k)})^2$ as a concentration scale for bounded marginal
averages. It should not, however, be interpreted automatically as the
effective sample size of a conditional Bellman regression problem.
Whenever a task-specific quantity $n_{k,\mathrm{eff}}$ has been
justified by the relevant concentration analysis, one may introduce
\(
    N_{\mathrm{eff}}
    =
    n_{0,\mathrm{eff}}
    +
    \sum_{k=1}^{J}n_{k,\mathrm{eff}}
\)
as a bookkeeping quantity. Additional transfer and coverage conditions
are required before this sum can be interpreted as target information
available for learning.
The results above therefore establish control of Bellman
bootstrapping, but do not by themselves imply consistent source
selection. The latter additionally requires alignment between target
risk and predictive scores, together with uniform control of the
corresponding score fluctuations.

The preceding results establish control of the Bellman bootstrap
remainder under dependent trajectory sampling. We next turn to source
selection and study whether the resulting predictive scores consistently
favor population-optimal source subsets.

\subsection{Predictive selection consistency and target-risk implications}
\label{subsec:predictive-selection}

The preceding analysis establishes control of Bellman-generated
responses under dependent trajectory sampling. We now study source
selection through the target predictive scores. To distinguish the
criterion used for selection from the criterion used for evaluation,
predictive optimality and target-risk optimality are treated separately.
For each $S\in\mathcal S_J$, let $\mathcal Q_S$ be a nonempty
deterministic class of target predictors. Suppose that the minimum
target risk over $\mathcal Q_S$ is finite and attained, and define
\[
    Q_S^\dagger
    =
    \arg\min_{Q\in\mathcal Q_S}R_0(Q),
    \quad
    R_0^\dagger(S)
    =
    R_0(Q_S^\dagger),
\]
where $\arg\min$ denotes a fixed deterministic choice when the
minimizer is not unique. The oracle target risk and the corresponding
oracle subset class are
\[
    R_0^*
    =
    \min_{S\in\mathcal S_J}R_0^\dagger(S),
    \quad
    \mathcal S^{\mathrm{risk}}
    =
    \left\{
        S\in\mathcal S_J:
        R_0^\dagger(S)=R_0^*
    \right\}.
\]
For a deterministic tolerance $\epsilon_N>0$, define the
target-risk near-oracle class
\[
    \mathcal G_N^{\mathrm{risk}}(\epsilon_N)
    =
    \left\{
        S\in\mathcal S_J:
        R_0^\dagger(S)<R_0^*+\epsilon_N
    \right\}.
\]
These quantities specify the population target of transfer and are
not themselves used to construct the predictive selection rule.

Using the Bellman decomposition from Section~3.1, write
\[
    \mathbf Y_e
    =
    \overline{\mathbf Y}
    +
    \mathbf b_e,
    \quad
    \overline Y_t^{(k)}
    =
    Q_*^{(k)}(X_t^{(k)})+\epsilon_t^{(k)}.
\]
The vector $\overline{\mathbf Y}$ removes the Bellman bootstrap
remainder and serves only as an analytical reference.
For each $S\in\mathcal S_J$, define the predictive score based on the
actual Bellman responses and its oracle-response counterpart by
\[
    \ell_{e,N}(S)
    =
    \log p_{S,N}\!\left(
        \mathbf Y_e^{(0)}
        \mid X,\mathbf Y_e^{(S)}
    \right),\qquad
    \ell_N^\circ(S)
    =
    \log p_{S,N}\!\left(
        \overline{\mathbf Y}^{(0)}
        \mid X,\overline{\mathbf Y}^{(S)}
    \right).
\]
All candidate subsets are evaluated using the same target-response
vector and the same predictive construction. The comparison between
$\ell_{e,N}$ and $\ell_N^\circ$ is made with the predictive
hyperparameters held fixed. If hyperparameters are refitted after
perturbing the responses, the resulting change must be included in the
score perturbation considered below.
Let $\Pi_N$ be a deterministic prior satisfying
$\Pi_N(S)>0$ for every $S\in\mathcal S_J$. Define the working subset
posterior by
\begin{equation}\nonumber
    \Pi_{e,N}(S\mid\mathbf Y_e,X)
    =
    \frac{
        \Pi_N(S)\exp\{\ell_{e,N}(S)\}
    }{
        \displaystyle
        \sum_{U\in\mathcal S_J}
        \Pi_N(U)\exp\{\ell_{e,N}(U)\}
    }.
    \label{eq:selection_working_posterior}
\end{equation}
This is a predictive-score weighting rule and need not coincide with
the Bayesian posterior under the full trajectory likelihood
\citep{bissiri2016general}.
Assume that $\ell_N^\circ(S)$ is integrable under the true trajectory
law $\mathbb P_\star$ for every $S\in\mathcal S_J$, and define
\(
    \mathcal L_N^\circ(S)
    =
    \mathbb E_\star[\ell_N^\circ(S)].
\)
The optimal population predictive score and its oracle subset class are
\[
    \mathcal L_N^*
    =
    \max_{S\in\mathcal S_J}\mathcal L_N^\circ(S),
    \quad
    \mathcal S_N^{\mathrm{pred}}
    =
    \left\{
        S\in\mathcal S_J:
        \mathcal L_N^\circ(S)=\mathcal L_N^*
    \right\}.
\]
Choose the predictive oracle reference
\begin{equation}\nonumber
    S_N^\star
    =
    \arg\max_{S\in\mathcal S_N^{\mathrm{pred}}}\Pi_N(S),
    \quad
    \pi_N^\star
    =
    \Pi_N(S_N^\star),
    \label{eq:selection_reference}
\end{equation}
where $\arg\max$ denotes a fixed deterministic choice when the
maximizer is not unique.
For a deterministic predictive tolerance $d_N>0$, define
\[
    \mathcal B_N^{\mathrm{pred}}(d_N)
    =
    \left\{
        S\in\mathcal S_J:
        \mathcal L_N^*
        -
        \mathcal L_N^\circ(S)
        \geq d_N
    \right\}.
\]
Thus,
$\mathcal S_J\setminus\mathcal B_N^{\mathrm{pred}}(d_N)$
is the corresponding predictive near-oracle class.
For each candidate subset, define the centered oracle-score contrast
\[
    Z_N(S)
    =
    \ell_N^\circ(S)
    -
    \ell_N^\circ(S_N^\star)
    -
    \left\{
        \mathcal L_N^\circ(S)
        -
        \mathcal L_N^*
    \right\}.
\]
By construction,
\(
    \mathbb E_\star[Z_N(S)]=0.
\)
Define the perturbation of the relative predictive score by
\[
\begin{aligned}
    u_{e,N}(S)
    =
    \bigl\{
        \ell_{e,N}(S)-\ell_{e,N}(S_N^\star)
    \bigr\}
    -
    \bigl\{
        \ell_N^\circ(S)-\ell_N^\circ(S_N^\star)
    \bigr\},
\end{aligned}
\]
and set
\(
    \mathfrak B_{e,N}
    =
    \max_{S\in\mathcal S_J}|u_{e,N}(S)|.
\)
The actual relative predictive score then admits the exact
decomposition
\begin{equation}\nonumber
\begin{aligned}
    \ell_{e,N}(S)-\ell_{e,N}(S_N^\star)
    ={}
    -\left\{
        \mathcal L_N^*
        -
        \mathcal L_N^\circ(S)
    \right\}
    +
    Z_N(S)
    +
    u_{e,N}(S).
\end{aligned}
\label{eq:selection_score_decomposition}
\end{equation}
Hence, selection is governed by three components: population
predictive separation, stochastic fluctuation of the oracle-response
score, and perturbation caused by Bellman bootstrapping. Define the population tail envelope
\[
    \Psi_N(x)
    =
    \sup_{S\in\mathcal S_J}
    \mathbb P_\star
    \left(
        |Z_N(S)|>x
    \right),
    \quad x>0.
\]
This is a property of the population distribution of the
oracle-response scores and involves no fitted quantity.
For every deterministic $a_N>0$, the union bound gives
\begin{equation}\nonumber
    \mathbb P_\star\!\left(
        \max_{S\in\mathcal S_J}|Z_N(S)|>a_N
    \right)
    \leq
    |\mathcal S_J|\Psi_N(a_N).
    \label{eq:selection_uniform_score}
\end{equation}
Specific models may be used to obtain an explicit form for
$\Psi_N$, but the selection result below only requires the resulting
uniform tail bound.
The prior enters through the deterministic quantity
\(
    q_N
    =
    \max_{S\in\mathcal S_J}
    \log
    \frac{\Pi_N(S)}{\pi_N^\star}.
\)

The following result transfers the Bellman-bootstrap control obtained
in Section~3.1 to the predictive scores.

\begin{lemma}[Pathwise predictive-score perturbation under Bellman bootstrapping]
\label{lem:selection_bootstrap_scores}

For each $S\in\mathcal S_J$, suppose that, conditional on the input
locations $X$, the working target predictive density has the Gaussian
form
\[
    p_{S,N}
    \bigl(
        \mathbf y^{(0)}
        \mid X,\mathbf y^{(S)}
    \bigr)
    =
    \mathcal N
    \left(
        \mathbf a_S+A_S\mathbf y^{(S)},
        V_S
    \right).
\]
Conditional on $X$, the quantities $\mathbf a_S$, $A_S$, and $V_S$
are determined by a fixed predictive specification and do not depend
on the response values. In particular, no response-dependent
hyperparameter refitting is performed when passing from the
oracle-response score to the actual score.
Suppose there exist deterministic sequences
$\underline v_N>0$ and $L_{A,N}<\infty$ such that, almost surely,
\[
    \inf_{S\in\mathcal S_J}
    \lambda_{\min}(V_S)
    \geq
    \underline v_N,
    \quad
    \sup_{S\in\mathcal S_J}
    \|A_S\|_{\mathrm{op}}
    \leq
    L_{A,N}.
\]
Define
\[
\begin{aligned}
    \mathbf r_S^\circ
    =
    \overline{\mathbf Y}^{(0)}
    -
    \mathbf a_S
    -
    A_S\overline{\mathbf Y}^{(S)},\quad
    \mathbf d_{S,e}
    =
    \mathbf b_e^{(0)}
    -
    A_S\mathbf b_e^{(S)},\quad
    \mathcal R_N
    =
    \max_{S\in\mathcal S_J}
    \|\mathbf r_S^\circ\|_2,\quad
    \mathcal D_{e,N}
    =
    \max_{S\in\mathcal S_J}
    \|\mathbf d_{S,e}\|_2.
\end{aligned}
\]
Then
\begin{equation}
    \mathfrak B_{e,N}
    \leq
    \frac{
        2\mathcal R_N\mathcal D_{e,N}
        +
        \mathcal D_{e,N}^2
    }{
        \underline v_N
    }.
    \label{eq:selection_bootstrap_score_bound}
\end{equation}
Moreover, if $\mathbf b_e$ denotes the vector obtained by stacking
the Bellman bootstrap remainders from all tasks, then
\begin{equation}
    \mathcal D_{e,N}
    \leq
    (1+L_{A,N})
    \|\mathbf b_e\|_2,
    \label{eq:selection_d_bound}
\end{equation}
and hence
\begin{equation}
\begin{aligned}
    \mathfrak B_{e,N}
    \leq
    \frac{1}{\underline v_N}
    \Bigl\{
        2(1+L_{A,N})
        \mathcal R_N
        \|\mathbf b_e\|_2
        +
        (1+L_{A,N})^2
        \|\mathbf b_e\|_2^2
    \Bigr\}.
\end{aligned}
\label{eq:selection_bootstrap_score_pathwise}
\end{equation}
\end{lemma}

\noindent The proof is provided in Appendix~\ref{app:proof:lem:selection_bootstrap_scores}.

\begin{corollary}[Rate of the bootstrap predictive-score perturbation]
\label{cor:selection_bootstrap_score_rate}

Let
\(
    n_{\mathrm{tot}}
    =
    \sum_{k=0}^{J}n_k,
\)
and define
\(
    \beta_{e,N}^2
    =
    \gamma^2
    \left\{
        r_{e-1,N}^2
        +
        4B^2\delta_N
    \right\},
\)
where $r_{e-1,N}$ and $\delta_N$ are the deterministic quantities
appearing in
Corollary~\ref{cor:bootstrap_stability}. Suppose there exist deterministic sequences
$s_N>0$ and $\eta_{R,N}\downarrow0$ such that
\(
    \mathbb P_\star
    \left(
        \mathcal R_N>s_N
    \right)
    \leq
    \eta_{R,N}.
\)
Then
\begin{equation}
    \mathfrak B_{e,N}
    =
    O_{\mathbb P_\star}
    \left[
        \frac{
            (1+L_{A,N})
            s_N\sqrt{n_{\mathrm{tot}}}\,
            \beta_{e,N}
            +
            (1+L_{A,N})^2
            n_{\mathrm{tot}}\beta_{e,N}^2
        }{
            \underline v_N
        }
    \right].
    \label{eq:selection_bootstrap_score_rate}
\end{equation}
\end{corollary}

\noindent The proof is provided in Appendix~\ref{app:proof:cor:selection_bootstrap_score_rate}.

For later use, define
\begin{equation}
    \rho_{e,N}^{\mathrm{bt}}
    =
    \frac{
        (1+L_{A,N})
        s_N\sqrt{n_{\mathrm{tot}}}\,
        \beta_{e,N}
        +
        (1+L_{A,N})^2
        n_{\mathrm{tot}}\beta_{e,N}^2
    }{
        \underline v_N
    }.
    \label{eq:selection_bootstrap_rate_scale}
\end{equation}
Then Corollary~\ref{cor:selection_bootstrap_score_rate} gives
\(
    \mathfrak B_{e,N}
    =
    O_{\mathbb P_\star}
    \bigl(
        \rho_{e,N}^{\mathrm{bt}}
    \bigr).
\)
In addition, if
\(
    s_N=O(\sqrt{n_0}),
    ~
    L_{A,N}=O(1)
\) and 
\(   
    \underline v_N^{-1}=O(1),
\)
then
\(
    \rho_{e,N}^{\mathrm{bt}}
    =
    O\!\left(
        \sqrt{n_0n_{\mathrm{tot}}}\,
        \beta_{e,N}
        +
        n_{\mathrm{tot}}\beta_{e,N}^2
    \right).
\)

\begin{lemma}[Deterministic posterior comparison]
\label{lem:selection_posterior_bound}

Let $\varnothing\neq\mathcal B\subseteq\mathcal S_J$. Then
\begin{equation}
\begin{aligned}
    \Pi_{e,N}
    (\mathcal B\mid\mathbf Y_e,X)
    \leq
    \min\Bigg\{
        1,\,
        \exp\Bigg(
            \log|\mathcal B|
            +
            q_N
            +
            \max_{S\in\mathcal B}
            \bigl\{
                \ell_{e,N}(S)
                -
                \ell_{e,N}(S_N^\star)
            \bigr\}
        \Bigg)
    \Bigg\}.
\end{aligned}
\label{eq:selection_mass_bound}
\end{equation}
Moreover,
\begin{equation}
    \max_{S\in\mathcal B}
    \frac{
        \Pi_{e,N}(S\mid\mathbf Y_e,X)
    }{
        \Pi_{e,N}(S_N^\star\mid\mathbf Y_e,X)
    }
    \leq
    \exp\Bigg(
        q_N
        +
        \max_{S\in\mathcal B}
        \bigl\{
            \ell_{e,N}(S)
            -
            \ell_{e,N}(S_N^\star)
        \bigr\}
    \Bigg).
    \label{eq:selection_odds_bound}
\end{equation}
For $\mathcal B=\varnothing$, the posterior-mass conclusion is
immediate.
\end{lemma}

\noindent The proof is provided in Appendix~\ref{app:proof:lem:selection_posterior_bound}.

\begin{theorem}[Predictive near-oracle posterior concentration and selection]
\label{thm:predictive_selection}

Let $d_N>0$ be deterministic, and let $e=e_N$ be a deterministic
iteration index for which
Corollary~\ref{cor:selection_bootstrap_score_rate} applies. Write
\(
    \mathcal B_N
    =
    \mathcal B_N^{\mathrm{pred}}(d_N).
\)
For $a>0$, define the upper-tail fluctuation probability
\(
    \Xi_N^+(a;d_N)
    =
    \mathbb P_\star
    \left(
        \max_{S\in\mathcal B_N}
        Z_N(S)>a
    \right),
\)
with the convention that $\Xi_N^+(a;d_N)=0$ when
$\mathcal B_N=\varnothing$.
Suppose there exists a deterministic sequence $a_N>0$ such that
\begin{equation}
    \Xi_N^+(a_N;d_N)
    \longrightarrow0.
    \label{eq:selection_score_fluctuation}
\end{equation}
Define the prior-mass penalty
\(
    \Lambda_N(d_N)
    =
    \log
    \left(
        1\vee
        \frac{
            \Pi_N(\mathcal B_N)
        }{
            \pi_N^\star
        }
    \right),
\)
and the effective predictive margin
\(
    \kappa_N
    =
    d_N-a_N-\Lambda_N(d_N).
    \label{eq:selection_effective_margin}
\)
Assume
\(
    \kappa_N\longrightarrow\infty,
    ~
    \rho_{e,N}^{\mathrm{bt}}
    =
    o(\kappa_N),
\)
where $\rho_{e,N}^{\mathrm{bt}}$ is the deterministic rate in
Corollary~\ref{cor:selection_bootstrap_score_rate}.
\\
Then
\begin{equation}
    \Pi_{e,N}
    \left(
        \mathcal B_N
        \mid
        \mathbf Y_e,X
    \right)
    \xrightarrow{\mathbb P_\star}0.
    \label{eq:selection_predictive_concentration}
\end{equation}
Moreover,
\begin{equation}
    \mathbb E_\star
    \left[
        \Pi_{e,N}
        \left(
            \mathcal B_N
            \mid
            \mathbf Y_e,X
        \right)
    \right]
    \longrightarrow0.
    \label{eq:selection_predictive_mean_concentration}
\end{equation}
If $\widehat S$ is drawn from the working subset posterior, with the
auxiliary sampling randomness included in the probability, then
\begin{equation}
    \mathbb P_\star
    \left(
        \widehat S\in\mathcal B_N
    \right)
    \longrightarrow0.
    \label{eq:selection_posterior_draw}
\end{equation}
The same conclusion holds for any maximum-a-posteriori selection.
\end{theorem}

\noindent The proof is provided in Appendix~\ref{app:proof:thm:predictive_selection}.
Theorem~\ref{thm:predictive_selection} establishes concentration with
respect to the population predictive criterion. Predictive optimality,
however, need not coincide with target-risk optimality. To quantify
their population-level discrepancy, define the calibration radius
\begin{equation}\nonumber
    \omega_N(d)
    =
    \max_{
        S\in
        \mathcal S_J
        \setminus
        \mathcal B_N^{\mathrm{pred}}(d)
    }
    \left\{
        R_0^\dagger(S)-R_0^*
    \right\},
    \qquad d>0.
    \label{eq:risk_score_calibration_radius}
\end{equation}
The maximum is well defined because
$\mathcal S_J$ is finite and
$\mathcal S_J\setminus
\mathcal B_N^{\mathrm{pred}}(d)$
contains the population predictive oracle class.
By construction,
\begin{equation}\nonumber
    S\notin
    \mathcal B_N^{\mathrm{pred}}(d)
    \quad\Longrightarrow\quad
    R_0^\dagger(S)-R_0^*
    \leq
    \omega_N(d).
    \label{eq:risk_score_calibration}
\end{equation}
Thus, $\omega_N(d)$ measures the largest target-risk excess among
subsets that are within $d$ of the optimal population predictive
score. It is a deterministic population quantity and does not involve
a fitted estimator.

\begin{corollary}[Target-risk implication of predictive selection]
\label{cor:selection_target_risk}

Under the conditions of
Theorem~\ref{thm:predictive_selection},
\begin{equation}
\begin{aligned}
    \Pi_{e,N}
    \Bigl(
        \bigl\{
            S\in\mathcal S_J:
            R_0^\dagger(S)-R_0^*
            \leq
            \omega_N(d_N)
        \bigr\}
        \,\Bigm|\,
        \mathbf Y_e,X
    \Bigr)
    \xrightarrow{\mathbb P_\star}1.
\end{aligned}
\label{eq:selection_calibrated_risk_posterior}
\end{equation}
Moreover,
\begin{equation}
\begin{aligned}
    \mathbb E_\star
    \Bigg[
        \Pi_{e,N}
        \Bigl(
            \bigl\{
                S:
                R_0^\dagger(S)-R_0^*
                \leq
                \omega_N(d_N)
            \bigr\}
            \,\Bigm|\,
            \mathbf Y_e,X
        \Bigr)
    \Bigg]
    \longrightarrow1.
\end{aligned}
\label{eq:selection_calibrated_risk_mean}
\end{equation}
For either a posterior draw or a maximum-a-posteriori selection,
\begin{equation}
    \mathbb P_\star
    \left\{
        R_0^\dagger(\widehat S)-R_0^*
        \leq
        \omega_N(d_N)
    \right\}
    \longrightarrow1.
    \label{eq:selection_calibrated_risk_rule}
\end{equation}
Consequently, if
\(
    \omega_N(d_N)\longrightarrow0,
\)
then the selected subset is target-risk near-oracle:
\begin{equation}
    R_0^\dagger(\widehat S)-R_0^*
    \xrightarrow{\mathbb P_\star}0.
    \label{eq:selection_target_risk_consistency}
\end{equation}
More generally, for any deterministic sequence
$\epsilon_N>\omega_N(d_N)$,
\begin{equation}
    \Pi_{e,N}
    \left(
        \mathcal G_N^{\mathrm{risk}}(\epsilon_N)
        \mid
        \mathbf Y_e,X
    \right)
    \xrightarrow{\mathbb P_\star}1,
    \label{eq:selection_target_risk_posterior}
\end{equation}
and
\begin{equation}
    \mathbb P_\star
    \left\{
        R_0^\dagger(\widehat S)
        <
        R_0^*+\epsilon_N
    \right\}
    \longrightarrow1.
    \label{eq:selection_target_risk_rule}
\end{equation}
\end{corollary}

\noindent The proof is provided in Appendix~\ref{app:proof:cor:selection_target_risk}.

Theorem~\ref{thm:predictive_selection} establishes concentration on
subsets that are near-optimal under the population predictive
criterion. Corollary~\ref{cor:selection_target_risk} quantifies the
target-risk consequence through the population calibration radius
$\omega_N(d_N)$. This quantity measures the largest target-risk excess
within the predictive near-oracle class.
The predictive selection result therefore does not require predictive
and target-risk optimality to coincide. Target-risk consistency follows
when the discrepancy between the two criteria vanishes at the selected
predictive tolerance, namely when
$\omega_N(d_N)\to0$. All quantities entering this calibration statement
are defined at the population level and do not involve fitted
estimators.

\section{Transfer-adaptive posterior contraction under predictive source selection}
\label{sec:posterior_contraction}

The preceding analysis establishes two ingredients needed for transfer:
control of Bellman-generated responses and concentration of the
source-subset weights under the population predictive criterion.
Neither result, however, determines how much statistical information
the selected source observations provide about the target Q-function.
Posterior contraction requires an additional analysis of the target
marginal posterior conditional on a source subset.
This section therefore separates three questions: contraction of the
target working posterior, frequentist accuracy of its posterior mean,
and decision quality of the policy induced by that posterior. We first
identify the information available to the target under the separable
multi-task GP model and show why generic full-rank task covariance does
not justify additive pooling of source and target sample sizes. We then
derive finite-design perturbation bounds and a selection-to-contraction
result. Finally, an explicit shared-function experiment is used to
establish a pooled contraction rate and its corresponding minimax lower
bound.

Throughout this section, source relevance is distinguished from source
information. Predictive selection determines which subsets receive
asymptotically nonnegligible weight, whereas the amount of transferable
information is determined by the conditional covariance structure of the
working GP model. In particular, an additive effective sample size is
not assumed merely because a subset is predictively near-oracle. The
rank-one shared-function model considered below is an additional
structural specialization of the separable GP model and is not implied
by bounded eigenvalues of a general task covariance matrix. Unless
stated otherwise, all probability statements are under the true
trajectory distribution $\mathbb P_\star$.

\subsection{Setup, selection weights and transferable information}
\label{subsec:pc-setup}

Let
\(
    f_0=Q_*^{(0)}
\) and
\(
    \mathbb H=L^2(P_X^{(0)})
\)
with norm $\|\cdot\|_{\mathbb H}$. For
$S\in\mathcal S_J$, define the active task set
\(
    \mathcal I(S)=\{0\}\cup S.
\)
The source-subset weights are those obtained from the predictive
selection step:
\begin{equation}\nonumber
    \omega_{e,N}(S)
    =
    \Pi_{e,N}
    \left(
        S\mid\mathbf Y_e,X
    \right),
    \quad
    S\in\mathcal S_J.
    \label{eq:pc-selection-weight}
\end{equation}
For a fixed subset $S$, let
$\Pi_{S,e,N}^{(0)}$ denote the corresponding target marginal working
posterior on $\mathbb H$. To avoid confusing this function-space
posterior with the subset posterior
$\Pi_{e,N}(S\mid\mathbf Y_e,X)$, define the model-averaged target
working posterior by
\begin{equation}\nonumber
    \widetilde\Pi_{e,N}^{(0)}(A)
    =
    \sum_{S\in\mathcal S_J}
    \omega_{e,N}(S)
    \Pi_{S,e,N}^{(0)}(A),
    \quad
    A\subseteq\mathbb H,
    \label{eq:pc-mixture}
\end{equation}
where
\(
    \sum_{S\in\mathcal S_J}\omega_{e,N}(S)=1.
\)
Let
$\mathcal G_N^{\mathrm{ctr}}\subseteq\mathcal S_J$
be a nonempty deterministic family of subsets for which the contraction
conditions introduced below hold uniformly, and define the posterior
weight assigned outside this family by
\begin{equation}\nonumber
    \Omega_N^{\mathrm{ctr}}
    =
    \sum_{S\notin\mathcal G_N^{\mathrm{ctr}}}
    \omega_{e,N}(S).
    \label{eq:pc-bad-weight}
\end{equation}
To transfer subsetwise contraction to
$\widetilde\Pi_{e,N}^{(0)}$, it is sufficient that
\(
    \Omega_N^{\mathrm{ctr}}
    \xrightarrow{\mathbb P_\star}0.
\)
This condition can be linked directly to the predictive selection
result. In particular, if for some deterministic $d_N>0$,
\begin{equation}
    \mathcal S_J
    \setminus
    \mathcal B_N^{\mathrm{pred}}(d_N)
    \subseteq
    \mathcal G_N^{\mathrm{ctr}},
    \label{eq:pc-selection-contraction-link1}
\end{equation}
then
\[
\begin{aligned}
    \Omega_N^{\mathrm{ctr}}
    &\leq
    \Pi_{e,N}
    \left(
        \mathcal B_N^{\mathrm{pred}}(d_N)
        \mid
        \mathbf Y_e,X
    \right)
    \xrightarrow{\mathbb P_\star}0
\end{aligned}
\]
by Theorem~\ref{thm:predictive_selection}. If
\eqref{eq:pc-selection-contraction-link1} does not hold, concentration on
$\mathcal G_N^{\mathrm{ctr}}$ requires a separate argument. Thus,
predictive near-oracle selection and posterior contraction are related
but logically distinct.
For every active task, retain the Bellman decomposition
\begin{equation}\nonumber
    \mathbf Y_e^{(k)}
    =
    \overline{\mathbf Y}^{(k)}
    +
    \mathbf b_e^{(k)},
    \quad
    \overline Y_t^{(k)}
    =
    Q_*^{(k)}(X_t^{(k)})
    +
    \epsilon_t^{(k)}.
    \label{eq:pc-responses}
\end{equation}
The oracle response
$\overline{\mathbf Y}^{(k)}$
is an analytical device and is not an additional observed sample.
The Bellman remainder
$\mathbf b_e^{(k)}$
need not be centered, Gaussian, or independent of the oracle response.
For conservative Bellman targets it may also contain the coverage and
pessimism terms identified in Section~3.1. Consequently, its effect
cannot in general be represented by simply increasing a Gaussian
observation-noise variance.

\begin{proposition}[Frequency-wise target information and the obstruction to generic pooling]
\label{prop:pc-information}

Fix $S\in\mathcal S_J$ and let
\(
    B_S
    =
    K_{\mathrm{task}}
    \bigl(
        \mathcal I(S),\mathcal I(S)
    \bigr).
\)
For a fixed input-kernel eigenvalue $\lambda_j>0$, consider the Gaussian
sequence working experiment
\begin{equation}
    \boldsymbol\theta_j
    \sim
    \mathcal N
    \left(
        0,
        \lambda_j B_S
    \right),
    \quad
    \mathbf Z_j
    =
    \boldsymbol\theta_j
    +
    D_S^{-1/2}\boldsymbol\xi_j,
    \label{eq:pc-sequence-experiment}
\end{equation}
where
\(
    D_S
    =
    \operatorname{diag}
    \left(
        \tau_k:
        k\in\mathcal I(S)
    \right),
    ~
    \tau_k>0,
\)
and
$\boldsymbol\xi_j\sim\mathcal N(0,I)$
is independent of
$\boldsymbol\theta_j$.
\\
Partition $B_S$ according to the target and source coordinates as
\[
    B_S
    =
    \begin{pmatrix}
        B_{00} & B_{0S}\\
        B_{S0} & B_{SS}
    \end{pmatrix},
\]
and let
\(
    D_{S,\mathrm{src}}
    =
    \operatorname{diag}(\tau_k:k\in S).
\)
For $S\neq\varnothing$, define
\begin{equation}
\begin{aligned}
    c_{j,S}
    &=
    \lambda_j B_{00}
    -
    \lambda_j^2
    B_{0S}
    \left(
        \lambda_jB_{SS}
        +
        D_{S,\mathrm{src}}^{-1}
    \right)^{-1}
    B_{S0}.
\end{aligned}
\label{eq:pc-source-conditional-variance}
\end{equation}
For $S=\varnothing$, set
$c_{j,\varnothing}=\lambda_jB_{00}$.
Then the conditional posterior variance of the target coefficient is
\begin{equation}
    v_{0j}(S)
    =
    \operatorname{Var}
    \left(
        \theta_{0j}\mid\mathbf Z_j
    \right)
    =
    \left(
        \tau_0+c_{j,S}^{-1}
    \right)^{-1}.
    \label{eq:pc-frequency-information}
\end{equation}
Suppose now that $S\neq\varnothing$ and $B_S\succ0$. Define the
task-specific conditional-variance factor
\(
    \chi_S
    =
    B_{00}
    -
    B_{0S}
    B_{SS}^{-1}
    B_{S0}.
\)
Then $\chi_S>0$ and
\begin{equation}
    c_{j,S}
    \geq
    \lambda_j\chi_S.
    \label{eq:pc-source-information-floor}
\end{equation}
Consequently,
\begin{equation}
    v_{0j}(S)
    \geq
    \left\{
        \tau_0
        +
        (\lambda_j\chi_S)^{-1}
    \right\}^{-1}.
    \label{eq:pc-target-variance-floor}
\end{equation}
Moreover, as all source precisions diverge,
\(
    \min_{k\in S}\tau_k\longrightarrow\infty,
\)
the lower bound is attained in the limit:
\begin{equation}
    c_{j,S}
    \longrightarrow
    \lambda_j\chi_S,
    \quad
    v_{0j}(S)
    \longrightarrow
    \left\{
        \tau_0
        +
        (\lambda_j\chi_S)^{-1}
    \right\}^{-1}.
    \label{eq:pc-oracle-source-limit}
\end{equation}
In particular, for a two-task model
\[
    B_S
    =
    \begin{pmatrix}
        1 & \varrho\\
        \varrho & 1
    \end{pmatrix},
    \quad
    0<|\varrho|<1,
\]
we have
$\chi_S=1-\varrho^2$, therefore complete knowledge of the source function
leaves
\begin{equation}
    v_{0j}^{\mathrm{oracle\ source}}
    =
    \left\{
        \tau_0
        +
        \bigl[
            (1-\varrho^2)\lambda_j
        \bigr]^{-1}
    \right\}^{-1}.
    \label{eq:pc-innovation-floor}
\end{equation}
Hence, with fixed target precision $\tau_0$, arbitrarily precise source
observations alone cannot force the target posterior to contract to a
point.
By contrast, suppose the task covariance is rank one,
\begin{equation}
    B_S
    =
    \mathbf g_{\mathcal I(S)}
    \mathbf g_{\mathcal I(S)}^\top,
    \quad
    g_0=1.
    \label{eq:pc-rank-one-task-kernel}
\end{equation}
Equivalently,
\[
    \theta_{kj}=g_k\theta_j,
    \quad
    \theta_j\sim\mathcal N(0,\lambda_j).
\]
Then
\begin{equation}
    \operatorname{Var}
    \left(
        \theta_{0j}\mid\mathbf Z_j
    \right)
    =
    \left(
        \lambda_j^{-1}
        +
        \mathfrak n_S
    \right)^{-1},
    \quad
    \mathfrak n_S
    =
    \sum_{k\in\mathcal I(S)}
    \tau_k g_k^2.
    \label{eq:pc-pooled-information}
\end{equation}
Thus $\mathfrak n_S$ is an additive pooled precision in the rank-one
experiment. It should not be interpreted as an effective sample size
for the original trajectory data unless such an identification is
established separately.
\end{proposition}

\noindent The proof is provided in Appendix~\ref{app:proof:prop:pc-information}.

Proposition~\ref{prop:pc-information} distinguishes two fundamentally
different transfer regimes. Under a full-rank task covariance,
conditioning on arbitrarily precise source observations leaves a
positive task-specific innovation variance
$\lambda_j\chi_S$ for the target. Source observations can reduce
uncertainty, but they cannot replace increasing target information.
Under the rank-one shared-function model, this innovation component
vanishes and the taskwise precisions combine additively through
$\mathfrak n_S$.
The quantity $\mathfrak n_S$ is therefore justified as a pooled
precision only within the stated shared-function experiment. An
identification such as
\(
    \tau_k
    \asymp
    \frac{n_{k,\mathrm{eff}}}{\sigma_k^2}
\)
for dependent reinforcement-learning trajectories requires a separate
concentration or experiment-reduction argument. It does not follow
from source selection, task-kernel correlation, or the raw observation
counts alone.

\subsection{Spectral regularity, RKHS geometry and posterior perturbation}
\label{subsec:pc-regularity-perturbation}

Posterior contraction requires two distinct forms of control. The first
concerns approximation of the target truth by the GP prior and the
amount of prior mass assigned to small neighborhoods of that truth.
The second concerns the perturbation introduced when oracle responses
are replaced by Bellman-generated responses. We develop these two
ingredients in a common notation and keep them separate from the
source-selection argument of the preceding section.
Let
\[
    T_{\mathrm{in}}g(x)
    =
    \int
    k_{\mathrm{in}}(x,x')
    g(x')
    P_X^{(0)}(\mathrm dx')
\]
denote the covariance operator induced by the input kernel on
\(
    \mathbb H=L^2(P_X^{(0)}).
\)
Let
$\{(\lambda_j,\varphi_j)\}_{j\geq1}$
be its eigenpairs, with
$\{\varphi_j\}_{j\geq1}$
orthonormal in $\mathbb H$. Assume
\begin{equation}
    c_\lambda j^{-1-2\alpha/d}
    \leq
    \lambda_j
    \leq
    C_\lambda j^{-1-2\alpha/d},
    \quad
    \alpha>0,
    \label{eq:pc-eigenvalue-decay}
\end{equation}
and let
\(
    \theta_{0j}
    =
    \langle f_0,\varphi_j\rangle_{\mathbb H},
    ~
    f_0=Q_*^{(0)}.
\)
We impose the spectral smoothness condition
\begin{equation}
    \sum_{j\geq1}
    j^{2\alpha/d}
    \theta_{0j}^2
    \leq
    L^2.
    \label{eq:pc-smoothness}
\end{equation}
The corresponding spectral smoothness class is denoted by
\[
    \Theta_\alpha(L)
    :=
    \left\{
        f\in\mathbb H:
        \sum_{j\geq1}
        j^{2\alpha/d}
        \langle f,\varphi_j\rangle_{\mathbb H}^2
        \leq L^2
    \right\}.
\]
Let $\mathcal H_{\mathrm{in}}$ denote the RKHS associated with
$k_{\mathrm{in}}$. In the above spectral representation,
\(
    \|h\|_{\mathcal H_{\mathrm{in}}}^2
    =
    \sum_{j\geq1}
    \frac{
        \langle h,\varphi_j\rangle_{\mathbb H}^2
    }{
        \lambda_j
    }
\)
whenever the series is finite.
Condition~\eqref{eq:pc-smoothness} describes regularity of the truth
relative to the eigenbasis of the covariance operator. It should not
be identified automatically with membership in the native RKHS.
For example, under the standard Mat\'ern parameterization on a regular
$d$-dimensional domain, Mat\'ern smoothness $\alpha$ corresponds to a
native Sobolev order $\alpha+d/2$ rather than $\alpha$
\citep{kanagawa2018gaussian}.

\begin{lemma}[Separable multi-task RKHS geometry]
\label{lem:pc-vector-rkhs}

Fix $S\in\mathcal S_J$, let
\(
    \mathcal I(S)=\{0\}\cup S,
\)
and define
\(
    B_S
    =
    K_{\mathrm{task}}
    \bigl(
        \mathcal I(S),\mathcal I(S)
    \bigr).
\)
Consider the separable kernel
\[
    K_S\bigl((k,x),(\ell,x')\bigr)
    =
    (B_S)_{k\ell}
    k_{\mathrm{in}}(x,x'),
    \quad
    k,\ell\in\mathcal I(S).
\]
If $B_S\succ0$, the associated vector-valued RKHS
$\mathcal H_{B_S}$ consists of vectors
\(
    \mathbf f_S
    =
    \bigl(
        f^{(k)}
    \bigr)_{k\in\mathcal I(S)}
\)
with components in $\mathcal H_{\mathrm{in}}$, and its squared norm is
\begin{equation}
    \|\mathbf f_S\|_{\mathcal H_{B_S}}^2
    =
    \sum_{k,\ell\in\mathcal I(S)}
    (B_S^{-1})_{k\ell}
    \left\langle
        f^{(k)},f^{(\ell)}
    \right\rangle_{\mathcal H_{\mathrm{in}}}.
    \label{eq:pc-vector-norm}
\end{equation}
If, for deterministic constants
$0<b_-\leq b_+<\infty$,
\(
    b_-I
    \preceq
    B_S
    \preceq
    b_+I,
\)
then
\begin{equation}
    b_+^{-1}
    \sum_{k\in\mathcal I(S)}
    \|f^{(k)}\|_{\mathcal H_{\mathrm{in}}}^2
    \leq
    \|\mathbf f_S\|_{\mathcal H_{B_S}}^2
    \leq
    b_-^{-1}
    \sum_{k\in\mathcal I(S)}
    \|f^{(k)}\|_{\mathcal H_{\mathrm{in}}}^2.
    \label{eq:pc-vector-norm-equivalence}
\end{equation}
If $B_S$ is positive semidefinite but singular, the same expression
holds with $B_S^\dagger$ in place of $B_S^{-1}$, provided
$\mathbf f_S$ belongs to the range of $B_S$ in the task coordinate.
\end{lemma}

\noindent The proof is provided in Appendix~\ref{app:proof:lem:pc-vector-rkhs}.
The preceding lemma describes the geometry induced by the task
covariance. Norm equivalence controls the cost of approximating a
vector of task functions, but it does not imply that source
observations act as independent target observations. In particular,
the full-rank and rank-one information structures identified in
Proposition~\ref{prop:pc-information} remain fundamentally different.

\begin{lemma}[Spectral approximation and GP prior concentration]
\label{lem:pc-spectral}

Suppose
\eqref{eq:pc-eigenvalue-decay} and
\eqref{eq:pc-smoothness} hold. Then there exists $C<\infty$ such that,
for all sufficiently large $R$,
\begin{equation}
    \inf_{
        h\in\mathcal H_{\mathrm{in}}:
        \|h\|_{\mathcal H_{\mathrm{in}}}\leq R
    }
    \|h-f_0\|_{\mathbb H}
    \leq
    C R^{-2\alpha/d}.
    \label{eq:pc-approximation}
\end{equation}
Let
\(
    W
    =
    \sum_{j\geq1}
    \sqrt{\lambda_j}\,
    Z_j\varphi_j,
    ~
    Z_j
    \stackrel{\mathrm{iid}}{\sim}
    \mathcal N(0,1),
\)
be the centered $\mathbb H$-valued GP induced by the input covariance
operator. Then there exist constants
$C_1,C_2<\infty$ such that, for all sufficiently small
$\varepsilon>0$,
\begin{align}
    -\log
    \mathbb P
    \left(
        \|W\|_{\mathbb H}\leq\varepsilon
    \right)
    &\leq
    C_1\varepsilon^{-d/\alpha},
    \label{eq:pc-centered-small-ball}\\
    -\log
    \mathbb P
    \left(
        \|W-f_0\|_{\mathbb H}\leq\varepsilon
    \right)
    &\leq
    C_2\varepsilon^{-d/\alpha}.
    \label{eq:pc-shifted-small-ball}
\end{align}
The constants may depend on
$L,\alpha,d,c_\lambda$, and $C_\lambda$.
\end{lemma}

\noindent The proof is provided in Appendix~\ref{app:proof:lem:pc-spectral}.
Lemma~\ref{lem:pc-spectral} provides approximation and prior-mass
bounds for the target component. These are ingredients for posterior
contraction, but they do not by themselves establish contraction under
Bellman-generated responses. We next quantify the perturbation induced
by replacing the analytical oracle responses with the actual responses
used by the learning algorithm.

Fix $S\in\mathcal S_J$ and stack the active-task responses as
\(
    \mathbf Y_{e,S}
    =
    \bigl(
        \mathbf Y_e^{(k)}
    \bigr)_{k\in\mathcal I(S)},
    ~
    \overline{\mathbf Y}_{S}
    =
    \bigl(
        \overline{\mathbf Y}^{(k)}
    \bigr)_{k\in\mathcal I(S)},
\)
and
\(
    \mathbf b_{e,S}
    =
    \bigl(
        \mathbf b_e^{(k)}
    \bigr)_{k\in\mathcal I(S)}.
\)
Then
\[
    \mathbf Y_{e,S}
    =
    \overline{\mathbf Y}_{S}
    +
    \mathbf b_{e,S}.
\]
Let $\mathbf K_{S,N}$ denote the GP Gram matrix associated with the
active task-tagged inputs and let
$\Lambda_{S,N}\succ0$
be the corresponding working-noise covariance. Throughout this
comparison, the kernel, prior mean, and working-noise specification
are held fixed. In particular, no response-dependent hyperparameter
refitting is performed between the oracle and actual response fits.
For a target query $x$, let
$\mathbf k_{0,S,N}(x)$ denote the covariance vector between $(0,x)$
and the active task-tagged inputs. Define the target smoothing operator
\(
    \mathcal P_{S,N}:
    \mathbb R^{n(S)}
    \longrightarrow
    \mathbb H
\)
by
\begin{equation}
    (\mathcal P_{S,N}v)(x)
    =
    \mathbf k_{0,S,N}(x)^\top
    (\mathbf K_{S,N}+\Lambda_{S,N})^{-1}v,
    \label{eq:pc-smoothing-operator}
\end{equation}
where
\(
    n(S)
    =
    \sum_{k\in\mathcal I(S)}n_k.
\)
Let $\mathbf m_{S,N}$ be the stacked prior-mean vector at the active
inputs. The subset-specific target marginal working posterior introduced
in Section~\ref{subsec:pc-setup} is
\[
    \Pi_{S,e,N}^{(0)}
    =
    \mathcal N_{\mathbb H}
    \left(
        \mu_{S,e,N}^{(0)},
        C_{S,N}^{(0)}
    \right),
\]
where
\begin{equation}
    \mu_{S,e,N}^{(0)}
    =
    m^{(0)}
    +
    \mathcal P_{S,N}
    \left(
        \mathbf Y_{e,S}
        -
        \mathbf m_{S,N}
    \right),
    \label{eq:pc-posterior-mean}
\end{equation}
and
\begin{equation}\nonumber
\begin{aligned}
    C_{S,N}^{(0)}(x,x')
    =
    K\bigl((0,x),(0,x')\bigr)
    -
    \mathbf k_{0,S,N}(x)^\top
    (\mathbf K_{S,N}+\Lambda_{S,N})^{-1}
    \mathbf k_{0,S,N}(x').
\end{aligned}
\label{eq:pc-posterior-covariance}
\end{equation}
Assume that $C_{S,N}^{(0)}$ defines a positive trace-class covariance
operator on $\mathbb H$.
Define the analytical oracle posterior mean by
\begin{equation}
    \overline\mu_{S,N}^{(0)}
    =
    m^{(0)}
    +
    \mathcal P_{S,N}
    \left(
        \overline{\mathbf Y}_{S}
        -
        \mathbf m_{S,N}
    \right).
    \label{eq:pc-oracle-posterior-mean}
\end{equation}
The actual and oracle fits use the same covariance operator
$C_{S,N}^{(0)}$.

\begin{lemma}[Pathwise target-posterior perturbation by Bellman errors]
\label{lem:pc-perturbation}

For every $S\in\mathcal S_J$,
\begin{equation}
    \mu_{S,e,N}^{(0)}
    -
    \overline\mu_{S,N}^{(0)}
    =
    \mathcal P_{S,N}\mathbf b_{e,S}.
    \label{eq:pc-shift}
\end{equation}
Consequently,
\begin{equation}
    \|\mu_{S,e,N}^{(0)}-f_0\|_{\mathbb H}
    \leq
    \|\overline\mu_{S,N}^{(0)}-f_0\|_{\mathbb H}
    +
    d_{S,e,N}^{\mathrm{bt}},
    \label{eq:pc-shift-bound}
\end{equation}
where
\(
    d_{S,e,N}^{\mathrm{bt}}
    =
    \|
        \mathcal P_{S,N}
        \mathbf b_{e,S}
    \|_{\mathbb H}.
\)
Moreover,
\begin{equation}
    d_{S,e,N}^{\mathrm{bt}}
    \leq
    \|
        \mathcal P_{S,N}
        \Lambda_{S,N}^{1/2}
    \|_{\mathrm{op}}
    \|
        \Lambda_{S,N}^{-1/2}
        \mathbf b_{e,S}
    \|_2.
    \label{eq:pc-bootstrap-operator-bound}
\end{equation}
\end{lemma}

\noindent The proof is provided in Appendix~\ref{app:proof:lem:pc-perturbation}.

\begin{corollary}[Deterministic-rate control of the posterior perturbation]
\label{cor:pc-bootstrap-perturbation-rate}

Let $\mathcal G_N^{\mathrm{ctr}}$ be the deterministic contraction
family introduced in Section~\ref{subsec:pc-setup}. Suppose there exist
deterministic sequences
$\underline\nu_N>0$ and
$L_{\mathcal P,N}<\infty$ such that, almost surely,
\begin{equation}
    \Lambda_{S,N}
    \succeq
    \underline\nu_N I,
    \quad
    \|
        \mathcal P_{S,N}
        \Lambda_{S,N}^{1/2}
    \|_{\mathrm{op}}
    \leq
    L_{\mathcal P,N}
    \label{eq:pc-operator-envelope}
\end{equation}
for every
$S\in\mathcal G_N^{\mathrm{ctr}}$.
Let $\beta_{e,N}$ denote the deterministic Bellman-bootstrap scale
obtained in Corollary~\ref{cor:bootstrap_stability}, so that
\begin{equation}\nonumber
    \sup_{k,t}
    \mathbb E_\star
    \left[
        (b_{t,e}^{(k)})^2
    \right]
    \leq
    \beta_{e,N}^2.
    \label{eq:pc-bootstrap-coordinate-bound}
\end{equation}
Then, for every
$S\in\mathcal G_N^{\mathrm{ctr}}$,
\begin{equation}
    \mathbb E_\star
    \left[
        \bigl(
            d_{S,e,N}^{\mathrm{bt}}
        \bigr)^2
    \right]
    \leq
    L_{\mathcal P,N}^2
    \underline\nu_N^{-1}
    n(S)
    \beta_{e,N}^2.
    \label{eq:pc-bootstrap-moment}
\end{equation}
Moreover, for every deterministic $u_N>0$,
\begin{equation}
\begin{aligned}
    \mathbb P_\star
    \left(
        \max_{S\in\mathcal G_N^{\mathrm{ctr}}}
        d_{S,e,N}^{\mathrm{bt}}
        >
        u_N
    \right)
    \leq
    \frac{
        L_{\mathcal P,N}^2
        \beta_{e,N}^2
    }{
        \underline\nu_N u_N^2
    }
    \sum_{S\in\mathcal G_N^{\mathrm{ctr}}}
    n(S).
\end{aligned}
\label{eq:pc-bootstrap-uniform-tail}
\end{equation}
\end{corollary}

\noindent The proof is provided in Appendix~\ref{app:proof:cor:pc-bootstrap-perturbation-rate}.

\begin{lemma}[Gaussian concentration for the target working posterior]
\label{lem:pc-gaussian-tail}

Let
\(
    G
    \sim
    \mathcal N_{\mathbb H}(0,C),
\)
where $C$ is a positive trace-class covariance operator on
$\mathbb H$. Then, for every $t>0$,
\begin{equation}
    \mathbb P
    \left\{
        \|G\|_{\mathbb H}
        >
        \sqrt{\operatorname{tr}(C)}
        +
        \sqrt{
            2\|C\|_{\mathrm{op}}t
        }
    \right\}
    \leq
    e^{-t}.
    \label{eq:pc-gaussian-tail}
\end{equation}
Consequently, if
\(
    \Pi
    =
    \mathcal N_{\mathbb H}(\mu,C),
\)
then, for every $f_0\in\mathbb H$ and $t>0$,
\begin{equation}
\begin{aligned}
    \Pi
    \Bigl(
        \|f-f_0\|_{\mathbb H}
        >
        \|\mu-f_0\|_{\mathbb H}
        +
        \sqrt{\operatorname{tr}(C)}
        +
        \sqrt{
            2\|C\|_{\mathrm{op}}t
        }
    \Bigr)
    \leq
    e^{-t}.
\end{aligned}
\label{eq:pc-posterior-tail}
\end{equation}
Moreover,
\begin{equation}
    \int
    \|f-f_0\|_{\mathbb H}^2
    \,\Pi(\mathrm df)
    =
    \|\mu-f_0\|_{\mathbb H}^2
    +
    \operatorname{tr}(C).
    \label{eq:pc-posterior-moment}
\end{equation}
\end{lemma}

\noindent The proof is provided in Appendix~\ref{app:proof:lem:pc-gaussian-tail}.
The results above isolate the ingredients required by the contraction
analysis. Lemma~\ref{lem:pc-spectral} controls approximation and prior
mass around the target truth. Lemma~\ref{lem:pc-perturbation} and
Corollary~\ref{cor:pc-bootstrap-perturbation-rate} quantify the
additional displacement caused by Bellman-generated responses.
Lemma~\ref{lem:pc-gaussian-tail} converts bounds on the posterior mean
and covariance into posterior concentration in the target prediction
norm. None of these results alone implies a contraction rate. Their
role is to provide verifiable inputs for the selection-adaptive
contraction theorem that follows.

\subsection{Main posterior contraction results}
\label{subsec:pc-main}

We now combine the subset-selection weights, the oracle posterior
approximation, and the Bellman perturbation bounds developed above.
The first result is a lifting theorem: it converts uniform contraction
of the subset-specific target posteriors into contraction of the
model-averaged target posterior. The subsequent results identify the
rate in two explicitly specified Gaussian sequence experiments. The
full-rank task model exhibits a target-information bottleneck, whereas
the rank-one shared-function model permits genuine precision pooling.

For deterministic sequences $\varepsilon_N>0$ and
$\mathfrak i_N>0$, define
\[
    \mathcal R_N^{\mathrm{orc}}
    =
    \max_{S\in\mathcal G_N^{\mathrm{ctr}}}
    \max
    \left\{
        \frac{
            \|\overline\mu_{S,N}^{(0)}-f_0\|_{\mathbb H}
        }{
            \varepsilon_N
        },
        \frac{
            \sqrt{
                \operatorname{tr}
                (C_{S,N}^{(0)})
            }
        }{
            \varepsilon_N
        },
        \sqrt{
            \mathfrak i_N
            \|C_{S,N}^{(0)}\|_{\mathrm{op}}
        }
    \right\}.
\]
For a fixed constant $C_0<\infty$, let
\(
    \zeta_N^{\mathrm{orc}}(C_0)
    =
    \mathbb P_\star
    \left(
        \mathcal R_N^{\mathrm{orc}}>C_0
    \right).
    \label{eq:pc-oracle-failure}
\)
This is a population failure probability associated with the
oracle-response experiment.
For $u>0$, let
\(
    \zeta_N^{\mathrm{bt}}(u)
    =
    \frac{
        L_{\mathcal P,N}^2
        \beta_{e,N}^2
    }{
        \underline\nu_N u^2
    }
    \sum_{S\in\mathcal G_N^{\mathrm{ctr}}}
    n(S),
    \label{eq:pc-bootstrap-failure}
\)
which is the deterministic upper bound supplied by
Corollary~\ref{cor:pc-bootstrap-perturbation-rate}.

\begin{theorem}[Selection-adaptive target posterior contraction]
\label{thm:posterior-contraction}

Let
\(
    \varepsilon_N\downarrow0,
~
    u_N\downarrow0
\) and
\(
    t_N
    =
    \mathfrak i_N\varepsilon_N^2
    \longrightarrow\infty.
\)
Suppose that, for some fixed $C_0<\infty$,
\(
    \zeta_N^{\mathrm{orc}}(C_0)
    +
    \zeta_N^{\mathrm{bt}}(u_N)
    \longrightarrow0.
\)
Assume further that, for the deterministic predictive tolerance
$d_N>0$ used in
Theorem~\ref{thm:predictive_selection},
\begin{equation}
    \mathcal S_J
    \setminus
    \mathcal B_N^{\mathrm{pred}}(d_N)
    \subseteq
    \mathcal G_N^{\mathrm{ctr}}
    \label{eq:pc-selection-contraction-link}
\end{equation}
for all sufficiently large $N$.
\\
Then, with $\mathbb P_\star$-probability tending to one,
\begin{equation}
\begin{aligned}
    \widetilde\Pi_{e,N}^{(0)}
    \left(
        \|f-f_0\|_{\mathbb H}
        >
        R_N
    \right)
    \leq
    \Pi_{e,N}
    \left(
        \mathcal B_N^{\mathrm{pred}}(d_N)
        \mid
        \mathbf Y_e,X
    \right)
    +
    e^{-t_N},
\end{aligned}
\label{eq:pc-mixture-contraction-bound}
\end{equation}
where
\(
    R_N
    =
    (2+\sqrt{2})C_0\varepsilon_N
    +
    u_N.
    \label{eq:pc-generic-radius}
\)
Consequently,
\begin{equation}
    \widetilde\Pi_{e,N}^{(0)}
    \left(
        \|f-f_0\|_{\mathbb H}
        >
        R_N
    \right)
    \xrightarrow{\mathbb P_\star}0.
    \label{eq:pc-mixture-contraction}
\end{equation}
Hence the model-averaged target posterior contracts at rate
$O(\varepsilon_N+u_N)$. If
$u_N=O(\varepsilon_N)$, its contraction rate is
$O(\varepsilon_N)$.
\\
Moreover, with $\mathbb P_\star$-probability tending to one,
\begin{equation}
\begin{aligned}
    \sup_{S\in\mathcal G_N^{\mathrm{ctr}}}
    \int
        \|f-f_0\|_{\mathbb H}^2
        \,
        \Pi_{S,e,N}^{(0)}(\mathrm df)
    \leq
    (C_0\varepsilon_N+u_N)^2
    +
    C_0^2\varepsilon_N^2.
\end{aligned}
\label{eq:pc-good-moment}
\end{equation}
\end{theorem}

\noindent The proof is provided in Appendix~\ref{app:proof:thm:posterior-contraction}.

\begin{theorem}[Full-rank task covariance yields a target-information rate]
\label{thm:pc-fullrank}

Fix a deterministic subset $S\in\mathcal S_J$ and let
\(
    B_S
    =
    K_{\mathrm{task}}
    \bigl(
        \mathcal I(S),\mathcal I(S)
    \bigr)
\)
be a deterministic task covariance matrix satisfying
\begin{equation}
    b_-I
    \preceq
    B_S
    \preceq
    b_+I,
    \quad
    0<b_-\leq b_+<\infty.
    \label{eq:pc-fullrank-task-bounds}
\end{equation}
Consider the Gaussian sequence experiment
\(
    \mathbf Z_j
    =
    \boldsymbol\theta_j^\star
    +
    D_S^{-1/2}\boldsymbol\xi_j
\) and
\(
    D_S
    =
    \operatorname{diag}
    \left(
        \tau_k:
        k\in\mathcal I(S)
    \right),
\)
where the precisions $\tau_k>0$ are deterministic,
\(
    \boldsymbol\xi_j
    \stackrel{\mathrm{ind}}{\sim}
    \mathcal N(0,I),
\)
and the population coefficient sequence satisfies
\begin{equation}
    \sum_{j\geq1}
    j^{2\alpha/d}
    \|
        \boldsymbol\theta_j^\star
    \|_2^2
    \leq
    L_*^2.
    \label{eq:pc-joint-smoothness}
\end{equation}
Consider the fixed working prior
\(
    \boldsymbol\theta_j
    \stackrel{\mathrm{ind}}{\sim}
    \mathcal N
    \left(
        0,
        \lambda_jB_S
    \right),
\)
 where the prior variance sequence $\left\{\lambda_j\right\}_{j \geq 1}$ satisfies condition \eqref{eq:pc-eigenvalue-decay}.
Suppose
\(
    \tau_0\longrightarrow\infty,
\)
while the source precisions may vary arbitrarily. Define
\[
    a_\alpha
    =
    \frac{\alpha}{2\alpha+d},
    \quad
    \varepsilon_{0,N}
    =
    \tau_0^{-a_\alpha},
    \quad
    t_{0,N}
    =
    \tau_0\varepsilon_{0,N}^2.
\]
Let
\(
    \overline\Pi_{S,N}^{(0)}
    =
    \mathcal N_{\mathbb H}
    \left(
        \overline\mu_{S,N}^{(0)},
        C_{S,N}^{(0)}
    \right)
\)
denote the corresponding oracle target posterior. 
\\
Then there exist
constants $0<c<C<\infty$, independent of the source precisions, such
that
\begin{equation}
    c\varepsilon_{0,N}^2
    \leq
    \operatorname{tr}
    \bigl(
        C_{S,N}^{(0)}
    \bigr)
    \leq
    C\varepsilon_{0,N}^2,
    \quad
    \|
        C_{S,N}^{(0)}
    \|_{\mathrm{op}}
    \leq
    \tau_0^{-1},
    \label{eq:pc-fullrank-covariance}
\end{equation}
and
\begin{equation}
    \mathbb E_\star
    \left[
        \|
            \overline\mu_{S,N}^{(0)}-f_0
        \|_{\mathbb H}^2
    \right]
    \leq
    C\varepsilon_{0,N}^2.
    \label{eq:pc-fullrank-mean-risk}
\end{equation}
Moreover, for some fixed $M<\infty$,
\begin{equation}
\begin{aligned}
    \mathbb P_\star
    \Biggl[
        \overline\Pi_{S,N}^{(0)}
        \left(
            \|f-f_0\|_{\mathbb H}
            >
            M\varepsilon_{0,N}
        \right)
        >
        e^{-t_{0,N}}
    \Biggr]
    \leq
    e^{-t_{0,N}}
\end{aligned}
\label{eq:pc-fullrank-contraction}
\end{equation}
for all sufficiently large $N$. Hence
$\overline\Pi_{S,N}^{(0)}$ contracts around $f_0$ at rate
$\varepsilon_{0,N}$.
\\
Conversely, for every deterministic
$r_N=o(\varepsilon_{0,N})$,
\begin{equation}
    \sup_{h\in\mathbb H}
    \mathcal N_{\mathbb H}
    \left(
        h,
        C_{S,N}^{(0)}
    \right)
    \left\{
        f:
        \|f-f_0\|_{\mathbb H}
        \leq
        r_N
    \right\}
    \longrightarrow0.
    \label{eq:pc-fullrank-lower-radius}
\end{equation}
Thus, under a uniformly positive-definite task covariance, increasing
source precision may improve constants but cannot improve the
contraction order determined by the target precision $\tau_0$.
\end{theorem}

\noindent The proof is provided in Appendix~\ref{app:proof:thm:pc-fullrank}.

\begin{corollary}[Full-rank contraction under Bellman perturbation]
\label{cor:pc-fullrank-bootstrap}

Under Theorem~\ref{thm:pc-fullrank}, let $u_N>0$ be deterministic.
If the deterministic bound from
Corollary~\ref{cor:pc-bootstrap-perturbation-rate} satisfies
\[
    \frac{
        L_{\mathcal P,N}^2
        \beta_{e,N}^2
        n(S)
    }{
        \underline\nu_N u_N^2
    }
    \longrightarrow0,
\]
then the actual target working posterior contracts around $f_0$ at
radius
\[
    O
    \left(
        \tau_0^{-\alpha/(2\alpha+d)}
        +
        u_N
    \right)
\]
in $\mathbb P_\star$-probability.
\end{corollary}

\noindent The proof is provided in Appendix~\ref{app:proof:cor:pc-fullrank-bootstrap}.

For each deterministic
$S\in\mathcal G_N^{\mathrm{ctr}}$, consider the oracle coefficient
experiment
\begin{equation}
    \overline Z_{kj}
    =
    g_k\theta_{0j}^\star
    +
    h_{kj}
    +
    \tau_k^{-1/2}\xi_{kj},
    \quad
    k\in\mathcal I(S),
    \quad
    j\geq1,
    \label{eq:pc-shared-sequence-model}
\end{equation}
where
\(
    g_0=1,
    ~
    \tau_k>0,
    ~
    \xi_{kj}
    \stackrel{\mathrm{ind}}{\sim}
    \mathcal N(0,1),
\)
and
\(
    \theta_{0j}^\star
    =
    \langle f_0,\varphi_j\rangle_{\mathbb H}.
\)
The deterministic coefficients $h_{kj}$ represent departures from
exact sharing, with $h_{0j}=0$.
The working model ignores the discrepancy terms and uses
\[
    \theta_j
    \stackrel{\mathrm{ind}}{\sim}
    \mathcal N(0,\lambda_j),
    \quad
    \overline Z_{kj}\mid\theta_j
    \sim
    \mathcal N(g_k\theta_j,\tau_k^{-1}).
\]
Equivalently, the working multi-task covariance has the rank-one form
$\mathbf g\mathbf g^\top$.
This experiment is either taken as an exact Gaussian sequence model or
obtained from the original observation scheme through a separately
validated reduction. In particular, the precisions $\tau_k$ are not
identified with trajectory sample sizes without an additional
statistical argument. Bellman perturbations are not included in
\eqref{eq:pc-shared-sequence-model}; their effect is already handled
by the perturbation analysis preceding
Theorem~\ref{thm:posterior-contraction}.
For $S\in\mathcal G_N^{\mathrm{ctr}}$, define
\begin{equation}
    \mathfrak n_S
    =
    \sum_{k\in\mathcal I(S)}
    \tau_k g_k^2
    \label{eq:pc-pooled-precision}
\end{equation}
and
\begin{equation}
    \overline Z_{S,j}
    =
    \frac{1}{\mathfrak n_S}
    \sum_{k\in\mathcal I(S)}
    \tau_k g_k\overline Z_{kj}.
    \label{eq:pc-pooled-observation}
\end{equation}
Also set
\[
    w_j(n)
    =
    \frac{n\lambda_j}{1+n\lambda_j},
    \quad
    v_j(n)
    =
    \frac{\lambda_j}{1+n\lambda_j}.
\]
The pooled discrepancy coefficients are
\(
    h_{S,j}
    =
    \frac{1}{\mathfrak n_S}
    \sum_{k\in\mathcal I(S)}
    \tau_k g_k h_{kj},
\)
and define the deterministic smoothed discrepancy
\begin{equation}
    \Delta_{S,N}^{\mathrm{sh}}
    =
    \left\{
        \sum_{j\geq1}
        w_j(\mathfrak n_S)^2
        h_{S,j}^2
    \right\}^{1/2},
    \quad
    \Delta_N^{\mathrm{sh}}
    =
    \max_{S\in\mathcal G_N^{\mathrm{ctr}}}
    \Delta_{S,N}^{\mathrm{sh}}.
    \label{eq:pc-shared-discrepancy}
\end{equation}
Because $h_{kj}$, $\tau_k$, and $g_k$ are fixed population quantities,
$\Delta_N^{\mathrm{sh}}$ is deterministic.

\begin{lemma}[Conjugate pooling and spectral bounds]
\label{lem:pc-pooling}

Under the working model,
\begin{equation}
    \theta_j
    \mid
    \overline{\mathbf Z},S
    \sim
    \mathcal N
    \left(
        w_j(\mathfrak n_S)\overline Z_{S,j},
        v_j(\mathfrak n_S)
    \right).
    \label{eq:pc-conjugate}
\end{equation}
Suppose
$f_0\in\Theta_\alpha(L)$ and
\eqref{eq:pc-eigenvalue-decay} holds. Define
\(
    a_\alpha
    =
    \frac{\alpha}{2\alpha+d}.
\)
Then there exists $C<\infty$ such that, for every $n\geq2$,
\begin{equation}
\begin{aligned}
    \sum_{j\geq1}
    \bigl(1-w_j(n)\bigr)^2
    (\theta_{0j}^\star)^2
    +
    \sum_{j\geq1}v_j(n)
    +
    \sum_{j\geq1}\frac{w_j(n)^2}{n}  
    \leq
    Cn^{-2a_\alpha},
\end{aligned}
\label{eq:pc-pooled-spectral-bounds}
\end{equation}
and
\begin{equation}
    \sup_{j\geq1}
    \max
    \left\{
        v_j(n),
        \frac{w_j(n)^2}{n}
    \right\}
    \leq
    n^{-1}.
    \label{eq:pc-pooled-operator-bounds}
\end{equation}
\end{lemma}

\noindent The proof is provided in Appendix~\ref{app:proof:lem:pc-pooling}.

\begin{theorem}[Uniform oracle contraction in the shared-function experiment]
\label{thm:pc-pooled-rate}

Suppose
\eqref{eq:pc-eigenvalue-decay},
\eqref{eq:pc-smoothness}, and
\eqref{eq:pc-shared-sequence-model}
hold. Let
$\mathcal G_N^{\mathrm{ctr}}\subseteq\mathcal S_J$
be nonempty and deterministic, and define
\(
    \underline{\mathfrak n}_N
    =
    \min_{S\in\mathcal G_N^{\mathrm{ctr}}}
    \mathfrak n_S
    \longrightarrow\infty.
\)
Set
\(
    \varepsilon_N
    =
    \underline{\mathfrak n}_N^{-a_\alpha},
~
    t_N
    =
    \underline{\mathfrak n}_N\varepsilon_N^2
\) and
\(
    \ell_N
    =
    \log
    \left(
        2|\mathcal G_N^{\mathrm{ctr}}|
    \right).
\)
Assume the deterministic conditions
\(
    \ell_N=O(t_N)
\) and
\(
    \Delta_N^{\mathrm{sh}}
    =
    O(\varepsilon_N).
\)
Then there exists a constant $C<\infty$ such that
\begin{equation}
\begin{aligned}
    \mathbb P_\star
    \Bigg(
        \max_{S\in\mathcal G_N^{\mathrm{ctr}}}
        \|
            \overline\mu_{S,N}^{(0)}-f_0
        \|_{\mathbb H}
        >
        C\varepsilon_N
    \Bigg)
    \leq
    e^{-t_N}
\end{aligned}
\label{eq:pc-shared-mean-bound}
\end{equation}
for all sufficiently large $N$, while
\begin{equation}
    \sup_{S\in\mathcal G_N^{\mathrm{ctr}}}
    \operatorname{tr}
    \bigl(
        C_{S,N}^{(0)}
    \bigr)
    \leq
    C\varepsilon_N^2,
    \quad
    \sup_{S\in\mathcal G_N^{\mathrm{ctr}}}
    \|
        C_{S,N}^{(0)}
    \|_{\mathrm{op}}
    \leq
    \underline{\mathfrak n}_N^{-1}.
    \label{eq:pc-shared-covariance-bound}
\end{equation}
Consequently, the oracle conditions of
Theorem~\ref{thm:posterior-contraction}
hold with information scale
$\mathfrak i_N=\underline{\mathfrak n}_N$
and oracle rate $\varepsilon_N$.
\end{theorem}

\noindent The proof is provided in Appendix~\ref{app:proof:thm:pc-pooled-rate}.

\begin{corollary}[Transfer-adaptive contraction in the shared-function experiment]
\label{cor:pc-pooled-transfer}

Under Theorem~\ref{thm:pc-pooled-rate}, suppose that, for the
deterministic predictive tolerance $d_N$ used in
Theorem~\ref{thm:predictive_selection},
\[
    \mathcal S_J
    \setminus
    \mathcal B_N^{\mathrm{pred}}(d_N)
    \subseteq
    \mathcal G_N^{\mathrm{ctr}}
\]
for all sufficiently large $N$.
\\
If the deterministic bootstrap bound in
Theorem~\ref{thm:posterior-contraction} satisfies
\(
    \zeta_N^{\mathrm{bt}}(u_N)\to0
\)
for some deterministic
$u_N=O(\varepsilon_N)$,
then
\[
    \widetilde\Pi_{e,N}^{(0)}
    \left(
        \|f-f_0\|_{\mathbb H}
        >
        M\varepsilon_N
    \right)
    \xrightarrow{\mathbb P_\star}0
\]
for a sufficiently large constant $M$.
Therefore,
\(
    \widetilde\Pi_{e,N}^{(0)}
\)
contracts at rate
\(
    \varepsilon_N
    =
    \underline{\mathfrak n}_N^{
        -\alpha/(2\alpha+d)
    }.
\)
\end{corollary}

\noindent The proof is provided in Appendix~\ref{app:proof:cor:pc-pooled-transfer}.

\subsection{Posterior mean risk, minimax optimality and transfer gains}
\label{subsec:pc-risk}

Posterior contraction and frequentist prediction risk are related but
distinct. Contraction in probability does not by itself imply a
second-moment risk bound for the unmodified posterior mean. We first
derive a risk bound for an explicitly projected posterior mean. We then
establish the minimax rate in the exact shared-function sequence
experiment and clarify the distinction between improvement in
contraction scale and strict frequentist risk improvement. Finally, we
give an exact prior-averaged transfer-gain identity under correct
specification.

\begin{corollary}[Risk of the projected posterior mean]
\label{cor:pc-mean-risk}

Suppose the conditions of
Theorem~\ref{thm:posterior-contraction} hold and
\(
    \|f_0\|_{\mathbb H}\leq L
\)
for some fixed $L<\infty$. Let
\(
    \mathbb B_L
    =
    \left\{
        f\in\mathbb H:
        \|f\|_{\mathbb H}\leq L
    \right\},
\)
and let
$T_L:\mathbb H\to\mathbb B_L$
be the metric projection onto $\mathbb B_L$.
Define the projected model-averaged posterior by
\[
    \widetilde\Pi_{e,N}^{(0),L}
    =
    \widetilde\Pi_{e,N}^{(0)}
    \circ
    T_L^{-1}
\]
and its posterior mean by
\[
    \widehat f_{e,N}^{L}
    =
    \int_{\mathbb H}
    f\,
    \widetilde\Pi_{e,N}^{(0),L}(\mathrm df).
\]
Let
\(
    r_N=\varepsilon_N+u_N,
\)
and define the deterministic population error bounds
\[
    \delta_N^{\mathrm{ctr}}
    =
    \zeta_N^{\mathrm{orc}}(C_0)
    +
    \zeta_N^{\mathrm{bt}}(u_N)
\]
and
\[
    \delta_N^{\mathrm{sel}}
    =
    \mathbb E_\star
    \left[
        \Pi_{e,N}
        \left(
            \mathcal B_N^{\mathrm{pred}}(d_N)
            \mid
            \mathbf Y_e,X
        \right)
    \right].
\]
Then, for some constant $C<\infty$ independent of $N$,
\begin{equation}
    \mathbb E_\star
    \left[
        \|
            \widehat f_{e,N}^{L}-f_0
        \|_{\mathbb H}^2
    \right]
    \leq
    Cr_N^2
    +
    4L^2
    \left(
        \delta_N^{\mathrm{ctr}}
        +
        \delta_N^{\mathrm{sel}}
    \right).
    \label{eq:pc-projected-risk}
\end{equation}
Consequently, if
\(
    r_N\to0,
    ~
    \delta_N^{\mathrm{ctr}}\to0,
    ~
    \delta_N^{\mathrm{sel}}\to0,
\)
then
\[
    \mathbb E_\star
    \left[
        \|
            \widehat f_{e,N}^{L}-f_0
        \|_{\mathbb H}^2
    \right]
    \longrightarrow0.
\]
If, in addition,
\(
    \delta_N^{\mathrm{ctr}}
    +
    \delta_N^{\mathrm{sel}}
    =
    O(r_N^2),
\)
then
\[
    \mathbb E_\star
    \left[
        \|
            \widehat f_{e,N}^{L}-f_0
        \|_{\mathbb H}^2
    \right]
    =
    O(r_N^2).
\]
\end{corollary}

\noindent The proof is provided in Appendix~\ref{app:proof:cor:pc-mean-risk}.
The projection in
Corollary~\ref{cor:pc-mean-risk}
is an explicit modification of the posterior mean. No corresponding
risk rate is asserted here for the unprojected posterior mean.
Such a result requires additional uniform-integrability or posterior
moment control.

\begin{theorem}[Minimax optimality in the exact shared-function experiment]
\label{thm:pc-minimax}

Fix a deterministic subset $S$ and consider the exact shared-function
sequence experiment
\eqref{eq:pc-shared-sequence-model}
with
\[
    h_{kj}=0,
    \quad
    k\in\mathcal I(S),
    \quad
    j\geq1.
\]
Let
\[
    n
    =
    \mathfrak n_S
    =
    \sum_{k\in\mathcal I(S)}
    \tau_k g_k^2
    \longrightarrow\infty,
\]
and suppose
\eqref{eq:pc-eigenvalue-decay}
holds. Then, for every fixed $L>0$, there exist constants
$0<c<C<\infty$ such that
\begin{equation}
    c
    n^{-2\alpha/(2\alpha+d)}
    \leq
    \inf_{\widehat f}
    \sup_{f_0\in\Theta_\alpha(L)}
    \mathbb E_{f_0}
    \left[
        \|
            \widehat f-f_0
        \|_{\mathbb H}^2
    \right]
    \leq
    C
    n^{-2\alpha/(2\alpha+d)}.
    \label{eq:pc-minimax}
\end{equation}
Here, the infimum is taken over all measurable estimators based on the observations in the sequence experiment.
The posterior mean associated with
\eqref{eq:pc-conjugate}
attains the upper bound up to a constant factor.
\end{theorem}

\noindent The proof is provided in Appendix~\ref{app:proof:thm:pc-minimax}.

Theorem~\ref{thm:pc-minimax} is a minimax result for the stated
Gaussian shared-function experiment. It does not imply the same lower
bound for an unrestricted heterogeneous reinforcement-learning model
without an explicit reduction to this experiment. Related conjugate
Gaussian sequence analyses are discussed by
\citet{knapik2011bayesian}.

Let $\tau_0$ denote the target-only precision in the same Gaussian
sequence experiment. Since the contraction radius has order
$n^{-a_\alpha}$, where
\(
    a_\alpha
    =
    \frac{\alpha}{2\alpha+d},
\)
the nominal ratio between the pooled and target-only contraction
scales is
\begin{equation}\nonumber
    \frac{
        \mathfrak n_S^{-a_\alpha}
    }{
        \tau_0^{-a_\alpha}
    }
    =
    \left(
        \frac{
            \mathfrak n_S
        }{
            \tau_0
        }
    \right)^{-a_\alpha}.
    \label{eq:pc-rate-ratio}
\end{equation}
Thus, if
\(
    \frac{\mathfrak n_S}{\tau_0}
    \longrightarrow
    \rho
    \in
    (1,\infty),
\)
the limiting ratio is $\rho^{-a_\alpha}<1$, corresponding to a
constant-factor improvement in the nominal contraction scale.
If
\(
    \frac{\mathfrak n_S}{\tau_0}
    \longrightarrow\infty,
\)
the ratio tends to zero.
These comparisons concern contraction and minimax scales. They do not
by themselves imply strict frequentist risk improvement for a fixed target truth.
In the scalar sequence experiment
\eqref{eq:pc-equivalent-scalar-sequence}, the posterior mean at
precision $n$ is
\[
    \widehat\theta_{j,n}
    =
    w_j(n)
    \left(
        \theta_{0j}^\star
        +
        n^{-1/2}\xi_j
    \right).
\]
Its exact frequentist risk at a fixed
$f_0$ is therefore
\begin{equation}
\begin{aligned}
    \mathcal R_n(f_0)
    :=
    \mathbb E_{f_0}
    \left[
        \|
            \widehat f_n-f_0
        \|_{\mathbb H}^2
    \right]
    =
    \sum_{j\geq1}
    \left[
        \frac{
            (\theta_{0j}^\star)^2
        }{
            (1+n\lambda_j)^2
        }
        +
        \frac{
            n\lambda_j^2
        }{
            (1+n\lambda_j)^2
        }
    \right].
    \label{eq:pc-exact-fixed-risk}
\end{aligned}
\end{equation}
Hence, for deterministic precisions $n_1>n_0$, strict frequentist
improvement at the fixed truth $f_0$ occurs precisely when
\(
    \mathcal R_{n_1}(f_0)
    <
    \mathcal R_{n_0}(f_0).
\)
This truth-specific comparison cannot be inferred from the minimax
rate alone.

\begin{proposition}[Exact Bayesian transfer gain from additional information]
\label{prop:pc-bayes-gain}

Let $F$ be an $\mathbb H$-valued random element satisfying
\(
    \mathbb E
    \|F\|_{\mathbb H}^2
    <
    \infty,
\)
and let
\(
    \mathscr D_0
    \subseteq
    \mathscr D_{0S}
\)
be nested information sigma-fields. Define
\[
    m_0
    =
    \mathbb E
    \left[
        F
        \mid
        \mathscr D_0
    \right],
    \quad
    m_{0S}
    =
    \mathbb E
    \left[
        F
        \mid
        \mathscr D_{0S}
    \right].
\]
Then
\begin{equation}
\begin{aligned}
    \mathbb E
    \left[
        \|F-m_0\|_{\mathbb H}^2
    \right]
    -
    \mathbb E
    \left[
        \|F-m_{0S}\|_{\mathbb H}^2
    \right]
    =
    \mathbb E
    \left[
        \|m_{0S}-m_0\|_{\mathbb H}^2
    \right]
    \geq
    0.
\end{aligned}
\label{eq:pc-bayes-identity}
\end{equation}
The inequality is strict if and only if
\[
    \mathbb P
    \left(
        m_{0S}\neq m_0
    \right)
    >
    0.
\]
If the two conditional laws are Gaussian with positive trace-class
conditional covariance operators $C_0$ and $C_{0S}$, then
\begin{equation}
\begin{aligned}
    \mathbb E
    \left[
        \|F-m_0\|_{\mathbb H}^2
    \right]
    -
    \mathbb E
    \left[
        \|F-m_{0S}\|_{\mathbb H}^2
    \right]
    =
    \mathbb E
    \left[
        \operatorname{tr}
        \left(
            C_0-C_{0S}
        \right)
    \right].
\end{aligned}
\label{eq:pc-bayes-covariance-gain}
\end{equation}
\end{proposition}

\noindent The proof is provided in Appendix~\ref{app:proof:prop:pc-bayes-gain}.

Proposition~\ref{prop:pc-bayes-gain} is a prior-averaged statement
under correct specification. It does not imply pointwise frequentist
risk dominance for a fixed target truth. Such a conclusion requires a
truth-specific comparison such as
\eqref{eq:pc-exact-fixed-risk}, rather than a minimax rate alone.

\subsection{Offline policy performance under coverage and pessimism}
\label{subsec:pc-offline}

We return to the target MDP and retain the notation
\(
    Q_*^{(0)}, 
    V_*^{(0)},
    \mathcal C_\beta^{(0)}(s)
\)
from the preceding sections. Recall that
\(
    f_0:=Q_*^{(0)}
\)
is viewed as a function on the common input space
$\mathcal X$, with
\(
    \mathbb H=L^2(P_X^{(0)}).
\)
For a state--action pair $(s,a)$, write
\(
    x=\phi^{(0)}(s,a).
\)
Let
$\widetilde\mu_{e,N}^{(0)}$
and
$\widetilde\sigma_{e,N}^{(0)}$
denote, respectively, the mean and marginal standard deviation of the
model-averaged target working posterior
$\widetilde\Pi_{e,N}^{(0)}$.
Define the pessimism penalty
\begin{equation}\nonumber
    \mathfrak p_{e,N}^{(0)}(x)
    =
    \alpha_{\mathrm{pess}}
    \widetilde\sigma_{e,N}^{(0)}(x),
    \quad
    \alpha_{\mathrm{pess}}\geq0.
    \label{eq:pc-pessimism-penalty}
\end{equation}
The corresponding supported pessimistic policy is
\begin{equation}
\begin{aligned}
    \widehat\pi_{e,N}(s)
    =
    \arg\max_{a\in\mathcal C_\beta^{(0)}(s)}
    \Bigl\{
        \widetilde\mu_{e,N}^{(0)}
        \bigl(
            \phi^{(0)}(s,a)
        \bigr)
        -
        \mathfrak p_{e,N}^{(0)}
        \bigl(
            \phi^{(0)}(s,a)
        \bigr)
    \Bigr\},
\end{aligned}
\label{eq:pc-pessimistic-policy}
\end{equation}
where ties are resolved by a fixed measurable rule.
The penalty in
\eqref{eq:pc-pessimistic-policy}
is an algorithmic pessimism adjustment. Interpreting it as a valid
lower-confidence correction requires an additional calibration result
and is not needed for the deterministic performance bounds below.
Define the best population action available within the supported
action set by
\begin{equation}\nonumber
    a_{\beta,*}^{(0)}(s)
    =
    \arg\max_{a\in\mathcal C_\beta^{(0)}(s)}
    Q_*^{(0)}(s,a),
    \label{eq:pc-supported-optimal-action}
\end{equation}
with the same convention for measurable tie-breaking. The associated
coverage gap is
\begin{equation}
    c_\beta^{(0)}(s)
    =
    V_*^{(0)}(s)
    -
    Q_*^{(0)}
    \bigl(
        s,a_{\beta,*}^{(0)}(s)
    \bigr).
    \label{eq:pc-coverage-gap}
\end{equation}
Since
$\mathcal C_\beta^{(0)}(s)\subseteq\mathcal A^{(0)}$,
\(
    c_\beta^{(0)}(s)\geq0.
\)
Define the target posterior-mean error on the common input space by
\begin{equation}\nonumber
    e_{e,N}^{(0)}(x)
    =
    \widetilde\mu_{e,N}^{(0)}(x)-f_0(x).
    \label{eq:pc-target-mean-error}
\end{equation}
For the uniform bound below, let
\[
    \|e_{e,N}^{(0)}\|_{\infty,\beta}
    =
    \sup_s
    \sup_{a\in\mathcal C_\beta^{(0)}(s)}
    \left|
        e_{e,N}^{(0)}
        \bigl(
            \phi^{(0)}(s,a)
        \bigr)
    \right|
\]
and
\[
    \|\mathfrak p_{e,N}^{(0)}\|_{\infty,\beta}
    =
    \sup_s
    \sup_{a\in\mathcal C_\beta^{(0)}(s)}
    \mathfrak p_{e,N}^{(0)}
    \bigl(
        \phi^{(0)}(s,a)
    \bigr).
\]
For an initial-state distribution $\rho_0$ and a measurable target
policy $\pi$, define its normalized discounted state-occupancy measure
by
\begin{equation}\nonumber
    d_{\rho_0}^{\pi}
    =
    (1-\gamma)
    \sum_{t=0}^{\infty}
    \gamma^t
    \mathcal L_{\rho_0,\pi}(S_t).
    \label{eq:pc-discounted-occupancy}
\end{equation}
Let
\[
    \Pi_\beta^{(0)}
    =
    \left\{
        \pi:
        \pi(s)\in\mathcal C_\beta^{(0)}(s)
        \ \text{for every }s
    \right\}
\]
denote the deterministic supported-policy class.
For $\pi\in\Pi_\beta^{(0)}$, define the two probability measures on
$\mathcal X$
\begin{align}\nonumber
    \nu_\pi^\pi
    =
    \mathcal L
    \left(
        \phi^{(0)}
        \bigl(
            S,\pi(S)
        \bigr)
    \right),
    \quad
    S\sim d_{\rho_0}^{\pi}, \quad
    \nu_\beta^\pi
    =
    \mathcal L
    \left(
        \phi^{(0)}
        \bigl(
            S,a_{\beta,*}^{(0)}(S)
        \bigr)
    \right),
    \quad
    S\sim d_{\rho_0}^{\pi}.
    \label{eq:pc-supported-input-law}
\end{align}
Define the population concentrability coefficients
\begin{equation}\nonumber
\begin{aligned}
    C_{\pi,\beta}^{(0)}
    =
    \sup_{\pi\in\Pi_\beta^{(0)}}
    \left\|
        \frac{
            \mathrm d\nu_\pi^\pi
        }{
            \mathrm dP_X^{(0)}
        }
    \right\|_\infty, \quad
    C_\beta^{(0)}
    =
    \sup_{\pi\in\Pi_\beta^{(0)}}
    \left\|
        \frac{
            \mathrm d\nu_\beta^\pi
        }{
            \mathrm dP_X^{(0)}
        }
    \right\|_\infty,
\end{aligned}
\label{eq:pc-concentrability}
\end{equation}
whenever the corresponding absolute-continuity relations hold.
These are population quantities associated with the target MDP,
supported action sets, representation, and evaluation distribution.

\begin{theorem}[Offline policy loss under coverage and pessimism]
\label{thm:offline-safe-transfer-performance}

Suppose
\(
    C_{\pi,\beta}^{(0)}<\infty
    \)
    and
    \(
    C_\beta^{(0)}<\infty.
\)
Then the policy
$\widehat\pi_{e,N}$
defined in
\eqref{eq:pc-pessimistic-policy}
satisfies, pathwise, for every state $s$,
\begin{equation}
\begin{aligned}
    &
    V_*^{(0)}(s)
    -
    Q_*^{(0)}
    \bigl(
        s,\widehat\pi_{e,N}(s)
    \bigr)\\
    &\quad\leq
    c_\beta^{(0)}(s)
    +
    \left|
        e_{e,N}^{(0)}
        \bigl(
            \phi^{(0)}
            (
                s,a_{\beta,*}^{(0)}(s)
            )
        \bigr)
    \right|
    +
    \left|
        e_{e,N}^{(0)}
        \bigl(
            \phi^{(0)}
            (
                s,\widehat\pi_{e,N}(s)
            )
        \bigr)
    \right|
    +
    \mathfrak p_{e,N}^{(0)}
    \bigl(
        \phi^{(0)}
        (
            s,a_{\beta,*}^{(0)}(s)
        )
    \bigr).
\end{aligned}
\label{eq:pc-greedy-gap}
\end{equation}
Consequently,
\begin{equation}
    \|
        V_*^{(0)}
        -
        V^{\widehat\pi_{e,N}}
    \|_\infty
    \leq
    \frac{
        \|c_\beta^{(0)}\|_\infty
        +
        2\|e_{e,N}^{(0)}\|_{\infty,\beta}
        +
        \|\mathfrak p_{e,N}^{(0)}\|_{\infty,\beta}
    }{
        1-\gamma
    }.
    \label{eq:pc-offline-sup}
\end{equation}
Moreover,
\begin{equation}
\begin{aligned}
    \mathbb E_{\rho_0}
    \left[
        V_*^{(0)}(S_0)
        -
        V^{\widehat\pi_{e,N}}(S_0)
    \right]
    \leq
    \frac{1}{1-\gamma}
    \Bigg[
        \mathbb E_{
            d_{\rho_0}^{\widehat\pi_{e,N}}
        }
        c_\beta^{(0)}(S)
        +
        \left(
            \sqrt{C_\beta^{(0)}}
            +
            \sqrt{C_{\pi,\beta}^{(0)}}
        \right)
        \|e_{e,N}^{(0)}\|_{\mathbb H}
        +
        \sqrt{C_\beta^{(0)}}
        \|
            \mathfrak p_{e,N}^{(0)}
        \|_{\mathbb H}
    \Bigg].
\end{aligned}
\label{eq:pc-offline-l2}
\end{equation}
\end{theorem}

\noindent The proof is provided in Appendix~\ref{app:proof:thm:offline-safe-transfer-performance}.

Theorem~\ref{thm:offline-safe-transfer-performance} separates three
sources of offline policy loss: restricted-action coverage, target
Q-function estimation error, and pessimism. The coverage gap
$c_\beta^{(0)}$ and the concentrability coefficients
$C_\beta^{(0)}$ and $C_{\pi,\beta}^{(0)}$ are population quantities.
The theorem is pathwise in the posterior mean and uncertainty
functions. No assumption is imposed on the realized sizes of
$e_{e,N}^{(0)}$ or
$\mathfrak p_{e,N}^{(0)}$.
Their statistical rates, when available from the preceding posterior
analysis, may be substituted directly into
\eqref{eq:pc-offline-l2}.
In particular, an
$L^2(P_X^{(0)})$
posterior-contraction result can support the average-performance bound
only after the corresponding posterior-mean and posterior-uncertainty
norms have been controlled. It does not by itself imply the uniform
quantities appearing in
\eqref{eq:pc-offline-sup}. Likewise,
$c_\beta^{(0)}\equiv0$
eliminates the restricted-action approximation error but does not by
itself establish either estimation accuracy or concentrability.

The preceding results address distinct stages of the transfer-learning
procedure. The source-selection theory controls posterior weight on
predictively inferior subsets. Theorem~\ref{thm:posterior-contraction}
transfers verified oracle and Bellman-perturbation bounds to the target
working posterior. Theorem~\ref{thm:pc-fullrank} identifies the
full-rank regime in which the target precision determines the
contraction order, while
Theorem~\ref{thm:pc-pooled-rate} and
Theorem~\ref{thm:pc-minimax}
establish the pooled rate and its minimax optimality in the exact
shared-function experiment.
The offline result
Theorem~\ref{thm:offline-safe-transfer-performance}
adds support coverage and norm conversion to translate target
Q-function accuracy into policy performance.

\section{Experiments}
\label{sim:sec}

We evaluate Bayesian transfer Gaussian-process Q-learning
(BT-GP-OQL) in a unified synthetic benchmark designed to
investigate the effects of task heterogeneity, imbalanced
behavior coverage, Bellman bootstrapping, and target-aware
source selection in offline reinforcement learning.
By varying a common per-task sample size and evaluating
all performance metrics within the same Monte Carlo
replications, the experimental design enables systematic
comparisons of source-selection accuracy, target Q-function
estimation, negative-transfer behavior, uncertainty-aware
pessimism, and offline policy quality. This unified setting
also allows us to examine how these aspects of transfer
learning evolve as the amount of available training data
increases.
The implementation retains the multi-task GP covariance
structure and conservative target Bellman pseudo-responses
specified in Algorithm~2, while incorporating practical
procedures for predictive source scoring, numerical
stabilization, and final transfer acceptance. 

\subsection{Common experimental setup}
\label{sim:setup}

The state space is $\mathcal S=[0,1]^2$, and the action
space is $\mathcal A=\{1,2,3,4\}$. Each GP input is
represented by a state-action pair $x=(s,a)$, with state
dependence modeled through a continuous kernel and action
dependence captured by a discrete-action covariance.
We consider one target task and five source tasks,
\[
\mathcal T=\{M^{(0)},M^{(1)},M^{(2)},M^{(3)},M^{(4)},M^{(5)}\},
\]
where the source tasks are labeled \texttt{good1},
\texttt{good2}, \texttt{weak}, \texttt{bad1}, and
\texttt{bad2}, respectively. For each experimental
configuration, all tasks have the same trajectory length:
\begin{equation}\nonumber
n_0=n_1=\cdots=n_5=n,
\qquad n\in\{30,50,80,120,170\}.
\label{sim:sample-grid}
\end{equation}
Combining the five sample sizes with 30 Monte Carlo seeds
yields 150 offline task-family runs. For each seed, the
underlying latent functions are held fixed across sample
sizes, and trajectory generation uses random seeds
independent of $n$. As a result, shorter trajectories
form prefixes of the corresponding longer trajectories.
The final target evaluation states are also held fixed
across sample sizes within each seed.

To construct the random spectral functions, we set
$d=2$, $\alpha=1.5$, and $M=80$. The nonzero frequency
vectors in $\{0,\ldots,8\}^2$ are ordered by increasing
squared Euclidean norm, with ties resolved
lexicographically, and denoted by $m_1,\ldots,m_M$.
The corresponding basis functions are defined as
\begin{equation}\nonumber
\phi_j(s)=\prod_{\ell=1}^{2}
\begin{cases}
1, & m_{j\ell}=0,\\
\sqrt{2}\cos(\pi m_{j\ell}s_\ell),
& m_{j\ell}>0.
\end{cases}
\label{sim:basis}
\end{equation}
For each component function and Monte Carlo seed,
independent random signs $\xi_j\in\{-1,1\}$ and
amplitudes $u_j\sim\operatorname{Unif}(0.8,1.2)$
are used to generate the spectral coefficients:
\begin{equation}\nonumber
\widetilde\theta_j
=\xi_j u_j j^{-(\alpha/d+1/2)},
\qquad
\theta_j
=\frac{R\widetilde\theta_j}
{\left(
\sum_{\ell=1}^{M}
\ell^{2\alpha/d}\widetilde\theta_\ell^2
\right)^{1/2}},
\qquad
F(s)=\sum_{j=1}^{M}\theta_j\phi_j(s).
\label{sim:truth}
\end{equation}
This normalization ensures that
$\sum_{j=1}^{M}j^{2\alpha/d}\theta_j^2=R^2$.
We set $R=2$ for the shared state component $F_0$
and the source-specific perturbation functions $H_k$,
and $R=1.5$ for the four action-dependent components
$F_a$. This construction generates finite-dimensional
random functions satisfying a Sobolev-type weighted
coefficient constraint. Since the functions are finite
cosine expansions, they are smooth and are not claimed
to possess exact regularity $\alpha$. Moreover,
averaging estimation errors over this random function
ensemble does not constitute an estimate of the
worst-case Sobolev risk.

The target optimal action-value function is constructed as
\begin{equation}\nonumber
Q_\star^{(0)}(s,a)
=-0.35+0.10F_0(s)+b_a+0.035F_a(s),
\qquad
b=(0.13,0.10,0.07,-0.10).
\label{sim:target-q}
\end{equation}
The action-specific offsets assign a lower baseline
value to the fourth action, which is also sampled
infrequently under the target behavior policy.
To introduce varying degrees of task mismatch,
the source optimal action-value functions are defined by
\begin{equation}\nonumber
Q_\star^{(k)}(s,a)
=Q_\star^{(0)}(s,a)+c_kH_k(s)+d_{k,a},
\qquad k=1,\ldots,5,
\label{sim:source-q}
\end{equation}
where the task-specific coefficients are given in
Table~\ref{sim:tab:tasks}. Sources 1 and 2 are designed
to be closely related to the target task, whereas
source 3 introduces a larger perturbation and remains
eligible for inclusion in the empirical near-oracle
reference class. Sources 4 and 5 are designated as
potentially harmful because they combine substantial
Q-function mismatch with action-specific offsets
favoring the fourth action. These design-based
classifications are distinguished from the empirical
assessment of source usefulness through target
prediction risk and policy performance.

\begin{table}[htbp]
\centering
\caption{Source-task parameter settings. The coefficients
$c_k$ and $d_{k,a}$ specify the source Q-function
perturbations, while $v_k$ denotes the task-specific
transition shift in Equation~\eqref{sim:transition},
with $v_0=(0,0)$ for the target task.}
\label{sim:tab:tasks}
\small
\setlength{\tabcolsep}{4pt}
\begin{tabular}{clrll}
\toprule
$k$ & Source & $c_k$ & $d_k=(d_{k,1},\ldots,d_{k,4})$ & $v_k$\\
\midrule
1 & good1 & $0.030$ & $(0,0,0,0)$ & $(0.004,-0.003)$\\
2 & good2 & $-0.045$ & $(0,0,0,0)$ & $(-0.004,0.004)$\\
3 & weak & $0.180$ & $(0.04,-0.03,0.02,0.08)$ & $(0.018,-0.012)$\\
4 & bad1 & $0.750$ & $(-0.30,-0.24,-0.18,0.70)$ & $(0.060,0.045)$\\
5 & bad2 & $-0.900$ & $(-0.38,-0.29,-0.20,0.90)$ & $(-0.065,0.050)$\\
\bottomrule
\end{tabular}
\end{table}

For each task $k$, the initial state is sampled uniformly
from $[0,1]^2$, and subsequent states evolve according to
\begin{equation}
S_{t+1}^{(k)}
=\operatorname{clip}_{[0,1]^2}
\left\{
0.55S_t^{(k)}
+0.45(1/2,1/2)
+u_{A_t^{(k)}}+v_k+\zeta_t
\right\},
\qquad
\zeta_t\sim N(0,0.06^2I_2),
\label{sim:transition}
\end{equation}
where $u_1=(-0.05,0)$, $u_2=(0.05,0)$,
$u_3=(0,-0.05)$, and $u_4=(0,0.05)$ specify
the action-dependent state shifts, while $v_k$
represents the task-specific transition shift.
The behavior policy selects actions independently
of the current state, with action probabilities
$
\beta_0=(0.323,0.323,0.324,0.030)
$, $
\beta_k=(0.315,0.315,0.320,0.050)
$, and $\geq 1$.
With discount factor $\gamma=0.95$, the reward
is constructed from the prescribed optimal
Q-function as
\begin{equation}\nonumber
r_t^{(k)}
=Q_\star^{(k)}(S_t^{(k)},A_t^{(k)})
-\gamma\max_{a'}
Q_\star^{(k)}(S_{t+1}^{(k)},a')
+\epsilon_t,
\qquad
\epsilon_t\sim N(0,0.08^2).
\label{sim:reward}
\end{equation}
Taking conditional expectations verifies that
$Q_\star^{(k)}$ satisfies the Bellman optimality
equation for the constructed task. These
prespecified optimal Q-functions are used
to generate rewards and evaluate estimation
errors. During training, however, Bellman
pseudo-responses are constructed using
estimated continuation values rather than
oracle $Q_\star$ labels.
The implementation uses the known behavior
probabilities with a support threshold of
$\tau_\beta=0.01$. Since all four actions have
probabilities above this threshold,
$C_\beta(s)=\mathcal A$ throughout the benchmark,
despite the limited coverage of the fourth
action. This setting allows the effect of
uncertainty-aware pessimism to be evaluated
without excluding the rarely sampled action
from the admissible action set.

For the sample-size comparisons, we use the
nominal reference scale
$
N_{\mathrm{nom}}=n_0+n_1+n_2=3n,
$
corresponding to the target task and the two
core source tasks. This quantity represents
the nominal number of observations rather
than the effective information size appearing
in the theoretical analysis. The final GP model
based on the selected source subset uses
$n(1+|\widehat S|)$ observations, where
$\widehat S$ may include the weak source
or be empty.
To provide an additional diagnostic of
temporal dependence, we estimate an
autocorrelation-based effective sample size
for each task. Specifically, effective sample
sizes are computed for the two state coordinates
$s_1$ and $s_2$, the reward, and the indicator
of selecting action 4, using an initial-positive-
autocorrelation sum with at most
$\min(30,\lfloor n/4\rfloor)$ lags. The median
of these four estimates is recorded as the
task-level diagnostic effective sample size.
This quantity is used only for descriptive
purposes and does not replace the mixing
conditions or effective sample sizes required
by the theoretical analysis.

\subsection{Gaussian-process implementation and source selection}
\label{sim:implementation}

All fits use zero prior mean and a separable state-action kernel
\begin{equation}\nonumber
 k_{\mathrm{in}}((s,a),(s',a'))
 =0.5^2(1+z)\exp(-z)
 \bigl[\mathbf 1\{a=a'\}+0.03\mathbf 1\{a\ne a'\}\bigr],
 \quad z=\frac{\sqrt{3}\|s-s'\|_2}{0.30}.
 \label{sim:kernel}
\end{equation}
The state factor corresponds to a  Mat\'ern-$3 / 2$ kernel, with smoothness parameter $\nu=1.5$ and length scale $\ell=0.30$.
For an active task set, the multi-task covariance is
\begin{equation}\nonumber
 k_{\mathrm{MT}}((k,x),(\ell,x'))
 =K_{\mathrm{task}}(k,\ell)k_{\mathrm{in}}(x,x'),
 \qquad
 K_{\mathrm{task}}=(1-\rho)I+\rho\mathbf 1\mathbf 1^\top.
 \label{sim:mtkernel}
\end{equation}
For source selection, we fit a separate two-task Gaussian process
jointly to the target task and each candidate source, using a
task-correlation parameter of $\rho=0.88$. For the final multi-task
GP fits, we set $\rho=0.92$. The target-only baseline uses the
corresponding single-task kernel, which is independent of $\rho$.
Posterior means and latent-function variances are computed using
exact GP regression with Cholesky factorization and diagonal
pseudo-observation noise covariance matrices. To improve numerical stability, an initial diagonal stabilization term of $10^{-8}$ is added to the GP covariance matrix before Cholesky factorization. All available training observations are retained without imposing a predefined sample-size cap. At the largest sample size, the all-source GP fit therefore uses $6\times170=1020$ observations.

During the Bellman bootstrapping stage, a separate GP model
is fitted for each task, with $\mu_{e-1}^{(k)}$ and
$\sigma_{e-1}^{(k)}$ denoting the corresponding posterior mean
and latent standard deviation, respectively. At epoch $e$, the raw target and source responses are
\begin{align}
 z_{i,e}^{(0)}&=r_i^{(0)}+\gamma\max_{a'\in C_\beta(s_i')}
 \left\{\mu^{(0)}_{e-1}(s_i',a')-0.75\sigma^{(0)}_{e-1}(s_i',a')\right\},\label{sim:target-bootstrap}\\
 z_{i,e}^{(k)}&=r_i^{(k)}+\gamma\max_{a'\in\mathcal A}\mu^{(k)}_{e-1}(s_i',a'),\quad k\ge1.\label{sim:source-bootstrap}
\end{align}
In the first epoch, the continuation-value estimates
are set to zero, yielding $z_{i,1}^{(k)}=r_i^{(k)}$.
To improve numerical stability during iterative Bellman
bootstrapping, subsequent pseudo-responses are updated
using the following relaxation scheme:
\begin{equation}\nonumber
y_{i,1}^{(k)}=z_{i,1}^{(k)}, \qquad
y_{i,e}^{(k)}
=0.55y_{i,e-1}^{(k)}+0.45z_{i,e}^{(k)},
\quad e\geq 2.
\end{equation}
The target pseudo-observation variance follows the uncertainty weighting used in Algorithm~2,
\begin{equation}\nonumber
 v_{i,e}^{(0)}=0.08^2+\bigl[\sigma_{e-1}^{(0)}(s_i,a_i)\bigr]^2.
 \label{sim:targetvariance}
\end{equation}
For source tasks the implementation additionally uses
$v_{i,e}^{(k)}=3(0.08^2)+0.10[\sigma_{e-1}^{(k)}(s_i,a_i)]^2$.
At initialization, the current-state variance is set to $0.5^2$. Relaxation and the source variance rule are numerical choices of this simulation implementation, rather than extra theoretical guarantees.

To monitor the stability of the taskwise Bellman updates,
we construct an auxiliary evaluation grid $\mathcal G$
by pairing 180 states sampled under the target behavior
policy with all four actions. At epoch $e$, the maximum
change in posterior mean predictions across tasks and
grid points is defined as
\begin{equation}
d_e=\max_{0\le k\le5}\max_{x\in\mathcal G}
\left|\mu_e^{(k)}(x)-\mu_{e-1}^{(k)}(x)\right|.
\label{sim:change}
\end{equation}
Each run performs between four and 20 epochs. After the
minimum of four epochs, the procedure terminates when
$d_e\le 0.015$, or when $d_e$ exceeds 1.6 times the
smallest change observed among eligible epochs for two
consecutive epochs. The pseudo-responses and associated
working variances are retained from the eligible epoch
with the smallest $d_e$, with rollback to an earlier
epoch if necessary. This selection criterion relies
exclusively on changes in fitted posterior means and
does not require knowledge of the true optimal
Q-functions $Q_\star^{(k)}$. Source selection is then
performed using the retained pseudo-data, followed by
a single final joint multi-task GP fit. The resulting
implementation therefore adopts a staged approximation
to the joint Bellman recursion, rather than updating
source-selection decisions at every bootstrap epoch.

To assess the predictive contribution of each candidate source $k$
to the target task, we compare a target-only GP with a two-task GP
fitted jointly to the target and source $k$. The retained target
pseudo-data are randomly partitioned into three folds. For each
fold $f$, $\mu_{-f,k}$ and $s^2_{-f,k}$ denote the target predictive
mean and latent variance obtained by excluding the target
observations in fold $f$ while retaining all observations from
source $k$. The subscript $\varnothing$ denotes the corresponding
target-only fit. Using the retained target pseudo-observation
variances $v_i^{(0)}$, we define the cross-validated predictive
score difference as
\begin{equation}
L_k=\sum_{f=1}^{3}\sum_{i\in f}
\log\frac{
\varphi(y_i^{(0)};\mu_{-f,k}(x_i),s^2_{-f,k}(x_i)+v_i^{(0)})
}{
\varphi(y_i^{(0)};\mu_{-f,\varnothing}(x_i),
s^2_{-f,\varnothing}(x_i)+v_i^{(0)})
},
\label{sim:predictivescore}
\end{equation}
where $\varphi(\cdot;m,v)$ denotes the Gaussian density with
mean $m$ and variance $v$.

To account for distributional differences between the target
and source tasks, we compute the squared maximum mean
discrepancy $\widehat{\mathrm{MMD}}_k^2$ using a Gaussian kernel
on the normalized state-action representation $(s_1,s_2,a/4)$.
The kernel bandwidth is set to the median positive pairwise
distance among the combined target and source observations.
We use the biased all-pairs MMD estimator, restricting this
calculation to the first $\min(n,120)$ observations from each
task. The resulting source inclusion score and preliminary
selected subset are defined by
\begin{equation}
\widehat p_k=\operatorname{logit}^{-1}\left\{
\operatorname{logit}(0.35)+0.60L_k
-0.08n\widehat{\mathrm{MMD}}_k^2
\right\},\qquad
\widehat S_{\mathrm{pre}}
=\{k:\widehat p_k\ge0.5\}.
\label{sim:inclusion}
\end{equation}
Although the figures use the term ``posterior inclusion
probability'' for consistency with the implementation,
$\widehat p_k$ represents an empirical-Bayes-style inclusion
weight based on tempered predictive scores, rather than an
exact spike-and-slab posterior obtained by integrating over
all source subsets. Furthermore, the Bellman pseudo-responses
and their working variances are computed before the
cross-validation folds are formed and remain fixed throughout
the scoring procedure. The Bellman regression stage is not
repeated within individual folds. Consequently, these scores
serve as conditional predictive model-comparison diagnostics
rather than independent validation of the complete adaptive
learning procedure.

For a nonempty preliminary subset
$\widehat S_{\mathrm{pre}}$, we perform an additional two-fold
target-predictive comparison between the selected source
subset and the target-only model. Let $L_{\mathrm{gate}}$
denote the corresponding aggregate predictive score
difference, computed analogously to
\eqref{sim:predictivescore}. Assigning equal prior
weights to the two candidates, we define the final
transfer acceptance score and selected source subset as
\begin{equation}\nonumber
p_{\mathrm{tr}}
=\operatorname{logit}^{-1}(0.70L_{\mathrm{gate}}),
\qquad
\widehat S=
\begin{cases}
\widehat S_{\mathrm{pre}}, & p_{\mathrm{tr}}\ge0.5,\\
\varnothing, & p_{\mathrm{tr}}<0.5.
\end{cases}
\label{sim:gate}
\end{equation}
If $\widehat S_{\mathrm{pre}}=\varnothing$, the final
comparison is omitted, $p_{\mathrm{tr}}$ is recorded as
missing, and the target-only model is retained. We report
the frequencies of empty preliminary subsets, explicit
transfer rejections, and overall target-only selection
separately. This data-dependent acceptance rule provides
a mechanism for reverting to target-only learning when
the preliminary transfer model fails to improve the
predictive score. However, it does not guarantee
non-degradation of population target Q-function risk
or policy performance in finite samples.

\subsection{Evaluation protocol and connection to theory}
\label{sim:evaluation}

We evaluate three GP-based estimators: Target-only
($S=\varnothing$), All-source ($S=\{1,\ldots,5\}$),
and BT-GP-OQL ($S=\widehat S$). All three estimators
use the same retained taskwise pseudo-responses for their
respective active tasks. As a simulation-based benchmark,
we additionally consider an Oracle-safe reference
that evaluates all eight subsets of $\{1,2,3\}$, including
the empty subset, and selects the one minimizing the
empirical target Q-function risk on an independent set
of 700 evaluation states. The selected oracle model is
subsequently assessed on the common final test set,
which is distinct from the states used for oracle
selection. This benchmark provides a restricted
finite-sample reference rather than the unrestricted
population oracle considered in the theoretical analysis.
For all methods except the no-pessimism ablation,
the offline policy is obtained by maximizing the
pessimistic posterior value estimate over the
behavior-supported action set:
\begin{equation}
\widehat\pi_S(s)
=\arg\max_{a\in C_\beta(s)}
\left\{
\mu_S^{(0)}(s,a)
-0.75\sigma_S^{(0)}(s,a)
\right\}.
\label{sim:policy}
\end{equation}
The Selected-source, no pessimism ablation
instead uses
$\arg\max_a\mu_{\widehat S}^{(0)}(s,a)$,
based on the same final selected GP posterior.
This comparison isolates the effect of pessimism
during policy extraction while retaining the
conservative Bellman updates used during training.

For each Monte Carlo seed, we generate $m=1600$
evaluation states from a target-behavior trajectory
following a burn-in period of 60 transitions.
All four actions are evaluated at each state.
For a fitted posterior mean $\widehat Q$,
the empirical target Q-function mean squared error
is defined as
\begin{equation}
\widehat{\mathrm{MSE}}_Q(\widehat Q)
=\frac{1}{4m}\sum_{j=1}^{m}\sum_{a=1}^{4}
\left[
\widehat Q(s_j,a)
-Q_\star^{(0)}(s_j,a)
\right]^2,
\qquad m=1600.
\label{sim:mse}
\end{equation}
This metric assigns equal weight to all four actions,
regardless of their frequencies under the target
behavior policy.
To assess decision quality, we report the average
optimal-Q score of the actions selected by a policy
and the corresponding optimal-action Q gap:
\begin{equation}
\begin{aligned}
\widehat J(\pi)
&=\frac{1}{m}\sum_{j=1}^{m}
Q_\star^{(0)}(s_j,\pi(s_j)),\\
\widehat G_Q(\pi)
&=\frac{1}{m}\sum_{j=1}^{m}
\left[
\max_a Q_\star^{(0)}(s_j,a)
-Q_\star^{(0)}(s_j,\pi(s_j))
\right].
\end{aligned}
\label{sim:policy-metrics}
\end{equation}
The quantity $\widehat G_Q(\pi)$ measures the
average optimal-action Q gap at the evaluation
states and should not be interpreted as the
infinite-horizon policy value loss
$V^\star-V^\pi$.
To complement this statewise comparison,
we compute a separate rollout diagnostic
$\widehat G_{80}$ based on 80-step discounted
returns. The learned and true greedy policies
are evaluated from 120 common initial states
using identical transition-noise realizations.
Reward noise is omitted during evaluation,
and no terminal-value correction is applied
to the truncated returns.

The designated harmful-source set is
$\mathcal U=\{4,5\}$. We report the empirical
frequency with which the final selected subset
$\widehat S$ contains at least one source in
$\mathcal U$, together with the frequency of
retaining both core sources $\{1,2\}$.
To summarize the pre-gate inclusion scores,
we also compute the factorized quantity
\begin{equation}\nonumber
w_{\mathcal U}
=1-\prod_{k\in\mathcal U}(1-\widehat p_k).
\label{sim:unsafe-score}
\end{equation}
The complementary quantity $1-w_{\mathcal U}$
is labeled ``posterior no unsafe'' in the
figures. These quantities are constructed
from the approximate marginal inclusion scores
and do not represent posterior probabilities
obtained from a joint source-subset model
after the final transfer gate.
To evaluate proximity to the restricted
oracle reference, we define the empirical
target-risk gap on the final test states as
\[
\Delta_R
=\widehat{\mathrm{MSE}}_Q(\widehat Q_{\mathrm{BT}})
-\widehat{\mathrm{MSE}}_Q(\widehat Q_{\mathrm{oracle}}).
\]
A run is classified as near-oracle when
\begin{equation}
\Delta_R
\le
\max\left\{
0.010,\;
0.15\widehat{\mathrm{MSE}}_Q
(\widehat Q_{\mathrm{oracle}})
\right\}.
\label{sim:nearoracle-flag}
\end{equation}
We report the continuous risk gap alongside
this binary indicator, since the fixed absolute
tolerance may be relatively permissive when
the oracle estimation error is small.
To assess negative transfer in policy quality,
we compare each transfer policy with the
target-only baseline using the paired difference
\[
\Delta_J(S)
=\widehat J(\widehat\pi_S)
-\widehat J(\widehat\pi_{\varnothing}).
\]
We report both the frequency of negative
transfer, defined by $\Delta_J<0$, and the
frequency of material degradation, defined
by $\Delta_J<-0.010$. The latter threshold
specifies a practical effect-size criterion
and does not replace the zero-threshold
definition of negative transfer.

At epoch $e$, the continuation-value function
used by the algorithm for task $k$ is denoted
by $V_{e-1}^{(k),\mathrm{alg}}$, including the
pessimistic adjustment for the target task
in Equations~\eqref{sim:target-bootstrap}--
\eqref{sim:source-bootstrap}.
The bootstrap discrepancy is measured by
\begin{equation}\nonumber
B_e
=\frac{1}{6}\sum_{k=0}^{5}\frac{1}{n}
\sum_{i=1}^{n}
\gamma^2
\left[
V_{e-1}^{(k),\mathrm{alg}}(s_i')
-\max_a Q_\star^{(k)}(s_i',a)
\right]^2.
\label{sim:bootstrap-diagnostic}
\end{equation}
This quantity measures the squared discrepancy
between the continuation values used by the
algorithm and the corresponding optimal
continuation values.
On the monitoring grid consisting of 180 states
and four actions, we approximate the algorithm-
aligned, unrelaxed Bellman update using 12 fresh
transitions for each state-action pair.
The mean squared discrepancy between this
Monte Carlo Bellman approximation and the
fitted posterior mean $\mu_e^{(k)}$ is denoted
by $D_e$ and averaged across tasks.
We additionally record the successive-iterate
change $d_e$ defined in Equation~\eqref{sim:change}
and the maximum grid-based Q-function error
\[
U_e
=\max_{k,x\in\mathcal G}
\left|
\mu_e^{(k)}(x)-Q_\star^{(k)}(x)
\right|.
\]
These diagnostics characterize the taskwise
Bellman bootstrapping stage rather than the
final selected multi-task GP posterior.
Moreover, the grid-based maxima and
Monte Carlo Bellman approximations should
not be identified with exact uniform-norm
errors or population Bellman residuals.

To characterize empirical scaling over the
simulated sample-size range, we fit the
following log-linear model for each positive
outcome $E_{r,n}$ and Monte Carlo seed $r$:
\begin{equation}\nonumber
\log E_{r,n}
=c_r-\beta\log(3n)+\varepsilon_{r,n},
\label{sim:ratefit}
\end{equation}
where $c_r$ denotes a seed-specific fixed
effect. The quantity $3n$ is the nominal
combined sample size of the target task and
the two core sources, rather than a
theoretically established effective sample size.
In addition to the primary seed-fixed-effect
regressions, we report pooled estimates and
descriptive fits restricted to runs without
selected harmful sources or satisfying the
near-oracle criterion. Since these restrictions
depend on observed outcomes, the corresponding
estimates are treated as supplementary
descriptive summaries rather than the
primary rate analysis.
The theoretical reference exponents are
$2\alpha/(2\alpha+d)=0.6$ for squared
estimation error and
$\alpha/(2\alpha+d)=0.3$ for root
estimation error. The exponent $0.3$
is also displayed as a reference for
policy-related outcomes, without asserting
that it provides an exact decay rate
for the empirical optimal-action Q gap
in Equation~\eqref{sim:policy-metrics}.

All reported means are computed over
30 Monte Carlo seeds at each sample size
$n$. Within each seed, method comparisons
are based on common underlying trajectories
and evaluation states.
Error bars for source-inclusion scores
and pessimism gains reproduce the
implementation's mean $\pm1.96$ Monte Carlo
standard errors. The rate tables report
nominal 95\% confidence intervals obtained
from ordinary linear regression with
seed fixed effects. These intervals are
neither seed-cluster-adjusted nor
bootstrap-based, although the nested
sample-size design induces dependence
among observations from the same seed.
The available experimental summaries
provide standard errors for BT-GP-OQL
Q-function estimation error, source-inclusion
scores, and pessimism gains, but do not
include paired standard errors for every
comparison between methods. We therefore
interpret the reported differences as
empirical average improvements unless
the available uncertainty estimates
support a stronger statistical conclusion.

\FloatBarrier
\subsection{Simulation results}
\label{sim:results}

\subsubsection{Target-aware source selection and the target-only gate}

\begin{figure}[p]
\centering
\begin{subfigure}{0.86\linewidth}
\includegraphics[width=\linewidth]{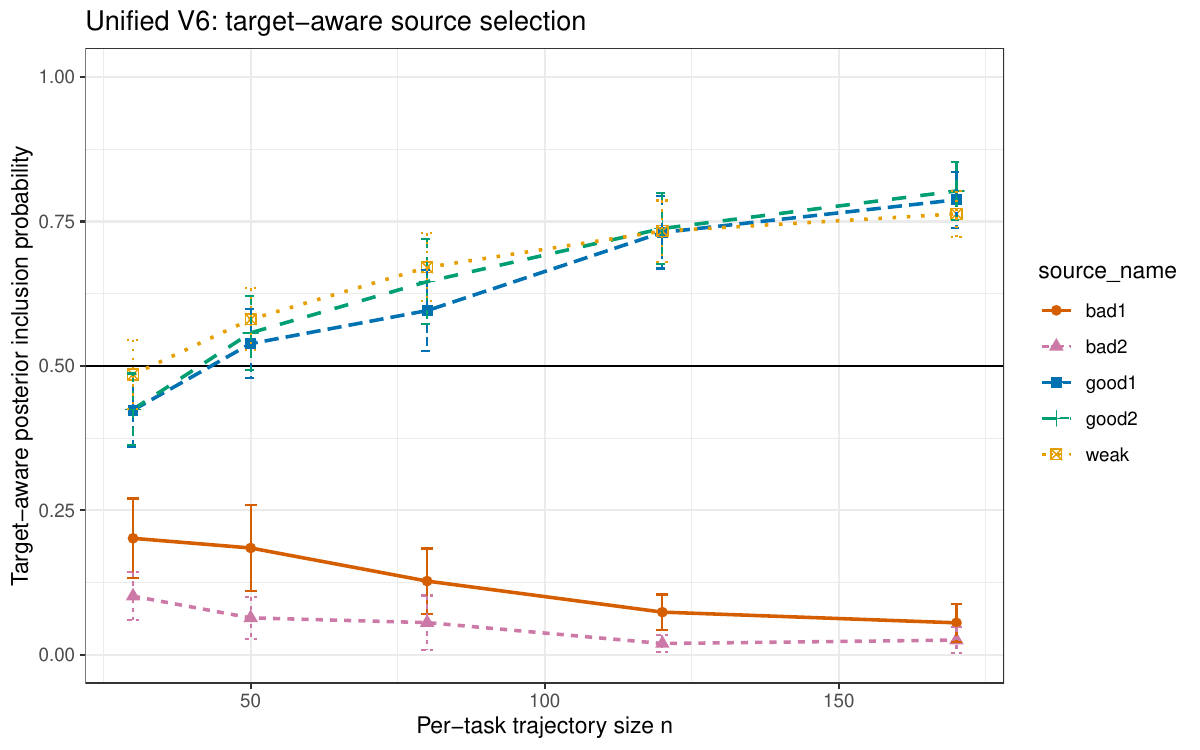}
\caption{Marginal inclusion scores; bars are mean $\pm1.96$ SE.}
\end{subfigure}
\par\medskip
\begin{subfigure}{0.86\linewidth}
\includegraphics[width=\linewidth]{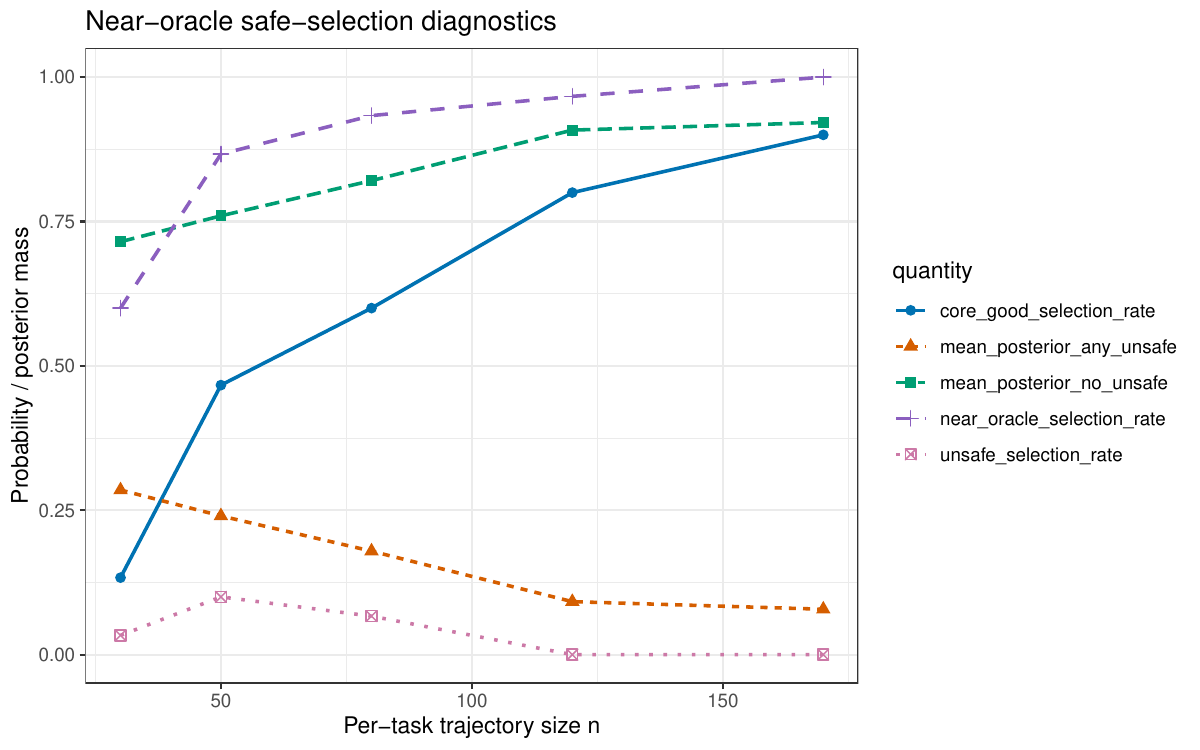}
\caption{Final selection frequencies and pre-gate factorized scores.}
\end{subfigure}
\caption{Source selection in the unified 30-seed experiment. Similar sources become more frequently retained as $n$ increases, while designated harmful sources are increasingly excluded. The near-oracle flag is tolerance-based and does not require a unique recovered subset.}
\label{sim:fig:selection}
\end{figure}

Table~\ref{sim:tab:inclusion} and Figure~\ref{sim:fig:selection} show increasing separation between similar and designated harmful sources. The mean inclusion score of good1 increases from 0.423 to 0.788, and that of good2 increases from 0.425 to 0.803. In contrast, the scores of bad1 and bad2 decrease from 0.201 and 0.101 to 0.055 and 0.025. The weak source remains competitive, with its mean score rising from 0.485 to 0.763. This behavior is consistent with selecting sources for their predictive contribution rather than imposing exact recovery of $\{1,2\}$.

\begin{table}[htbp]
\centering
\caption{Target-aware marginal inclusion scores, reported as mean $\pm$ Monte Carlo standard error over 30 seeds. These are the empirical-Bayes-style scores in Equation~\eqref{sim:inclusion}.}
\label{sim:tab:inclusion}
\small\setlength{\tabcolsep}{4pt}
\begin{tabular}{rccccc}
\toprule
$n$ & good1 & good2 & weak & bad1 & bad2 \\
\midrule
30 & $0.423 \pm 0.033$ & $0.425 \pm 0.032$ & $0.485 \pm 0.031$ & $0.201 \pm 0.035$ & $0.101 \pm 0.021$ \\
50 & $0.538 \pm 0.030$ & $0.557 \pm 0.033$ & $0.581 \pm 0.027$ & $0.185 \pm 0.038$ & $0.064 \pm 0.018$ \\
80 & $0.596 \pm 0.036$ & $0.646 \pm 0.037$ & $0.671 \pm 0.030$ & $0.127 \pm 0.029$ & $0.055 \pm 0.024$ \\
120 & $0.731 \pm 0.032$ & $0.738 \pm 0.031$ & $0.733 \pm 0.027$ & $0.074 \pm 0.015$ & $0.019 \pm 0.008$ \\
170 & $0.788 \pm 0.025$ & $0.803 \pm 0.025$ & $0.763 \pm 0.021$ & $0.055 \pm 0.017$ & $0.025 \pm 0.011$ \\
\bottomrule
\end{tabular}

\end{table}

The frequency of selecting both core sources rises from 13.3\% to 90.0\%, whereas the final harmful-source selection frequencies are 3.3\%, 10.0\%, 6.7\%, 0\%, and 0\% over the five sample sizes. Thus no harmful source is selected in the 30 runs at either $n=120$ or $n=170$, although this observed zero frequency is not a zero population probability. The mean factorized harmful-source score decreases from 0.285 to 0.079. Table~\ref{sim:tab:selection} also shows that the near-oracle success frequency rises from 60.0\% to 100\%. More informatively, the mean continuous risk gap decreases from 0.01196 to 0.00045; the mean relative gap decreases from 45.8\% to 12.4\%. The 100\% success rate should be read together with the tolerance in Equation~\eqref{sim:nearoracle-flag}, rather than as exact oracle recovery.

\begin{table}[htbp]
\centering
\caption{Final selection diagnostics. Core: both good1 and good2 retained. Harmful: at least one of bad1 and bad2 retained. Near-oracle: Equation~\eqref{sim:nearoracle-flag}. $w_{\mathcal U}$ is the pre-gate factorized score. Relative gaps are means of per-run relative gaps, not ratios of aggregate risks.}
\label{sim:tab:selection}
\small\setlength{\tabcolsep}{3.5pt}
\begin{tabular}{rrrrrrr}
\toprule
$n$ & Core (\%) & Harmful (\%) & Near-oracle (\%) & $w_{\mathcal U}$ & $\overline{\Delta_R}$ & Rel. gap (\%) \\
\midrule
30 & 13.3 & 3.3 & 60.0 & 0.285 & 0.01196 & 45.8 \\
50 & 46.7 & 10.0 & 86.7 & 0.240 & 0.00548 & 31.0 \\
80 & 60.0 & 6.7 & 93.3 & 0.179 & 0.00296 & 26.1 \\
120 & 80.0 & 0.0 & 96.7 & 0.092 & 0.00137 & 19.1 \\
170 & 90.0 & 0.0 & 100.0 & 0.079 & 0.00045 & 12.4 \\
\bottomrule
\end{tabular}

\end{table}

\FloatBarrier

\begin{figure}[htbp]
\centering
\includegraphics[width=0.90\linewidth]{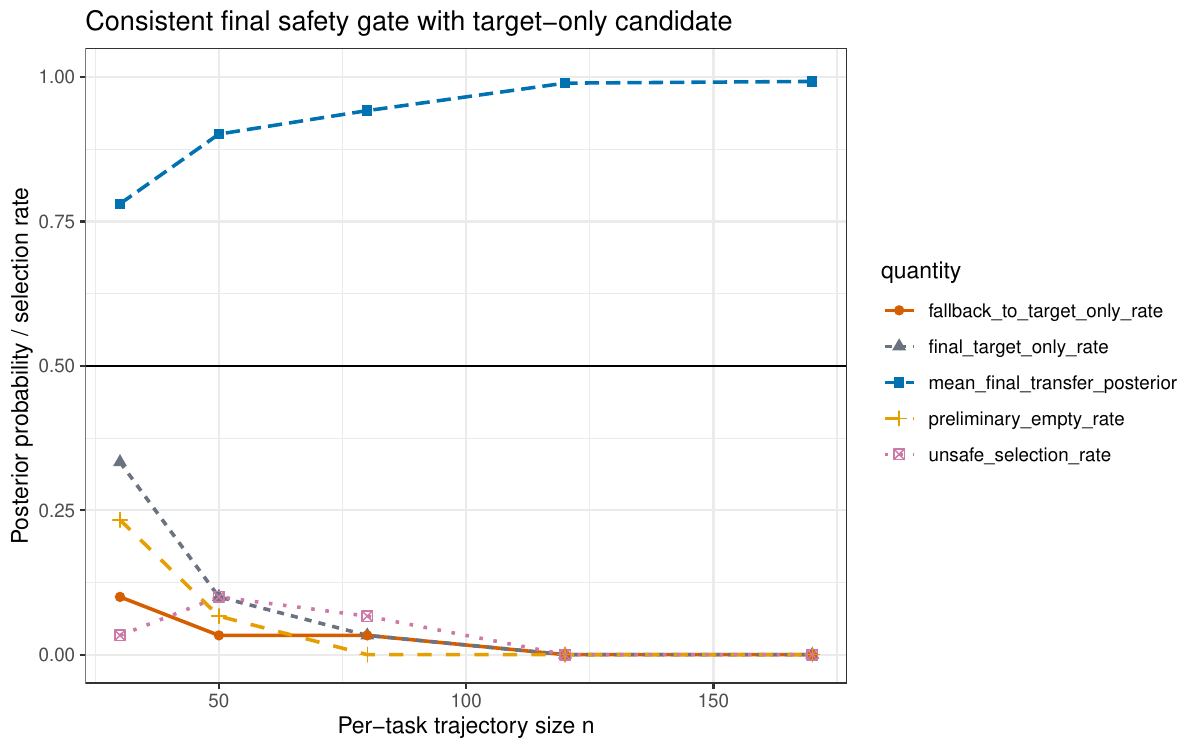}
\caption{Final transfer acceptance and target-only usage. Preliminary-empty runs are not included in the average transfer score. The threshold at 0.5 applies to evaluated proposals, not to an unconditional average over all runs.}
\label{sim:fig:gate}
\end{figure}

The final target-only comparison is summarized in Table~\ref{sim:tab:gate} and Figure~\ref{sim:fig:gate}. At $n=30$, seven runs have no preliminary transfer proposal and three additional runs reject a nonempty proposal, giving ten target-only runs in total. At $n=50$ the corresponding numbers are two and one; at $n=80$ there is one explicit rejection and no preliminary-empty run. At $n=120$ and $n=170$, all runs retain a transfer proposal. The mean final transfer score, conditional on gate evaluation, increases from 0.781 to 0.993. These observations document the fallback mechanism; they do not show that it guarantees policy safety, since its criterion is conditional predictive density rather than policy return.

\begin{table}[htbp]
\centering
\caption{Final target-only gate. Gate runs counts nonempty proposals. Empty, Rejected, and Target-only are percentages of all 30 runs, with Target-only $=$ Empty $+$ Rejected. Mean $p_{\mathrm{tr}}$ is conditional on evaluating the gate.}
\label{sim:tab:gate}
\small\setlength{\tabcolsep}{4pt}
\begin{tabular}{rrrrrr}
\toprule
$n$ & Gate runs & Empty (\%) & Rejected (\%) & Target-only (\%) & Mean $p_{\mathrm{tr}}$ \\
\midrule
30 & 23 & 23.3 & 10.0 & 33.3 & 0.781 \\
50 & 28 & 6.7 & 3.3 & 10.0 & 0.901 \\
80 & 30 & 0.0 & 3.3 & 3.3 & 0.942 \\
120 & 30 & 0.0 & 0.0 & 0.0 & 0.990 \\
170 & 30 & 0.0 & 0.0 & 0.0 & 0.993 \\
\bottomrule
\end{tabular}

\end{table}

\FloatBarrier

\subsubsection{Target-function estimation and transfer gain}

\begin{figure}[htbp]
\centering
\includegraphics[width=0.90\linewidth]{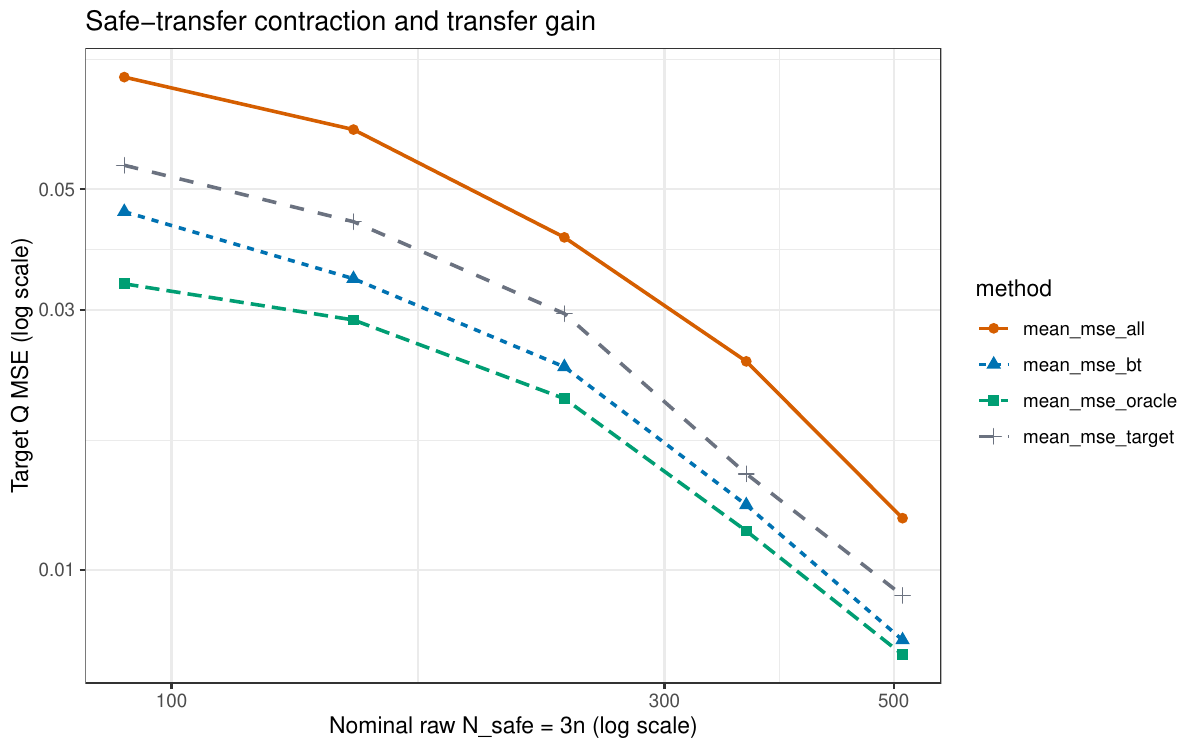}
\caption{Target-function estimation versus the nominal raw scale $3n$. The plotted quantities are mean squared errors of posterior means, not posterior tail probabilities. Mean BT error is lower than target-only error at each reported sample size.}
\label{sim:fig:q}
\end{figure}

The final target-function errors in Table~\ref{sim:tab:q} decrease with sample size for all four methods. Mean BT MSE decreases from 0.04550 at $n=30$ to 0.00744 at $n=170$, an 83.7\% reduction over the simulated range. It is below the target-only mean at every sample size. The reductions relative to the target-only aggregate MSE are 17.8\%, 21.4\%, 20.2\%, 12.4\%, and 17.2\%, respectively. The restricted oracle reference has the lowest mean error, and the BT--oracle gap becomes small at the largest sample size. All-source regression has higher mean MSE than target-only throughout this sequence. Figure~\ref{sim:fig:q} therefore displays a persistent average advantage for selective transfer over indiscriminate pooling. The supplied summaries do not establish a positive paired gain in every seed or provide paired uncertainty for each MSE contrast.

\begin{table}[htbp]
\centering
\caption{Target $Q$ estimation on the common test states. All entries are means; only BT also reports its supplied Monte Carlo SE. Reduction is $100(1-\overline{\mathrm{MSE}}_{\mathrm{BT}}/\overline{\mathrm{MSE}}_{\mathrm{Target}})$. The oracle subset is selected on a separate state sample.}
\label{sim:tab:q}
\small\setlength{\tabcolsep}{3.5pt}
\begin{tabular}{rrrrrr}
\toprule
$n$ & Target-only & All-source & BT (mean $\pm$ SE) & Oracle-safe & Reduction \\
\midrule
30 & 0.05536 & 0.08038 & $0.04550 \pm 0.00482$ & 0.03353 & 17.8\% \\
50 & 0.04361 & 0.06438 & $0.03427 \pm 0.00442$ & 0.02878 & 21.4\% \\
80 & 0.02956 & 0.04081 & $0.02360 \pm 0.00344$ & 0.02064 & 20.2\% \\
120 & 0.01502 & 0.02416 & $0.01316 \pm 0.00216$ & 0.01179 & 12.4\% \\
170 & 0.00898 & 0.01245 & $0.00744 \pm 0.00108$ & 0.00699 & 17.2\% \\
\bottomrule
\end{tabular}

\end{table}

\FloatBarrier

\subsubsection{Negative-transfer safety and offline policy quality}

\begin{figure}[p]
\centering
\begin{subfigure}{0.86\linewidth}
\includegraphics[width=\linewidth]{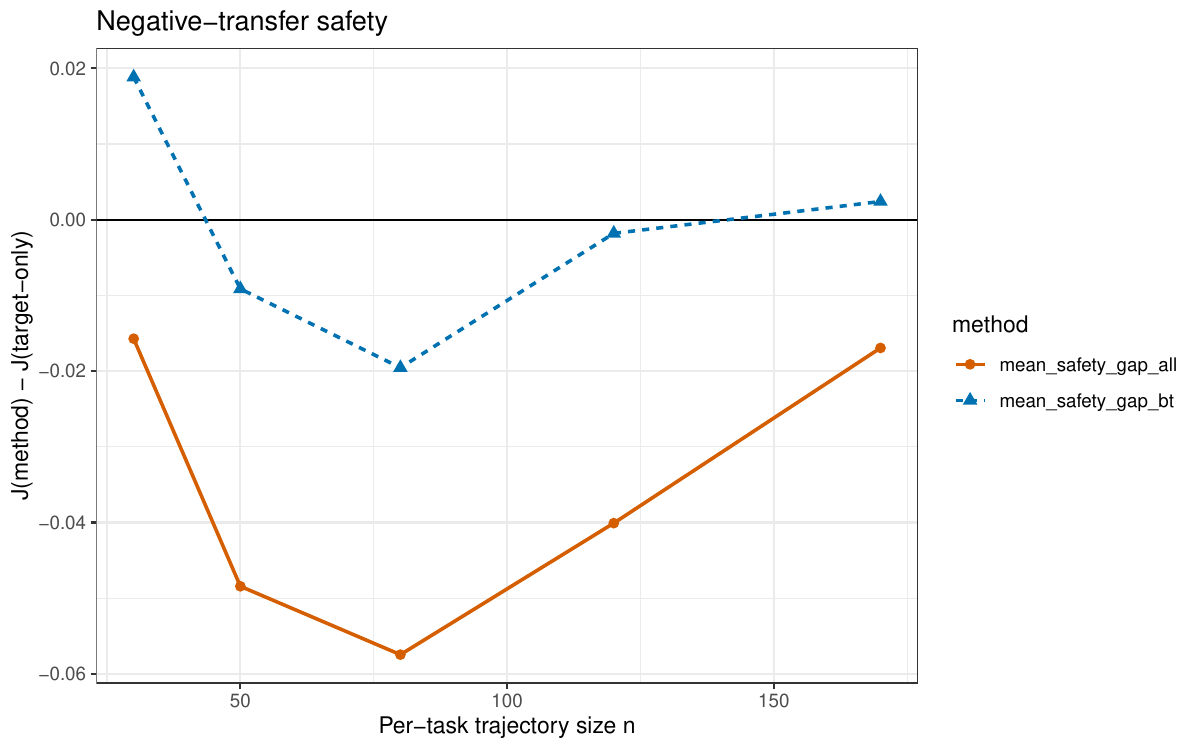}
\caption{Mean paired gap $J(\pi_S)-J(\pi_{\varnothing})$.}
\end{subfigure}
\par\medskip
\begin{subfigure}{0.86\linewidth}
\includegraphics[width=\linewidth]{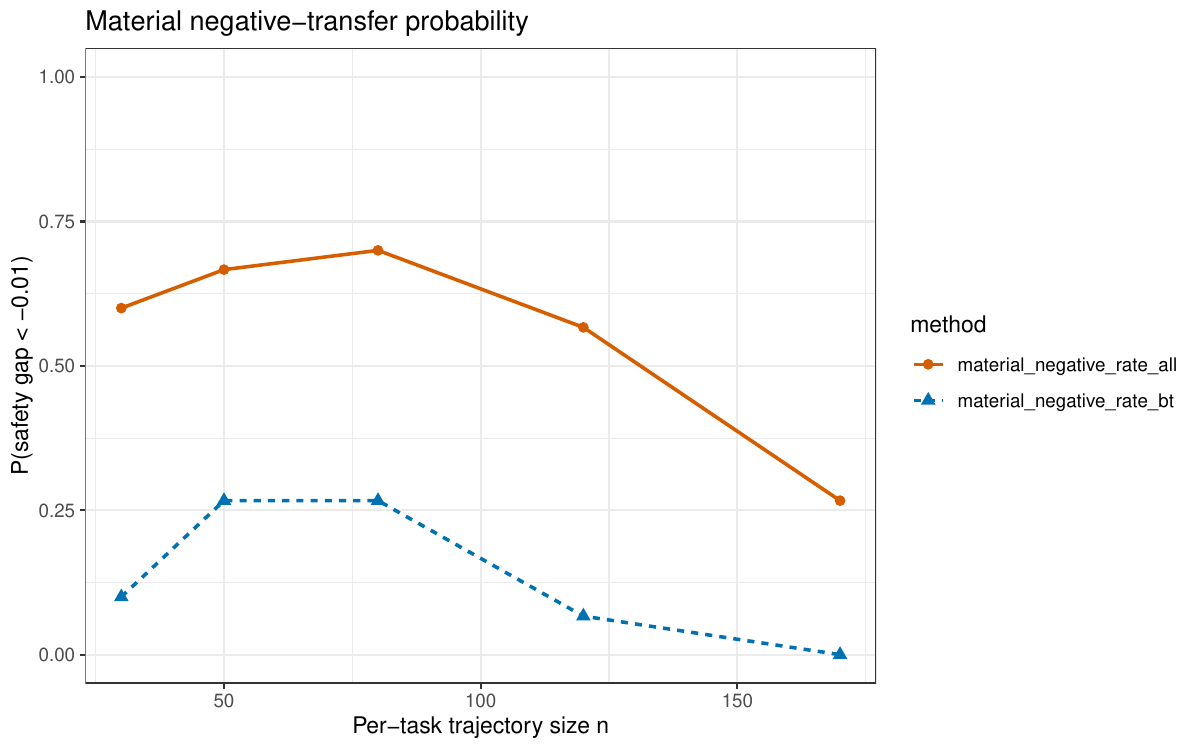}
\caption{Empirical frequency of $\Delta_J<-0.010$.}
\end{subfigure}
\caption{Negative transfer in the same unified benchmark. Selective transfer reduces harm relative to all-source transfer, while finite-sample negative gaps remain. The material-risk threshold is a stated effect-size convention.}
\label{sim:fig:safety}
\end{figure}

Table~\ref{sim:tab:safety} and Figure~\ref{sim:fig:safety} show that all-source transfer can harm the target policy despite using more observations. Its mean safety gap is negative at all five sample sizes, reaching $-0.05745$ at $n=80$. BT reduces this average harm substantially, but does not eliminate it in every setting: its mean gaps at $n=50$, $80$, and $120$ are $-0.00913$, $-0.01954$, and $-0.00180$. In particular, lower average $Q$-MSE does not ensure a positive policy-value gap, because the two metrics weight errors differently and policy extraction depends on action ranking and uncertainty.
The material negative-transfer frequencies correspond to 18, 20, 21, 17, and 8 all-source failures out of 30 runs, compared with 3, 8, 8, 2, and 0 BT failures. These results support a reduction in the observed frequency and magnitude of material harm, not finite-sample non-degradation. At $n=170$, the zero-threshold BT negative-transfer rate is still 16.7\% even though no run crosses the material threshold $-0.010$; reporting both thresholds prevents small negative gaps from being hidden.

\begin{table}[htbp]
\centering
\caption{Policy safety and suboptimality. Safety gaps are paired against the target-only policy. Risk denotes the percentage of runs with $\Delta_J<-0.010$. $G_Q$ is the action gap in Equation~\eqref{sim:policy-metrics}; $G_{80}$ is the separate truncated rollout gap. Entries without uncertainty are reported means or empirical frequencies.}
\label{sim:tab:safety}
\small\setlength{\tabcolsep}{3.0pt}
\begin{tabular}{rrrrrrr}
\toprule
$n$ & $\overline\Delta_J$ (All) & $\overline\Delta_J$ (BT) & Risk All (\%) & Risk BT (\%) & $G_Q$ (BT) & $G_{80}$ (BT) \\
\midrule
30 & -0.01572 & +0.01884 & 60.0 & 10.0 & 0.06509 & 1.2449 \\
50 & -0.04842 & -0.00913 & 66.7 & 26.7 & 0.06264 & 1.2365 \\
80 & -0.05745 & -0.01954 & 70.0 & 26.7 & 0.04067 & 0.7876 \\
120 & -0.04008 & -0.00180 & 56.7 & 6.7 & 0.01596 & 0.3732 \\
170 & -0.01694 & +0.00241 & 26.7 & 0.0 & 0.00905 & 0.1761 \\
\bottomrule
\end{tabular}

\end{table}

\FloatBarrier

\begin{figure}[htbp]
\centering
\includegraphics[width=0.90\linewidth]{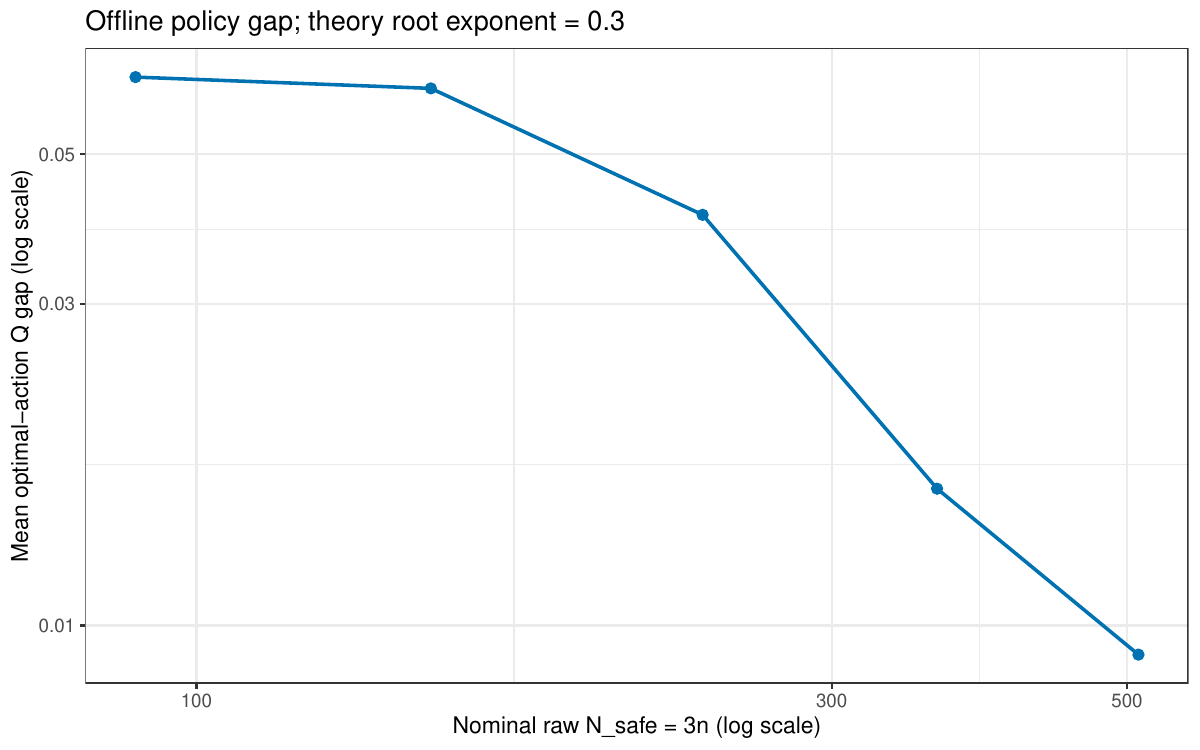}
\caption{Mean optimal-action $Q$ gap of BT. The title's exponent 0.3 is a theoretical reference from the manuscript; the displayed statistic is an empirical action gap, not a direct measurement of an infinite-horizon value-loss rate.}
\label{sim:fig:policygap}
\end{figure}

For completeness, Table~\ref{sim:tab:policyvalues} gives the policy-quality means themselves. The BT action gap decreases from 0.06509 to 0.00905, an 86.1\% reduction, and the 80-step rollout gap decreases from 1.2449 to 0.1761. The two measures have the same overall direction, but different definitions and scales. Figure~\ref{sim:fig:policygap} provides descriptive support for improving target decision quality with additional data, without identifying the plotted $Q$ gap with the value-loss quantity in Theorem~5.

\begin{table}[htbp]
\centering
\caption{Mean policy-quality score $\widehat J$ from Equation~\eqref{sim:policy-metrics}; larger is better. The score evaluates a selected action using $Q_\star$, and is distinct from a full policy rollout return.}
\label{sim:tab:policyvalues}
\small
\begin{tabular}{rrrrr}
\toprule
$n$ & Target-only & All-source & BT & Oracle-safe \\
\midrule
30 & -0.30416 & -0.31988 & -0.28532 & -0.27671 \\
50 & -0.27373 & -0.32215 & -0.28286 & -0.26808 \\
80 & -0.24135 & -0.29880 & -0.26090 & -0.24656 \\
120 & -0.23439 & -0.27447 & -0.23619 & -0.23209 \\
170 & -0.23169 & -0.24863 & -0.22928 & -0.22960 \\
\bottomrule
\end{tabular}

\end{table}

\FloatBarrier

\subsubsection{Contribution of uncertainty-aware pessimism}

\begin{figure}[p]
\centering
\begin{subfigure}{0.86\linewidth}
\includegraphics[width=\linewidth]{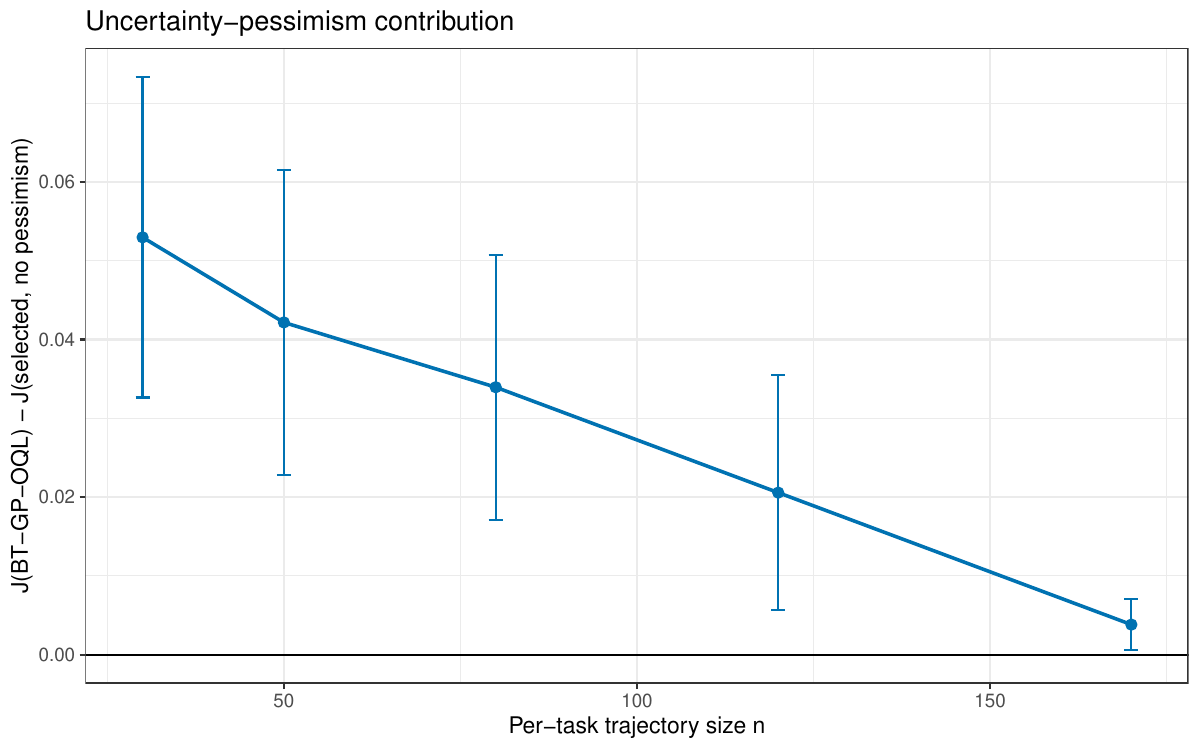}
\caption{Paired policy-quality gain with nominal uncertainty bars.}
\end{subfigure}
\par\medskip
\begin{subfigure}{0.86\linewidth}
\includegraphics[width=\linewidth]{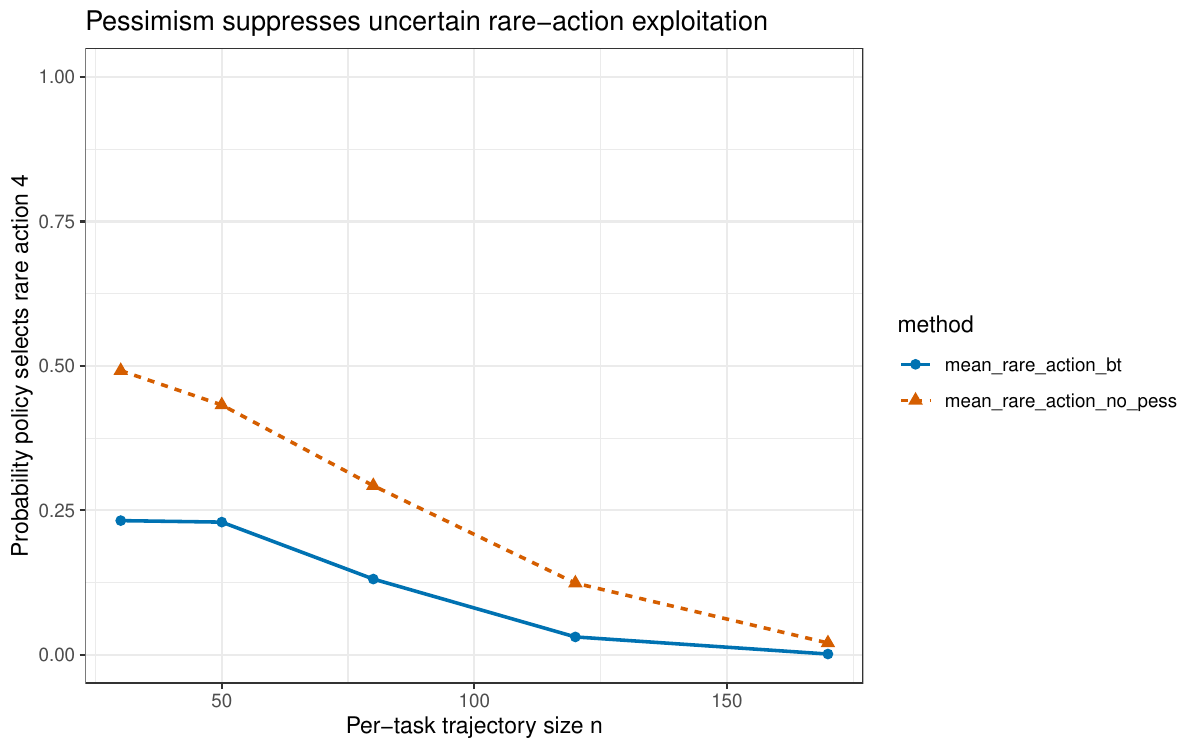}
\caption{Selection frequency of the poorly covered fourth action.}
\end{subfigure}
\caption{Decision-stage pessimism in the unified experiment. Both policies share the same fitted posterior and training-stage conservative responses; only their final action-selection rule differs.}
\label{sim:fig:pessimism}
\end{figure}

Table~\ref{sim:tab:pessimism} reports the policy-extraction ablation on the same selected GP posterior. The mean gain from using $\mu-0.75\sigma$ rather than $\mu$ is positive at every sample size and decreases from 0.05297 to 0.00381. The program's nominal mean $\pm1.96$ SE intervals are above zero for all five displayed comparisons. These are unadjusted Monte Carlo intervals, not a simultaneous multiple-comparison statement.
Figure~\ref{sim:fig:pessimism} relates these gains to the action-4 selection frequencies. Without pessimism, the frequencies decrease from 49.17\% to 2.02\%; with pessimism, they decrease from 23.21\% to 0.10\%. Thus the conservative rule reduces the selection of the poorly covered fourth action, and its mean advantage becomes small when the action is rarely selected by either rule. This is evidence for the decision-stage contribution of uncertainty penalization in this benchmark. It does not isolate the effect of conservative pseudo-labels, which are shared by both policies.

\begin{table}[htbp]
\centering
\caption{Pessimism ablation. Gain is $\widehat J(\pi_{\mathrm{BT}})-\widehat J(\pi_{\mathrm{no\ pess}})$. Nominal intervals reproduce mean $\pm1.96$ SE. The final columns are mean action-4 selection frequencies on the test states.}
\label{sim:tab:pessimism}
\small\setlength{\tabcolsep}{4pt}
\begin{tabular}{rccrr}
\toprule
$n$ & Gain (mean $\pm$ SE) & Nominal 95\% interval & No pess. (\%) & Pess. (\%) \\
\midrule
30 & $0.05297 \pm 0.01038$ & $[0.03263,\,0.07331]$ & 49.17 & 23.21 \\
50 & $0.04217 \pm 0.00989$ & $[0.02278,\,0.06156]$ & 43.23 & 22.95 \\
80 & $0.03395 \pm 0.00858$ & $[0.01713,\,0.05076]$ & 29.25 & 13.09 \\
120 & $0.02057 \pm 0.00761$ & $[0.00566,\,0.03548]$ & 12.39 & 3.07 \\
170 & $0.00381 \pm 0.00165$ & $[0.00059,\,0.00704]$ & 2.02 & 0.10 \\
\bottomrule
\end{tabular}

\end{table}

\FloatBarrier

\subsubsection{Numerical Bellman stability and bootstrap error}

\begin{figure}[htbp]
\centering
\includegraphics[width=\linewidth]{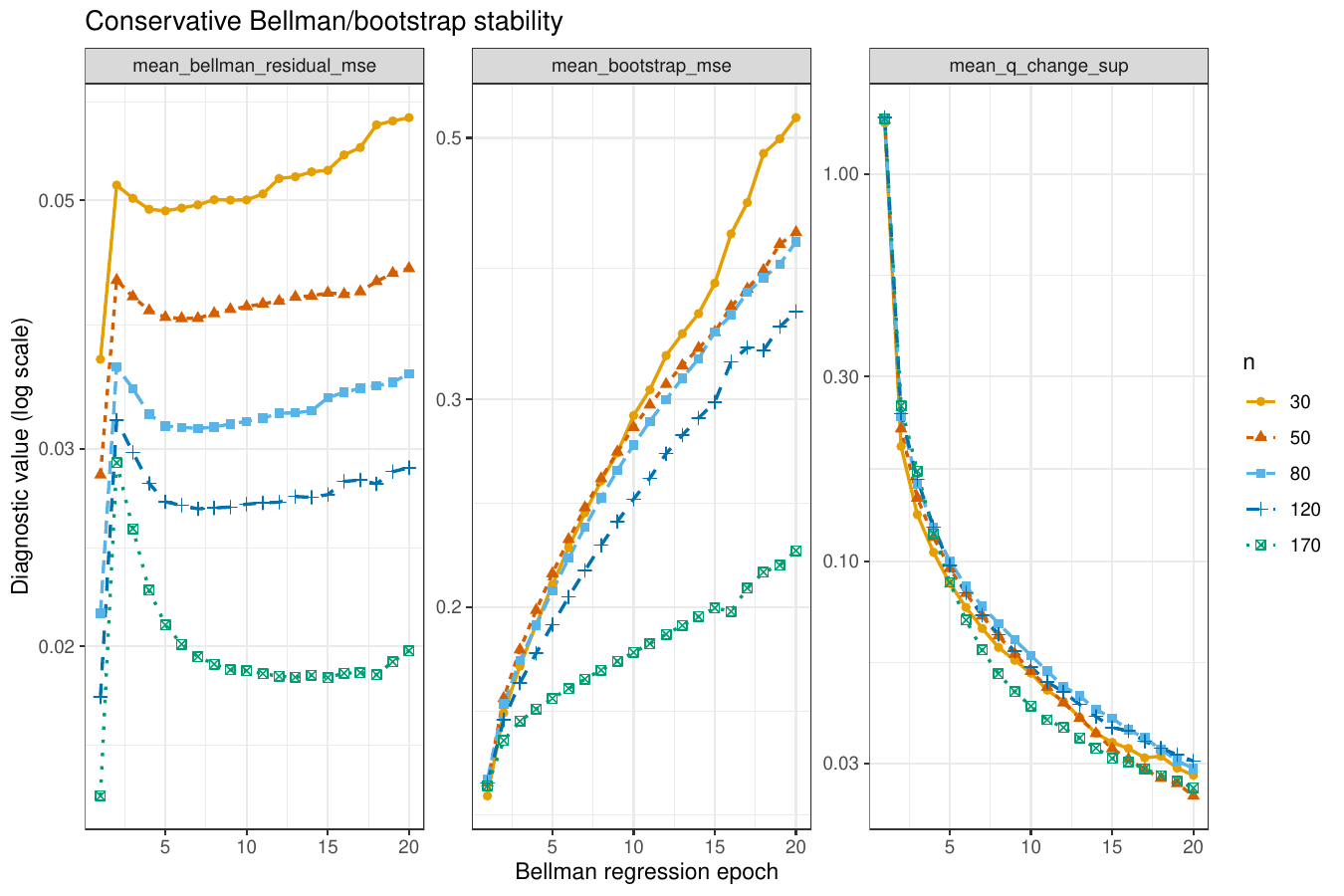}
\caption{Taskwise Bellman diagnostics over epochs. Panels display the unrelaxed algorithm-aligned operator residual, the discrepancy between bootstrapped continuation values and the known optimum, and successive fitted-function change. Epoch-specific averages include runs reaching that epoch, whereas Table~\ref{sim:tab:stability} averages retained-epoch diagnostics over all runs.}
\label{sim:fig:bootstrap}
\end{figure}

Figure~\ref{sim:fig:bootstrap} shows that successive fitted-function changes decrease sharply from their initial values and reach approximately $0.02$--$0.03$ near the end of the iterations. Table~\ref{sim:tab:stability} reports the diagnostics at the retained epoch. The mean algorithm-aligned Bellman residual decreases from 0.0546 to 0.0191 as $n$ increases. The mean bootstrap discrepancy also decreases across sample sizes, from 0.3993 to 0.2149, although it generally increases with epoch within a fixed sample size. The corresponding pooled descriptive exponents are 0.560 (nominal interval $[0.447,0.673]$) for $D_e$ and 0.297 ($[0.100,0.493]$) for $B_e$.
These quantities must be distinguished. Small $d_e$ indicates small successive updates; it does not imply small error relative to $Q_\star$. Here $B_e$ includes target pessimism as well as function-estimation and bootstrap error, and the reported summaries do not separate these contributions. Moreover, only 20.0\%--40.0\% of runs meet the stopping tolerance. No run activates the growth guard, and one run at $n=120$ retains an earlier epoch. The results support improved numerical behavior and smaller residuals with more data, but not verification that the bootstrap error converges to zero or that every fitted iteration satisfies the assumptions of Theorem~1.

\begin{table}[htbp]
\centering
\caption{Taskwise diagnostics at the retained epoch, averaged over 30 runs. Stop is the percentage satisfying $d_e\le0.015$. $B_e$, $D_e$, and $U_e$ are defined in Section~\ref{sim:evaluation}; $U_e$ is a finite-grid maximum averaged over runs.}
\label{sim:tab:stability}
\small\setlength{\tabcolsep}{5pt}
\begin{tabular}{rrrrrrr}
\toprule
$n$ & Epoch & Stop (\%) & $d_e$ & $B_e$ & $D_e$ & $U_e$ \\
\midrule
30 & 17.93 & 40.0 & 0.0227 & 0.3993 & 0.0546 & 1.641 \\
50 & 19.47 & 23.3 & 0.0230 & 0.3862 & 0.0424 & 1.662 \\
80 & 19.37 & 20.0 & 0.0272 & 0.3832 & 0.0338 & 1.618 \\
120 & 18.90 & 26.7 & 0.0273 & 0.3310 & 0.0278 & 1.507 \\
170 & 19.03 & 23.3 & 0.0231 & 0.2149 & 0.0191 & 1.247 \\
\bottomrule
\end{tabular}

\end{table}

\FloatBarrier

\subsubsection{Finite-range contraction slopes}

\begin{figure}[htbp]
\centering
\includegraphics[width=\linewidth]{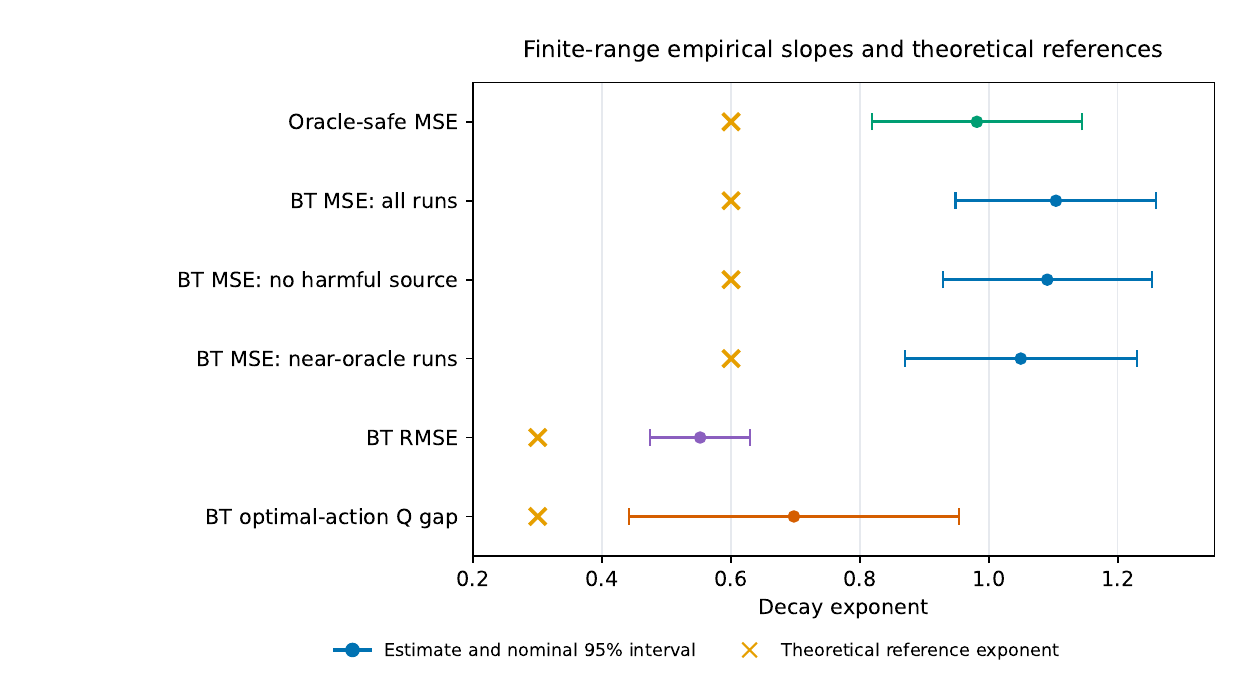}
\caption{Empirical finite-range slopes and theoretical references, redrawn from the supplied rate CSV without changing estimates or intervals. The MSE reference is 0.6 and the root-error/policy reference is 0.3. Faster observed slopes for this finite spectral ensemble do not establish a minimax rate or posterior contraction in probability.}
\label{sim:fig:rates}
\end{figure}

Table~\ref{sim:tab:rates} and Figure~\ref{sim:fig:rates} summarize the seed-fixed-effect regressions. Over the nominal range $N_{\mathrm{nom}}=90$--$510$, the estimated BT MSE exponent is 1.104, with a reported nominal interval of $[0.948,1.260]$. The restricted oracle exponent is 0.981 ($[0.819,1.144]$). The BT root-error exponent is exactly half the MSE exponent, 0.552 ($[0.474,0.630]$), because the corresponding run-level response is the square root of the same MSE. It is therefore not an independent rate confirmation. The action-gap exponent is 0.698 ($[0.442,0.953]$).

\begin{table}[htbp]
\centering
\caption{Seed-fixed-effect log--log fits. Ref. gives the displayed theoretical reference, not an equality constraint. Intervals and $R^2$ are reproduced from the ordinary linear-model output; intervals are not seed-cluster-adjusted. The conditional run subsets are secondary descriptive analyses.}
\label{sim:tab:rates}
\small\setlength{\tabcolsep}{4pt}
\begin{tabular}{lrrcrr}
\toprule
Outcome / run subset & Ref. & $\widehat\beta$ & Nominal 95\% CI & $R^2$ & Rows \\
\midrule
Oracle-safe MSE & 0.600 & 0.981 & $[0.819,\,1.144]$ & 0.706 & 150 \\
BT MSE (all runs) & 0.600 & 1.104 & $[0.948,\,1.260]$ & 0.734 & 150 \\
BT MSE (no harmful) & 0.600 & 1.091 & $[0.928,\,1.253]$ & 0.731 & 144 \\
BT MSE (near-oracle) & 0.600 & 1.050 & $[0.869,\,1.230]$ & 0.712 & 131 \\
BT RMSE & 0.300 & 0.552 & $[0.474,\,0.630]$ & 0.734 & 150 \\
BT action gap & 0.300 & 0.698 & $[0.442,\,0.953]$ & 0.543 & 150 \\
\bottomrule
\end{tabular}

\end{table}

The slopes exceed the displayed nonparametric references over this finite range. This is not a test that the estimates equal the theoretical exponents. Theorem~3 concerns a posterior contraction guarantee under its assumptions, whereas the plotted curves measure the error of posterior means after finite, conservative bootstrapping. The truth ensemble has a fixed 80-term truncation and represents random functions rather than a worst-case class. In addition, the mean final selected source count rises from 1.2 to 2.9, so the information used by the estimator does not remain a constant multiple of the nominal core-source count across the finite grid. These features preclude interpreting the fitted exponent as an exact minimax exponent. The increased $R^2$ of the fixed-effect fit also includes the seed intercepts; by itself it does not measure the quality of a universal power law. The result establishes a decreasing average estimation-error pattern in the reported range, not a separate proof of the posterior theorem.
\FloatBarrier

\subsection{Discussion of simulation results}

The unified simulation benchmark provides empirical evidence
for the benefits of target-aware Bayesian transfer in offline
Q-learning with heterogeneous source tasks. As the available
training data increase, the source-selection mechanism becomes
more effective at distinguishing informative sources from
those exhibiting substantial task mismatch. Closely related
sources are increasingly retained, while the inclusion of
designated harmful sources becomes less frequent. This
improved source discrimination is accompanied by more
accurate target Q-function estimation. Across all evaluated
sample sizes, BT-GP-OQL achieves lower mean estimation error
than both target-only learning and indiscriminate all-source
transfer. Moreover, its estimation error progressively
approaches that of the restricted oracle reference, suggesting
that target-predictive source selection can exploit useful
cross-task information while limiting the adverse effects
of task heterogeneity.

The policy-level results further highlight the importance
of combining source selection with uncertainty-aware
pessimism. Compared with indiscriminate source sharing,
BT-GP-OQL substantially reduces the observed frequency
and magnitude of material negative transfer, although
performance degradation relative to target-only learning
is not eliminated in every finite-sample setting.
The pessimistic policy-extraction rule also achieves
higher average decision quality than its non-pessimistic
counterpart when applied to the same fitted GP posterior.
In particular, penalizing posterior uncertainty reduces
the selection of the rarely observed fourth action,
which is more susceptible to estimation uncertainty.
Both the optimal-action Q gap and the truncated rollout
gap decrease as the training sample size increases,
indicating improved decision quality. These findings
illustrate the complementary roles of target-aware
source selection in controlling cross-task mismatch
and posterior pessimism in addressing uncertainty
associated with limited data coverage.

The experiments also reveal decreasing estimation errors
and improved Bellman-update diagnostics as more data
become available. The observed empirical decay rates
are favorable over the investigated sample-size range,
while the Bellman residual and bootstrap discrepancy
generally decrease with increasing sample size.
Nevertheless, these numerical patterns should be
distinguished from the asymptotic guarantees established
in the theoretical analysis. In particular, the empirical
inclusion scores are approximate predictive weights
rather than exact source-subset posterior probabilities,
and the observed error-decay exponents are based on
a finite-dimensional function ensemble and nominal
sample sizes. They therefore do not directly verify
posterior contraction or minimax optimality under
the general assumptions of the theoretical framework.
Furthermore, the finite-sample results do not establish
unconditional protection against negative transfer.
Taken together, the simulations support the practical
value of integrating target-aware source selection,
multi-task GP estimation, and uncertainty-aware
conservative policy learning, while clarifying the
statistical and empirical limitations of the proposed
offline transfer framework.

\section{Real data analysis}
\label{sec:real-data}

To complement the synthetic benchmark, we assess BT-GP-OQL using the
human decision-making data released with \citet{BenArtzi2023}. We address
two distinct empirical questions: whether cross-participant transfer
improves decision quality relative to learning from target data alone,
and whether target-aware selection identifies more useful sources than
uniformly choosing the same number from a fixed candidate pool. The
comparators are Target-only, BT-GP-OQL, and Random-$m$, evaluated at $\gamma\in\{0,0.5,0.95\}$. The actions
and payoffs are genuine behavioral observations, although the reward
schedules were generated by the experimental task rather than observed
in a financial market. Accordingly, this application evaluates immediate
action-selection quality on recorded histories; it does not identify
the counterfactual long-horizon return of deploying a learned policy.

The original study analyzed 178 participants in a four-armed bandit
experiment. Each participant completed four blocks of 50 trials, with
two of four cards offered on each trial. Reward probabilities drifted
over time, and feedback was available for the chosen card. Blocks
alternated between gain and loss conditions, with recorded payoffs in
$\{0,1\}$ and $\{-1,0\}$, respectively; different blocks used different
cards \citep{BenArtzi2023}. We start from the archived
\texttt{df\_raw.csv}, containing 40,600 trial records and 203 participant
identifiers, rather than the processed subset used for the original
behavioral comparisons \citep{BenArtzi2023Data}. Thus, the raw-data count
and the original study's final analysis sample are different quantities.
We check participant identifiers, trial order, offered and chosen cards,
and condition-specific reward ranges. Invalid choices and missing
rewards do not contribute fitting responses. For $\gamma>0$,
nonterminal responses require consecutive valid observations within the
same participant and block: missing trials and block boundaries are not
bridged, and an unavailable successor is not treated as a terminal state.
At $\gamma=0$, a valid immediate-reward observation can be used without a
valid successor. These eligibility rules distinguish usable reward
observations from usable transitions.

The preceding development protocol excluded 23 participants involved in
exploratory screening and used a separate 24-participant cohort for
calibration. The remaining 156 participants form the evaluation cohort.
The configuration denoted C10 was locked using development data and is
not recalibrated here. For each evaluation target $i$, a frozen,
outcome-independent assignment supplies seven distinct candidate source
participants, $C_i$, from the other evaluation participants. The target
never appears in its own source pool. The same assignments are used
across methods and discounts; development and excluded participants are not reintroduced as sources.
Each participant defines a constructed transfer task, with blocks
retained as separate episodes. Local card identities are combined with
block context without assuming that equally numbered cards have
identical reward schedules across participants. Every comparison uses
$K=7$ candidate sources, even though 156 targets are evaluated. Sources
can be shared between targets, so target-level analyses are not fully
independent. The matched-size comparison and the additional
$\gamma=0.5$ setting were introduced after the evaluation histories had
been examined; they are therefore reported as post-hoc mechanism and
sensitivity analyses, not as a new independent confirmation.

Within each block, using zero-based indices $t=0,\ldots,49$, trials
0--29 are used for GP fitting, 30--34 for source screening and the final
transfer gate, 35--38 for a later-validation diagnostic, and 40--49 for
target evaluation. Trial 39 is not a fitting or validation-scoring
response, but remains in the sequence and may supply the successor
state for trial 38. Each participant contributes at most 120 fitting
observations, so a joint fit with $m$ sources contains at most
$120(1+m)$ observations. Sources contribute only their fitting
partitions.
A fitting-partition boundary is not an episode terminal: the response
at trial 29 can use the pre-choice state at trial 30, but not trial 30's
current reward. Only trial 49 is a true block terminal. Early and late
validation use disjoint scoring responses, although their histories and
successor states remain temporally related. Earlier recorded rewards
may enter later pre-choice history summaries, but GP parameters remain
frozen during evaluation. Test scores and the archived reward
probabilities are excluded from fitting, screening, and gating.

For task $k$, block $b$, and trial $t$, let
$\mathcal O_{b,t}^{(k)}\subseteq\{0,1,2,3\}$ denote the offered pair.
The working state is
\begin{equation}\nonumber
 s_{b,t}^{(k)}=
 \left(
 \frac{t}{49},\ c_b^{(k)},\ \mathbf e_b,\
 \mathbf o_{b,t}^{(k)},\
 \frac{\mathbf n_{b,t}^{(k)}}{50},\
 \overline{\mathbf r}_{b,t}^{(k)}
 \right),
 \label{real:state}
\end{equation}
where $c_b^{(k)}$ indicates the gain condition, $\mathbf e_b$ encodes
the block, $\mathbf o_{b,t}^{(k)}$ indicates available cards, and
$\mathbf n_{b,t}^{(k)}$ and $\overline{\mathbf r}_{b,t}^{(k)}$ contain
past selection counts and mean observed rewards. Histories reset at
block boundaries. Unselected cards have count and mean initialized to
zero, with the count distinguishing no experience from an observed zero
mean. Only pre-choice information enters the state; this representation
is not an assertion of a fully observed stationary Markov decision
process.
All maximizations are restricted to offered cards. A ridge-penalized
choice model, fitted only to target training choices, retains offered
actions whose estimated choice probability is at least $0.01$. If
thresholding removes both cards, the offered card with highest estimated
probability is retained. All three target comparators use this same
supported set, denoted $\mathcal C_i(s)$. Sources use their offered pair
without the target-specific support restriction. Random presentation of
a pair is not interpreted as a uniform participant choice propensity.

Taskwise models, source scores, and selected subsets are recomputed
separately for each $\gamma\in\{0,0.5,0.95\}$. The unrelaxed Bellman
response is
\begin{equation}\nonumber
 z_{b,t,e,\gamma}^{(k)}=
 \begin{cases}
 r_{b,t}^{(k)}, & t=49,\\[2pt]
 r_{b,t}^{(k)}+\gamma
 \displaystyle\max_{a\in\mathcal C_{b,t+1}^{(k)}}
 \left\{
 \mu_{e-1,\gamma}^{(k)}(s_{b,t+1}^{(k)},a)
 -\alpha_k\sigma_{e-1,\gamma}^{(k)}(s_{b,t+1}^{(k)},a)
 \right\}, & t<49,
 \end{cases}
 \label{real:bellman}
\end{equation}
where $\mathcal C^{(0)}$ is the supported target offer set,
$\mathcal C^{(k)}=\mathcal O^{(k)}$ for sources, $\alpha_0=0.75$, and
$\alpha_k=0$ for $k\geq1$. The nonterminal expression requires a valid
successor, and continuation is initialized to zero. At $\gamma=0$, a
single GP regression fits immediate rewards. Positive discounts use
response relaxation with weight $0.45$, at most 20 epochs, and the
change-based stopping and rollback rules of the implementation. In
particular, the $\gamma=0.5$ results arise from actual refitting, not
interpolation between the other settings.
The input kernel combines a Mat\'ern-$3/2$ state kernel (length scale
$0.30$, signal standard deviation $0.50$) with an action factor equal to
one for matching actions and $0.03$ otherwise. The joint task
correlation matrix has unit diagonal and constant off-diagonal
correlation: $0.88$ for single-source screening and $0.92$ for final
joint fitting. Exact GP linear algebra is used. Target working
variances are $0.25+\sigma_{e-1,\gamma}^2(s,a)$; source working
variances are $0.75+0.10\sigma_{e-1,\gamma}^2(s,a)$. These are
pseudo-response working variances, not a Gaussian generative model for
binary payoffs. Final joint fits combine retained taskwise
pseudo-responses and variances without adding validation or test
responses to GP training.

For each target and discount, early-validation pseudo-responses and
working variances are constructed from the training-only target
continuation estimate and held fixed across candidates. Let
$\ell_{i,\gamma}^{0}$ be the summed univariate Gaussian log predictive
density on trials 30--34 under Target-only, and
$\ell_{ik,\gamma}^{\mathrm{pair}}$ the corresponding score after
including source $k$ with the screening task correlation. Write
$\Delta\ell_{ik,\gamma}=\ell_{ik,\gamma}^{\mathrm{pair}}
-\ell_{i,\gamma}^{0}$. With $n_{i,\gamma}^{\mathrm{fit}}$ usable target
fitting responses, C10 uses
\begin{equation}\nonumber
 \widetilde p_{ik,\gamma}=
 \operatorname{logit}^{-1}\!\left[
 \operatorname{logit}(0.50)+1.20\,\Delta\ell_{ik,\gamma}
 -0.04\,n_{i,\gamma}^{\mathrm{fit}}\widehat D_{ik,\gamma}
 \right],
 \label{real:source-score}
\end{equation}
where $\widehat D_{ik,\gamma}$ is the nonnegative empirical squared
maximum mean discrepancy between target and source training inputs,
using the radial-basis kernel and median bandwidth rule in the
implementation. Sources with $\widetilde p_{ik,\gamma}\geq0.5$ form
$S_{i,\gamma}^{\mathrm{pre}}$. This quantity is a tempered predictive
inclusion score, not an exact full spike-and-slab subset posterior.
For a nonempty preliminary subset, a final joint fit is compared with
Target-only on the same early-validation responses:
\begin{equation}\nonumber
 q_{i,\gamma}=
 \operatorname{logit}^{-1}\!\left[
 \operatorname{logit}(0.50)+0.70\left\{
 \ell_{i,\gamma}^{\mathrm{joint}}(S_{i,\gamma}^{\mathrm{pre}})
 -\ell_{i,\gamma}^{0}\right\}
 \right].
 \label{real:gate}
\end{equation}
The final subset $\widehat S_{i,\gamma}$ equals
$S_{i,\gamma}^{\mathrm{pre}}$ when $q_{i,\gamma}\geq0.5$ and is empty
otherwise. An empty preliminary subset also yields Target-only, without
computing a gate score. Screening and gating reuse early validation, so
the gate is a model-selection rule, not independent evidence of
finite-sample safety. C10 and all thresholds are fixed across discounts.

Target-only uses target fitting data; BT-GP-OQL additionally uses
$\widehat S_{i,\gamma}$. To isolate the contribution of source identity
at a matched source count, set $m_{i,\gamma}=|\widehat S_{i,\gamma}|$ and
let
\begin{equation}\nonumber
 \mathfrak S_{i,\gamma}
 =\{S\subseteq C_i:|S|=m_{i,\gamma}\},
 \qquad |\mathfrak S_{i,\gamma}|=\binom{7}{m_{i,\gamma}}.
 \label{real:matched-subsets}
\end{equation}
Random-$m$ fits every subset in $\mathfrak S_{i,\gamma}$ as a separate
policy, without an additional screening or gate step. All methods use
identical target test records, GP settings, supported actions, and the
decision rule
\begin{equation}\nonumber
 \widehat\pi_{i,S,\gamma}(s)
 =\arg\max_{a\in\mathcal C_i(s)}
 \left\{\widehat\mu_{i,S,\gamma}(s,a)
 -0.75\widehat\sigma_{i,S,\gamma}(s,a)\right\}.
 \label{real:policy}
\end{equation}
Thus, this design compares transfer and source identity selection; it
does not separately identify the effect of decision-stage pessimism.
Setting \texttt{MECH\_RANDOM\_DRAWS = 35L} permits complete enumeration
because $\max_{0\leq m\leq7}\binom7m=35$. Actual subset counts and
uniqueness are checked. Random-$m$ is the exact uniform mean of the
separate subset-policy scores, not a posterior-mixture policy, majority
vote, or test-selected best subset. It uses real source participants.
When $m_{i,\gamma}=0$, both BT-GP-OQL and Random-$m$ coincide with
Target-only.

The archived \texttt{prob1} and \texttt{prob2} fields contain the
favorable-outcome probabilities of the historically chosen and unchosen
cards \citep{BenArtzi2023Data}. They are mapped to the actual offered
card identities and accessed only after the fitted model and its test
actions have been frozen. They are reward probabilities, not
behavior-policy propensities. For test record $j$ of target $i$,
\begin{equation}\nonumber
 m_{ij}(a)=
 \begin{cases}
 p_{ij}(a), & \text{gain condition},\\
 p_{ij}(a)-1, & \text{loss condition},
 \end{cases}
 \label{real:expected-reward}
\end{equation}
where the favorable loss outcome is zero. On the common evaluable test
set $\mathcal I_i$, the expected immediate payoff and optimal-action gap
are
\[
 \widehat J_{i,\gamma}(\pi)
 =\frac{1}{|\mathcal I_i|}\sum_{j\in\mathcal I_i}
 m_{ij}\!\left(\pi(s_{ij})\right),
\qquad
 \widehat G_{i,\gamma}(\pi)
 =\frac{1}{|\mathcal I_i|}\sum_{j\in\mathcal I_i}
 \left[\max_{a\in\mathcal O_{ij}}m_{ij}(a)
 -m_{ij}\!\left(\pi(s_{ij})\right)\right].
\]
The reference maximum uses the full offered pair, so the gap includes
any loss caused by support restrictions. On identical test records,
$\widehat J+\widehat G$ is the method-independent offered-pair oracle
score. Payoff and gap are therefore complementary expressions of the
same immediate action-selection quality, not independent confirmations.
For Random-$m$, define
\[
 \overline J_{i,\gamma}^{\mathrm{rand}}
 =\binom{7}{m_{i,\gamma}}^{-1}
 \sum_{S\in\mathfrak S_{i,\gamma}}
 \widehat J_{i,\gamma}(\widehat\pi_{i,S,\gamma}),
\]
with an analogous gap. Subsets are averaged within target before targets
receive equal weight. With $N_{\mathrm{eval}}=156$, our principal
contrasts are
\[
 \widehat\Delta_{\mathrm{transfer}}(\gamma)
 =\frac{1}{N_{\mathrm{eval}}}\sum_{i=1}^{N_{\mathrm{eval}}}
 \left(\widehat J_{i,\gamma}^{\mathrm{BT}}
 -\widehat J_{i,\gamma}^{\mathrm{target}}\right),
 \label{real:transfer-contrast}\qquad
 \widehat\Delta_{\mathrm{selection}}(\gamma)
 =\frac{1}{N_{\mathrm{eval}}}\sum_{i=1}^{N_{\mathrm{eval}}}
 \left(\widehat J_{i,\gamma}^{\mathrm{BT}}
 -\overline J_{i,\gamma}^{\mathrm{rand}}\right).
\]
The first measures transfer relative to target-only fitting; the second
measures source identity selection conditional on the source count
chosen by BT-GP-OQL.
On trials 35--38, we evaluate single-source log predictive density (LPD)
gains relative to Target-only, using frozen target-only continuation
responses for each discount. Within each target, these per-trial gains
are averaged separately over finally selected and excluded sources,
and their difference is then averaged across eligible targets. Cases
lacking either group are omitted from this diagnostic, not coded as
zero; they remain in the main policy comparisons. This assesses
prediction of working Bellman responses rather than accuracy against a
known long-horizon value function. The early and later windows have
disjoint scoring responses but retain temporal dependence.
The intervals displayed in this section are 95\% percentile intervals
from 2,000 fixed-fit target-participant bootstrap draws. Subset scores
are averaged within target before resampling, and each draw uses the
same target indices across methods and discounts. These intervals are
descriptive and conditional on the fitted models and frozen source
assignments. They do not propagate all uncertainty from shared sources
or model refitting, and the enumerated subsets are not treated as
additional participants. Although the implementation also supports a
separate participant-cohort resampling and joint-refitting procedure,
those source-aware intervals are not the intervals shown in the present
three-discount figures. In particular, no endpoint-discount interval is
reused for $\gamma=0.5$.

\begin{figure}[!htbp]
\centering
\includegraphics[width=0.92\linewidth]{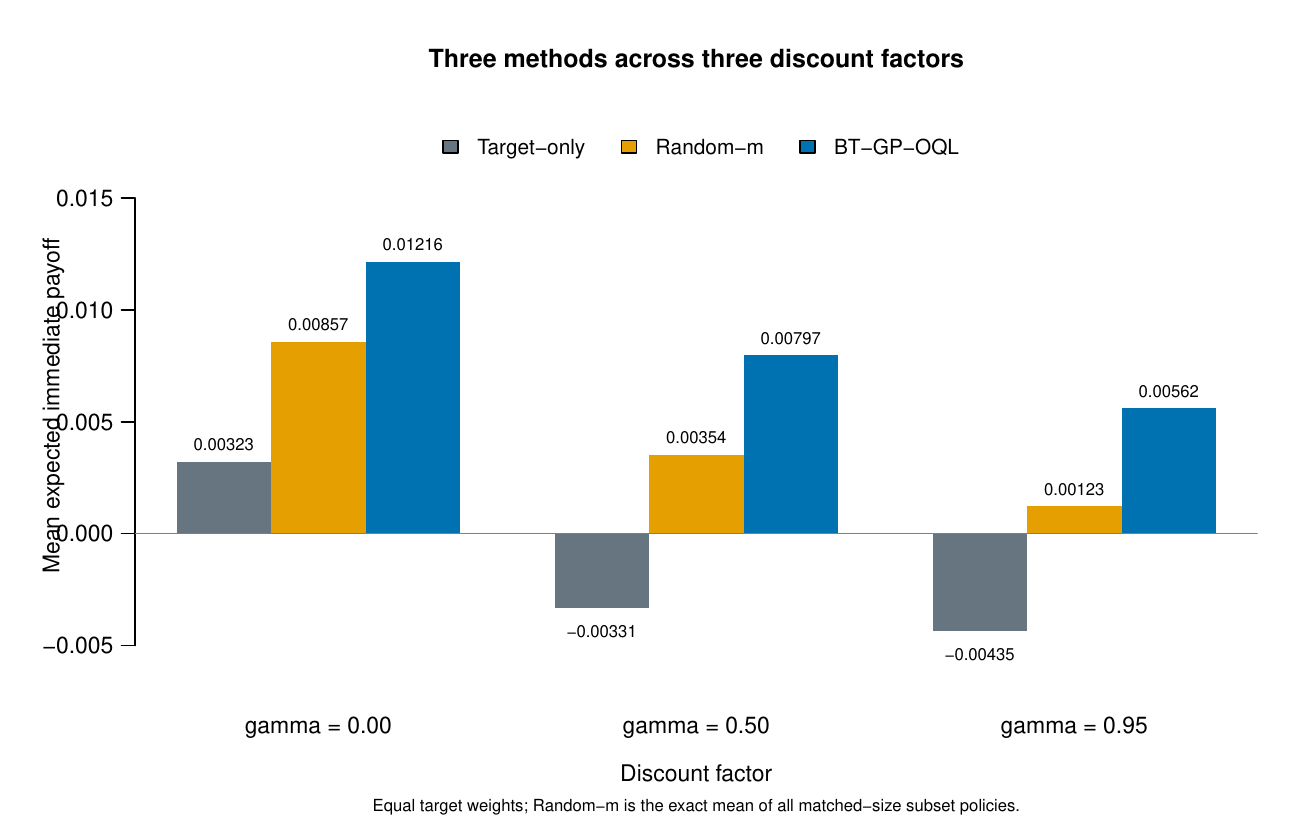}
\caption{Mean expected immediate payoff for Target-only (gray),
Random-$m$ (orange), and BT-GP-OQL (blue). Each discount is fitted
separately with C10 and the seven-source assignments fixed. These bars
show equal-target means; they do not carry uncertainty intervals.}
\label{fig:tg-payoff}
\end{figure}

The analysis covers 156 evaluation targets at each of the three
discounts, yielding 468 target--discount groups before subset
enumeration. Table~\ref{tab:tg-methods} and
Figure~\ref{fig:tg-payoff} show the same ordering of mean expected
immediate payoff at every discount:
\begin{equation}
 \overline J_{\mathrm{BT}}(\gamma)
 >\overline J_{\mathrm{Random}\text{-}m}(\gamma)
 >\overline J_{\mathrm{Target-only}}(\gamma),
 \qquad \gamma\in\{0,0.5,0.95\}.
 \label{real:observed-ordering}
\end{equation}
At $\gamma=0$, the corresponding means are $0.01216$, $0.00857$, and
$0.00323$. At the added middle discount, they are $0.00797$, $0.00354$,
and $-0.00331$, respectively. At $\gamma=0.95$, they are $0.00562$,
$0.00123$, and $-0.00435$. Thus, the observed advantage is not restricted
to the two previously examined endpoint settings.

\begin{table}[!htbp]
\centering
\small
\caption{Three-method performance on 156 evaluation targets at each discount.}
\label{tab:tg-methods}
\begin{adjustbox}{max width=\linewidth}
\begin{tabular}{clrr}
\toprule
$\gamma$ & Method & Mean immediate payoff & Mean immediate gap \\
\midrule
0.00 & Target-only & $0.00323$ & $0.07968$ \\
     & Random-$m$ & $0.00857$ & $0.07433$ \\
     & BT-GP-OQL & $0.01216$ & $0.07075$ \\
\addlinespace
0.50 & Target-only & $-0.00331$ & $0.08622$ \\
     & Random-$m$ & $0.00354$ & $0.07937$ \\
     & BT-GP-OQL & $0.00797$ & $0.07494$ \\
\addlinespace
0.95 & Target-only & $-0.00435$ & $0.08725$ \\
     & Random-$m$ & $0.00123$ & $0.08168$ \\
     & BT-GP-OQL & $0.00562$ & $0.07729$ \\
\bottomrule
\end{tabular}
\end{adjustbox}
\par\smallskip
\begin{minipage}{0.97\linewidth}
\footnotesize
Targets receive equal weight. Random-$m$ averages all matched-size
subset-policy scores within each target first. Higher payoff and lower
gap are preferable. Values are rounded to five decimal places; bold
entries identify the best cohort mean among the three reported methods,
not uniform superiority across individuals. The outcomes are immediate
scores on logged histories, not closed-loop discounted returns.
\end{minipage}
\end{table}

\begin{figure}[!htbp]
\centering
\includegraphics[width=0.92\linewidth]{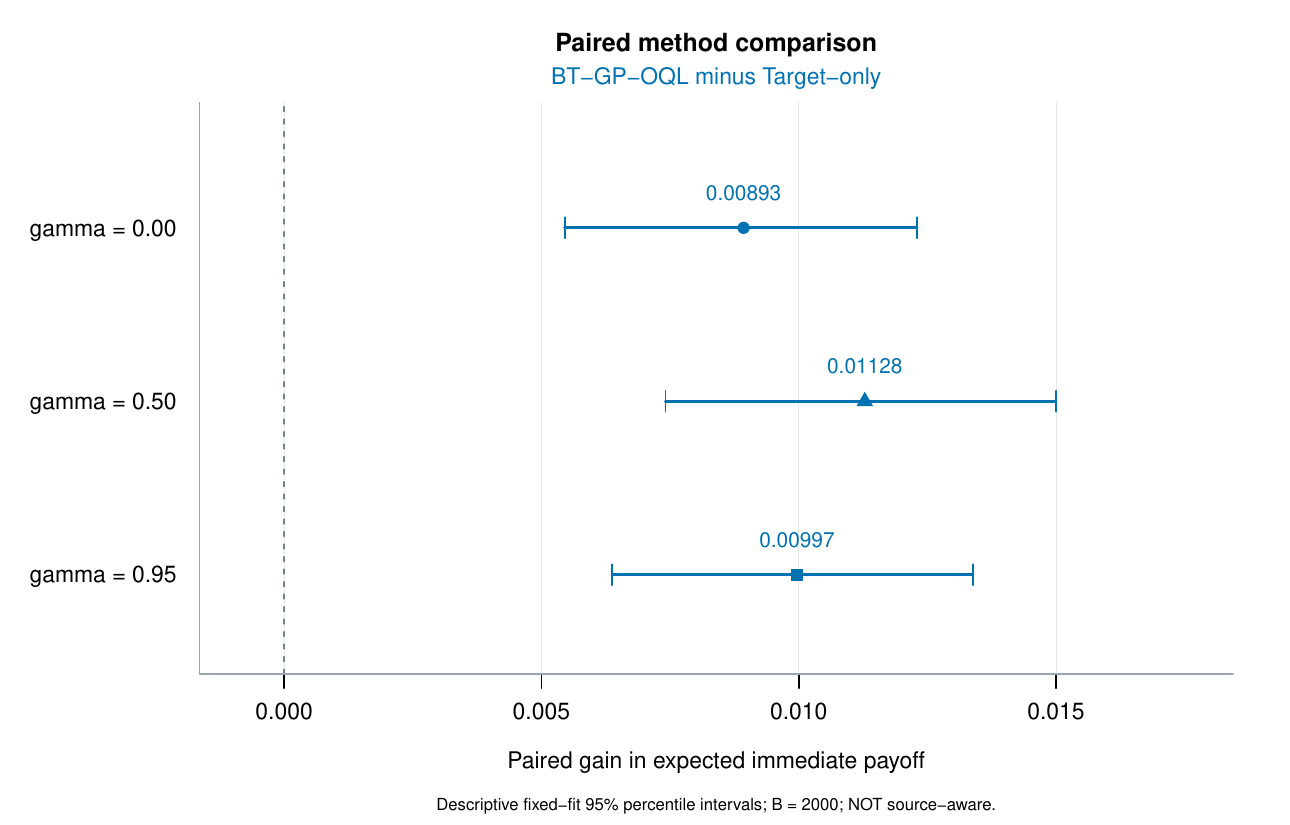}
\caption{BT-GP-OQL minus Target-only at the three discount settings.
Points show mean paired gains in expected immediate payoff; horizontal
bars are descriptive fixed-fit target-participant bootstrap 95\%
percentile intervals based on 2,000 draws. The display does not account
fully for shared-source and model-refitting uncertainty.}
\label{fig:tg-target}
\end{figure}

Table~\ref{tab:tg-contrasts} and Figure~\ref{fig:tg-target} isolate the
first research question. The mean BT-GP-OQL gains over Target-only are
$0.00893$, $0.01128$, and $0.00997$ at $\gamma=0$, $0.5$, and $0.95$,
respectively. The displayed fixed-fit percentile intervals lie above
zero in all three cases. These results provide descriptive support for
positive average transfer gains under the frozen source assignments
and fitted models. They do not imply that every participant benefits
or that the transfer gate guarantees finite-sample safety.

\begin{table}[!htbp]
\centering
\small
\caption{The two principal mean payoff contrasts at each discount.}
\label{tab:tg-contrasts}
\begin{adjustbox}{max width=\linewidth}
\begin{tabular}{crr}
\toprule
$\gamma$ & BT-GP-OQL $-$ Target-only & BT-GP-OQL $-$ Random-$m$ \\
\midrule
0.00 & $0.00893$ & $0.00358$ \\
0.50 & $0.01128$ & $0.00443$ \\
0.95 & $0.00997$ & $0.00439$ \\
\bottomrule
\end{tabular}
\end{adjustbox}
\par\smallskip
\begin{minipage}{0.97\linewidth}
\footnotesize
All contrasts use 156 target-level differences. Their 95\% descriptive
fixed-fit percentile intervals ($B=2{,}000$) appear in
Figures~\ref{fig:tg-target} and~\ref{fig:tg-random}; these are not
source-aware cohort-refit intervals. Contrasts are calculated before rounding; their last reported digit
can differ from the difference of the separately rounded method means
in Table~\ref{tab:tg-methods}.
\end{minipage}
\end{table}

\begin{figure}[!htbp]
\centering
\includegraphics[width=0.92\linewidth]{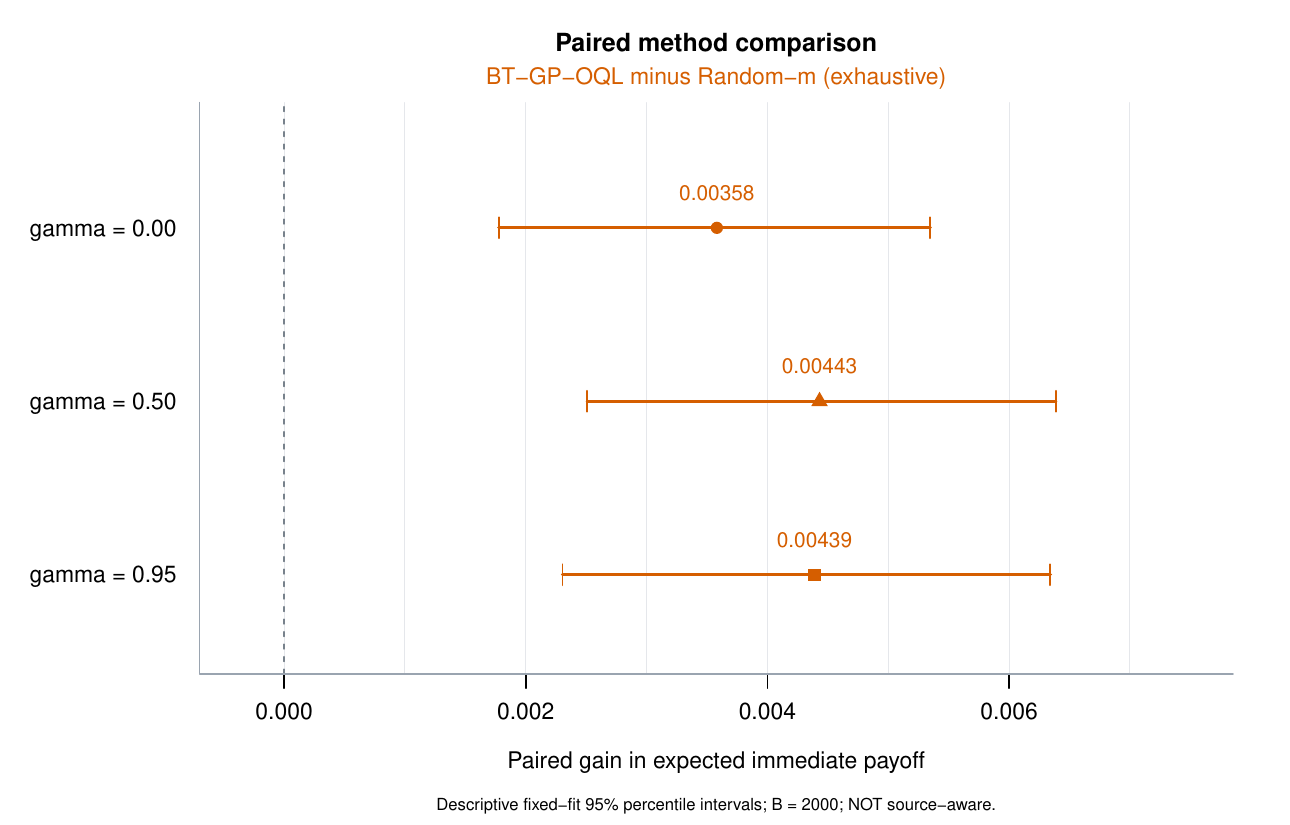}
\caption{BT-GP-OQL minus the exhaustive matched-size Random-$m$
comparator. Positive differences favor target-aware source selection at
the source count chosen by BT-GP-OQL. Horizontal bars are 95\%
descriptive fixed-fit target-participant bootstrap percentile intervals
($B=2{,}000$), not source-aware intervals. This mechanism comparison is
post-hoc and exploratory.}
\label{fig:tg-random}
\end{figure}

The second contrast holds the candidate pool and the number of sources
fixed within each target--discount group. BT-GP-OQL exceeds the exact
uniform matched-size Random-$m$ mean by $0.00358$, $0.00443$, and
$0.00439$ at the three respective discounts
(Table~\ref{tab:tg-contrasts}). The descriptive intervals in
Figure~\ref{fig:tg-random} also lie above zero. Thus, the selected source
identities yield better cohort-average immediate payoff than uniform
same-size selection in each examined setting. Because every matched-size
subset is included, this result is not an artifact of drawing a
particularly unfavorable Monte Carlo sample of random subsets.
Nevertheless, it remains a post-hoc comparison on previously examined
histories: exhaustive enumeration removes subset-sampling error, not
participant dependence or selection of the analysis itself. The result
concerns the mean over matched-size subsets, not the best subset for
each participant.

\begin{figure}[!htbp]
\centering
\includegraphics[width=0.92\linewidth]{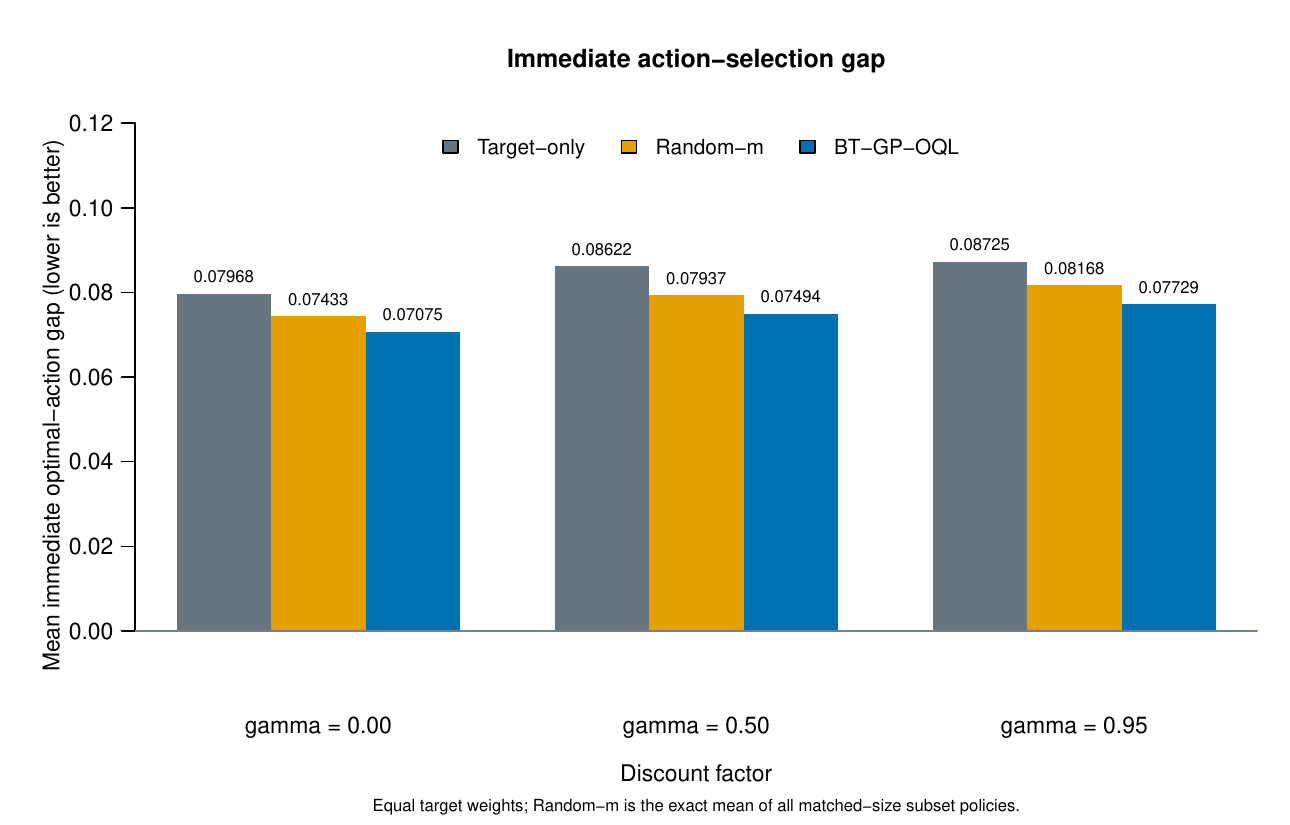}
\caption{Mean immediate optimal-action gap for Target-only (gray),
Random-$m$ (orange), and BT-GP-OQL (blue); lower values are preferable.
The reference is the best of the full offered pair, so the gap includes
any loss caused by behavior-support restrictions. On identical records,
payoff and gap sum to the same method-independent oracle score.}
\label{fig:immediate_gap}
\end{figure}

The gap results in Table~\ref{tab:tg-methods} and
Figure~\ref{fig:immediate_gap} are consistent with the payoff ordering.
BT-GP-OQL has the smallest mean gap at each discount: $0.07075$,
$0.07494$, and $0.07729$, compared with $0.07968$, $0.08622$, and
$0.08725$ for Target-only. Random-$m$ lies between them. As noted above,
this gap reduction is the complementary expression of higher payoff on
the same test records, not a statistically independent outcome.
The middle discount preserves both positive mean contrasts, with a
transfer gain of $0.01128$ and a matched-size selection gain of
$0.00443$. It therefore extends the observed direction of the results
to a setting between the original endpoints. It does not identify an
optimal discount: BT-GP-OQL's absolute mean immediate payoff is higher
at $\gamma=0$ than at $0.5$ or $0.95$, while its largest observed gain
relative to the discount-specific Target-only comparator occurs at
$\gamma=0.5$. Absolute payoff and improvement over a moving baseline
are different comparisons. No conclusion about the best long-horizon
discount follows from these immediate-reward scores.

\begin{figure}[!htbp]
\centering
\includegraphics[width=0.92\linewidth]{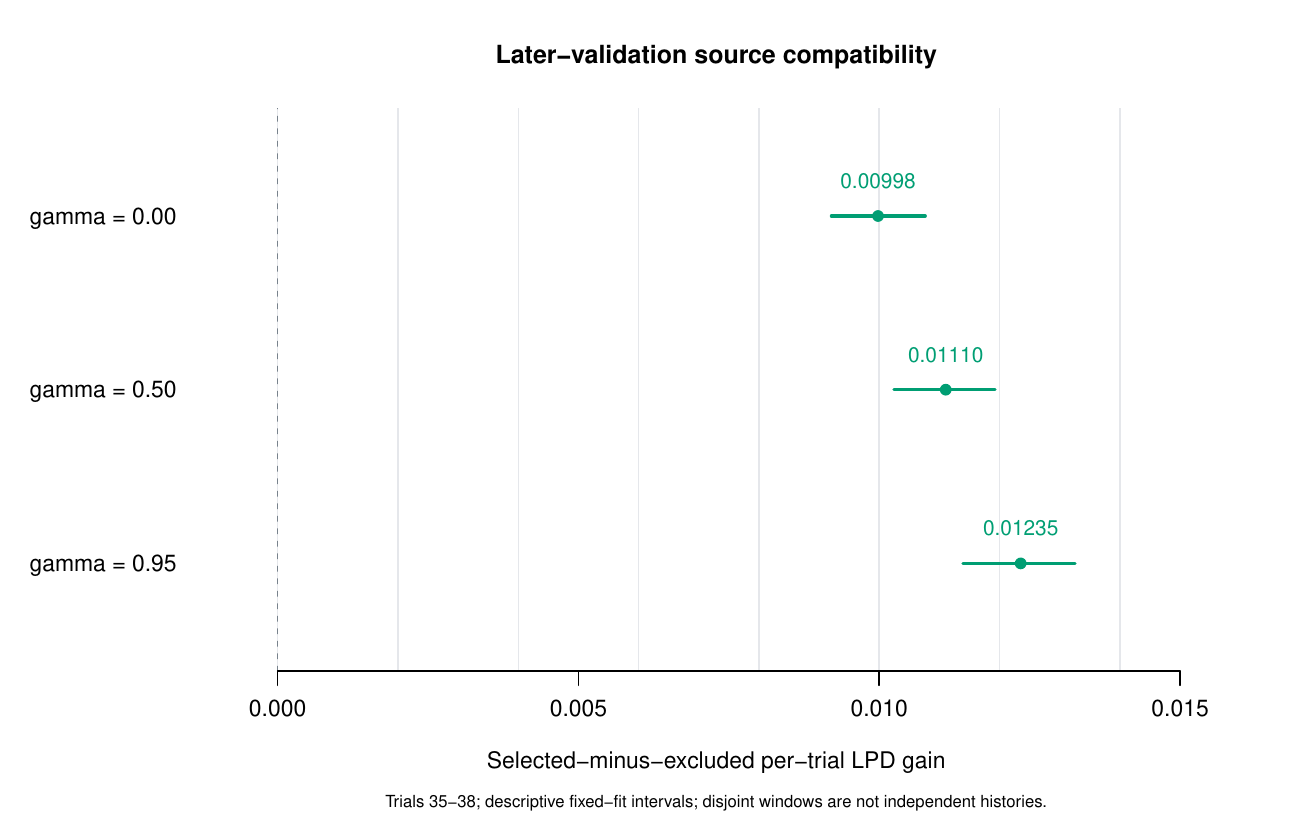}
\caption{Later-validation predictive compatibility: mean selected-minus-excluded
single-source LPD gain per trial, with descriptive fixed-fit 95\%
percentile intervals. Positive values favor the selected sources on
responses not used for their early screening. The comparison uses
disjoint scoring windows from the same histories and does not classify
all excluded sources as harmful.}
\label{fig:validation_compatibility}
\end{figure}

Table~\ref{tab:tg-late} and Figure~\ref{fig:validation_compatibility}
provide an auxiliary check on the source-selection mechanism. Selected
sources exceed excluded sources in mean later-validation per-trial LPD
gain by $0.00998$, $0.01110$, and $0.01235$ at $\gamma=0$, $0.5$, and
$0.95$, respectively. The displayed descriptive intervals remain above
zero. Thus, the ranking obtained from early validation also separates
sources with stronger and weaker predictive compatibility in the later
window, rather than reflecting only fit to the screening responses.
This is supporting predictive evidence, not proof that excluded sources
are harmful or that the inclusion-score construction dominates other
ranking rules. The later responses concern working Bellman targets,
and the two windows remain temporally dependent.

\begin{table}[!htbp]
\centering
\small
\caption{Selected-minus-excluded predictive gain on later-validation trials.}
\label{tab:tg-late}
\begin{adjustbox}{max width=\linewidth}
\begin{tabular}{cr}
\toprule
$\gamma$ & Mean selected-minus-excluded per-trial LPD gain \\
\midrule
0.00 & $0.00998$ \\
0.50 & $0.01110$ \\
0.95 & $0.01235$ \\
\bottomrule
\end{tabular}
\end{adjustbox}
\par\smallskip
\begin{minipage}{0.97\linewidth}
\footnotesize
Single-source gains are relative to Target-only on trials 35--38 after
screening on trials 30--34. Source gains are averaged separately within
target over selected and excluded sources before computing the contrast.
Only targets with both groups enter this diagnostic. The descriptive
fixed-fit intervals are shown in
Figure~\ref{fig:validation_compatibility}; temporal dependence between
the windows remains.
\end{minipage}
\end{table}

\clearpage

The three-discount results support two focused observations: cross-participant
information sharing improves cohort-average immediate payoff relative
to Target-only, and target-aware source choice improves upon the exact
uniform matched-size comparator. The direction of both contrasts is
preserved at the supplementary $\gamma=0.5$ setting, with no
recalibration of C10. These are average, protocol-specific findings,
not claims of improvement for every participant or subset-size stratum.
They do not establish universal negative-transfer protection,
computational savings, superiority to unreported methods, or a distinct
benefit of decision-stage pessimism. The fixed-fit intervals are
pointwise descriptive summaries, not simultaneous or source-aware
coverage guarantees. Finally, because evaluation conditions on logged
histories while fitted parameters remain frozen, it does not identify
a counterfactual closed-loop policy return. Within these limits, the
application supplies complementary behavioral evidence for transfer
and target-adaptive source selection alongside the known-value
synthetic experiments.

\FloatBarrier

\newpage

\bibliographystyle{plainnat}
\bibliography{reference}

% ============================================================
% Complete proofs (moved verbatim from the main text).
% ============================================================
\clearpage
\appendix
\numberwithin{equation}{section}
\section{Proofs of Bellman bootstrap stability and source selection}
\label{app:selection-proofs}

\subsection{Proof of Lemma~\ref{lem:markov_bellman_concentration}}
\label{app:proof:lem:markov_bellman_concentration}

\begin{proof}
For a bounded measurable function $f$, write
\[
    \operatorname{osc}(f)=\sup_x f(x)-\inf_x f(x),
    \quad
    (P_i f)(x)=\int f(y)\,P_i(\mathrm{d}y\mid x).
\]
The total-variation convention in the statement gives
\[
    |(P_i f)(u)-(P_i f)(v)|
    \leq
    \operatorname{osc}(f)\,
    \bigl\|P_i(\cdot\mid u)-P_i(\cdot\mid v)\bigr\|_{\mathrm{TV}}.
\]
Taking the supremum over $u$ and $v$ yields
\[
    \operatorname{osc}(P_i f)
    \leq\theta_i\operatorname{osc}(f).
\]
Define the expected remaining sum, as a function of the current state,
by
\[
    H_i(x)
    =
    h_i(x)
    +\sum_{j=i+1}^{N}
    (P_iP_{i+1}\cdots P_{j-1}h_j)(x),
    \quad i=1,\ldots,N.
\]
In particular, $H_N=h_N$.
Repeated application of the oscillation inequality gives
\[
\begin{aligned}
    \operatorname{osc}(H_i)
    \leq
    \operatorname{osc}(h_i)
    +\sum_{j=i+1}^{N}
    \left(\prod_{\ell=i}^{j-1}\theta_\ell\right)
    \operatorname{osc}(h_j)
    \leq
    2B\left(
        1+\sum_{j=i+1}^{N}
        \prod_{\ell=i}^{j-1}\theta_\ell
    \right)
    =2Bc_{i,N}.
\end{aligned}
\]
Let
\[
    \mathcal{G}_i=\sigma(U_1,\ldots,U_i),
    \quad i=1,\ldots,N,
\]
and let $\mathcal{G}_0$ be the trivial sigma-field. Define
\[
    M_i=\mathbb{E}[F_N\mid\mathcal{G}_i],
    \quad i=0,\ldots,N.
\]
By the tower property of conditional expectation,
$\{M_i\}_{i=0}^{N}$ is a martingale with respect to
$\{\mathcal{G}_i\}_{i=0}^{N}$. Moreover,
\[
    M_0=\mathbb{E}F_N,
    \quad
    M_N=F_N.
\]
Define the martingale differences
\[
    D_i=M_i-M_{i-1},
    \quad i=1,\ldots,N.
\]
Using the Markov property and the definition of $H_i$, we obtain
\[
    D_i
    =
    \frac{
        H_i(U_i)
        -
        \mathbb{E}[H_i(U_i)\mid\mathcal{G}_{i-1}]
    }{N}.
\]
Since
$\mathbb{E}[H_i(U_i)\mid\mathcal{G}_{i-1}]$
lies between $\inf_x H_i(x)$ and $\sup_x H_i(x)$,
\[
    |D_i|
    \leq
    \frac{\operatorname{osc}(H_i)}{N}
    \leq
    \frac{2Bc_{i,N}}{N}.
\]
Furthermore,
\[
    F_N-\mathbb{E}F_N
    =
    M_N-M_0
    =
    \sum_{i=1}^{N}D_i.
\]
Azuma--Hoeffding's inequality therefore gives
\[
    \mathbb{P}\!\left(
        |F_N-\mathbb{E}F_N|\geq u
    \right)
    \leq
    2\exp\!\left\{
        -\frac{u^2}
        {2\sum_{i=1}^{N}(2Bc_{i,N}/N)^2}
    \right\},
\]
which yields the first inequality in
\eqref{eq:boot-markov-concentration}.
The second follows from
$\sum_{i=1}^{N}c_{i,N}^2\leq NM_N^2$.
\\
Under the uniform contraction condition,
\[
    c_{i,N}
    \leq
    \sum_{r=0}^{N-i}\theta^r
    \leq
    \frac{1}{1-\theta},
\]
which establishes the asserted bound on $M_N$.
Finally, stationarity and $h_i=h$ imply
\[
    \mathbb{E}F_N
    =
    \frac{1}{N}\sum_{i=1}^{N}\mathbb{E}[h(U_i)]
    =
    \mathbb{E}[h(U_1)].
\]
\end{proof}

\subsection{Proof of Lemma~\ref{lem:posterior_regression}}
\label{app:proof:lem:posterior_regression}

\begin{proof}
Choose a $1/4$-net $\mathcal V$ of the Euclidean unit sphere with
$|\mathcal V|\leq9^m$. For any symmetric matrix $D$, approximating
a unit vector attaining $\|D\|_{\mathrm{op}}$ by a point of
$\mathcal V$ gives
\[
    \|D\|_{\mathrm{op}}
    \leq\max_{v\in\mathcal V}|v^\top Dv|
         +\frac12\|D\|_{\mathrm{op}}.
\]
Hence $\|D\|_{\mathrm{op}}
\leq2\max_{v\in\mathcal V}|v^\top Dv|$.
For each fixed $v\in\mathcal V$, the function
\(
    w\longmapsto
    \bigl(v^\top\psi(\phi^{(k)}(w))\bigr)^2
\)
is bounded by $\kappa^2$ and is evaluated along the native Markov
chain. Stationarity identifies its mean with $v^\top\Sigma_kv$.
Applying Lemma~\ref{lem:markov_bellman_concentration} at deviation
level $\nu/4$ and taking a union bound over tasks and net points
therefore yields
\[
\begin{aligned}
    \mathbb P\!\left(
        \max_k\|A_k/N-\Sigma_k\|_{\mathrm{op}}>\nu/2
    \right)
    \leq
    2(J+1)9^m
    \exp\!\left\{-\frac{N\nu^2}{128\kappa^4M_N^2}\right\}
    \leq\delta/2.
\end{aligned}
\]
Let $\mathcal E_D$ denote the complementary event.
On $\mathcal E_D$,
$\lambda_{\min}(A_k)\geq N\nu/2$ for every $k$.
Write $\xi_k=(\xi_1^{(k)},\ldots,\xi_N^{(k)})^\top$.
For a fixed unit vector $v$, the coefficient
$v^\top\psi(X_t^{(k)})$ is $\mathscr F_t^{(k)}$-measurable.
The conditional noise assumption therefore implies
\[
    \mathbb E\!\left[
        \exp\!\left(u v^\top\psi(X_t^{(k)})\xi_t^{(k)}\right)
        \mid\mathscr F_t^{(k)}
    \right]
    \leq\exp\!\left(\frac{u^2\sigma^2\kappa^2}{2}\right).
\]
Repeated conditioning gives
\[
    \mathbb E\exp\!\left(
        u\sum_{t=1}^{N}
        v^\top\psi(X_t^{(k)})\xi_t^{(k)}
    \right)
    \leq\exp\!\left(\frac{u^2\sigma^2\kappa^2N}{2}\right).
\]
Choose a $1/2$-net $\mathcal U$ of the unit sphere with
$|\mathcal U|\leq5^m$, so that
$\|z\|_2\leq2\max_{v\in\mathcal U}|v^\top z|$.
The two-sided exponential tail bound and a union bound then give
an event $\mathcal E_\xi$ of probability at least $1-\delta/2$
on which
\begin{equation}\nonumber
    \max_k\|\Phi_k^\top\xi_k\|_2
    \leq2\sigma\kappa\sqrt{2NL_N(\delta)}
    =Ns_N(\delta).
\label{eq:boot-noise-event}
\end{equation}
Indeed, the failure probability is at most
$2(J+1)5^m\exp\{-L_N(\delta)\}=\delta/2$.
Neither this union bound nor the design argument requires
independence between tasks.
On $\mathcal E_D$, the block-diagonal structure of $A$ gives
\[
    \|A^{-1}\|_{\mathrm b}\leq\frac{2}{N\nu},
    \quad
    \|\lambda_NA^{-1}\Omega\|_{\mathrm b}
    \leq\frac{2\lambda_N\Lambda_N}{N\nu}\leq\frac12.
\]
Using
$(A+\lambda_N\Omega)^{-1}
=(I+\lambda_NA^{-1}\Omega)^{-1}A^{-1}$
and the convergent Neumann series, we obtain
\begin{equation}
    \|(A+\lambda_N\Omega)^{-1}\|_{\mathrm b}
    \leq
    \frac{\|A^{-1}\|_{\mathrm b}}
         {1-\|\lambda_NA^{-1}\Omega\|_{\mathrm b}}
    \leq\frac{4}{N\nu}.
\label{eq:boot-block-inverse}
\end{equation}
Stacking the coefficient and noise vectors as $\vartheta$ and
$\xi$, respectively, the regression model yields the exact identity
\[
    \widehat\vartheta-\vartheta
    =(A+\lambda_N\Omega)^{-1}
      (\Phi^\top\xi-\lambda_N\Omega\vartheta).
\]
Consequently, on $\mathcal E_D\cap\mathcal E_\xi$,
\[
    \|\widehat\vartheta-\vartheta\|_{2,\infty}
    \leq\frac{4}{N\nu}
         \left\{Ns_N(\delta)+\lambda_N\Lambda_NR\right\}.
\]
The feature bound implies
\[
    \max_k\|\widehat f^{(k)}-f^{(k)}\|_\infty
    \leq\kappa\|\widehat\vartheta-\vartheta\|_{2,\infty}.
\]
Since $\mathbb P(\mathcal E_D\cap\mathcal E_\xi)\geq1-\delta$,
this proves \eqref{eq:boot-regression-bound}.

Keeping the design and regularization fixed gives
\[
    \widehat\vartheta_d-\widehat\vartheta
    =(A+\lambda_N\Omega)^{-1}\Phi^\top d.
\]
The bound $\|\Phi_k^\top d_k\|_2\leq N\kappa a$, together with
\eqref{eq:boot-block-inverse}, implies
\[
    \max_k\|\widehat f_d^{(k)}-\widehat f^{(k)}\|_\infty
    \leq
    \kappa\frac{4}{N\nu}N\kappa a
    =\frac{4\kappa^2a}{\nu}.
\]
This argument is pathwise on $\mathcal E_D$ and is therefore valid
for data-dependent perturbations.
\end{proof}

\subsection{Proof of Lemma~\ref{lem:bellman_residual}}
\label{app:proof:lem:bellman_residual}

\begin{proof}
Fix a task $k$ and a deterministic $q\in\mathcal Q_k$. Write
$V_q(s)=\max_{a\in\mathcal A^{(k)}}q(s,a)$ and define
\[
    \xi_t^{(k)}(q)
    =Y_t^{(k)}(q)-(\mathcal T^{(k)}q)(S_t^{(k)},A_t^{(k)}).
\]
The Markov reward/transition model gives
$\mathbb E[\xi_t^{(k)}(q)\mid\mathscr F_t^{(k)}]=0$.
Let $d_q=V_q-V_*^{(k)}$, so that $\|d_q\|_\infty\leq2B$, and set
\[
    \zeta_t(q)=d_q(S_{t+1}^{(k)})
        -\mathbb E[d_q(S_{t+1}^{(k)})\mid\mathscr F_t^{(k)}].
\]
The Bellman identity implies
\[
    \xi_t^{(k)}(q)=\epsilon_t^{(k)}+\gamma\zeta_t(q).
\]
Since $d_q$ takes values in $[-2B,2B]$, the conditional form of
Hoeffding's lemma yields
\[
    \mathbb E[\exp(u\zeta_t(q))\mid\mathscr F_t^{(k)}]
    \leq\exp(2B^2u^2).
\]
Conditional Cauchy--Schwarz therefore gives
\[
\begin{aligned}
    \mathbb E[e^{u\xi_t^{(k)}(q)}\mid\mathscr F_t^{(k)}]
    &\leq
    \left\{\mathbb E[e^{2u\epsilon_t^{(k)}}
                         \mid\mathscr F_t^{(k)}]\right\}^{1/2}
    \left\{\mathbb E[e^{2u\gamma\zeta_t(q)}
                         \mid\mathscr F_t^{(k)}]\right\}^{1/2}\\
    &\leq\exp(u^2\sigma^2+4u^2\gamma^2B^2)\\
    & =\exp\!\left(\frac{u^2\sigma_B^2}{2}\right).
\end{aligned}
\]
No independence between the innovation and the continuation-value
fluctuation is used.

Choose an $\varepsilon$-net $\mathcal V_k\subseteq\mathcal Q_k$
with $|\mathcal V_k|\leq e^{H_N(\varepsilon)}$ for each task, and
a $1/2$-net $\mathcal U$ of the Euclidean unit sphere with
$|\mathcal U|\leq5^m$.
Applying the conditional exponential-moment argument used in
\eqref{eq:boot-noise-event} to every fixed pair
$(v,q)\in\mathcal U\times\mathcal V_k$ gives, with probability
at least $1-\delta/2$,
\[
    \max_k\max_{q\in\mathcal V_k}
    \left\|
        \sum_{t=1}^{N}\psi(X_t^{(k)})\xi_t^{(k)}(q)
    \right\|_2
    \leq2\sigma_B\kappa\sqrt{2N\ell_N}.
\]
Indeed, the union bound has failure probability at most
$2(J+1)5^m e^{H_N(\varepsilon)-\ell_N}=\delta/2$.
Only taskwise nets enter this bound because the noise is controlled
in the block-maximum norm.
For any $q\in\mathcal Q_k$, choose $q'\in\mathcal V_k$ with
$\|q-q'\|_\infty\leq\varepsilon$.
The Lipschitz property of action maximization and the contraction
of $\mathcal T^{(k)}$ imply
\[
\begin{aligned}
    |\xi_t^{(k)}(q)-\xi_t^{(k)}(q')|
    \leq\gamma\|V_q-V_{q'}\|_\infty
            +\|\mathcal T^{(k)}q-\mathcal T^{(k)}q'\|_\infty
    \leq2\gamma\varepsilon.
\end{aligned}
\]
The feature bound then extends the finite-net inequality to the
entire function classes. Thus, on a single event $\mathcal E_U$
of probability at least $1-\delta/2$,
\begin{equation}
    \max_k\sup_{q\in\mathcal Q_k}\frac{1}{N}
    \left\|
        \sum_{t=1}^{N}\psi(X_t^{(k)})\xi_t^{(k)}(q)
    \right\|_2
    \leq
    2\sigma_B\kappa\sqrt{\frac{2\ell_N}{N}}
    +2\gamma\kappa\varepsilon.
\label{eq:boot-uniform-noise}
\end{equation}
All concentration arguments were applied to deterministic net
elements. The subsequent extension is pathwise, so the resulting
inequality also holds at data-dependent members of the classes.
For each $\boldsymbol q\in\mathcal Q$, choose the coefficients in
\eqref{eq:boot-bellman-approximation} and define
\[
    d_t^{(k)}(q_k)
    =(\mathcal T^{(k)}q_k)(S_t^{(k)},A_t^{(k)})
      -\psi(X_t^{(k)})^\top\vartheta_k(q_k).
\]
Then $\max_{k,t}|d_t^{(k)}(q_k)|\leq a_N$, and the stacked
responses satisfy
\[
    Y(\boldsymbol q)
    =\Phi\vartheta(\boldsymbol q)
       +\xi(\boldsymbol q)+d(\boldsymbol q).
\]
On the design event $\mathcal E_D$ from the preceding proof,
\eqref{eq:boot-block-inverse} holds. Therefore,
\[
\begin{aligned}
    \|\widehat\vartheta(\boldsymbol q)
           -\vartheta(\boldsymbol q)\|_{2,\infty}
    \leq\frac{4}{N\nu}\Bigl\{
        \|\Phi^\top\xi(\boldsymbol q)\|_{2,\infty}
        +N\kappa a_N
        +\lambda_N\Lambda_NR_N\Bigr\}.
\end{aligned}
\]
Combining this inequality with \eqref{eq:boot-uniform-noise} and
the approximation condition gives, on $\mathcal E_D\cap\mathcal E_U$,
\[
\begin{aligned}
    \max_k\|\mathfrak F_N^{(k)}(\boldsymbol q)
                   -\mathcal T^{(k)}q_k\|_\infty
    &\leq a_N
          +\kappa\|\widehat\vartheta(\boldsymbol q)
                          -\vartheta(\boldsymbol q)\|_{2,\infty}\\
    &\leq a_N+\frac{4\kappa}{\nu}
       \left\{
           2\sigma_B\kappa\sqrt{\frac{2\ell_N}{N}}
           +2\gamma\kappa\varepsilon+\kappa a_N
           +\frac{\lambda_N\Lambda_NR_N}{N}
       \right\}\\
    &=\overline\eta_N(\delta,\varepsilon),
\end{aligned}
\]
uniformly over $\boldsymbol q\in\mathcal Q$.
The intersection has probability at least $1-\delta$, which proves
\eqref{eq:boot-uniform-bellman}.
Substituting $\boldsymbol q=\widehat{\boldsymbol Q}_{e-1}$ on this
same event proves the assertion for every admissible fitted iteration.
\end{proof}

\subsection{Proof of Theorem~\ref{the:bootstrap_consistency}}
\label{app:proof:the:bootstrap_consistency}

\begin{proof}
Since $Q_*^{(k)}$ is the fixed point of $\mathcal T^{(k)}$,
\(
    \mathcal T^{(k)}Q_*^{(k)}=Q_*^{(k)}.
\)
Hence, for every task $k$ and iteration $e\geq1$,
\[
\begin{aligned}
    \|\widehat Q_e^{(k)}-Q_*^{(k)}\|_\infty
    &=
    \|\widehat Q_e^{(k)}
      -\mathcal T^{(k)}Q_*^{(k)}\|_\infty \\
    &\leq
    \|\widehat Q_e^{(k)}
      -\mathcal T^{(k)}\widehat Q_{e-1}^{(k)}\|_\infty 
    +
    \|\mathcal T^{(k)}\widehat Q_{e-1}^{(k)}
      -\mathcal T^{(k)}Q_*^{(k)}\|_\infty .
\end{aligned}
\]
By the $\gamma$-contraction property,
\[
    \|\mathcal T^{(k)}\widehat Q_{e-1}^{(k)}
      -\mathcal T^{(k)}Q_*^{(k)}\|_\infty
    \leq
    \gamma
    \|\widehat Q_{e-1}^{(k)}-Q_*^{(k)}\|_\infty .
\]
Therefore,
\[
    \|\widehat Q_e^{(k)}-Q_*^{(k)}\|_\infty
    \leq
    \eta_{e,N}+\gamma D_{e-1,N}.
\]
Taking the maximum over $k$ gives
\begin{equation}
    D_{e,N}
    \leq
    \gamma D_{e-1,N}+\eta_{e,N}.
    \label{eq:boot-one-step-recursion}
\end{equation}
For $e=1$, inequality \ref{eq:boot-one-step-recursion} gives
\[
D_{1,N}\leq \gamma D_{0,N}+\eta_{1,N},
\]
which establishes the base case.
Suppose that \ref{eq:boot-recursion} holds at iteration $e-1$ for
some integer $e\geq 2$, that is,
\[
D_{e-1,N}
\leq
\gamma^{e-1}D_{0,N}
+
\sum_{j=1}^{e-1}
\gamma^{e-1-j}\eta_{j,N}.
\]
Substituting this induction hypothesis into \ref{eq:boot-one-step-recursion}, we have
\begin{align*}
D_{e,N}
&\leq \gamma D_{e-1,N}+\eta_{e,N}\\
&\leq
\gamma\left(
\gamma^{e-1}D_{0,N}
+
\sum_{j=1}^{e-1}
\gamma^{e-1-j}\eta_{j,N}
\right)
+\eta_{e,N}\\
&=
\gamma^eD_{0,N}
+
\sum_{j=1}^{e}
\gamma^{e-j}\eta_{j,N}.
\end{align*}
Thus, the induction step is established, and \ref{eq:boot-recursion} holds for every integer $e\geq 1$.

By Lemma~\ref{lem:bellman_residual}, there exists an event
$\mathcal E_N$ with
\(
    \mathbb P(\mathcal E_N)\geq1-\delta_N
\)
such that, simultaneously for all $1\leq j\leq E_N$,
\(
    \eta_{j,N}\leq\overline\eta_N.
\)
On $\mathcal E_N$, using $D_{0,N}\leq D_0$ in
\eqref{eq:boot-recursion} gives
\[
\begin{aligned}
    D_{e,N}
    \leq
    \gamma^eD_0
    +
    \overline\eta_N
    \sum_{j=1}^{e}\gamma^{e-j} 
    =
    \gamma^eD_0
    +
    \frac{1-\gamma^e}{1-\gamma}\,
    \overline\eta_N,
\end{aligned}
\]
which proves \eqref{eq:boot-high-probability}.
\\
Finally, at $e=E_N$,
\(
    D_{E_N,N}
    \leq
    \gamma^{E_N}D_0
    +
    \frac{\overline\eta_N}{1-\gamma}
\)
on $\mathcal E_N$. The right-hand side converges to zero by assumption.
Hence, for any $u>0$, it is smaller than $u$ for all sufficiently large
$N$, and therefore
\[
    \mathbb P(D_{E_N,N}>u)
    \leq
    \mathbb P(\mathcal E_N^c)
    \leq
    \delta_N
    \longrightarrow0.
\]
Thus,
\[
    D_{E_N,N}\xrightarrow{\mathbb P}0.
\]
\end{proof}

\subsection{Proof of Corollary~\ref{cor:bootstrap_stability}}
\label{app:proof:cor:bootstrap_stability}

\begin{proof}
Fix a task $k$, an iteration $e\geq1$, and an index $t$, and write
$s'=S_{t+1}^{(k)}$. Define
\[
    \Delta_{e-1}^{(k)}(s')
    =
    \sup_{a\in\mathcal A^{(k)}}
    \left|
        \widehat Q_{e-1}^{(k)}(s',a)
        -
        Q_*^{(k)}(s',a)
    \right|.
\]
For every $a\in\mathcal A^{(k)}$,
\[
    \widehat Q_{e-1}^{(k)}(s',a)
    \leq
    Q_*^{(k)}(s',a)
    +\Delta_{e-1}^{(k)}(s').
\]
Taking the maximum over $a$ gives
\[
    \max_{a\in\mathcal A^{(k)}}
    \widehat Q_{e-1}^{(k)}(s',a)
    \leq
    \max_{a\in\mathcal A^{(k)}}
    Q_*^{(k)}(s',a)
    +\Delta_{e-1}^{(k)}(s').
\]
Similarly,
\[
    \max_{a\in\mathcal A^{(k)}}
    Q_*^{(k)}(s',a)
    \leq
    \max_{a\in\mathcal A^{(k)}}
    \widehat Q_{e-1}^{(k)}(s',a)
    +\Delta_{e-1}^{(k)}(s').
\]
Combining the two inequalities yields
\[
\begin{aligned}
    \left|
        \max_{a\in\mathcal A^{(k)}}
        \widehat Q_{e-1}^{(k)}(s',a)
        -
        \max_{a\in\mathcal A^{(k)}}
        Q_*^{(k)}(s',a)
    \right| 
    \leq
    \Delta_{e-1}^{(k)}(s').
\end{aligned}
\]
Therefore,
\[
\begin{aligned}
    |b_{t,e}^{(k)}|
    &\leq
    \gamma
    \sup_{a\in\mathcal A^{(k)}}
    \left|
        \widehat Q_{e-1}^{(k)}(S_{t+1}^{(k)},a)
        -
        Q_*^{(k)}(S_{t+1}^{(k)},a)
    \right| \\
    &\leq
    \gamma
    \|\widehat Q_{e-1}^{(k)}-Q_*^{(k)}\|_\infty \\
    &\leq
    \gamma D_{e-1,N}.
\end{aligned}
\]
Taking the supremum over $k$ and $t$ proves
\eqref{eq:boot-pathwise-stability}. The argument is pathwise and
therefore does not require independence between the successor state
and the fitted iterate.
Let $\mathcal E_N$ denote the event supplied by
Theorem~\ref{the:bootstrap_consistency}. Then
\(
    \mathbb P(\mathcal E_N)\geq1-\delta_N,
\)
and simultaneously for $0\leq e\leq E_N$,
\(
    D_{e,N}\leq r_{e,N}
    ~\text{on }\mathcal E_N.
\)
Hence, for every $1\leq e\leq E_N$,
\[
    \sup_{k,t}|b_{t,e}^{(k)}|
    \leq
    \gamma D_{e-1,N}
    \leq
    \gamma r_{e-1,N}
    \quad\text{on }\mathcal E_N,
\]
which proves \eqref{eq:boot-high-probability-remainder}.
Now let $e_N\leq E_N$ be deterministic and suppose that
\[
    r_{e_N-1,N}\to0,
    \quad
    \delta_N\to0.
\]
For any $u>0$, there exists $N_u$ such that
\(
    \gamma r_{e_N-1,N}<u
\)
for all $N\geq N_u$. On $\mathcal E_N$ we then have
\(
    \sup_{k,t}|b_{t,e_N}^{(k)}|<u.
\)
Consequently,
\[
\begin{aligned}
    \mathbb P\!\left(
        \sup_{k,t}|b_{t,e_N}^{(k)}|>u
    \right)
    \leq
    \mathbb P(\mathcal E_N^c)
    \leq
    \delta_N
    \longrightarrow0.
\end{aligned}
\]
Thus,
\[
    \sup_{k,t}|b_{t,e_N}^{(k)}|
    \xrightarrow{\mathbb P}0.
\]
It remains to establish the mean-square bound.
Suppose that the fitted update is deterministically restricted to the
supremum-norm ball of radius $B$ and that
$\|Q_*^{(k)}\|_\infty\leq B$ for every task.
Then, for every realization,
\[
\begin{aligned}
    D_{e-1,N}
    &=
    \max_k
    \|\widehat Q_{e-1}^{(k)}-Q_*^{(k)}\|_\infty \\
    &\leq
    \max_k
    \left\{
        \|\widehat Q_{e-1}^{(k)}\|_\infty
        +
        \|Q_*^{(k)}\|_\infty
    \right\} \\
    &\leq 2B.
\end{aligned}
\]
Decomposing according to $\mathcal E_N$ gives
\[
\begin{aligned}
    \mathbb E[D_{e-1,N}^2]
    &=
    \mathbb E\!\left[
        D_{e-1,N}^2\mathbf 1_{\mathcal E_N}
    \right]
    +
    \mathbb E\!\left[
        D_{e-1,N}^2\mathbf 1_{\mathcal E_N^c}
    \right] \\
    &\leq
    r_{e-1,N}^2\,\mathbb P(\mathcal E_N)
    +
    4B^2\,\mathbb P(\mathcal E_N^c) \\
    &\leq
    r_{e-1,N}^2+4B^2\delta_N.
\end{aligned}
\]
On the other hand, the pathwise inequality already proved implies
\[
    (b_{t,e}^{(k)})^2
    \leq
    \gamma^2D_{e-1,N}^2.
\]
Taking expectations,
\[
    \mathbb E\!\left[(b_{t,e}^{(k)})^2\right]
    \leq
    \gamma^2\mathbb E[D_{e-1,N}^2].
\]
The right-hand side does not depend on $k$ or $t$, so
\[
\begin{aligned}
    \sup_{k,t}
    \mathbb E\!\left[(b_{t,e}^{(k)})^2\right]
    \leq
    \gamma^2\mathbb E[D_{e-1,N}^2]
    \leq
    \gamma^2
    \left\{
        r_{e-1,N}^2+4B^2\delta_N
    \right\}.
\end{aligned}
\]
This proves \eqref{eq:boot-mean-square-stability}.
Finally, if
$r_{e_N-1,N}\to0$ and $\delta_N\to0$, then
\[
    \gamma^2
    \left\{
        r_{e_N-1,N}^2+4B^2\delta_N
    \right\}
    \longrightarrow0,
\]
and therefore
\[
    \sup_{k,t}
    \mathbb E\!\left[(b_{t,e_N}^{(k)})^2\right]
    \longrightarrow0.
\]
\end{proof}

\subsection{Proof of Lemma~\ref{lem:selection_bootstrap_scores}}
\label{app:proof:lem:selection_bootstrap_scores}

\begin{proof}
Fix $S\in\mathcal S_J$. Since the same predictive specification is
used for the actual and oracle-response scores, the predictive
covariance $V_S$ is unchanged when
$\overline{\mathbf Y}$ is replaced by
$\mathbf Y_e=\overline{\mathbf Y}+\mathbf b_e$.
The log-determinant terms therefore cancel.
By definition,
\[
\begin{aligned}
    \mathbf Y_e^{(0)}
    -
    \mathbf a_S
    -
    A_S\mathbf Y_e^{(S)}
    =
    \overline{\mathbf Y}^{(0)}
    +
    \mathbf b_e^{(0)}
    -
    \mathbf a_S
    -
    A_S
    \bigl(
        \overline{\mathbf Y}^{(S)}
        +
        \mathbf b_e^{(S)}
    \bigr)
    =
    \mathbf r_S^\circ
    +
    \mathbf d_{S,e}.
\end{aligned}
\]
Expanding the Gaussian quadratic form gives
\[
\begin{aligned}
    \ell_{e,N}(S)
    -
    \ell_N^\circ(S)
    &=
    -\frac12
    \left[
        (\mathbf r_S^\circ+\mathbf d_{S,e})^\top
        V_S^{-1}
        (\mathbf r_S^\circ+\mathbf d_{S,e})
        -
        (\mathbf r_S^\circ)^\top
        V_S^{-1}
        \mathbf r_S^\circ
    \right]\\
    &=
    -
    (\mathbf r_S^\circ)^\top
    V_S^{-1}
    \mathbf d_{S,e}
    -
    \frac12
    \mathbf d_{S,e}^\top
    V_S^{-1}
    \mathbf d_{S,e}.
\end{aligned}
\]
Since
\(
    \lambda_{\min}(V_S)
    \geq
    \underline v_N,
\)
we have
\(
    \|V_S^{-1}\|_{\mathrm{op}}
    \leq
    \underline v_N^{-1}.
\)
Therefore,
\[
\begin{aligned}
    \left|
        \ell_{e,N}(S)-\ell_N^\circ(S)
    \right|
    \leq
    \frac{1}{\underline v_N}
    \left\{
        \|\mathbf r_S^\circ\|_2
        \|\mathbf d_{S,e}\|_2
        +
        \frac12
        \|\mathbf d_{S,e}\|_2^2
    \right\}
    \leq
    \frac{
        \mathcal R_N\mathcal D_{e,N}
        +
        \tfrac12\mathcal D_{e,N}^2
    }{
        \underline v_N
    }.
\end{aligned}
\]
Recall that
\[
\begin{aligned}
    u_{e,N}(S)
    =
    \bigl\{
        \ell_{e,N}(S)
        -
        \ell_N^\circ(S)
    \bigr\}
    -
    \bigl\{
        \ell_{e,N}(S_N^\star)
        -
        \ell_N^\circ(S_N^\star)
    \bigr\}.
\end{aligned}
\]
The triangle inequality thus yields
\[
    |u_{e,N}(S)|
    \leq
    \frac{
        2\mathcal R_N\mathcal D_{e,N}
        +
        \mathcal D_{e,N}^2
    }{
        \underline v_N
    }.
\]
Taking the maximum over
$S\in\mathcal S_J$
proves
\eqref{eq:selection_bootstrap_score_bound}.
\\
Next,
\[
\begin{aligned}
    \|\mathbf d_{S,e}\|_2
    =
    \|
        \mathbf b_e^{(0)}
        -
        A_S\mathbf b_e^{(S)}
    \|_2
    \leq
    \|\mathbf b_e^{(0)}\|_2
    +
    \|A_S\|_{\mathrm{op}}
    \|\mathbf b_e^{(S)}\|_2
    \leq
    (1+L_{A,N})
    \|\mathbf b_e\|_2.
\end{aligned}
\]
Taking the maximum over $S$ proves
\eqref{eq:selection_d_bound}. Substitution into
\eqref{eq:selection_bootstrap_score_bound} yields
\eqref{eq:selection_bootstrap_score_pathwise}.
\end{proof}

\subsection{Proof of Corollary~\ref{cor:selection_bootstrap_score_rate}}
\label{app:proof:cor:selection_bootstrap_score_rate}

\begin{proof}
Set
\(
    t_{e,N}
    =
    \sqrt{n_{\mathrm{tot}}}\,\beta_{e,N},
\)
and define the deterministic rate
\[
    \rho_{e,N}^{\mathrm{bt}}
    =
    \frac{
        (1+L_{A,N})s_N t_{e,N}
        +
        (1+L_{A,N})^2 t_{e,N}^2
    }{
        \underline v_N
    }.
\]
It is enough to show that
\(
    \mathfrak B_{e,N}
    =
    O_{\mathbb P_\star}
    \bigl(
        \rho_{e,N}^{\mathrm{bt}}
    \bigr).
\)
By Corollary~\ref{cor:bootstrap_stability},
\(
    \sup_{k,t}
    \mathbb E_\star
    \left[
        (b_{t,e}^{(k)})^2
    \right]
    \leq
    \beta_{e,N}^2.
\)
Since $\mathbf b_e$ contains exactly
$n_{\mathrm{tot}}$ bootstrap-remainder coordinates,
\[
\begin{aligned}
    \mathbb E_\star
    \|\mathbf b_e\|_2^2
    =
    \sum_{k=0}^{J}
    \sum_{t=1}^{n_k}
    \mathbb E_\star
    \left[
        (b_{t,e}^{(k)})^2
    \right]
    \leq
    n_{\mathrm{tot}}
    \beta_{e,N}^2
    =
    t_{e,N}^2.
\end{aligned}
\]
No independence among the coordinates of $\mathbf b_e$ is required
for this inequality.
If $t_{e,N}=0$, then
$\mathbb E_\star\|\mathbf b_e\|_2^2=0$, and hence
$\mathbf b_e=0$ almost surely. In this case,
Lemma~\ref{lem:selection_bootstrap_scores} gives
$\mathfrak B_{e,N}=0$ almost surely, so the conclusion is immediate.
We therefore consider $t_{e,N}>0$.
For any constant $M\geq1$, Markov's inequality gives
\begin{equation}
\begin{aligned}
    \mathbb P_\star
    \left(
        \|\mathbf b_e\|_2
        >
        M t_{e,N}
    \right)
    =
    \mathbb P_\star
    \left(
        \|\mathbf b_e\|_2^2
        >
        M^2t_{e,N}^2
    \right)
    \leq
    \frac{
        \mathbb E_\star
        \|\mathbf b_e\|_2^2
    }{
        M^2t_{e,N}^2
    }
    \leq
    M^{-2}.
\end{aligned}
\label{eq:bootstrap_vector_markov}
\end{equation}
By assumption,
\(
    \mathbb P_\star
    \left(
        \mathcal R_N>s_N
    \right)
    \leq
    \eta_{R,N}.
\)
Define
\(
    \mathcal E_N(M)
    =
    \left\{
        \mathcal R_N\leq s_N
    \right\}
    \cap
    \left\{
        \|\mathbf b_e\|_2
        \leq
        M t_{e,N}
    \right\}.
\)
The union bound and
\eqref{eq:bootstrap_vector_markov} imply
\(
    \mathbb P_\star
    \bigl(
        \mathcal E_N(M)^c
    \bigr)
    \leq
    \eta_{R,N}+M^{-2}.
\)
On $\mathcal E_N(M)$, the pathwise bound
\eqref{eq:selection_bootstrap_score_pathwise} yields
\[
\begin{aligned}
    \mathfrak B_{e,N}
    \leq
    \frac{1}{\underline v_N}
    \Bigl\{
        2M(1+L_{A,N})
        s_Nt_{e,N}
        +
        M^2(1+L_{A,N})^2
        t_{e,N}^2
    \Bigr\}.
\end{aligned}
\]
Since $M\geq1$,
\(
    2M a+M^2b
    \leq
    3M^2(a+b),
    ~
    a,b\geq0.
\)
Therefore,
\(
    \mathfrak B_{e,N}
    \leq
    3M^2
    \rho_{e,N}^{\mathrm{bt}}
    ~
    \text{on }\mathcal E_N(M).
\)
Consequently,
\begin{equation}
    \mathbb P_\star
    \left(
        \mathfrak B_{e,N}
        >
        3M^2
        \rho_{e,N}^{\mathrm{bt}}
    \right)
    \leq
    \eta_{R,N}+M^{-2}.
    \label{eq:bootstrap_rate_probability}
\end{equation}
Now fix $\varepsilon>0$. Choose $M\geq1$ sufficiently large that
\(
    M^{-2}<\frac{\varepsilon}{2}.
\)
Since $\eta_{R,N}\to0$, for all sufficiently large $N$,
\(
    \eta_{R,N}<\frac{\varepsilon}{2}.
\)
It follows from
\eqref{eq:bootstrap_rate_probability} that
\(
    \mathbb P_\star
    \left(
        \mathfrak B_{e,N}
        >
        3M^2
        \rho_{e,N}^{\mathrm{bt}}
    \right)
    <
    \varepsilon
\)
for all sufficiently large $N$. Hence,
\(
    \mathfrak B_{e,N}
    =
    O_{\mathbb P_\star}
    \bigl(
        \rho_{e,N}^{\mathrm{bt}}
    \bigr).
\)
Substituting
$t_{e,N}
=\sqrt{n_{\mathrm{tot}}}\beta_{e,N}$
into the definition of
$\rho_{e,N}^{\mathrm{bt}}$ gives
\[
    \mathfrak B_{e,N}
    =
    O_{\mathbb P_\star}
    \left[
        \frac{
            (1+L_{A,N})
            s_N\sqrt{n_{\mathrm{tot}}}\,
            \beta_{e,N}
            +
            (1+L_{A,N})^2
            n_{\mathrm{tot}}
            \beta_{e,N}^2
        }{
            \underline v_N
        }
    \right],
\]
which is
\eqref{eq:selection_bootstrap_score_rate}.
\end{proof}

\subsection{Proof of Lemma~\ref{lem:selection_posterior_bound}}
\label{app:proof:lem:selection_posterior_bound}

\begin{proof}
By the definition of the working subset posterior,
\[
\begin{aligned}
    \Pi_{e,N}
    (\mathcal B\mid\mathbf Y_e,X)
    &=
    \frac{
        \displaystyle
        \sum_{S\in\mathcal B}
        \Pi_N(S)
        \exp\{\ell_{e,N}(S)\}
    }{
        \displaystyle
        \sum_{U\in\mathcal S_J}
        \Pi_N(U)
        \exp\{\ell_{e,N}(U)\}
    }.
\end{aligned}
\]
The denominator is bounded below by the contribution of the reference
subset:
\[
    \sum_{U\in\mathcal S_J}
    \Pi_N(U)\exp\{\ell_{e,N}(U)\}
    \geq
    \pi_N^\star
    \exp\{\ell_{e,N}(S_N^\star)\}.
\]
Therefore,
\[
\begin{aligned}
    \Pi_{e,N}
    (\mathcal B\mid\mathbf Y_e,X)
    &\leq
    \sum_{S\in\mathcal B}
    \frac{\Pi_N(S)}{\pi_N^\star}
    \exp\left\{
        \ell_{e,N}(S)
        -
        \ell_{e,N}(S_N^\star)
    \right\}\\
    &\leq
    |\mathcal B|
    \exp\Bigg(
        q_N
        +
        \max_{S\in\mathcal B}
        \bigl\{
            \ell_{e,N}(S)
            -
            \ell_{e,N}(S_N^\star)
        \bigr\}
    \Bigg).
\end{aligned}
\]
Since posterior mass is at most one,
\eqref{eq:selection_mass_bound} follows.
For each $S\in\mathcal B$,
\[
    \frac{
        \Pi_{e,N}(S\mid\mathbf Y_e,X)
    }{
        \Pi_{e,N}(S_N^\star\mid\mathbf Y_e,X)
    }
    =
    \frac{\Pi_N(S)}{\pi_N^\star}
    \exp\left\{
        \ell_{e,N}(S)
        -
        \ell_{e,N}(S_N^\star)
    \right\}.
\]
Using
\(
    \log
    \frac{\Pi_N(S)}{\pi_N^\star}
    \leq q_N
\)
and taking the maximum over $S\in\mathcal B$ proves
\eqref{eq:selection_odds_bound}.
\end{proof}

\subsection{Proof of Theorem~\ref{thm:predictive_selection}}
\label{app:proof:thm:predictive_selection}

\begin{proof}
If $\mathcal B_N=\varnothing$, all conclusions are immediate. Hence
consider $N$ for which $\mathcal B_N\neq\varnothing$.
Define
\(
    \mathcal E_{1,N}
    =
    \left\{
        \max_{S\in\mathcal B_N}
        Z_N(S)
        \leq
        a_N
    \right\}.
\)
By \eqref{eq:selection_score_fluctuation},
\(
    \mathbb P_\star
    \left(
        \mathcal E_{1,N}^c
    \right)
    =
    \Xi_N^+(a_N;d_N)
    \longrightarrow0.
\)
Since $\kappa_N\to\infty$, we have $\kappa_N>0$ for all sufficiently
large $N$. Define
\(
    \mathcal E_{2,N}
    =
    \left\{
        \mathfrak B_{e,N}
        \leq
        \frac{\kappa_N}{2}
    \right\}.
\)
Corollary~\ref{cor:selection_bootstrap_score_rate} gives
\(
    \mathfrak B_{e,N}
    =
    O_{\mathbb P_\star}
    \left(
        \rho_{e,N}^{\mathrm{bt}}
    \right).
\)
Together with
$\rho_{e,N}^{\mathrm{bt}}=o(\kappa_N)$,
this implies
\[
    \frac{\mathfrak B_{e,N}}{\kappa_N}
    \xrightarrow{\mathbb P_\star}0.
\]
To verify this implication explicitly, fix $\varepsilon>0$. By
$\mathfrak B_{e,N}
=O_{\mathbb P_\star}(\rho_{e,N}^{\mathrm{bt}})$,
there exists $C_\varepsilon<\infty$ such that, for all sufficiently
large $N$,
\[
    \mathbb P_\star
    \left(
        \mathfrak B_{e,N}
        >
        C_\varepsilon
        \rho_{e,N}^{\mathrm{bt}}
    \right)
    <
    \varepsilon.
\]
Since
\(
    \frac{
        \rho_{e,N}^{\mathrm{bt}}
    }{
        \kappa_N
    }
    \longrightarrow0,
\)
we have
\(
    C_\varepsilon
    \rho_{e,N}^{\mathrm{bt}}
    <
    \frac{\kappa_N}{2}
\)
for all sufficiently large $N$. Consequently,
\[
\begin{aligned}
    \mathbb P_\star(\mathcal E_{2,N}^c)
    =
    \mathbb P_\star
    \left(
        \mathfrak B_{e,N}
        >
        \frac{\kappa_N}{2}
    \right)
    &\leq
    \mathbb P_\star
    \left(
        \mathfrak B_{e,N}
        >
        C_\varepsilon
        \rho_{e,N}^{\mathrm{bt}}
    \right)
    <
    \varepsilon.
\end{aligned}
\]
Because $\varepsilon>0$ is arbitrary,
\(
    \mathbb P_\star(\mathcal E_{2,N}^c)
    \longrightarrow0.
\)
Let
\(
    \mathcal E_N
    =
    \mathcal E_{1,N}
    \cap
    \mathcal E_{2,N}.
\)
The union bound gives
\begin{equation}\nonumber
\begin{aligned}
    \mathbb P_\star(\mathcal E_N^c)
    \leq
    \Xi_N^+(a_N;d_N)
    +
    \mathbb P_\star
    \left(
        \mathfrak B_{e,N}
        >
        \frac{\kappa_N}{2}
    \right)
    \longrightarrow0.
\end{aligned}
\label{eq:selection_good_event_probability}
\end{equation}
Now fix $S\in\mathcal B_N$. By definition of
$\mathcal B_N^{\mathrm{pred}}(d_N)$,
\begin{equation}
    \mathcal L_N^*
    -
    \mathcal L_N^\circ(S)
    \geq
    d_N.
    \label{eq:selection_bad_population_gap}
\end{equation}
The exact score decomposition
\eqref{eq:selection_score_decomposition} gives
\[
\begin{aligned}
    \ell_{e,N}(S)
    -
    \ell_{e,N}(S_N^\star)
    &=
    -
    \left\{
        \mathcal L_N^*
        -
        \mathcal L_N^\circ(S)
    \right\}
    +
    Z_N(S)
    +
    u_{e,N}(S).
\end{aligned}
\]
On $\mathcal E_N$,
\(
    Z_N(S)\leq a_N
\)
and
\(
    u_{e,N}(S)
    \leq
    |u_{e,N}(S)|
    \leq
    \mathfrak B_{e,N}
    \leq
    \frac{\kappa_N}{2}.
\)
Combining these inequalities with
\eqref{eq:selection_bad_population_gap} yields, simultaneously for
all $S\in\mathcal B_N$,
\begin{equation}\nonumber
    \ell_{e,N}(S)
    -
    \ell_{e,N}(S_N^\star)
    \leq
    -d_N
    +
    a_N
    +
    \frac{\kappa_N}{2}.
    \label{eq:selection_relative_score_event}
\end{equation}
We next bound the posterior mass of $\mathcal B_N$ directly. From
\eqref{eq:selection_working_posterior},
\[
\begin{aligned}
    \Pi_{e,N}
    \left(
        \mathcal B_N
        \mid
        \mathbf Y_e,X
    \right)
    =
    \frac{
        \displaystyle
        \sum_{S\in\mathcal B_N}
        \Pi_N(S)
        \exp\{\ell_{e,N}(S)\}
    }{
        \displaystyle
        \sum_{U\in\mathcal S_J}
        \Pi_N(U)
        \exp\{\ell_{e,N}(U)\}
    }.
\end{aligned}
\]
The denominator is at least
\(
    \pi_N^\star
    \exp\{\ell_{e,N}(S_N^\star)\}.
\)
Therefore, on $\mathcal E_N$,
\[
\begin{aligned}
    \Pi_{e,N}
    \left(
        \mathcal B_N
        \mid
        \mathbf Y_e,X
    \right)
    \leq
    \frac{
        \Pi_N(\mathcal B_N)
    }{
        \pi_N^\star
    }
    \exp
    \left\{
        -d_N
        +
        a_N
        +
        \frac{\kappa_N}{2}
    \right\}.
\end{aligned}
\]
By the definition of $\Lambda_N(d_N)$,
\(
    \frac{
        \Pi_N(\mathcal B_N)
    }{
        \pi_N^\star
    }
    \leq
    \exp\{\Lambda_N(d_N)\}.
\)
Hence
\[
\begin{aligned}
    \Pi_{e,N}
    \left(
        \mathcal B_N
        \mid
        \mathbf Y_e,X
    \right)
    \leq
    \exp
    \left\{
        \Lambda_N(d_N)
        -
        d_N
        +
        a_N
        +
        \frac{\kappa_N}{2}
    \right\}.
\end{aligned}
\]
Since
\(
    \kappa_N
    =
    d_N-a_N-\Lambda_N(d_N),
\)
the exponent equals $-\kappa_N/2$. Thus,
\begin{equation}
    \Pi_{e,N}
    \left(
        \mathcal B_N
        \mid
        \mathbf Y_e,X
    \right)
    \leq
    \exp
    \left(
        -\frac{\kappa_N}{2}
    \right)
    \quad
    \text{on }\mathcal E_N.
    \label{eq:selection_mass_on_good_event}
\end{equation}
Because $\kappa_N\to\infty$,
\(
    \exp(-\kappa_N/2)\longrightarrow0.
\)
For any fixed $\varepsilon>0$, the right-hand side of
\eqref{eq:selection_mass_on_good_event} is smaller than $\varepsilon$
for all sufficiently large $N$. Therefore,
\[
\begin{aligned}
    \mathbb P_\star
    \left\{
        \Pi_{e,N}
        \left(
            \mathcal B_N
            \mid
            \mathbf Y_e,X
        \right)
        >
        \varepsilon
    \right\}
    \leq
    \mathbb P_\star(\mathcal E_N^c)
    \longrightarrow0,
\end{aligned}
\]
which proves
\eqref{eq:selection_predictive_concentration}.
Since posterior mass is bounded by one,
\[
\begin{aligned}
    &
    \mathbb E_\star
    \left[
        \Pi_{e,N}
        \left(
            \mathcal B_N
            \mid
            \mathbf Y_e,X
        \right)
    \right]\\
    &=
    \mathbb E_\star
    \left[
        \Pi_{e,N}
        \left(
            \mathcal B_N
            \mid
            \mathbf Y_e,X
        \right)
        \mathbf 1_{\mathcal E_N}
    \right]
    +
    \mathbb E_\star
    \left[
        \Pi_{e,N}
        \left(
            \mathcal B_N
            \mid
            \mathbf Y_e,X
        \right)
        \mathbf 1_{\mathcal E_N^c}
    \right]\\
    &\leq
    \exp(-\kappa_N/2)
    +
    \mathbb P_\star(\mathcal E_N^c)
    \longrightarrow0.
\end{aligned}
\]
This proves
\eqref{eq:selection_predictive_mean_concentration}.
\\
Suppose next that $\widehat S$ is drawn from the working subset
posterior. Conditional on $(\mathbf Y_e,X)$,
\[
    \mathbb P
    \left(
        \widehat S\in\mathcal B_N
        \mid
        \mathbf Y_e,X
    \right)
    =
    \Pi_{e,N}
    \left(
        \mathcal B_N
        \mid
        \mathbf Y_e,X
    \right).
\]
Taking expectation over the data and auxiliary posterior-sampling
randomness gives
\[
\begin{aligned}
    \mathbb P_\star
    \left(
        \widehat S\in\mathcal B_N
    \right)
    =
    \mathbb E_\star
    \left[
        \Pi_{e,N}
        \left(
            \mathcal B_N
            \mid
            \mathbf Y_e,X
        \right)
    \right]
    \longrightarrow0.
\end{aligned}
\]
This proves
\eqref{eq:selection_posterior_draw}.
\\
Finally, let $\widehat S_{\mathrm{MAP}}$ be any
maximum-a-posteriori subset. For every $S\in\mathcal B_N$,
\[
\begin{aligned}
    \frac{
        \Pi_{e,N}(S\mid\mathbf Y_e,X)
    }{
        \Pi_{e,N}(S_N^\star\mid\mathbf Y_e,X)
    }
    &=
    \frac{
        \Pi_N(S)
    }{
        \pi_N^\star
    }
    \exp
    \left\{
        \ell_{e,N}(S)
        -
        \ell_{e,N}(S_N^\star)
    \right\}.
\end{aligned}
\]
Since
\(
    \Pi_N(S)
    \leq
    \Pi_N(\mathcal B_N),
\)
on $\mathcal E_N$,
\[
\begin{aligned}
    \frac{
        \Pi_{e,N}(S\mid\mathbf Y_e,X)
    }{
        \Pi_{e,N}(S_N^\star\mid\mathbf Y_e,X)
    }
    \leq
    \exp
    \left\{
        \Lambda_N(d_N)
        -
        d_N
        +
        a_N
        +
        \frac{\kappa_N}{2}
    \right\}
    &=
    \exp(-\kappa_N/2).
\end{aligned}
\]
Because $\kappa_N\to\infty$, this quantity is strictly smaller than
one for all sufficiently large $N$. Hence every subset in
$\mathcal B_N$ has strictly smaller posterior probability than
$S_N^\star$ on $\mathcal E_N$. Therefore no maximum-a-posteriori
subset belongs to $\mathcal B_N$ on this event, and
\[
    \mathbb P_\star
    \left(
        \widehat S_{\mathrm{MAP}}
        \in
        \mathcal B_N
    \right)
    \leq
    \mathbb P_\star(\mathcal E_N^c)
    \longrightarrow0.
\]
This completes the proof.
\end{proof}

\subsection{Proof of Corollary~\ref{cor:selection_target_risk}}
\label{app:proof:cor:selection_target_risk}

\begin{proof}
For brevity, write
\(
    \mathcal B_N
    =
    \mathcal B_N^{\mathrm{pred}}(d_N)
\)
and define
\[
    \mathcal C_N
    =
    \left\{
        S\in\mathcal S_J:
        R_0^\dagger(S)-R_0^*
        \leq
        \omega_N(d_N)
    \right\}.
\]
By the definition of
$\omega_N(d_N)$,
every subset outside $\mathcal B_N$ belongs to $\mathcal C_N$.
Hence
\(
    \mathcal S_J\setminus\mathcal B_N
    \subseteq
    \mathcal C_N.
\)
Taking complements within $\mathcal S_J$ gives
\begin{equation}
    \mathcal C_N^c
    \subseteq
    \mathcal B_N.
    \label{eq:selection_calibration_complement}
\end{equation}
It follows immediately that
\[
\begin{aligned}
    1-
    \Pi_{e,N}
    \left(
        \mathcal C_N
        \mid
        \mathbf Y_e,X
    \right)
    =
    \Pi_{e,N}
    \left(
        \mathcal C_N^c
        \mid
        \mathbf Y_e,X
    \right)
    &\leq
    \Pi_{e,N}
    \left(
        \mathcal B_N
        \mid
        \mathbf Y_e,X
    \right).
\end{aligned}
\]
The right-hand side converges to zero in
$\mathbb P_\star$-probability by
Theorem~\ref{thm:predictive_selection}. Therefore,
\[
    \Pi_{e,N}
    \left(
        \mathcal C_N
        \mid
        \mathbf Y_e,X
    \right)
    \xrightarrow{\mathbb P_\star}1,
\]
which proves
\eqref{eq:selection_calibrated_risk_posterior}.
\\
The same inclusion also gives
\[
\begin{aligned}
    1-
    \mathbb E_\star
    \left[
        \Pi_{e,N}
        \left(
            \mathcal C_N
            \mid
            \mathbf Y_e,X
        \right)
    \right]
    =
    \mathbb E_\star
    \left[
        \Pi_{e,N}
        \left(
            \mathcal C_N^c
            \mid
            \mathbf Y_e,X
        \right)
    \right]
    &\leq
    \mathbb E_\star
    \left[
        \Pi_{e,N}
        \left(
            \mathcal B_N
            \mid
            \mathbf Y_e,X
        \right)
    \right].
\end{aligned}
\]
The last quantity converges to zero by
Theorem~\ref{thm:predictive_selection}, proving
\eqref{eq:selection_calibrated_risk_mean}.
Now let $\widehat S$ be either selection rule covered by
Theorem~\ref{thm:predictive_selection}. From
\eqref{eq:selection_calibration_complement},
\(
    \left\{
        R_0^\dagger(\widehat S)-R_0^*
        >
        \omega_N(d_N)
    \right\}
    \subseteq
    \left\{
        \widehat S\in\mathcal B_N
    \right\}.
\)
Therefore,
\[
\begin{aligned}
    \mathbb P_\star
    \left\{
        R_0^\dagger(\widehat S)-R_0^*
        >
        \omega_N(d_N)
    \right\}
    \leq
    \mathbb P_\star
    \left(
        \widehat S\in\mathcal B_N
    \right)
    \longrightarrow0,
\end{aligned}
\]
where the convergence follows from
Theorem~\ref{thm:predictive_selection}.
This proves
\eqref{eq:selection_calibrated_risk_rule}.
\\
If $\omega_N(d_N)\to0$, then for every fixed $\varepsilon>0$,
$\omega_N(d_N)<\varepsilon$ for all sufficiently large $N$.
Consequently,
\[
\begin{aligned}
    \mathbb P_\star
    \left\{
        R_0^\dagger(\widehat S)-R_0^*
        >
        \varepsilon
    \right\}
    \leq
    \mathbb P_\star
    \left\{
        R_0^\dagger(\widehat S)-R_0^*
        >
        \omega_N(d_N)
    \right\}
    \longrightarrow0.
\end{aligned}
\]
Hence
\[
    R_0^\dagger(\widehat S)-R_0^*
    \xrightarrow{\mathbb P_\star}0,
\]
which proves
\eqref{eq:selection_target_risk_consistency}.
\\
Finally, let
$\epsilon_N>\omega_N(d_N)$.
For every
$S\notin\mathcal B_N$,
\(
    R_0^\dagger(S)-R_0^*
    \leq
    \omega_N(d_N)
    <
    \epsilon_N,
\)
therefore,
\(
    \mathcal S_J\setminus\mathcal B_N
    \subseteq
    \mathcal G_N^{\mathrm{risk}}(\epsilon_N).
\)
Equivalently,
\(
    \mathcal S_J
    \setminus
    \mathcal G_N^{\mathrm{risk}}(\epsilon_N)
    \subseteq
    \mathcal B_N.
\)
Thus
\[
\begin{aligned}
    1-
    \Pi_{e,N}
    \left(
        \mathcal G_N^{\mathrm{risk}}(\epsilon_N)
        \mid
        \mathbf Y_e,X
    \right)
    \leq
    \Pi_{e,N}
    \left(
        \mathcal B_N
        \mid
        \mathbf Y_e,X
    \right)
    \xrightarrow{\mathbb P_\star}0,
\end{aligned}
\]
which proves
\eqref{eq:selection_target_risk_posterior}.
\\
Similarly,
\(
    \left\{
        R_0^\dagger(\widehat S)
        \geq
        R_0^*+\epsilon_N
    \right\}
    \subseteq
    \left\{
        \widehat S\in\mathcal B_N
    \right\}.
\)
Hence
\[
\begin{aligned}
    \mathbb P_\star
    \left\{
        R_0^\dagger(\widehat S)
        \geq
        R_0^*+\epsilon_N
    \right\}
    \leq
    \mathbb P_\star
    \left(
        \widehat S\in\mathcal B_N
    \right)
    \longrightarrow0,
\end{aligned}
\]
which is equivalent to
\eqref{eq:selection_target_risk_rule}.
\end{proof}

\clearpage
\section{Proofs of posterior contraction, minimax optimality, and transfer gains}
\label{app:contraction-proofs}

\subsection{Proof of Proposition~\ref{prop:pc-information}}
\label{app:proof:prop:pc-information}

\begin{proof}
Fix a frequency $j$.
For $S\neq\varnothing$, let
$\mathbf Z_{j,S}$ denote the vector of source observations at frequency
$j$. Under \eqref{eq:pc-sequence-experiment},
the joint Gaussian covariance of
$\theta_{0j}$ and $\mathbf Z_{j,S}$ is
\[
    \operatorname{Cov}
    \begin{pmatrix}
        \theta_{0j}\\
        \mathbf Z_{j,S}
    \end{pmatrix}
    =
    \begin{pmatrix}
        \lambda_jB_{00}
        &
        \lambda_jB_{0S}\\
        \lambda_jB_{S0}
        &
        \lambda_jB_{SS}
        +
        D_{S,\mathrm{src}}^{-1}
    \end{pmatrix}.
\]
The Gaussian conditional-variance formula therefore gives
\[
\begin{aligned}
    \operatorname{Var}
    \left(
        \theta_{0j}
        \mid
        \mathbf Z_{j,S}
    \right)
    =
    \lambda_jB_{00}
    -
    \lambda_j^2
    B_{0S}
    \left(
        \lambda_jB_{SS}
        +
        D_{S,\mathrm{src}}^{-1}
    \right)^{-1}
    B_{S0}
    =
    c_{j,S}.
\end{aligned}
\]
For $S=\varnothing$, this conditional variance is simply
$\lambda_jB_{00}$, agreeing with the stated convention.
Conditional on the source observations, the target observation is
\(
    Z_{0j}
    =
    \theta_{0j}
    +
    \tau_0^{-1/2}\xi_{0j}.
\)
Thus the source-conditioned Gaussian prior for $\theta_{0j}$ has
variance $c_{j,S}$, while the target observation contributes precision
$\tau_0$. Gaussian conjugacy gives
\[
    \operatorname{Var}
    \left(
        \theta_{0j}
        \mid
        \mathbf Z_j
    \right)
    =
    \left(
        c_{j,S}^{-1}
        +
        \tau_0
    \right)^{-1},
\]
which proves
\eqref{eq:pc-frequency-information}.
\\
Suppose next that $B_S\succ0$. Its principal source block
$B_{SS}$ is then positive definite, and the Schur complement of
$B_{SS}$ in $B_S$ satisfies
\(
    \chi_S
    =
    B_{00}
    -
    B_{0S}B_{SS}^{-1}B_{S0}
    >
    0.
\)
Since
\(
    \lambda_jB_{SS}
    +
    D_{S,\mathrm{src}}^{-1}
    \succeq
    \lambda_jB_{SS},
\)
the order-reversing property of matrix inversion yields
\[
    \left(
        \lambda_jB_{SS}
        +
        D_{S,\mathrm{src}}^{-1}
    \right)^{-1}
    \preceq
    \lambda_j^{-1}B_{SS}^{-1}.
\]
Hence
\[
\begin{aligned}
    c_{j,S}
    =
    \lambda_jB_{00}
    -
    \lambda_j^2
    B_{0S}
    \left(
        \lambda_jB_{SS}
        +
        D_{S,\mathrm{src}}^{-1}
    \right)^{-1}
    B_{S0}
    \geq
    \lambda_jB_{00}
    -
    \lambda_j
    B_{0S}B_{SS}^{-1}B_{S0}
    &=
    \lambda_j\chi_S,
\end{aligned}
\]
proving
\eqref{eq:pc-source-information-floor}.
\\
Because the map
\(
    c\longmapsto
    (\tau_0+c^{-1})^{-1}
\)
is increasing on $(0,\infty)$,
\eqref{eq:pc-source-information-floor} implies
\[
    v_{0j}(S)
    \geq
    \left\{
        \tau_0
        +
        (\lambda_j\chi_S)^{-1}
    \right\}^{-1},
\]
which proves
\eqref{eq:pc-target-variance-floor}.
\\
If
$\min_{k\in S}\tau_k\to\infty$, then
\(
    D_{S,\mathrm{src}}^{-1}
    \longrightarrow0
\)
in operator norm. Continuity of matrix inversion on the positive
definite cone gives
\[
    \left(
        \lambda_jB_{SS}
        +
        D_{S,\mathrm{src}}^{-1}
    \right)^{-1}
    \longrightarrow
    \lambda_j^{-1}B_{SS}^{-1}.
\]
Substituting this limit into
\eqref{eq:pc-source-conditional-variance}
yields
\(
    c_{j,S}
    \longrightarrow
    \lambda_j\chi_S,
\)
and continuity of the scalar posterior-variance map proves
\eqref{eq:pc-oracle-source-limit}.
\\
For the two-task covariance matrix,
\(
    B_{00}=1,
    ~
    B_{SS}=1,
    ~
    B_{0S}=B_{S0}=\varrho,
\)
therefore,
\[
    \chi_S=1-\varrho^2.
\]
Equation~\eqref{eq:pc-innovation-floor} follows immediately.
\\
The variance floor also gives a direct obstruction to target posterior
contraction from source information alone. Fix any frequency $j_0$ with
$\lambda_{j_0}>0$ and keep $\tau_0$ fixed. By
\eqref{eq:pc-target-variance-floor},
\[
    v_{0j_0}(S)
    \geq
    v_*
    =
    \left\{
        \tau_0
        +
        (\lambda_{j_0}\chi_S)^{-1}
    \right\}^{-1}
    >
    0,
\]
uniformly over the source precisions. Conditional on the observations,
the $j_0$th target coefficient is therefore Gaussian with variance at
least $v_*$. For any center $h\in\mathbb H$ and any $r>0$,
\[
    \left\{
        \|f-h\|_{\mathbb H}\leq r
    \right\}
    \subseteq
    \left\{
        |\theta_{0j_0}-h_{j_0}|
        \leq r
    \right\},
\]
where $h_{j_0}$ is the $j_0$th coefficient of $h$. Since a Gaussian
density with variance at least $v_*$ is bounded above by
$(2\pi v_*)^{-1/2}$,
\[
    \Pi
    \left(
        \|f-h\|_{\mathbb H}\leq r
        \mid
        \mathbf Z
    \right)
    \leq
    \frac{2r}{\sqrt{2\pi v_*}}.
\]
The right-hand side tends to zero as $r\downarrow0$, uniformly in the
posterior center and in the source precisions. Thus source precision
alone cannot produce contraction to a point when the target innovation
component $\chi_S$ remains strictly positive.
\\
Finally, consider the rank-one model
\eqref{eq:pc-rank-one-task-kernel}. There exists a scalar coefficient
$\theta_j\sim\mathcal N(0,\lambda_j)$ such that
\(
    \theta_{kj}=g_k\theta_j,
    ~
    k\in\mathcal I(S).
\)
The observation model becomes
\[
    Z_{kj}
    =
    g_k\theta_j
    +
    \tau_k^{-1/2}\xi_{kj}.
\]
Up to terms independent of $\theta_j$, the negative log posterior is
\[
\begin{aligned}
    \frac{\theta_j^2}{2\lambda_j}
    +
    \frac12
    \sum_{k\in\mathcal I(S)}
    \tau_k
    \left(
        Z_{kj}-g_k\theta_j
    \right)^2
    =
    \frac12
    \left(
        \lambda_j^{-1}
        +
        \sum_{k\in\mathcal I(S)}
        \tau_kg_k^2
    \right)
    \theta_j^2
    -
    \left(
        \sum_{k\in\mathcal I(S)}
        \tau_kg_kZ_{kj}
    \right)
    \theta_j
    +
    \text{constant}.
\end{aligned}
\]
Hence the posterior precision of $\theta_j$ is
\[
    \lambda_j^{-1}
    +
    \sum_{k\in\mathcal I(S)}
    \tau_kg_k^2
    =
    \lambda_j^{-1}
    +
    \mathfrak n_S.
\]
Since $g_0=1$, the target coefficient satisfies
$\theta_{0j}=\theta_j$. Therefore
\[
    \operatorname{Var}
    \left(
        \theta_{0j}\mid\mathbf Z_j
    \right)
    =
    \left(
        \lambda_j^{-1}
        +
        \mathfrak n_S
    \right)^{-1},
\]
which proves
\eqref{eq:pc-pooled-information}.
\end{proof}

\subsection{Proof of Lemma~\ref{lem:pc-vector-rkhs}}
\label{app:proof:lem:pc-vector-rkhs}

\begin{proof}
Let
\(
    B_S
    =
    U
    \operatorname{diag}(b_1,\ldots,b_m)
    U^\top
\) and
\(
    m=|\mathcal I(S)|,
\)
with $U$ orthogonal and $b_r\geq0$. Introduce the transformed task
coordinates
\(
    g_r
    =
    \sum_{k\in\mathcal I(S)}
    U_{kr}f^{(k)}.
\)
Under this orthogonal transformation, the separable kernel decomposes
into the direct sum of the scalar kernels
\(
    b_r k_{\mathrm{in}},
    ~ r=1,\ldots,m.
\)
For $b_r>0$, the RKHS associated with
$b_rk_{\mathrm{in}}$ is $\mathcal H_{\mathrm{in}}$ with squared norm
$b_r^{-1}\|\cdot\|_{\mathcal H_{\mathrm{in}}}^2$. Hence
\[
    \|\mathbf f_S\|_{\mathcal H_{B_S}}^2
    =
    \sum_{r:b_r>0}
    b_r^{-1}
    \|g_r\|_{\mathcal H_{\mathrm{in}}}^2.
\]
When $B_S\succ0$, all $b_r$ are positive. Expanding the right-hand
side and using
\[
    B_S^{-1}
    =
    U
    \operatorname{diag}(b_r^{-1})
    U^\top
\]
gives
\eqref{eq:pc-vector-norm}.
Moreover,
\(
    b_+^{-1}I
    \preceq
    B_S^{-1}
    \preceq
    b_-^{-1}I.
\)
Applying these operator inequalities in the task coordinate and using
orthogonality of $U$ yields
\eqref{eq:pc-vector-norm-equivalence}.
If $B_S$ is singular, components corresponding to $b_r=0$ must vanish
for the vector-valued function to belong to the RKHS. On this range,
the same argument replaces $b_r^{-1}$ by zero for the null
eigenvalues, which is precisely the action of the Moore--Penrose
pseudoinverse $B_S^\dagger$.
\end{proof}

\subsection{Proof of Lemma~\ref{lem:pc-spectral}}
\label{app:proof:lem:pc-spectral}

\begin{proof}
Set
\(
    s=\frac{\alpha}{d}
\) and
\(
    p=1+2s.
\)
For $m\geq1$, define the truncated approximation
\(
    h_m
    =
    \sum_{j=1}^{m}
    \theta_{0j}\varphi_j.
\)
By Parseval's identity and
\eqref{eq:pc-smoothness},
\[
\begin{aligned}
    \|h_m-f_0\|_{\mathbb H}^2
    =
    \sum_{j>m}\theta_{0j}^2
    \leq
    (m+1)^{-2s}
    \sum_{j>m}
    j^{2s}\theta_{0j}^2
    &\leq
    L^2(m+1)^{-2s}.
\end{aligned}
\]
Using the lower eigenvalue bound in
\eqref{eq:pc-eigenvalue-decay},
\[
\begin{aligned}
    \|h_m\|_{\mathcal H_{\mathrm{in}}}^2
    =
    \sum_{j=1}^{m}
    \frac{\theta_{0j}^2}{\lambda_j}
    \leq
    c_\lambda^{-1}
    \sum_{j=1}^{m}
    j^{1+2s}\theta_{0j}^2
    \leq
    c_\lambda^{-1}
    m
    \sum_{j=1}^{m}
    j^{2s}\theta_{0j}^2
    \leq
    c_\lambda^{-1}L^2m.
\end{aligned}
\]
For $L>0$, choose
\(
    m
    =
    \left\lfloor
        \frac{c_\lambda R^2}{L^2}
    \right\rfloor,
\)
with $R$ large enough that $m\geq1$. Then
$\|h_m\|_{\mathcal H_{\mathrm{in}}}\leq R$, while
\[
    \|h_m-f_0\|_{\mathbb H}
    \leq
    C R^{-2s}
    =
    C R^{-2\alpha/d}.
\]
This proves
\eqref{eq:pc-approximation}. If $L=0$, then $f_0=0$ in
$\mathbb H$ and the result is immediate.

We next establish the centered small-ball bound. Choose
\(
    m
    =
    \left\lceil
        C_0\varepsilon^{-1/s}
    \right\rceil
\)
with $C_0$ sufficiently large. Since
$\lambda_j\lesssim j^{-1-2s}$,
\(
    \sum_{j>m}\lambda_j
    \leq
    C m^{-2s}
    \leq
    \frac{\varepsilon^2}{4}.
\)
Therefore, by Markov's inequality,
\[
    \mathbb P
    \left(
        \sum_{j>m}
        \lambda_jZ_j^2
        \leq
        \frac{\varepsilon^2}{2}
    \right)
    \geq
    \frac12.
\]
For the first $m$ coordinates, impose
\(
    |Z_j|
    \leq
    \frac{
        \varepsilon
    }{
        \sqrt{2m\lambda_j}
    },
    ~
    j=1,\ldots,m.
\)
On this event,
\(
    \sum_{j=1}^{m}
    \lambda_jZ_j^2
    \leq
    \frac{\varepsilon^2}{2}.
\)
Using
$\lambda_j\leq C_\lambda j^{-p}$
and
$m\asymp\varepsilon^{-1/s}$,
there exists $c>0$ such that
\(
    \frac{
        \varepsilon
    }{
        \sqrt{2m\lambda_j}
    }
    \geq
    c
    \left(
        \frac jm
    \right)^{p/2}.
\)
The thresholds are uniformly bounded above for $j\leq m$, so a
standard Gaussian density bound on a bounded interval gives
\[
    \mathbb P
    \left(
        |Z_j|
        \leq
        \frac{
            \varepsilon
        }{
            \sqrt{2m\lambda_j}
        }
    \right)
    \geq
    c_0
    \left(
        \frac jm
    \right)^{p/2}
\]
for some $c_0\in(0,1)$ independent of $j,m$, and $\varepsilon$.
The head and tail coordinates are independent. Hence
\[
\begin{aligned}
    \mathbb P
    \left(
        \|W\|_{\mathbb H}\leq\varepsilon
    \right)
    \geq
    \frac12
    \prod_{j=1}^{m}
    c_0
    \left(
        \frac jm
    \right)^{p/2}
    &=
    \frac12
    c_0^m
    \left(
        \frac{m!}{m^m}
    \right)^{p/2}.
\end{aligned}
\]
Stirling's inequality implies
\[
    \mathbb P
    \left(
        \|W\|_{\mathbb H}\leq\varepsilon
    \right)
    \geq
    \exp(-C m)
    \geq
    \exp
    \left(
        -C_1\varepsilon^{-1/s}
    \right).
\]
Since $1/s=d/\alpha$, this proves
\eqref{eq:pc-centered-small-ball}.
\\
For the shifted bound, the approximation argument above gives an
$h\in\mathcal H_{\mathrm{in}}$ such that
\(
    \|h-f_0\|_{\mathbb H}
    \leq
    \frac{\varepsilon}{2}
\) and
\(
    \|h\|_{\mathcal H_{\mathrm{in}}}^2
    \leq
    C\varepsilon^{-d/\alpha}.
\)
Let
\(
    A_\varepsilon
    =
    \left\{
        g\in\mathbb H:
        \|g\|_{\mathbb H}
        \leq
        \frac{\varepsilon}{2}
    \right\}.
\)
The Cameron--Martin formula gives
\[
\begin{aligned}
    \mathbb P(W\in h+A_\varepsilon)
    &=
    \exp
    \left(
        -\frac12
        \|h\|_{\mathcal H_{\mathrm{in}}}^2
    \right)
    \mathbb E
    \left[
        \mathbf 1_{A_\varepsilon}(W)
        \exp\{\ell_h(W)\}
    \right],
\end{aligned}
\]
where $\ell_h$ is the centered Gaussian linear functional associated
with $h$. Since $A_\varepsilon$ is symmetric and the law of $W$ is
symmetric,
\[
\begin{aligned}
    \mathbb E
    \left[
        \mathbf 1_{A_\varepsilon}(W)
        \exp\{\ell_h(W)\}
    \right]
    =
    \mathbb E
    \left[
        \mathbf 1_{A_\varepsilon}(W)
        \cosh\{\ell_h(W)\}
    \right]
    \geq
    \mathbb P(W\in A_\varepsilon).
\end{aligned}
\]
Moreover,
\(
    h+A_\varepsilon
    \subseteq
    \left\{
        g:
        \|g-f_0\|_{\mathbb H}
        \leq
        \varepsilon
    \right\}.
\)
Combining the Cameron--Martin bound with
\eqref{eq:pc-centered-small-ball} therefore gives
\[
    -\log
    \mathbb P
    \left(
        \|W-f_0\|_{\mathbb H}
        \leq
        \varepsilon
    \right)
    \leq
    C_2\varepsilon^{-d/\alpha},
\]
which proves
\eqref{eq:pc-shifted-small-ball}.
\end{proof}

\subsection{Proof of Lemma~\ref{lem:pc-perturbation}}
\label{app:proof:lem:pc-perturbation}

\begin{proof}
Subtracting
\eqref{eq:pc-oracle-posterior-mean}
from
\eqref{eq:pc-posterior-mean}
and using
\(
    \mathbf Y_{e,S}
    -
    \overline{\mathbf Y}_S
    =
    \mathbf b_{e,S}
\)
gives
\[
    \mu_{S,e,N}^{(0)}
    -
    \overline\mu_{S,N}^{(0)}
    =
    \mathcal P_{S,N}\mathbf b_{e,S},
\]
which proves
\eqref{eq:pc-shift}.
\\
The triangle inequality then gives
\[
\begin{aligned}
    \|\mu_{S,e,N}^{(0)}-f_0\|_{\mathbb H}
    \leq
    \|
        \overline\mu_{S,N}^{(0)}-f_0
    \|_{\mathbb H}
    +
    \|
        \mu_{S,e,N}^{(0)}
        -
        \overline\mu_{S,N}^{(0)}
    \|_{\mathbb H},
\end{aligned}
\]
and
\eqref{eq:pc-shift-bound}
follows from
\eqref{eq:pc-shift}.
\\
Finally,
\[
\begin{aligned}
    \mathcal P_{S,N}\mathbf b_{e,S}
    &=
    \left(
        \mathcal P_{S,N}
        \Lambda_{S,N}^{1/2}
    \right)
    \left(
        \Lambda_{S,N}^{-1/2}
        \mathbf b_{e,S}
    \right).
\end{aligned}
\]
Taking the operator norm from Euclidean space into $\mathbb H$ proves
\eqref{eq:pc-bootstrap-operator-bound}.
\end{proof}

\subsection{Proof of Corollary~\ref{cor:pc-bootstrap-perturbation-rate}}
\label{app:proof:cor:pc-bootstrap-perturbation-rate}

\begin{proof}
By
\eqref{eq:pc-bootstrap-operator-bound}
and
\eqref{eq:pc-operator-envelope},
\[
\begin{aligned}
    \bigl(
        d_{S,e,N}^{\mathrm{bt}}
    \bigr)^2
    \leq
    L_{\mathcal P,N}^2
    \|
        \Lambda_{S,N}^{-1/2}
        \mathbf b_{e,S}
    \|_2^2
    &\leq
    L_{\mathcal P,N}^2
    \underline\nu_N^{-1}
    \|
        \mathbf b_{e,S}
    \|_2^2.
\end{aligned}
\]
Taking expectation and expanding the Euclidean norm gives
\[
\begin{aligned}
    \mathbb E_\star
    \|
        \mathbf b_{e,S}
    \|_2^2
    =
    \sum_{k\in\mathcal I(S)}
    \sum_{t=1}^{n_k}
    \mathbb E_\star
    \left[
        (b_{t,e}^{(k)})^2
    \right]
    &\leq
    n(S)\beta_{e,N}^2.
\end{aligned}
\]
Combining the two displays proves
\eqref{eq:pc-bootstrap-moment}.
\\
This argument uses only linearity of expectation and does not require
independence.
For the uniform bound, the union bound and Markov's inequality give
\[
\begin{aligned}
    \mathbb P_\star
    \left(
        \max_{S\in\mathcal G_N^{\mathrm{ctr}}}
        d_{S,e,N}^{\mathrm{bt}}
        >
        u_N
    \right)
    \leq
    \sum_{S\in\mathcal G_N^{\mathrm{ctr}}}
    \mathbb P_\star
    \left(
        d_{S,e,N}^{\mathrm{bt}}
        >
        u_N
    \right)
    &\leq
    \frac{1}{u_N^2}
    \sum_{S\in\mathcal G_N^{\mathrm{ctr}}}
    \mathbb E_\star
    \left[
        \bigl(
            d_{S,e,N}^{\mathrm{bt}}
        \bigr)^2
    \right].
\end{aligned}
\]
Applying
\eqref{eq:pc-bootstrap-moment}
to every term yields
\eqref{eq:pc-bootstrap-uniform-tail}.
\end{proof}

\subsection{Proof of Lemma~\ref{lem:pc-gaussian-tail}}
\label{app:proof:lem:pc-gaussian-tail}

\begin{proof}
Let $\{\gamma_j\}_{j\geq1}$ denote the eigenvalues of $C$ and write
\[
    G
    =
    \sum_{j\geq1}
    \sqrt{\gamma_j}\,
    Z_j e_j,
    \quad
    Z_j
    \stackrel{\mathrm{iid}}{\sim}
    \mathcal N(0,1).
\]
Set
\[
    V=\operatorname{tr}(C)
    =
    \sum_{j\geq1}\gamma_j,
    \quad
    v=\|C\|_{\mathrm{op}}
    =
    \sup_j\gamma_j.
\]
For
$0<u<(2v)^{-1}$,
\[
\begin{aligned}
    \log
    \mathbb E
    \exp
    \left\{
        u
        \bigl(
            \|G\|_{\mathbb H}^2-V
        \bigr)
    \right\}
    &=
    \frac12
    \sum_{j\geq1}
    \left[
        -\log(1-2u\gamma_j)
        -
        2u\gamma_j
    \right].
\end{aligned}
\]
Using
\[
    -\log(1-x)-x
    \leq
    \frac{x^2}{2(1-x)},
    \quad
    0\leq x<1,
\]
we have
\[
\begin{aligned}
    \log
    \mathbb E
    \exp
    \left\{
        u
        \bigl(
            \|G\|_{\mathbb H}^2-V
        \bigr)
    \right\}
    \leq
    \frac{
        u^2
        \sum_j\gamma_j^2
    }{
        1-2uv
    }
    \leq
    \frac{
        u^2vV
    }{
        1-2uv
    }.
\end{aligned}
\]
The infinite-dimensional identity follows from the finite-dimensional
partial sums and monotone convergence.
The corresponding Chernoff bound gives
\[
    \mathbb P
    \left\{
        \|G\|_{\mathbb H}^2
        >
        V
        +
        2\sqrt{vVt}
        +
        2vt
    \right\}
    \leq
    e^{-t}.
\]
Since
\[
\begin{aligned}
    \left(
        \sqrt V
        +
        \sqrt{2vt}
    \right)^2
    =
    V
    +
    2\sqrt{2vVt}
    +
    2vt
    &\geq
    V
    +
    2\sqrt{vVt}
    +
    2vt,
\end{aligned}
\]
\eqref{eq:pc-gaussian-tail}
follows. The case $v=0$ is immediate.
\\
Now let
$f=\mu+G$. By the triangle inequality,
\[
    \|f-f_0\|_{\mathbb H}
    \leq
    \|\mu-f_0\|_{\mathbb H}
    +
    \|G\|_{\mathbb H}.
\]
Applying
\eqref{eq:pc-gaussian-tail}
to the final term proves
\eqref{eq:pc-posterior-tail}.
\\
Finally,
\[
\begin{aligned}
    \mathbb E
    \|f-f_0\|_{\mathbb H}^2
    &=
    \mathbb E
    \|
        \mu-f_0+G
    \|_{\mathbb H}^2\\
    &=
    \|\mu-f_0\|_{\mathbb H}^2
    +
    2
    \left\langle
        \mu-f_0,
        \mathbb E G
    \right\rangle_{\mathbb H}
    +
    \mathbb E
    \|G\|_{\mathbb H}^2\\
    &=
    \|\mu-f_0\|_{\mathbb H}^2
    +
    \operatorname{tr}(C),
\end{aligned}
\]
which proves
\eqref{eq:pc-posterior-moment}.
\end{proof}

\subsection{Proof of Theorem~\ref{thm:posterior-contraction}}
\label{app:proof:thm:posterior-contraction}

\begin{proof}
Define
\(
    \mathcal E_N^{\mathrm{orc}}
    =
    \left\{
        \mathcal R_N^{\mathrm{orc}}
        \leq
        C_0
    \right\}
\)
and
\(
    \mathcal E_N^{\mathrm{bt}}
    =
    \left\{
        \max_{S\in\mathcal G_N^{\mathrm{ctr}}}
        d_{S,e,N}^{\mathrm{bt}}
        \leq
        u_N
    \right\}.
\)
By definition,
\(
    \mathbb P_\star
    \left(
        (\mathcal E_N^{\mathrm{orc}})^c
    \right)
    =
    \zeta_N^{\mathrm{orc}}(C_0).
\)
Corollary~\ref{cor:pc-bootstrap-perturbation-rate} gives
\[
    \mathbb P_\star
    \left(
        (\mathcal E_N^{\mathrm{bt}})^c
    \right)
    \leq
    \zeta_N^{\mathrm{bt}}(u_N).
\]
Therefore, for
\(
    \mathcal E_N
    =
    \mathcal E_N^{\mathrm{orc}}
    \cap
    \mathcal E_N^{\mathrm{bt}},
\)
the union bound yields
\begin{equation}
    \mathbb P_\star(\mathcal E_N^c)
    \leq
    \zeta_N^{\mathrm{orc}}(C_0)
    +
    \zeta_N^{\mathrm{bt}}(u_N)
    \longrightarrow0.
    \label{eq:pc-good-event}
\end{equation}
Fix
$S\in\mathcal G_N^{\mathrm{ctr}}$.
On $\mathcal E_N$,
\(
    \|
        \overline\mu_{S,N}^{(0)}-f_0
    \|_{\mathbb H}
    \leq
    C_0\varepsilon_N
\)
and
\(
    d_{S,e,N}^{\mathrm{bt}}
    \leq
    u_N.
\)
Lemma~\ref{lem:pc-perturbation} therefore implies
\begin{equation}
    \|
        \mu_{S,e,N}^{(0)}-f_0
    \|_{\mathbb H}
    \leq
    C_0\varepsilon_N+u_N.
    \label{eq:pc-mean-bound}
\end{equation}
The definition of
$\mathcal E_N^{\mathrm{orc}}$
also gives
\begin{equation}
    \operatorname{tr}
    \bigl(
        C_{S,N}^{(0)}
    \bigr)
    \leq
    C_0^2\varepsilon_N^2
    \label{eq:pc-trace-bound}
\end{equation}
and
\begin{equation}\nonumber
    \|
        C_{S,N}^{(0)}
    \|_{\mathrm{op}}
    \leq
    \frac{
        C_0^2
    }{
        \mathfrak i_N
    }.
    \label{eq:pc-operator-bound}
\end{equation}
Apply
Lemma~\ref{lem:pc-gaussian-tail}
to
\(
    \Pi_{S,e,N}^{(0)}
    =
    \mathcal N_{\mathbb H}
    \left(
        \mu_{S,e,N}^{(0)},
        C_{S,N}^{(0)}
    \right)
\)
with $t=t_N$. Since
\(
    t_N
    =
    \mathfrak i_N\varepsilon_N^2,
\)
we have
\[
    \sqrt{
        \operatorname{tr}
        (C_{S,N}^{(0)})
    }
    \leq
    C_0\varepsilon_N
\]
and
\[
\begin{aligned}
    \sqrt{
        2
        \|
            C_{S,N}^{(0)}
        \|_{\mathrm{op}}
        t_N
    }
    \leq
    \sqrt{
        2
        \frac{
            C_0^2
        }{
            \mathfrak i_N
        }
        \mathfrak i_N
        \varepsilon_N^2
    }
    &=
    \sqrt{2}\,
    C_0\varepsilon_N.
\end{aligned}
\]
Combining these bounds with
\eqref{eq:pc-mean-bound}
gives
\[
    \Pi_{S,e,N}^{(0)}
    \left(
        \|f-f_0\|_{\mathbb H}
        >
        (2+\sqrt2)C_0\varepsilon_N+u_N
    \right)
    \leq
    e^{-t_N}
\]
simultaneously for all
$S\in\mathcal G_N^{\mathrm{ctr}}$.
Now use the mixture representation
\[
    \widetilde\Pi_{e,N}^{(0)}
    =
    \sum_{S\in\mathcal S_J}
    \omega_{e,N}(S)
    \Pi_{S,e,N}^{(0)}.
\]
On $\mathcal E_N$,
\[
\begin{aligned}
    \widetilde\Pi_{e,N}^{(0)}
    \left(
        \|f-f_0\|_{\mathbb H}
        >
        R_N
    \right)
    &\leq
    e^{-t_N}
    \sum_{S\in\mathcal G_N^{\mathrm{ctr}}}
    \omega_{e,N}(S)
    +
    \sum_{S\notin\mathcal G_N^{\mathrm{ctr}}}
    \omega_{e,N}(S)\\
    &\leq
    e^{-t_N}
    +
    \sum_{S\notin\mathcal G_N^{\mathrm{ctr}}}
    \omega_{e,N}(S).
\end{aligned}
\]
By
\eqref{eq:pc-selection-contraction-link},
\(
    S\notin\mathcal G_N^{\mathrm{ctr}}
    \Longrightarrow
    S\in
    \mathcal B_N^{\mathrm{pred}}(d_N)
\)
for all sufficiently large $N$. Hence
\[
\begin{aligned}
    \sum_{S\notin\mathcal G_N^{\mathrm{ctr}}}
    \omega_{e,N}(S)
    &\leq
    \Pi_{e,N}
    \left(
        \mathcal B_N^{\mathrm{pred}}(d_N)
        \mid
        \mathbf Y_e,X
    \right).
\end{aligned}
\]
This proves
\eqref{eq:pc-mixture-contraction-bound}.
\\
By
Theorem~\ref{thm:predictive_selection},
\[
    \Pi_{e,N}
    \left(
        \mathcal B_N^{\mathrm{pred}}(d_N)
        \mid
        \mathbf Y_e,X
    \right)
    \xrightarrow{\mathbb P_\star}0.
\]
Also,
$t_N\to\infty$, so $e^{-t_N}\to0$, while
\eqref{eq:pc-good-event} shows that
$\mathbb P_\star(\mathcal E_N^c)\to0$.
Therefore
\[
    \widetilde\Pi_{e,N}^{(0)}
    \left(
        \|f-f_0\|_{\mathbb H}
        >
        R_N
    \right)
    \xrightarrow{\mathbb P_\star}0,
\]
which proves
\eqref{eq:pc-mixture-contraction}.
\\
Finally, by the Gaussian posterior moment identity,
\[
\begin{aligned}
    \int
        \|f-f_0\|_{\mathbb H}^2
        \Pi_{S,e,N}^{(0)}(\mathrm df)
    =
    \|
        \mu_{S,e,N}^{(0)}-f_0
    \|_{\mathbb H}^2
    +
    \operatorname{tr}
    \bigl(
        C_{S,N}^{(0)}
    \bigr).
\end{aligned}
\]
Using
\eqref{eq:pc-mean-bound}
and
\eqref{eq:pc-trace-bound}
and taking the supremum over
$S\in\mathcal G_N^{\mathrm{ctr}}$
gives
\eqref{eq:pc-good-moment}.
\end{proof}

\subsection{Proof of Theorem~\ref{thm:pc-fullrank}}
\label{app:proof:thm:pc-fullrank}

\begin{proof}
For each frequency $j$, Gaussian conjugacy gives posterior covariance
\(
    V_{j,S}
    =
    \left(
        \lambda_j^{-1}B_S^{-1}
        +
        D_S
    \right)^{-1}.
\)
By
\eqref{eq:pc-fullrank-task-bounds},
\(
    b_+^{-1}I
    \preceq
    B_S^{-1}
    \preceq
    b_-^{-1}I.
\)
Hence
\[
\begin{aligned}
    \left(
        \lambda_j^{-1}b_-^{-1}I+D_S
    \right)^{-1}
    \preceq
    V_{j,S}
    \preceq
    \left(
        \lambda_j^{-1}b_+^{-1}I+D_S
    \right)^{-1},
\end{aligned}
\]
because matrix inversion reverses Loewner order. Taking the target
coordinate yields
\begin{equation}
    \frac{1}{
        \tau_0+(b_-\lambda_j)^{-1}
    }
    \leq
    (V_{j,S})_{00}
    \leq
    \frac{1}{
        \tau_0+(b_+\lambda_j)^{-1}
    }.
    \label{eq:pc-fullrank-frequency-bound}
\end{equation}
Set
\(
    s=\frac{\alpha}{d}
\) and
\(
    p=1+2s.
\)
Since
$\lambda_j\asymp j^{-p}$,
take
\(
    m_0
    \asymp
    \tau_0^{1/p}.
\)
Then
$\tau_0\lambda_{m_0}\asymp1$.
Using
\eqref{eq:pc-fullrank-frequency-bound},
\[
\begin{aligned}
    \operatorname{tr}
    \bigl(
        C_{S,N}^{(0)}
    \bigr)
    =
    \sum_{j\geq1}
    (V_{j,S})_{00}
    \asymp
    \frac{m_0}{\tau_0}
    +
    \sum_{j>m_0}\lambda_j
    \asymp
    \frac{m_0}{\tau_0}
    +
    m_0^{-2s}
    \asymp
    \tau_0^{-2s/p}
    =
    \varepsilon_{0,N}^2.
\end{aligned}
\]
The upper bound
\(
    (V_{j,S})_{00}
    \leq
    \tau_0^{-1}
\)
for every $j$ also gives
\(
    \|
        C_{S,N}^{(0)}
    \|_{\mathrm{op}}
    \leq
    \tau_0^{-1}.
\)
This proves
\eqref{eq:pc-fullrank-covariance}.
\\
We next control the oracle posterior mean. At frequency $j$,
\(
    \overline{\boldsymbol\mu}_j
    =
    V_{j,S}D_S\mathbf Z_j.
\)
Since
\[
    V_{j,S}
    \left(
        \lambda_j^{-1}B_S^{-1}
        +
        D_S
    \right)
    =
    I,
\]
we have
\[
    V_{j,S}D_S-I
    =
    -
    V_{j,S}
    \lambda_j^{-1}B_S^{-1}.
\]
The deterministic target bias is therefore
\(
    \beta_j
    =
    -
    \mathbf e_0^\top
    V_{j,S}
    \lambda_j^{-1}
    B_S^{-1}
    \boldsymbol\theta_j^\star.
\)
Using
\(
    \|B_S^{-1}\|_{\mathrm{op}}
    \leq
    b_-^{-1}
\) and
\(
    V_{j,S}
    \preceq
    b_+\lambda_j I,
\)
we obtain
\[
\begin{aligned}
    \|
        \mathbf e_0^\top
        V_{j,S}
        \lambda_j^{-1}
        B_S^{-1}
    \|_2^2
    \leq
    \lambda_j^{-2}
    b_-^{-2}
    \|
        V_{j,S}
    \|_{\mathrm{op}}
    (V_{j,S})_{00}
    \leq
    \frac{
        C
    }{
        1+\tau_0\lambda_j
    }.
\end{aligned}
\]
Consequently,
\[
\begin{aligned}
    \sum_{j\geq1}|\beta_j|^2
    &\leq
    C
    \sup_{j\geq1}
    \frac{
        j^{-2s}
    }{
        1+\tau_0\lambda_j
    }
    \sum_{j\geq1}
    j^{2s}
    \|
        \boldsymbol\theta_j^\star
    \|_2^2.
\end{aligned}
\]
The first factor satisfies
\[
    \sup_{j\geq1}
    \frac{
        j^{-2s}
    }{
        1+\tau_0\lambda_j
    }
    \leq
    C\tau_0^{-2s/p}
    =
    C\varepsilon_{0,N}^2.
\]
Indeed, for $j\leq m_0$,
\[
    \frac{j^{-2s}}{1+\tau_0\lambda_j}
    \leq
    C\frac{j}{\tau_0}
    \leq
    C\frac{m_0}{\tau_0},
\]
while for $j>m_0$ it is bounded by
$m_0^{-2s}$.
Thus
\begin{equation}
    \sum_{j\geq1}|\beta_j|^2
    \leq
    C\varepsilon_{0,N}^2.
    \label{eq:pc-fullrank-bias}
\end{equation}
The centered Gaussian noise in the oracle posterior mean has
frequency-wise covariance
\(
    V_{j,S}D_SV_{j,S}.
\)
Since
\[
    V_{j,S}D_SV_{j,S}
    =
    V_{j,S}
    -
    \lambda_j^{-1}
    V_{j,S}B_S^{-1}V_{j,S},
\]
we have
\[
    0
    \preceq
    V_{j,S}D_SV_{j,S}
    \preceq
    V_{j,S}.
\]
Hence its target covariance operator, denoted
$C_{S,N}^{\mathrm{mean}}$, satisfies
\[
    \operatorname{tr}
    \bigl(
        C_{S,N}^{\mathrm{mean}}
    \bigr)
    \leq
    C\varepsilon_{0,N}^2,
    \quad
    \|
        C_{S,N}^{\mathrm{mean}}
    \|_{\mathrm{op}}
    \leq
    \tau_0^{-1}.
\]
Combining this with
\eqref{eq:pc-fullrank-bias}
proves
\eqref{eq:pc-fullrank-mean-risk}.
\\
We now prove posterior contraction. Let
\(
    G_N^{\mathrm{mean}}
    \sim
    \mathcal N_{\mathbb H}
    \left(
        0,
        C_{S,N}^{\mathrm{mean}}
    \right).
\)
The oracle mean error can be written as
\[
    \overline\mu_{S,N}^{(0)}-f_0
    =
    b_N+G_N^{\mathrm{mean}},
\]
where, by
\eqref{eq:pc-fullrank-bias},
\(
    \|b_N\|_{\mathbb H}
    \leq
    C\varepsilon_{0,N}.
\)
Take
\[
    t=t_{0,N}
    =
    \tau_0\varepsilon_{0,N}^2
    =
    \tau_0^{d/(2\alpha+d)}
    \longrightarrow\infty.
\]
Lemma~\ref{lem:pc-gaussian-tail} gives
\[
    \mathbb P_\star
    \left(
        \|
            G_N^{\mathrm{mean}}
        \|_{\mathbb H}
        >
        C_1\varepsilon_{0,N}
    \right)
    \leq
    e^{-t_{0,N}}
\]
for a fixed $C_1<\infty$. Therefore,
\begin{equation}
    \mathbb P_\star
    \left(
        \|
            \overline\mu_{S,N}^{(0)}-f_0
        \|_{\mathbb H}
        >
        C_2\varepsilon_{0,N}
    \right)
    \leq
    e^{-t_{0,N}}.
    \label{eq:pc-fullrank-mean-tail}
\end{equation}
Conditional on the observations, write a draw from the oracle posterior
as
\[
    f
    =
    \overline\mu_{S,N}^{(0)}
    +
    G_N^{\mathrm{post}},
    \quad
    G_N^{\mathrm{post}}
    \sim
    \mathcal N_{\mathbb H}
    \left(
        0,
        C_{S,N}^{(0)}
    \right).
\]
Using
\eqref{eq:pc-fullrank-covariance}
and applying
Lemma~\ref{lem:pc-gaussian-tail}
again with $t=t_{0,N}$ gives
\[
    \overline\Pi_{S,N}^{(0)}
    \left(
        \|
            G_N^{\mathrm{post}}
        \|_{\mathbb H}
        >
        C_3\varepsilon_{0,N}
    \right)
    \leq
    e^{-t_{0,N}}.
\]
On the event in
\eqref{eq:pc-fullrank-mean-tail},
the triangle inequality therefore yields
\[
    \overline\Pi_{S,N}^{(0)}
    \left(
        \|f-f_0\|_{\mathbb H}
        >
        M\varepsilon_{0,N}
    \right)
    \leq
    e^{-t_{0,N}}
\]
for $M=C_2+C_3$. Together with
\eqref{eq:pc-fullrank-mean-tail}, this proves
\eqref{eq:pc-fullrank-contraction}.
\\
Finally, let
$r_N=o(\varepsilon_{0,N})$.
For any $h\in\mathbb H$, let
\(
    G_N
    \sim
    \mathcal N_{\mathbb H}
    \left(
        0,
        C_{S,N}^{(0)}
    \right)
\)
and put
\(
    q=h-f_0
\), 
\(
    M_q
    =
    \mathbb E
    \|
        G_N+q
    \|_{\mathbb H}^2.
\)
Then
\[
    M_q
    =
    \operatorname{tr}
    \bigl(
        C_{S,N}^{(0)}
    \bigr)
    +
    \|q\|_{\mathbb H}^2
    \geq
    c\varepsilon_{0,N}^2.
\]
Moreover,
\[
\begin{aligned}
    \operatorname{Var}
    \left(
        \|G_N+q\|_{\mathbb H}^2
    \right)
    =
    2
    \operatorname{tr}
    \left[
        \bigl(
            C_{S,N}^{(0)}
        \bigr)^2
    \right]
    +
    4
    \left\langle
        q,
        C_{S,N}^{(0)}q
    \right\rangle_{\mathbb H}
    &\leq
    4
    \|
        C_{S,N}^{(0)}
    \|_{\mathrm{op}}
    M_q.
\end{aligned}
\]
Since
$r_N^2=o(\varepsilon_{0,N}^2)$,
for all sufficiently large $N$,
\(
    r_N^2
    \leq
    \frac{M_q}{2}
\)
uniformly in $q$. Chebyshev's inequality gives
\[
\begin{aligned}
    \mathbb P
    \left(
        \|G_N+q\|_{\mathbb H}
        \leq
        r_N
    \right)
    \leq
    \frac{
        16
        \|
            C_{S,N}^{(0)}
        \|_{\mathrm{op}}
    }{
        M_q
    }
    \leq
    C
    \tau_0^{-1}
    \varepsilon_{0,N}^{-2}
    &=
    C
    \tau_0^{2a_\alpha-1}.
\end{aligned}
\]
Because
\[
    2a_\alpha-1
    =
    -\frac{d}{2\alpha+d}
    <
    0,
\]
the right-hand side tends to zero uniformly in $q$, proving
\eqref{eq:pc-fullrank-lower-radius}.
\end{proof}

\subsection{Proof of Corollary~\ref{cor:pc-fullrank-bootstrap}}
\label{app:proof:cor:pc-fullrank-bootstrap}

\begin{proof}
Corollary~\ref{cor:pc-bootstrap-perturbation-rate} gives
\[
    \mathbb P_\star
    \left(
        d_{S,e,N}^{\mathrm{bt}}>u_N
    \right)
    \longrightarrow0.
\]
On the complementary event,
Lemma~\ref{lem:pc-perturbation} shifts the oracle posterior mean by at
most $u_N$ while leaving the posterior covariance unchanged.
The result therefore follows from
Theorem~\ref{thm:pc-fullrank}
and the triangle inequality.
\end{proof}

\subsection{Proof of Lemma~\ref{lem:pc-pooling}}
\label{app:proof:lem:pc-pooling}

\begin{proof}
For fixed $j$, the negative log posterior, up to terms independent of
$\theta_j$, is
\[
    \frac{\theta_j^2}{2\lambda_j}
    +
    \frac12
    \sum_{k\in\mathcal I(S)}
    \tau_k
    \bigl(
        \overline Z_{kj}-g_k\theta_j
    \bigr)^2.
\]
Collecting the quadratic and linear terms in $\theta_j$ gives posterior
precision
\[
    \lambda_j^{-1}
    +
    \sum_{k\in\mathcal I(S)}
    \tau_k g_k^2
    =
    \lambda_j^{-1}+\mathfrak n_S
\]
and posterior mean
\[
\begin{aligned}
    \frac{
        \sum_k\tau_k g_k\overline Z_{kj}
    }{
        \lambda_j^{-1}+\mathfrak n_S
    }
    =
    \frac{
        \mathfrak n_S\lambda_j
    }{
        1+\mathfrak n_S\lambda_j
    }
    \overline Z_{S,j}
    =
    w_j(\mathfrak n_S)\overline Z_{S,j}.
\end{aligned}
\]
This proves
\eqref{eq:pc-conjugate}.
\\
It remains to establish the spectral bounds. Put
\(
    s=\frac{\alpha}{d},~
    p=1+2s,
\)
so that
\(
    \lambda_j\asymp j^{-p}.
\)
For $n\geq2$, take
\(
    m=\lfloor n^{1/p}\rfloor.
\)
Since
\(
    1-w_j(n)
    =
    \frac{1}{1+n\lambda_j},
\)
for $j\leq m$,
\(
    1-w_j(n)
    \leq
    Cn^{-1}j^p.
\)
Hence
\[
\begin{aligned}
    \sum_{j\leq m}
    \bigl(1-w_j(n)\bigr)^2
    (\theta_{0j}^\star)^2
    \leq
    Cn^{-2}
    m^{2p-2s}
    \sum_{j\leq m}
    j^{2s}
    (\theta_{0j}^\star)^2
    &\leq
    Cn^{-2}m^{2p-2s}.
\end{aligned}
\]
For $j>m$,
\[
\begin{aligned}
    \sum_{j>m}
    \bigl(1-w_j(n)\bigr)^2
    (\theta_{0j}^\star)^2
    \leq
    m^{-2s}
    \sum_{j>m}
    j^{2s}
    (\theta_{0j}^\star)^2
    &\leq
    Cm^{-2s}.
\end{aligned}
\]
Because $m\asymp n^{1/p}$, both terms are bounded by
\(
    Cn^{-2s/p}
    =
    Cn^{-2a_\alpha}.
\)
Next,
\(
    v_j(n)
    =
    \frac{\lambda_j}{1+n\lambda_j}
    \leq
    \min\{n^{-1},\lambda_j\}.
\)
Therefore,
\[
\begin{aligned}
    \sum_{j\geq1}v_j(n)
    \leq
    \frac{m}{n}
    +
    \sum_{j>m}\lambda_j
    \leq
    C
    \left(
        \frac{m}{n}
        +
        m^{-2s}
    \right)
    &\leq
    Cn^{-2a_\alpha}.
\end{aligned}
\]
Moreover,
\(
    v_j(n)\leq n^{-1}.
\)
Finally,
\[
    \frac{w_j(n)^2}{n}
    =
    w_j(n)v_j(n)
    \leq
    v_j(n),
\]
so both the sum and the supremum bounds for
$w_j(n)^2/n$ follow from the corresponding bounds for $v_j(n)$.
This proves
\eqref{eq:pc-pooled-spectral-bounds} and
\eqref{eq:pc-pooled-operator-bounds}.
\end{proof}

\subsection{Proof of Theorem~\ref{thm:pc-pooled-rate}}
\label{app:proof:thm:pc-pooled-rate}

\begin{proof}
Fix
$S\in\mathcal G_N^{\mathrm{ctr}}$.
From
\eqref{eq:pc-pooled-observation},
\(
    \overline Z_{S,j}
    =
    \theta_{0j}^\star
    +
    h_{S,j}
    +
    \eta_{S,j},
\)
where
\(
    \eta_{S,j}
    =
    \frac{1}{\mathfrak n_S}
    \sum_{k\in\mathcal I(S)}
    \tau_k g_k
    \tau_k^{-1/2}\xi_{kj}.
\)
Because the $\xi_{kj}$ are independent standard Gaussian variables,
\(
    \eta_{S,j}
    \sim
    \mathcal N
    \left(
        0,
        \mathfrak n_S^{-1}
    \right),
\)
and the variables $\{\eta_{S,j}\}_{j\geq1}$ are independent across
frequencies.
By Lemma~\ref{lem:pc-pooling}, the oracle posterior mean has
coefficient
\(
    w_j(\mathfrak n_S)\overline Z_{S,j}.
\)
Hence its coefficient error relative to $f_0$ is
\[
\begin{aligned}
    w_j(\mathfrak n_S)\overline Z_{S,j}
    -
    \theta_{0j}^\star
    =
    -
    \bigl(
        1-w_j(\mathfrak n_S)
    \bigr)
    \theta_{0j}^\star
    +
    w_j(\mathfrak n_S)h_{S,j}
    +
    w_j(\mathfrak n_S)\eta_{S,j}.
\end{aligned}
\]
Write these three terms as
\(
    B_{S,j},
    ~
    H_{S,j},
    ~
    G_{S,j},
\)
respectively. By Parseval's identity and
Lemma~\ref{lem:pc-pooling},
\[
    \sum_{j\geq1}B_{S,j}^2
    \leq
    C\mathfrak n_S^{-2a_\alpha}
    \leq
    C\varepsilon_N^2.
\]
By the definition
\eqref{eq:pc-shared-discrepancy},
\[
    \sum_{j\geq1}H_{S,j}^2
    =
    \bigl(
        \Delta_{S,N}^{\mathrm{sh}}
    \bigr)^2
    \leq
    \bigl(
        \Delta_N^{\mathrm{sh}}
    \bigr)^2
    \leq
    C\varepsilon_N^2.
\]
Thus the deterministic part of the oracle posterior-mean error has
$\mathbb H$-norm at most
\(
    C\varepsilon_N.
\)
The Gaussian term
\(
    G_S
    =
    \sum_{j\geq1}
    G_{S,j}\varphi_j
\)
has covariance eigenvalues
\(
    \frac{
        w_j(\mathfrak n_S)^2
    }{
        \mathfrak n_S
    }.
\)
Lemma~\ref{lem:pc-pooling} therefore gives
\[
    \operatorname{tr}
    \operatorname{Cov}(G_S)
    \leq
    C\mathfrak n_S^{-2a_\alpha}
    \leq
    C\varepsilon_N^2
\]
and
\[
    \|
        \operatorname{Cov}(G_S)
    \|_{\mathrm{op}}
    \leq
    \mathfrak n_S^{-1}
    \leq
    \underline{\mathfrak n}_N^{-1}.
\]
Apply
Lemma~\ref{lem:pc-gaussian-tail}
with
\(
    t=t_N+\ell_N.
\)
For each fixed $S$,
\[
\begin{aligned}
    \mathbb P_\star
    \Bigg(
        \|G_S\|_{\mathbb H}
        >
        C\varepsilon_N
        +
        \sqrt{
            \frac{
                2(t_N+\ell_N)
            }{
                \underline{\mathfrak n}_N
            }
        }
    \Bigg)
    \leq
    e^{-t_N-\ell_N}.
\end{aligned}
\]
Since
\(
    t_N
    =
    \underline{\mathfrak n}_N
    \varepsilon_N^2
\),~
$\ell_N=O(t_N)$,
\(
    \sqrt{
        \frac{
            t_N+\ell_N
        }{
            \underline{\mathfrak n}_N
        }
    }
    =
    O(\varepsilon_N).
\)
Hence there exists $C_1<\infty$ such that
\[
    \mathbb P_\star
    \left(
        \|G_S\|_{\mathbb H}
        >
        C_1\varepsilon_N
    \right)
    \leq
    e^{-t_N-\ell_N}.
\]
Taking a union bound over
$\mathcal G_N^{\mathrm{ctr}}$ gives
\[
\begin{aligned}
    \mathbb P_\star
    \left(
        \max_{S\in\mathcal G_N^{\mathrm{ctr}}}
        \|G_S\|_{\mathbb H}
        >
        C_1\varepsilon_N
    \right)
    \leq
    |
        \mathcal G_N^{\mathrm{ctr}}
    |
    e^{-t_N-\ell_N}
    &\leq
    \frac12e^{-t_N}
    \leq
    e^{-t_N}.
\end{aligned}
\]
Combining the deterministic bias, the deterministic discrepancy, and
the Gaussian mean-noise term yields
\[
    \max_{S\in\mathcal G_N^{\mathrm{ctr}}}
    \|
        \overline\mu_{S,N}^{(0)}-f_0
    \|_{\mathbb H}
    \leq
    C\varepsilon_N
\]
outside an event of probability at most $e^{-t_N}$. This proves
\eqref{eq:pc-shared-mean-bound}.
\\
Finally, the target posterior covariance for subset $S$ has eigenvalues
$v_j(\mathfrak n_S)$. Lemma~\ref{lem:pc-pooling} gives
\[
\begin{aligned}
    \operatorname{tr}
    \bigl(
        C_{S,N}^{(0)}
    \bigr)
    =
    \sum_{j\geq1}
    v_j(\mathfrak n_S)
    &\leq
    C\mathfrak n_S^{-2a_\alpha}
    \leq
    C\varepsilon_N^2,
\end{aligned}
\]
and
\[
\begin{aligned}
    \|
        C_{S,N}^{(0)}
    \|_{\mathrm{op}}
    =
    \sup_{j\geq1}
    v_j(\mathfrak n_S)
    &\leq
    \mathfrak n_S^{-1}
    \leq
    \underline{\mathfrak n}_N^{-1}.
\end{aligned}
\]
Taking the supremum over
$S\in\mathcal G_N^{\mathrm{ctr}}$
proves
\eqref{eq:pc-shared-covariance-bound}.
\\
Since
\(
    t_N
    =
    \underline{\mathfrak n}_N
    \varepsilon_N^2
    =
    \underline{\mathfrak n}_N^{
        d/(2\alpha+d)
    }
    \longrightarrow\infty,
\)
the failure probability tends to zero. Hence the oracle conditions of
Theorem~\ref{thm:posterior-contraction} hold with
$\mathfrak i_N=\underline{\mathfrak n}_N$.
\end{proof}

\subsection{Proof of Corollary~\ref{cor:pc-pooled-transfer}}
\label{app:proof:cor:pc-pooled-transfer}

\begin{proof}
Theorem~\ref{thm:pc-pooled-rate} verifies the oracle mean and covariance
conditions of
Theorem~\ref{thm:posterior-contraction}
with
\(
    \mathfrak i_N
    =
    \underline{\mathfrak n}_N
\)
and oracle rate $\varepsilon_N$. The deterministic inclusion above
allows Theorem~\ref{thm:predictive_selection} to control the posterior
weight outside
$\mathcal G_N^{\mathrm{ctr}}$.
Finally,
$\zeta_N^{\mathrm{bt}}(u_N)\to0$
controls the Bellman-induced posterior-mean perturbation, while
$u_N=O(\varepsilon_N)$ preserves the oracle rate.
Applying
Theorem~\ref{thm:posterior-contraction}
proves the claim.
\end{proof}

\subsection{Proof of Corollary~\ref{cor:pc-mean-risk}}
\label{app:proof:cor:pc-mean-risk}

\begin{proof}
Let $\mathcal E_N$ be the high-probability event constructed in
Theorem~\ref{thm:posterior-contraction}. By that theorem,
\(
    \mathbb P_\star(\mathcal E_N^c)
    \leq
    \delta_N^{\mathrm{ctr}}.
\)
Moreover, on $\mathcal E_N$,
\begin{equation}\nonumber
    \sup_{S\in\mathcal G_N^{\mathrm{ctr}}}
    \int
        \|f-f_0\|_{\mathbb H}^2
        \,
        \Pi_{S,e,N}^{(0)}(\mathrm df)
    \leq
    Cr_N^2
    \label{eq:pc-good-component-moment}
\end{equation}
for a fixed constant $C$.
Since $\mathbb B_L$ is closed and convex and
$f_0\in\mathbb B_L$, metric projection cannot increase distance from
$f_0$:
\(
    \|T_Lf-f_0\|_{\mathbb H}
    \leq
    \|f-f_0\|_{\mathbb H}.
\)
Also,
\(
    \|T_Lf-f_0\|_{\mathbb H}
    \leq
    2L
\)
for every $f\in\mathbb H$.
Using the mixture representation
\eqref{eq:pc-mixture}, on $\mathcal E_N$,
\[
\begin{aligned}
    &
    \int
        \|T_Lf-f_0\|_{\mathbb H}^2
        \,
        \widetilde\Pi_{e,N}^{(0)}(\mathrm df)
    \\
    &=
    \sum_{S\in\mathcal G_N^{\mathrm{ctr}}}
    \omega_{e,N}(S)
    \int
        \|T_Lf-f_0\|_{\mathbb H}^2
        \,
        \Pi_{S,e,N}^{(0)}(\mathrm df)
    +
    \sum_{S\notin\mathcal G_N^{\mathrm{ctr}}}
    \omega_{e,N}(S)
    \int
        \|T_Lf-f_0\|_{\mathbb H}^2
        \,
        \Pi_{S,e,N}^{(0)}(\mathrm df)
    \\
    &\leq
    Cr_N^2
    +
    4L^2
    \Omega_N^{\mathrm{ctr}}.
\end{aligned}
\]
On $\mathcal E_N^c$, the same integrated squared distance is bounded by
$4L^2$.
The population inclusion used in
Theorem~\ref{thm:posterior-contraction} implies, for all sufficiently
large $N$,
\(
    S\notin\mathcal G_N^{\mathrm{ctr}}
    \Longrightarrow
    S\in
    \mathcal B_N^{\mathrm{pred}}(d_N).
\)
Hence
\[
    \Omega_N^{\mathrm{ctr}}
    \leq
    \Pi_{e,N}
    \left(
        \mathcal B_N^{\mathrm{pred}}(d_N)
        \mid
        \mathbf Y_e,X
    \right),
\]
and therefore
\(
    \mathbb E_\star
    \Omega_N^{\mathrm{ctr}}
    \leq
    \delta_N^{\mathrm{sel}}.
\)
Finally, Hilbert-space Jensen's inequality gives
\[
\begin{aligned}
    \|
        \widehat f_{e,N}^{L}-f_0
    \|_{\mathbb H}^2
    &\leq
    \int
        \|f-f_0\|_{\mathbb H}^2
        \,
        \widetilde\Pi_{e,N}^{(0),L}(\mathrm df).
\end{aligned}
\]
Taking expectations and splitting over
$\mathcal E_N$ and $\mathcal E_N^c$ yields
\[
    \mathbb E_\star
    \left[
        \|
            \widehat f_{e,N}^{L}-f_0
        \|_{\mathbb H}^2
    \right]
    \leq
    Cr_N^2
    +
    4L^2
    \left(
        \delta_N^{\mathrm{ctr}}
        +
        \delta_N^{\mathrm{sel}}
    \right),
\]
which proves
\eqref{eq:pc-projected-risk}.
\end{proof}

\subsection{Proof of Theorem~\ref{thm:pc-minimax}}
\label{app:proof:thm:pc-minimax}

\begin{proof}
For each frequency $j$, define the whitened observation vector
\[
    \mathbf Y_j
    =
    \left(
        \sqrt{\tau_k}\,
        \overline Z_{kj}
    \right)_{k\in\mathcal I(S)}.
\]
Under exact sharing,
\[
    \mathbf Y_j
    =
    \mathbf q_S
    \theta_{0j}^\star
    +
    \boldsymbol\xi_j,
\]
where
\(
    \mathbf q_S
    =
    \left(
        \sqrt{\tau_k}g_k
    \right)_{k\in\mathcal I(S)}
\)
and
\(
    \|
        \mathbf q_S
    \|_2^2
    =
    n.
\)
Choose an orthogonal transformation whose first row is
$\mathbf q_S^\top/\sqrt n$. The first transformed coordinate is
\(
    \sqrt n\,
    \theta_{0j}^\star
    +
    \xi_j,
\)
while all remaining coordinates are standard Gaussian variables whose
joint law does not depend on $\theta_{0j}^\star$. Hence the full
experiment is statistically equivalent, for inference on $f_0$, to
the scalar Gaussian sequence
\begin{equation}
    Y_j
    =
    \theta_{0j}^\star
    +
    n^{-1/2}\xi_j,
    \quad
    \xi_j
    \stackrel{\mathrm{iid}}{\sim}
    \mathcal N(0,1).
    \label{eq:pc-equivalent-scalar-sequence}
\end{equation}
For the upper bound, the posterior mean is
\(
    \widehat\theta_{j,n}
    =
    w_j(n)Y_j.
\)
Its coordinatewise frequentist risk is
\[
\begin{aligned}
    \mathbb E_{f_0}
    \left[
        \left(
            \widehat\theta_{j,n}
            -
            \theta_{0j}^\star
        \right)^2
    \right]
    &=
    \left(
        1-w_j(n)
    \right)^2
    (\theta_{0j}^\star)^2
    +
    \frac{
        w_j(n)^2
    }{
        n
    }.
\end{aligned}
\]
Summing over $j$ and applying
Lemma~\ref{lem:pc-pooling} gives
\[
    \sup_{f_0\in\Theta_\alpha(L)}
    \mathbb E_{f_0}
    \left[
        \|
            \widehat f_n-f_0
        \|_{\mathbb H}^2
    \right]
    \leq
    C
    n^{-2\alpha/(2\alpha+d)}.
\]
This proves the upper bound.
\\
For the lower bound, let
\(
    s=\frac{\alpha}{d}
\)
and choose
\(
    m
    =
    \left\lfloor
        n^{1/(1+2s)}
    \right\rfloor.
\)
For all sufficiently large $n$, $m\geq1$. For
$\omega\in\{-1,1\}^{m}$, define the coefficient sequence
\[
    \theta_j^\omega
    =
    \begin{cases}
        h\omega_{j-m},
        & m<j\leq2m,\\
        0,
        & \text{otherwise},
    \end{cases}
\]
where
\(
    h
    =
    c_0m^{-s-1/2}.
\)
Then
\[
\begin{aligned}
    \sum_{j\geq1}
    j^{2s}
    (\theta_j^\omega)^2
    \leq
    m(2m)^{2s}h^2
    &=
    2^{2s}c_0^2.
\end{aligned}
\]
Choosing $c_0>0$ sufficiently small ensures
$f^\omega\in\Theta_\alpha(L)$ for every $\omega$.
Let $\omega^{(r)}$ differ from $\omega$ only in coordinate $r$.
The two corresponding scalar sequence experiments differ by $2h$ in
one coordinate, therefore
\[
    \operatorname{KL}
    \left(
        P_\omega,
        P_{\omega^{(r)}}
    \right)
    =
    2nh^2.
\]
Since
$m=\lfloor n^{1/(1+2s)}\rfloor$,
we have $nm^{-(1+2s)}\leq 2^{1+2s}$
for all sufficiently large $n$.
Consequently,
\[
\operatorname{KL}(P_\omega,P_{\omega^{(r)}})
=2nh^2
=2c_0^2nm^{-(1+2s)}
\leq 2^{2+2s}c_0^2.
\]
Choosing $c_0>0$ sufficiently small ensures that
\(
\operatorname{KL}(P_\omega,P_{\omega^{(r)}})
\leq \frac18
\)
uniformly over all
$\omega\in\{-1,1\}^m$
and $r\in\{1,\ldots,m\}$.
Pinsker's inequality then gives
\[
    \|
        P_\omega
        -
        P_{\omega^{(r)}}
    \|_{\mathrm{TV}}
    \leq
    \frac14.
\]
For any estimator $\widehat f$, let
$\widehat\theta_j$
denote its coefficients and define
\(
    \widehat\omega_r
    =
    \operatorname{sign}
    \left(
        \widehat\theta_{m+r}
    \right),
\)
with any fixed convention when the argument is zero.
Whenever
$\widehat\omega_r\neq\omega_r$,
\(
    \left|
        \widehat\theta_{m+r}
        -
        \theta_{m+r}^{\omega}
    \right|
    \geq
    h.
\)
Consequently,
\[
    \|
        \widehat f-f^\omega
    \|_{\mathbb H}^2
    \geq
    h^2
    \sum_{r=1}^{m}
    \mathbf 1
    \left\{
        \widehat\omega_r\neq\omega_r
    \right\}.
\]
Pairing neighboring vertices of the hypercube, the average testing
error for each coordinate is at least
\[
    \frac{
        1-
        \|
            P_\omega
            -
            P_{\omega^{(r)}}
        \|_{\mathrm{TV}}
    }{
        2
    }
    \geq
    \frac38.
\]
It follows that
\[
\begin{aligned}
    \sup_{\omega}
    \mathbb E_\omega
    \left[
        \|
            \widehat f-f^\omega
        \|_{\mathbb H}^2
    \right]
    \geq
    \frac38
    mh^2
    =
    c\,m^{-2s}
    \geq
    c\,
    n^{-2s/(1+2s)}
    =
    c\,
    n^{-2\alpha/(2\alpha+d)}.
\end{aligned}
\]
Taking the infimum over all estimators proves the lower bound.
\end{proof}

\subsection{Proof of Proposition~\ref{prop:pc-bayes-gain}}
\label{app:proof:prop:pc-bayes-gain}

\begin{proof}
Because
$\mathscr D_0\subseteq\mathscr D_{0S}$,
the tower property gives
\(
    \mathbb E
    \left[
        m_{0S}
        \mid
        \mathscr D_0
    \right]
    =
    m_0.
\)
Write
\[
    F-m_0
    =
    (F-m_{0S})
    +
    (m_{0S}-m_0).
\]
Expanding the squared norm yields
\[
\begin{aligned}
    \|F-m_0\|_{\mathbb H}^2
    =
    \|F-m_{0S}\|_{\mathbb H}^2
    +
    \|m_{0S}-m_0\|_{\mathbb H}^2
    +
    2
    \left\langle
        F-m_{0S},
        m_{0S}-m_0
    \right\rangle_{\mathbb H}.
\end{aligned}
\]
The cross term has expectation zero because
$m_{0S}-m_0$ is
$\mathscr D_{0S}$-measurable and
\[
    \mathbb E
    \left[
        F-m_{0S}
        \mid
        \mathscr D_{0S}
    \right]
    =
    0.
\]
Taking expectations proves
\eqref{eq:pc-bayes-identity}.
\\
The right-hand side of
\eqref{eq:pc-bayes-identity}
vanishes if and only if
$m_{0S}=m_0$ almost surely, proving the strictness criterion.
For the Gaussian case,
\[
    \mathbb E
    \left[
        \|F-m_0\|_{\mathbb H}^2
        \mid
        \mathscr D_0
    \right]
    =
    \operatorname{tr}(C_0)
\]
and
\[
    \mathbb E
    \left[
        \|F-m_{0S}\|_{\mathbb H}^2
        \mid
        \mathscr D_{0S}
    \right]
    =
    \operatorname{tr}(C_{0S}).
\]
Taking expectations and subtracting proves
\eqref{eq:pc-bayes-covariance-gain}.
\end{proof}

\clearpage
\section{Proof of the offline policy-performance bound}
\label{app:policy-proofs}

\subsection{Proof of Theorem~\ref{thm:offline-safe-transfer-performance}}
\label{app:proof:thm:offline-safe-transfer-performance}

\begin{proof}
Fix $s$ and abbreviate
\(
    a_c
    =
    a_{\beta,*}^{(0)}(s),
    ~
    \widehat a
    =
    \widehat\pi_{e,N}(s).
\)
By the definition of
$\widehat\pi_{e,N}$,
\[
\begin{aligned}
    \widetilde\mu_{e,N}^{(0)}
    \bigl(
        \phi^{(0)}(s,a_c)
    \bigr)
    -
    \mathfrak p_{e,N}^{(0)}
    \bigl(
        \phi^{(0)}(s,a_c)
    \bigr)
    \leq
    \widetilde\mu_{e,N}^{(0)}
    \bigl(
        \phi^{(0)}(s,\widehat a)
    \bigr)
    -
    \mathfrak p_{e,N}^{(0)}
    \bigl(
        \phi^{(0)}(s,\widehat a)
    \bigr).
\end{aligned}
\]
Using
\(
    \widetilde\mu_{e,N}^{(0)}
    =
    f_0+e_{e,N}^{(0)}
\)
and
\(
    f_0
    \bigl(
        \phi^{(0)}(s,a)
    \bigr)
    =
    Q_*^{(0)}(s,a),
\)
we obtain
\[
\begin{aligned}
    Q_*^{(0)}(s,a_c)
    -
    Q_*^{(0)}(s,\widehat a)
    \leq
    e_{e,N}^{(0)}
    \bigl(
        \phi^{(0)}(s,\widehat a)
    \bigr)
    -
    e_{e,N}^{(0)}
    \bigl(
        \phi^{(0)}(s,a_c)
    \bigr)
    +
    \mathfrak p_{e,N}^{(0)}
    \bigl(
        \phi^{(0)}(s,a_c)
    \bigr)
    -
    \mathfrak p_{e,N}^{(0)}
    \bigl(
        \phi^{(0)}(s,\widehat a)
    \bigr).
\end{aligned}
\]
Since
$\mathfrak p_{e,N}^{(0)}\geq0$,
\[
\begin{aligned}
    Q_*^{(0)}(s,a_c)
    -
    Q_*^{(0)}(s,\widehat a)
    \leq
    \left|
        e_{e,N}^{(0)}
        \bigl(
            \phi^{(0)}(s,a_c)
        \bigr)
    \right|
    +
    \left|
        e_{e,N}^{(0)}
        \bigl(
            \phi^{(0)}(s,\widehat a)
        \bigr)
    \right|
    +
    \mathfrak p_{e,N}^{(0)}
    \bigl(
        \phi^{(0)}(s,a_c)
    \bigr).
\end{aligned}
\]
By
\eqref{eq:pc-coverage-gap},
\[
\begin{aligned}
    V_*^{(0)}(s)
    -
    Q_*^{(0)}(s,\widehat a)
    =
    c_\beta^{(0)}(s)
    +
    Q_*^{(0)}(s,a_c)
    -
    Q_*^{(0)}(s,\widehat a),
\end{aligned}
\]
which proves
\eqref{eq:pc-greedy-gap}.
\\
Define
\[
    g_{e,N}(s)
    =
    V_*^{(0)}(s)
    -
    Q_*^{(0)}
    \bigl(
        s,\widehat\pi_{e,N}(s)
    \bigr).
\]
Optimality of $V_*^{(0)}$ implies
\(
    g_{e,N}(s)\geq0.
\)
The Bellman equations for
$Q_*^{(0)}$
and
$V^{\widehat\pi_{e,N}}$
give
\begin{equation}
\begin{aligned}
    V_*^{(0)}
    -
    V^{\widehat\pi_{e,N}}
    &=
    g_{e,N}
    +
    \gamma
    P^{\widehat\pi_{e,N}}
    \left(
        V_*^{(0)}
        -
        V^{\widehat\pi_{e,N}}
    \right).
\end{aligned}
\label{eq:pc-performance-recursion}
\end{equation}
Because the value functions are bounded, iterating
\eqref{eq:pc-performance-recursion}
gives
\begin{equation}
    V_*^{(0)}
    -
    V^{\widehat\pi_{e,N}}
    =
    \sum_{t=0}^{\infty}
    \gamma^t
    \left(
        P^{\widehat\pi_{e,N}}
    \right)^t
    g_{e,N}.
    \label{eq:pc-performance-resolvent}
\end{equation}
Hence
\[
    \|
        V_*^{(0)}
        -
        V^{\widehat\pi_{e,N}}
    \|_\infty
    \leq
    \frac{
        \|g_{e,N}\|_\infty
    }{
        1-\gamma
    }.
\]
Taking the supported suprema in
\eqref{eq:pc-greedy-gap}
therefore proves
\eqref{eq:pc-offline-sup}.

We next establish the average-performance bound.
Integrating
\eqref{eq:pc-performance-resolvent}
under the initial-state law $\rho_0$ gives
\begin{equation}
\begin{aligned}
    \mathbb E_{\rho_0}
    \left[
        V_*^{(0)}(S_0)
        -
        V^{\widehat\pi_{e,N}}(S_0)
    \right]
    =
    \frac{1}{1-\gamma}
    \mathbb E_{
        d_{\rho_0}^{\widehat\pi_{e,N}}
    }
    \left[
        g_{e,N}(S)
    \right].
\end{aligned}
\label{eq:pc-occupancy-identity}
\end{equation}
By construction,
\(
    \widehat\pi_{e,N}
    \in
    \Pi_\beta^{(0)}.
\)
Integrating
\eqref{eq:pc-greedy-gap}
with respect to
$d_{\rho_0}^{\widehat\pi_{e,N}}$
therefore gives
\[
\begin{aligned}
    \mathbb E_{
        d_{\rho_0}^{\widehat\pi_{e,N}}
    }
    g_{e,N}(S)
    \leq
    \mathbb E_{
        d_{\rho_0}^{\widehat\pi_{e,N}}
    }
    c_\beta^{(0)}(S)
    +
    \int
        |e_{e,N}^{(0)}(x)|
        \,
        \nu_\beta^{\widehat\pi_{e,N}}
        (\mathrm dx)
    +
    \int
        |e_{e,N}^{(0)}(x)|
        \,
        \nu_{\widehat\pi_{e,N}}^{
            \widehat\pi_{e,N}
        }
        (\mathrm dx)
    +
    \int
        \mathfrak p_{e,N}^{(0)}(x)
        \,
        \nu_\beta^{\widehat\pi_{e,N}}
        (\mathrm dx).
\end{aligned}
\]
Let $\nu\ll P_X^{(0)}$ and suppose
\(
    \left\|
        \frac{\mathrm d\nu}
             {\mathrm dP_X^{(0)}}
    \right\|_\infty
    \leq
    C.
\)
For every $q\in\mathbb H$,
Cauchy--Schwarz gives
\[
\begin{aligned}
    \int
        |q(x)|
        \,
        \nu(\mathrm dx)
    =
    \int
        |q(x)|
        \frac{\mathrm d\nu}
             {\mathrm dP_X^{(0)}}(x)
        P_X^{(0)}(\mathrm dx)
    \leq
    \|q\|_{\mathbb H}
    \left[
        \int
        \left(
            \frac{\mathrm d\nu}
                 {\mathrm dP_X^{(0)}}
        \right)^2
        \mathrm dP_X^{(0)}
    \right]^{1/2}.
\end{aligned}
\]
Since the density ratio is bounded by $C$ and integrates to one,
\[
    \int
    \left(
        \frac{\mathrm d\nu}
             {\mathrm dP_X^{(0)}}
    \right)^2
    \mathrm dP_X^{(0)}
    \leq
    C.
\]
Thus
\begin{equation}
    \int
        |q|
        \,
        \mathrm d\nu
    \leq
    \sqrt C\,
    \|q\|_{\mathbb H}.
    \label{eq:pc-density-conversion}
\end{equation}
Applying
\eqref{eq:pc-density-conversion}
with the population bounds in
\eqref{eq:pc-concentrability}
yields
\[
\begin{aligned}
    \mathbb E_{
        d_{\rho_0}^{\widehat\pi_{e,N}}
    }
    g_{e,N}(S)
    \leq
    \mathbb E_{
        d_{\rho_0}^{\widehat\pi_{e,N}}
    }
    c_\beta^{(0)}(S)
    +
    \left(
        \sqrt{C_\beta^{(0)}}
        +
        \sqrt{C_{\pi,\beta}^{(0)}}
    \right)
    \|e_{e,N}^{(0)}\|_{\mathbb H}
    +
    \sqrt{C_\beta^{(0)}}
    \|
        \mathfrak p_{e,N}^{(0)}
    \|_{\mathbb H}.
\end{aligned}
\]
Substitution into
\eqref{eq:pc-occupancy-identity}
proves
\eqref{eq:pc-offline-l2}.
\end{proof}

\end{document}